%% file: root.tex
\documentclass[journal]{IEEEtran}
\newcommand{\mb}[1]{\mathbf{#1}}
\newcommand{\behaviortext}[0]{\text{b}}

\usepackage[utf8]{inputenc}
\usepackage{pgfplots}
\usepackage{subcaption} 
\usepackage{graphicx}
\usepackage{siunitx}
\usepackage{tikz}
\usepackage{siunitx}
\usepackage{multicol}
\usepackage{etoolbox}
\usepackage[outline]{contour}
\makeatletter
\patchcmd{\Ginclude@eps}{"#1"}{#1}{}{}
\makeatother
\usetikzlibrary{fit,calc,trees,positioning,arrows,chains,shapes.geometric,decorations.pathreplacing,decorations.pathmorphing,shapes,matrix,shapes.symbols,matrix,backgrounds,spy}
\DeclareSIUnit{\mio}{Mio.}

\ifCLASSINFOpdf
\graphicspath{{Images/Tikz/}{Images/svgs/}}
\else
\fi
\usepackage{amsmath}
\usepackage{amssymb}

\newcommand\copyrighttext{%
	\scriptsize \copyright~2021 IEEE.  Personal use of this material is permitted.  Permission from IEEE must be obtained for all other uses, in any current or future media, including reprinting/republishing this material for advertising or promotional purposes, creating new collective works, for resale or redistribution to servers or lists, or reuse of any copyrighted component of this work in other works.}%

\newcommand\copyrightnotice{%
	\begin{tikzpicture}[remember picture,overlay]
	\node[anchor=south,yshift=10pt,xshift=0.25cm] at (current page.south) {{\parbox{\dimexpr\textwidth-\fboxsep-\fboxrule\relax}{\copyrighttext}}};
	\end{tikzpicture}%
}

\begin{document}
	\title{On-Road Motion Planning for Automated Vehicles at Ulm University}
	
	\author{Maximilian~Graf*,~Oliver~Speidel*,~Jona~Ruof~and~Klaus~Dietmayer% <-this % stops a space\
		\thanks{* M. Graf and O. Speidel contributed equally.}
		\thanks{M. Graf, O. Speidel, J. Ruof and K. Dietmayer are with the Institute 
			of Measurement, Control and Microtechnology of the University of Ulm,  
			89081 Ulm, Germany, e-mail: firstname.lastname@uni-ulm.de.}% <-this %
		%stops a space
%		\thanks{J. Doe and J. Doe are with Anonymous University.}% <-this % stops a
		%space
%		\thanks{Manuscript received April 19, 2005; revised August 26, 2015.}
	}

	\maketitle
	\copyrightnotice
	\begin{abstract}
		The Institute of Measurement, Control and Microtechnology at Ulm University 
		investigates advanced driver assistance systems for decades
		and concentrates in large parts on autonomous driving.
		It is well known that motion planning is a key technology for autonomous driving.
		It is first and foremost responsible for the safety of the vehicle passengers as
		well as of all surrounding traffic participants. However, a further task consists in
		providing a smooth and comfortable driving behavior. In Ulm,  we have the grateful
		opportunity to test our algorithms under real conditions in public traffic and
		diversified scenarios.
		In this paper, we would like to give the readers an insight of our work, about
		the vehicle, the test track, as well as of the related problems, challenges and
		solutions.  
		Therefore, we will describe the motion planning system and explain the implemented functionalities. 
		Furthermore, we will show how our vehicle moves through public road traffic and how it
		deals with challenging scenarios like e.g. driving through roundabouts and intersections.
	\end{abstract}
	
	\begin{IEEEkeywords}
		autonomous driving, automated vehicles, motion planning, trajectory planning.
	\end{IEEEkeywords}

	% \fi
	\IEEEpeerreviewmaketitle

	\section{Introduction}

\begin{figure}[t!]
		\includegraphics[width=\linewidth]{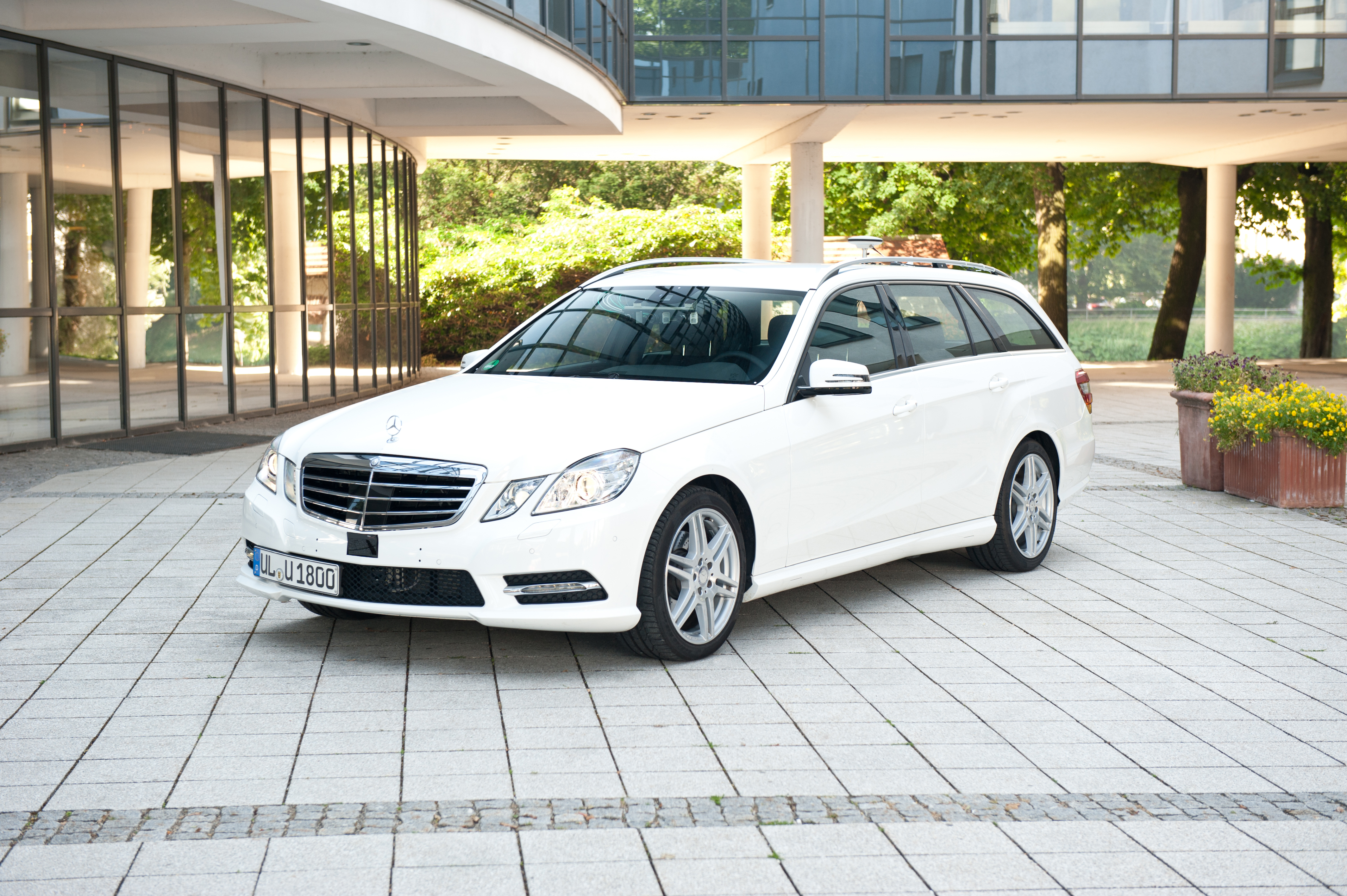}
		\caption{Experimental vehicle of the Institute of Measurement, Control and
			Microtechnology of the University of Ulm. Photo: Elvira Eberhardt, University of
			Ulm.}
		\label{fig:susi}
	\end{figure}

	\IEEEPARstart{S}{ince} several years, intensive research has been conducted on automated driving.
	In  2007, 11 vehicles were sent into a race through an urban environment in
	Victorville, California. Thereby, the goal was to complete a
	\SI{96}{\kilo\metre} course within 6 hours, whereas the vehicles had to
	navigate fully by themselves and obey Californian traffic rules. This event is
	well known as the Darpa Urban Challenge and  accepted to be a milestone of
	autonomous driving \cite{buehler2009darpa}. In 2013, the S-Class Bertha made its
	way through a $\SI{103}{\kilo\metre}$ route from Mannheim to Pforzheim in
	Germany \cite{makingberthadrive}.  Since 2017, the companies Waymo and Uber are
	providing taxi services in a small city close  to Phoenix in Arizona\texttt{}.
	But these are just the most recent and well known developments. Much effort is
	done in daily work behind closed doors of research institutes and companies.
	But still, automated driving functions are by far not yet included in every
	productive car.
	The problem arises from the complexity of road traffic due to the  large
	variety of situations, the huge number of maneuver and behavior options and the
	difficulty in predicting traffic participants.
	Thus, its not astonishing that the ECUs of modern productive cars contain up to
	$\SI{100}{\mio}$ lines of code. For comparison: The number of lines of code
	contained by the Boeing 787 is estimated to  approximately $\SI{6.5}{\mio}$ \cite{charette2009car}.
	\newline
	The Institute of Measurement Control and Microtechnology at Ulm University 
	deals with autonomous driving especially in structured environments and
	on-roads scenarios.
	First successes were reached in the year 2014: A Mercedes Benz E-Class drove a
	challenging route including roundabouts, pedestrian crosswalks, traffic lights
	etc., around the University fully autonomously
	\cite{autonomousdrivingatulmuniversity}.
	Meanwhile, the engineers of the institute equipped another vehicle, which is
	shown in figure~\ref{fig:susi}.
	Recently, a new motion planning system was designed and
	integrated, which enables the vehicle to navigate through longitudinal traffic,
	perform lane change and overtaking maneuvers as well as handle complicated
	intersection and roundabout scenarios. Furthermore, an emergency mode was developed to plan full braking and evasive trajectories in safety critical situations.
	In this paper we'd like to give the readers an insight of the motion planning
	system and show how the related driving functions apply under real conditions in
	public road traffic.	
	\newline
	The remainder of this paper is  structured as follows:
	Section~\ref{sec:related_work} can be seen as a summary of related work, in
	which we describe the requirements and challenges as well as a representative 
	excerpt of todays state of the art motion planning techniques.
	In section \ref{sec:Methodology}, we present our motion planning system which is composed of several operational modes and describe how these modes are realized to retrieve an overall system. 
	In the evaluation section, experiments are carried  out and the performance of
	our system is shown by different maneuvers in public road traffic. 	
	%\IEEEPARstart{T}{his} demo file is intended to serve as a ``starter file''
	%for IEEE journal papers produced under \LaTeX\ using
	%IEEEtran.cls version 1.8b and later.
	%I wish you the best of success.
	\section{On-Road Motion Planning - Requirements and Related Work}
	\label{sec:related_work}
\begin{figure}	
		\def\svgwidth{0.9\columnwidth} 	
		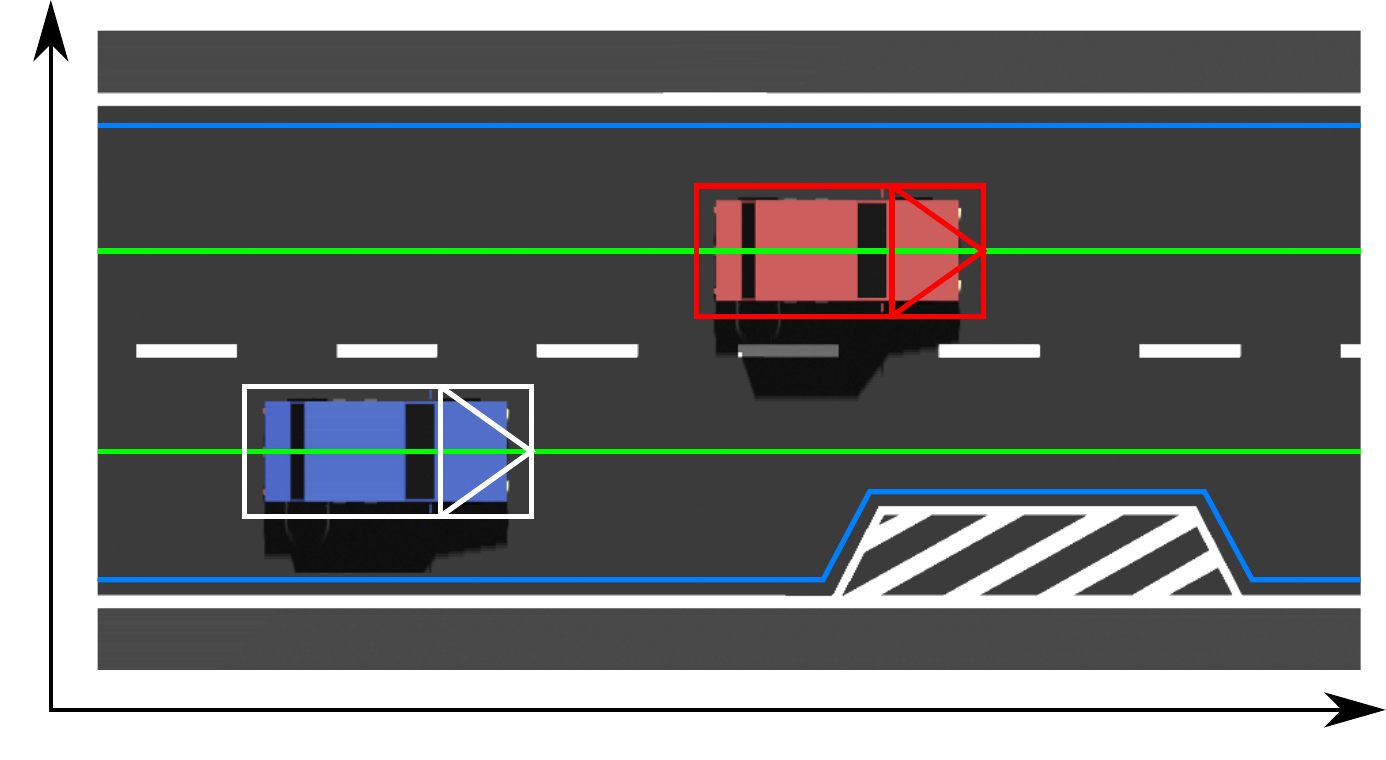
		\caption{Environmental model used in this work.
			Vehicle shapes are approximated by rectangle boxes. Lane boundaries are depicted in blue and the lane center lines in green.}
		\label{fig:env}
	\end{figure}

	Motion planners are key components for automated vehicles. They are mainly responsible for the resulting driving behavior and
	their main task is to ensure  safety for the vehicle passengers as well as of
	all surrounding traffic participants.
	In non-safety critical situations, they also have to provide non-jerky and
	comfortable driving behavior.
	Furthermore, the computation cycle time has to be small in order to react to
	dynamic changes of the environment in time.
	\newline
	 A common environmental representation often used for on-road driving is shown in figure \ref{fig:env}. Thereby, the vehicle environment	
	consists of multiple lanes which are represented by one center line and two
	associated freespace boundaries each. Static obstacles on the roadside are
	excluded from the drivable freespace by the boundary lines.
	Dynamic obstacles, such as cars are represented by rectangle boxes.
	Due to the importance of motion and trajectory planning, the corresponding research field is well advanced and many approaches  exist already. Therefore, we'd
	like to briefly summarize some of the most widely spread techniques.
	\newline
        We acknowledge that some recent planning approaches leverage learning based methods and partially observable Markov decision processes (POMDPs) with promising results (e.g. \cite{Kendall2019}, \cite{Codeville2019}, \cite{Eshed2020}, \cite{POMDPs}).
        This summary, however, will be focused on classical, optimization based approaches,
        which are, at the current state of research, still more easily and safely applicable to real world vehicles.
    \newline
	In general, trajectory planning methods can be classified into discrete and
	continuous approaches. 
	Using the former, the continuous space of trajectories gets discretized by
	generating a set of trajectory alternatives. This set might either consist of
	trajectory samples or corresponds to a search graph, whereas the graph edges
	represent motion alternatives from which the final trajectory has to be
	composed. 
	By using continuous approaches however, no such discretization is done and the
	continuous space is investigated directly by applying numerical optimization
	techniques for instance.
	A strength of discrete methods consists of the fact, that the best solution within the discrete set will always be found.
	However, since no continuous optimization is applied, this discrete solution will always be suboptimal and located in the neighborhood of the continuous global optimum at	best. 
	In contrast, continuous approaches will find a local optimal solution and in case of a
	convex problem even the global optimum. In presence of multiple local optima
	however, such methods are dependent on an appropriate initial solution to converge into the
	desired optimum.
	\newline
	A discrete method based on polynomial sampling is presented in
	\cite{werling2010optimal}. Therein, quinitic polynomials which are optimal in
	terms of the squared jerk are used to describe the longitudinal and lateral
	movement within a Fren\'et frame. The resulting trajectories are smooth and lead to comfortable driving
	behavior. However, the flexibility of the resulting trajectories is limited by
	the polynomial degree and the method is not well suited for large temporal
	planning horizons.
	In \cite{Hubmann2016} this problem is addressed, where a discrete action set is
	utilized in order to generate a state graph. Using graph-search methods, the best
	sequence of actions and associated states is extracted.
	As a result, a rough trajectory can be obtained which gets further optimized using
	the polynomial-based sampling approach presented in \cite{werling2010optimal}. 
	Another well-known discrete approach are the 
	Rapidly Exploring Random Trees (RRT) which generate a search graph by using a
	probabilistic discretization scheme. RRTs were e.g. used by the 
	MIT during the Darpa Urban Challenge~\cite{kuwata2009real}.
	These discrete approaches have proven great practicability in real world scenarios, but as discussed, always yield suboptimal trajectories.
	\newline
	A local continuous approach  which was used during the Bertha Benz Drive in 2013
	is presented in \cite{trajectoryplanningforbertha}. Therein, a static
	optimization problem is formulated, whereas the optimization variables are the
	spatio-temporal positions of the vehicle's reference position. In principle, arbitrary maneuvers or trajectories respectively can be planned, provided they satisfy  acceleration and curvature constraints.	
	Thus, the resulting trajectory model is flexible and the method is well suited for long term planning. 
	\newline
	In  \cite{nilsson2015longitudinal}, convex quadratic programming for planning lane change maneuvers is utilized.	
	Therein, the onedimensional longitudinal motion is stated as a convex quadratic
	optimization problem in the first step.  
	Based on the longitudinal motion, a further convex quadratic problem,  to compute
	the lateral movement, is solved.
	Convexity guarantees that there is only one optimum and thus, the optimal solution always corresponds to the global optimum. Furthermore, the quadratic nature of the problem formulations allows
	for fast convergence and small computation times.
	\newline
	Another approach based on convex quadratic optimization is the one in
	\cite{gutjahr2016lateral}.
	Thereby, a linearized model is used to describe the lateral vehicle dynamics. Subsequently, a time variant model predictive control problem is formulated to compute the optimal lateral movement.
	Convex quadratic optimization for planning fail safe trajectories in
	safety-critical situations is also used in \cite{PekChristian}.
	\newline
	For more information regarding on-road motion planning, the
	reader is referred to \cite{katrakazas2015real}.
	Nevertheless, it becomes obvious that much effort was spent already, many
	discretization schemes were applied and many continuous problem formulations stated. Thus,
	the question arises: How can the existing methods be combined to improve
	the driving performance while reducing the computation time?
	Subsequently, this question is also subject of current research work.	Accordingly, we'd like to describe two representative works:
	\newline
	A motion planning framework based on graph search and convex quadratic
	optimization is presented in \cite{zhan2017spatially}.
	In the first step, the algorithm of \cite{Hubmann2016} is applied to obtain a
	rough longitudinal motion prediction. This rough motion is then smoothed by
	solving a convex quadratic optimization problem.
	The lateral motion is computed via a graph search method as well and the final
	trajectory is smoothed by solving a convex quadratic problem to obtain the
	trajectory in two dimensions.
	\newline
	Another motion planning framework is the one described in \cite{meng2019decoupled}.
	Thereby, splines are used to create a path lattice in the first step. This path gets
	further optimized by solving a convex optimization problem.
	Subsequently, a speed profile on the computed path is planned by applying a
	lattice search using velocity primitives. The final trajectory gets smoothed by the solution of a convex quadratic optimization problem.
	\newline
	A crucial feature of our motion planning system is the deep integration of driver models.
    Therefore, we'd like to briefly summarize some state of the art motion planning techniques also utilizing driver models.
	Driver models, which are subject to the traffic flow science, are promising with
	regard to describing social interaction between vehicles. 
	In \cite{Evestedt2016}, the Intelligent Driver Model (IDM) is incorporated into  the
	cost functional of the  sampling based planner of \cite{werling2010optimal} to
	plan interaction aware trajectories in merging scenarios. 
	In \cite{graf1}, the IDM  was used together with the  quadratic cost function of
	\cite{zhan2017spatially} to compute human like trajectories and stopping maneuvers.	
	In \cite{Speidel2019}, a graph-based trajectory planning approach is presented
	using the IDM in order to estimate arising costs for other vehicles and to
	enable courteous behavior. \newline
	Based on the previous discussion, our motion planning framework is designed to combine 
	the flexibility and efficiency of continous optimization with driver models in order to enable social compliant behavior.
	\section{Methodology}
	\label{sec:Methodology}
	Our motion planning system is designed to enable level 4 automated driving according to the SAE\footnote{Society of Automotive Engineers.} 	\cite{sae2014taxonomy}.
	This means, the vehicle drives fully autonomously and does not expect the driver to intervene at any time.
	Therefore, the vehicle has to navigate through longitudinal traffic, keep safe
	distances to  preceding vehicles and stop if necessary.
	To react to congestions and to reduce travel time, the vehicle must be capable
	of performing lane changes and overtaking maneuvers. 
	To avoid handover requests, according functionalities to handle intersections
	and roundabouts must be implemented  as well.
	In safety critical situations, full braking or evasive maneuvers have to be planned.
	\newline
	In the following, all components of our system are described and discussed separately.
	Subsequently, we present our final system architecture and show how the
	components are put together to derive an effective overall framework for level 4 automated driving.
	\subsubsection{Central Optimization Problem}
\begin{figure}[!h]
		\begin{subfigure}[c]{0.35\linewidth}	
			\def\svgwidth{1\columnwidth} 	
			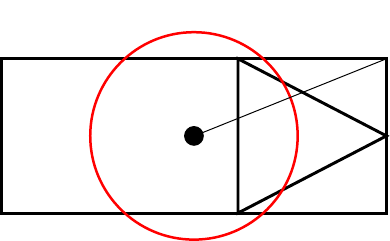	
			\caption{One circle is placed around the vehicles mass center.}
			\label{fig:onecircle}
		\end{subfigure}	
		\hspace{0.5cm}
		\begin{subfigure}[r]{0.55\linewidth}		
			\def\svgwidth{1\columnwidth} 	
			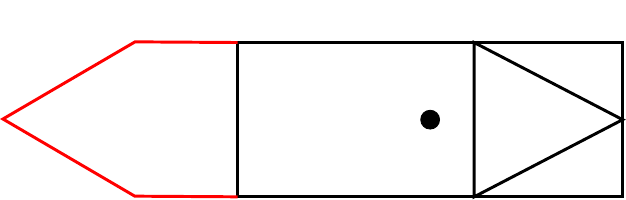	
			\caption{The preceding vehicle is virtually resized and a virtual triangle is appended at its rear.}
			\label{fig:virtresize}
		\end{subfigure}	
        \caption{One circle is placed around the shape of the ego vehicle (left figure). Since this circle does not cover the whole vehicle shape, the preceding vehicle is virtually resized (right figure). To "push" the ego vehicle laterally
		away, when it is reaching the preceding vehicle, a virtual triangle is appended at the preceding vehicles rear, as shown in
		\cite{wolf2008artificial}.}
	\end{figure}

	To exploit the advantages of continuous trajectory planning methods according to
	section \ref{sec:related_work}, we based our system on a corresponding continuous
	problem formulation.
	Precisely, we use the quadratic cost functional presented in
	\cite{zhan2017spatially}, which is essentially as follows
		\begin{equation}
	\begin{aligned}
	\min_{\boldsymbol \xi}      J&=\sum_i w_\behaviortext ||\mb x_i - \mb
	x_{\behaviortext,i}||_2^2+w_{a} ||\mb{\ddot x}_i||_2^2 \\ & \qquad + w_{\dot a} ||\mb{\dddot
		x}_i||_2^2 + w_{\ddot a} ||\mb{\ddddot x}_i||_2^2.
	\end{aligned}
	\label{eq:costfun}
	\end{equation}	
 The 2-dimensional positions $\mb x_i$ corrsepond to the trajectory support points of the vehicles gravity center at each time instant $t_i$ on the discretized planning horizon. These variables are to be optimized and are summarized in the vector of optimization variables $\boldsymbol \xi$.  The points  $\mb
		x_{\behaviortext,i}$ on the other hand have to be computed beforehand and remain constant during the optimization. Obviously, these points  strongly affect the pose of the global minimum of the functional (\ref{eq:costfun}) and, therefore, the computation result as well as the resulting driving behavior. This is why we will refer to the points $\mb
		x_{\behaviortext,i}$ as the "behavior~trajectory" throughout the remaining paper. In addition to \cite{zhan2017spatially}, we penalize the term  $||\mb{\ddddot x}_i||_2^2$ which can help to avoid steps within the derivative of the acceleration and thus contributes to smooth acceleration. Derivatives are approximated by using finite differences as described in
		\cite{trajectoryplanningforbertha} and \cite{zhan2017spatially}.
    For collision avoidance,  we use the strategy in \cite{trajectoryplanningforbertha}. Therein, obstacles are represented by polygons and a so called "pseudo distance" was designed to compute the distance to these polygons.
    Furthermore, circles are used to approximate the area covered by the vehicle along the trajectory.
    Finally, collision avoidance is achieved by requiring, through the constraints of an optimization problem, that the distances between the circle centers and the obstacle polygons are greater than or equal to the circle radii. 
In this work, our system slightly differs from \cite{trajectoryplanningforbertha}, since we use one circle per discrete time instant only which is placed around the vehicles mass center according to figure \ref{fig:onecircle}.
Since  this  circle  does  not  cover  the  whole  vehicle  shape, we  virtually  resize  the  preceding  vehicle by $r$,  according to figure  \ref{fig:virtresize}.  To obtain  the  same  ”push-away”  effect  as  in  \cite{trajectoryplanningforbertha},  we  append  a virtual triangle at the preceding vehicles rear as shown in \cite{wolf2008artificial}. Figure 4 illustrates how a rear-end collision can be avoided in this way. 
	\begin{figure}[!h]
		\def\svgwidth{1\columnwidth}
		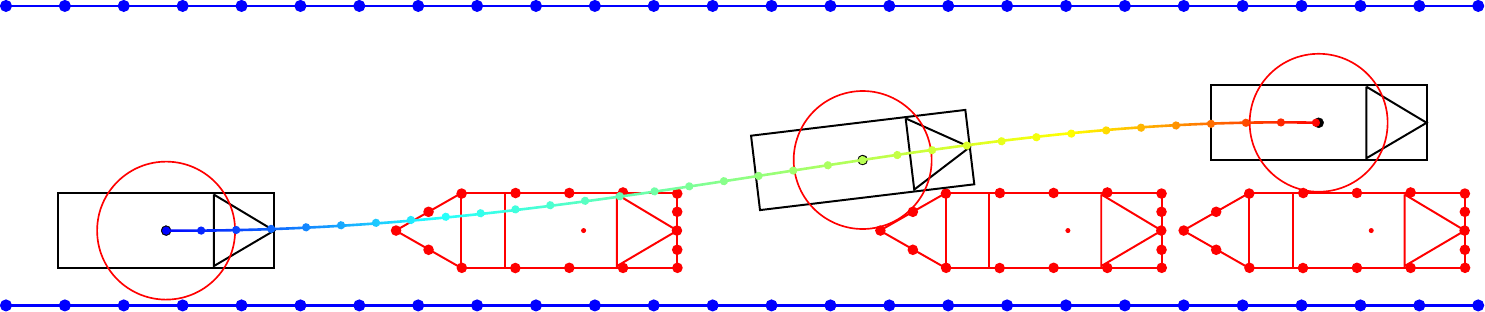
		\caption{Spatio temporal constraints for collision avoidance. At each time
			instant on the planning horizon, three scalar constraint functions are set up.
			One for the preceding vehicle, and two for the left and right free space
			boundary respectively.}
		\label{fig:spatialtempoconstraints}
	\end{figure}

		In addition to the constraints explained so far, further constraints are added to make sure, that the quadratic absolute acceleration stays within dedicated bounds \cite{trajectoryplanningforbertha}. Futher, spatio-temporal constraints, which guarantee safety in intersection scenarios (cmp. Section \ref{sec:long_driving}), are applied.
The constraints vector is then composed of five parts and can be written as follows:
	\begin{equation}
	         \boldsymbol h(\boldsymbol \xi)	=[\boldsymbol h_\text{fsbl}(\boldsymbol \xi)^\text{T},\boldsymbol h_\text{fsbr}(\boldsymbol \xi)^\text{T},\boldsymbol h_\text{dyn}(\boldsymbol \xi)^\text{T} \boldsymbol h_\text{st}(\boldsymbol \xi)^\text{T},\boldsymbol
	         h_{a}(\boldsymbol \xi)^\text{T}
	         ]^\text{T} \leq \boldsymbol 0.  
	         \label{eq:constraints}     
	\end{equation}
With the component due to the left free space boundary $\boldsymbol h_\text{fsbl}(\boldsymbol\xi)$, the right free space boundary $\boldsymbol h_\text{fsbr}(\boldsymbol\xi)$, dynamic obstacles $\boldsymbol h_\text{dyn}(\boldsymbol\xi)$, the spatio-temporal constraints $\boldsymbol h_\text{st}(\boldsymbol\xi)$ and the absolute acceleration $\boldsymbol h_a(\boldsymbol\xi)$ respectively.
	The constraints (\ref{eq:constraints}) together with the cost functional (\ref{eq:costfun}) are finally summarized to one overall optimization problem formulation.
	This problem  plays a central
	role within the architecture of our system and 
	we will refer to it as the ”central optimization
	problem” throughout the remaining paper.
	Each trajectory that gets forwarded to the vehicle controller corresponds to a
	solution of this problem. 
    What now remains to be clarified is the computation of the behavior trajectory $\mathbf x_\behaviortext$. In the following, we explain how to compute
	$\mathbf x_\behaviortext$ for different scenarios and maneuvers.
	
	\subsubsection{Longitudinal Traffic}
	\label{sec:long_driving}

    The longitudinal traffic mode (LTM) is the default driving mode and
    implements free driving and car following functionality. In addition, this
    mode is able to cope with intersections and roundabouts.
	\newline
    The behavior trajectory generation in the LTM mode is achieved through the
    incorporation of driver models. We'd like to briefly explain the
    corresponding model equations of the intelligent driver model (IDM):
	\begin{align}
	\dot{s} &= v \\
	\dot{v} &= a\left[1-\left(\frac{v}{v_\text{target}}\right)^\zeta - \left(\frac{s_\text{target}(v, \Delta v)}{\Delta s} \right)^2 \right]\,.
	\end{align}
    The longitudinal position on the ego lane is denoted by $s$ and the
    longitudinal velocity by $v$. The value $s_\text{target}$ corresponds to
    the desired distance to the vehicle in front and depends on the ego
    vehicle's velocity $v$ and the relative velocity $\Delta v$ between the
    vehicles. The current distance to the preceding vehicle is denoted by
    $\Delta s$. The parameter  $a$ is the maximum acceleration and $\zeta$ the
    acceleration exponent. Since the road curvature is not considered by the
    equations, the target velocity $v_\text{target}$ is extracted from a
    precomputed velocity profile (cmp.~\cite{Liebner2013}).
	\newline
    In our implementation we use an extension of the IDM, the "Enhanced
    Intelligent Driver Model" (EIDM), which builds on the equations above and
    provides improved braking behavior. Further explanations may be found in
    the related literature \cite{Treiber2000}, \cite{kesting2010enhanced}.
	\newline
	Accordingly, behavior trajectories $\mathbf x_\behaviortext$ utilized in following and free driving scenarios are directly generated using the EIDM.
	However, in intersection scenarios multiple maneuver options have to be regarded. Therefore, different behavior trajectory candidates are generated and subsequently, the behavior trajectory representing the best maneuver option is passed to the central optimization problem. 
	Especially in merging scenarios, it is further necessary to regard vehicles of different lanes, which is not viable by directly using car-following driver models. 
	Therefore, a virtual leading vehicle is modeled on the ego lane which represents a real vehicle on another lane in order to be able to generate smooth merging behavior by following the virtual leader.
	As a consequence, smooth merging as well as free driving, following and stopping behavior can be directly generated utilizing the EIDM in intersection scenarios.
	For further details regarding the generation of trajectory candidates and the modeling of the virtual leader, the interested reader is referred to one of our previous works \cite{speidel2020}.
	\newline
    To choose the best behavior trajectory $\mathbf x_\behaviortext$ in
    intersection scenarios, these are rated according to a cost function
    inspired by the MOBIL model \cite{Kesting2007}. The idea is to consider the
    ego costs and the costs of other vehicles arising through the ego behavior.
    The behavior trajectory which minimizes the overall costs for the ego
    vehicle and other vehicles is passed to the central optimization problem.
    As a consequence, the cost functional enables social compliant and
    courteous behavior. In general, the costs for a behavioral trajectory
    $\mathbf x_\behaviortext$ are defined as
	\begin{equation}
	J_\behaviortext = w_e j_e(\mathbf x_\behaviortext) + w_o \sum_{\iota\in \mathcal{O}} \sum_{k\in \mathcal{K}} P(\mathbf x^{\iota,k})
	j^{\iota,k}_{o}(\mathbf x_\behaviortext, \mathbf x^{\iota,k})\,,
	\label{eqn:lonbehcost}
	\end{equation}
    where $w_e$ and $w_o$ represent cost weights. In order to estimate $j_e$
    the negated average ego acceleration of the regarded behavior trajectory is
    utilized. Costs for other vehicles are represented by $j^{\iota,k}_{o}$.
    Thereby, for each vehicle $\iota \in \mathcal{O}$, multiple trajectory
    hypotheses $\mathbf{x}^{\iota,k}$ and their according probabilities $
    P(\mathbf{x}^{\iota,k})$ are considered. These hypotheses account for
    different behavior options $k \in \mathcal{K}$ of other vehicles in
    intersection scenarios.  To determine $j^{\iota,k}_{o}$ the vehicle $\iota$
    using trajectory hypothesis $k$ is regarded. The costs are derived from the
    acceleration deviation due to the estimated reaction on the ego vehicles
    trajectory. As a result, strong reactions to the trajectory of the ego vehicle are penalized encouraging courteous behavior towards other traffic participants. The reactions are modeled using the IDM, which enables a computationally efficient cost evaluation. 
    For further details the reader
    is referred to \cite{speidel2020}. % \textcolor{red}{Further explanations.}
	\newline
    In general, the optimization of longitudinal driving is subject to
    acceleration constraints as well as free space boundaries and collision
    constraints.			
	In addition, to guarantee safety in intersection scenarios and to make sure the optimized trajectory
		will stay in the vicinity of the behavior trajectory, spatio-temporal constraints according to   expression (\ref{eq:constraints}) are employed.
		These are derived based on the intersection geometry and the predictions of the vehicles, which are expected to occupy the intersection before and after the ego vehicle. In general, maximum and minimum distances are determined, which have to be traveled in a given time in order to enter and leave the intersection on time. Consequently, safety distances to the other vehicles are ensured while the deviation of the optimized trajectory to the behavior trajectory is limited. 
	\subsubsection{Lane Changing}
	\label{sec:lane_change}	
\begin{figure}
		\centering
		\begin{tikzpicture}
				\definecolor{compcolor}{RGB}{224,224,224};
			\tikzstyle{tyre} = [fill=compcolor,rectangle,rounded corners,draw,minimum width=1cm,minimum height=0.25cm];
		\def\steerangle{30};
			\def\massx{0cm};
			\def\massy{0cm};
			\def\fwx{2cm};
			\def\rwx{-2cm};
			\def\fwy{0cm};
			\def\rwy{0cm};
			\def\rponex{-3cm};
			\def\rponey{1cm};
			\def\rptwox{3cm};
			\def\rptwoy{1cm};
			\def\lapx{2.5cm};
			\def\lapy{1cm};	
			\def\arrowsoff{1cm};
			\begin{scope}
			\draw[green](\rponex,\rponey)--(\rptwox,\rptwoy);
			\node[tyre,rotate=\steerangle](fw) at (\fwx,\fwy) {};
			\node[tyre](rw) at (\rwx,\rwy) {};
			\node[circle,fill,inner sep=1pt](mass) at (\massx,\massy) {};
			\node[circle,fill,inner sep=1pt](lap) at (\lapx,\lapy) {};
			\node[circle,fill,inner sep=1pt] at (fw.center) {};
			\node[circle,fill,inner sep=1pt] at (rw.center) {};			
			\draw (fw.center)--(rw.center);
			\draw[dashed] (fw.center)--({\fwx+1cm},\fwy);
			\draw[dashed,rotate around={\steerangle:(fw)}] ({\fwx-1cm},\fwy)--({\fwx+1cm},\fwy);
			
			\draw[latex-latex'] (\rwx,{\rwy-\arrowsoff})--(\massx,{\massy-\arrowsoff});
			\draw[latex-latex'] (\fwx,{\fwy-\arrowsoff})--(\massx,{\massy-\arrowsoff});
			\node[yshift=-0.3cm] at ({\fwx/2+\massx/2},{\fwy-\arrowsoff}){$l_f$};
			\node[yshift=-0.3cm] at ({\rwx/2+\massx/2},{\rwy-\arrowsoff}){$l_r$};
			\draw[dashed] (\rwx,{\rwy-\arrowsoff})--(\rwx,\rwy);
			\draw[dashed] (\massx,{\rwy-\arrowsoff})--(\massx,\rwy);
			\draw[dashed] (\fwx,{\rwy-\arrowsoff})--(\fwx,\rwy);			
			\draw[latex-latex'] (mass)--(lap) node[midway,yshift=0.15cm,xshift=-0.15cm]{\footnotesize$L$};
			\draw ({\fwx+0.8cm},\fwy) arc (0:30:0.8cm) node[xshift=0.25cm,yshift=-0.15cm]{\footnotesize {$\delta$}};
			\draw[dashed,rotate around={-90:(\massx,\massy)}] (mass) arc (0:43:3.7);
			\end{scope}
		\end{tikzpicture}
		\caption{Functionality of the pure pursuit steering controller. The steering angle is computed such, that the mass center moves towards the look ahead point on a circular shaped trajectory.}
		\label{fig:purepursuit}
	\end{figure}
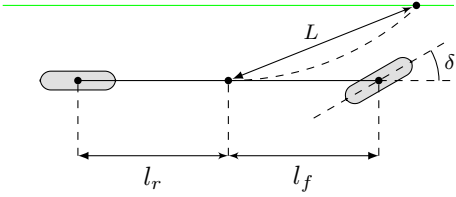

	In our previous work \cite{graf2019model}, we implemented lane change functionality by  numerically integrating the IDM on the  reference line to obtain the behavior trajectory. Subsequent minimization of the  cost functional (\ref{eq:costfun})  results in a transition to the target lane. To improve lateral guidance, we extend this procedure by additionally using a kinematic singletrack model.
	The model equations are given by:
	\begin{equation}
		\begin{aligned}
            \dot{x}&=v\cos{(\theta)}\\
            \dot{y}&=v\sin{(\theta)}\\
		\dot\theta &=\frac{v}{l_r+l_f} \underbrace{\delta }_{u_1}\\
		\dot v&= \underbrace{a }_{u_2}.
		\end{aligned}
	\end{equation}
	 For determining the steering angle $u_1=\delta$, we use a purepursuit steering controller \cite{kuwata2008motion}.
	For the sake of completeness, we'd like to briefly explain the control principle by the help of figure \ref{fig:purepursuit}.
	First, the "look ahead position" on the reference line must be computed. This corresponds to the one position on the reference line, whose distance to the mass center corresponds to the "look ahead distance" $L$, which has to be chosen in dependence of the velocity. Subsequently, the steering angle $\delta$ is chosen such, that the mass center moves towards the look ahead point on a circular arc. Obviously, smooth lateral guidance and low lateral accelerations can be obtained by increasing the look ahead distance.
	For more information and further details the reader is referred to \cite{kuwata2008motion}.
	The acceleration $u_2=a$ is computed by numerically integrating  the EIDM equations on  the reference line of  the target lane.
	\subsubsection{Emergency}
	\label{sec:emergency}
	The emergency mode is used in safety critical situations to compute full braking and evasive trajectories. Since the EIDM was not designed for safety critical situations, we use an optimization based approach to compute the behavior trajectory in such situations.
	To guarantee real-time capability,  two convex quadratic optimization problems for the longitudinal and lateral movement  are formulated and sequentially solved. 
	The problem formulations are very similar to the ones in \cite{qian2019synchronous}.
	As this work focuses on non-safety critical  scenarios,  further details are omitted.
	\subsubsection{Decision Making}
	\label{sec:decision_making}

\begin{figure}
		\centering
		\begin{tikzpicture}	
		\tikzstyle{decmaking} = [fill=compcolor,draw,rectangle,rounded corners,minimum
		width=4cm,minimum height=2cm];
		\tikzstyle{behgenerator} = [draw,rectangle,rounded corners,dashed,minimum
		width=1.5cm,minimum height=1cm];
		\tikzstyle{state} = [draw,rectangle,rounded corners,minimum width=2cm,minimum
		height=2cm];
		\tikzstyle{decmaking} = [draw,rectangle,rounded corners,minimum
		width=5cm,minimum height=2.5cm];
		
		\tikzstyle{decmakingcomps} = [draw,rectangle,rounded corners,dashed,minimum
		width=4cm,minimum height=0.5cm];

		\newcommand{\comp}[5]{
			\node[fill=compcolor,state,align=left] (#3) at (#1,#2) {};
			\node [anchor=north ,inner sep=0.2cm] at (#3.north){{#3}};	
			\node[fill=white,behgenerator,anchor=south,shift={(0,0.2)}] (innerbox) at
			(#3.south){#4};}
		\newcommand{\compdec}[6]{
			\node[fill=compcolor,decmaking,align=left] (#3) at (#1,#2) {};
			\node [anchor=north ,inner sep=0.2cm] at (#3.north){{#3}};	
			\node[fill=white,decmakingcomps,anchor=south,shift={(0,1)}] (innerboxone) at
			(#3.south){#4};
			\node[fill=white,decmakingcomps,anchor=south,shift={(0,0.2)}] (innerboxtwo) at
			(#3.south){#5};
			\path[-latex',draw](innerboxone)--(innerboxtwo);
		}
		\definecolor{compcolor}{RGB}{224,224,224};
		
		\tikzstyle{centraloptprob} = [rectangle,rounded corners,draw,fill=compcolor]

		\compdec{0}{0}{Decision Making}{Situation assessment}{State machine};

		\comp{-3cm}{-2.5cm}{Lane Change}{\footnotesize{Section
				\ref{sec:lane_change}}}{};
		\comp{0cm}{-2.5cm}{LTM}{\footnotesize{Section
				\ref{sec:long_driving}}}{};	
		\comp{3cm}{-2.5cm}{Emergency}{\footnotesize{Section \ref{sec:emergency}}}{};

		\node[fit=(LTM)(Emergency)(Lane Change)(Decision
		Making),dashed,draw,rounded corners,inner sep=0.5cm](behavior){};
		
		\node[centraloptprob,yshift=-2cm] at(behavior.south) (copt){
			$\begin{aligned}			
			\min_{\boldsymbol \xi} \ \ &J(\boldsymbol\xi)\\
			\text{s.t.}  \ \ &\boldsymbol h(\boldsymbol \xi)\leq \boldsymbol 0
			\end{aligned}$	};

		\node [anchor=south west,inner sep=0.0cm,yshift=-1cm,xshift=-1cm]  at
		(copt.south west)(optprobdescone){\rotatebox[]{90}{
				\begin{tabular}{l}
				\footnotesize{Central optimization} \\\footnotesize{ problem}\\ \phantom{a}  
				\end{tabular}	
		}};
		
		\node [anchor=south east,inner sep=0.0cm,yshift=-1cm,xshift=1cm]  at
		(copt.south east)(optprobdesctwophantom){\rotatebox[]{90}{\phantom{
					\begin{tabular}{l}
					\footnotesize{Central optimization} \\\footnotesize{ problem}\\ \phantom{a}  
					\end{tabular}	}
		}};
		
		\node[fit=(optprobdescone)(optprobdesctwophantom)(copt),dashed,draw,rounded
		corners,inner sep=0.0cm](coptframe){};
		\path (behavior) edge[-latex'] (coptframe);
		
		\node[rectangle,rounded corners,fill=compcolor,draw,yshift=-0.7cm]
		at(coptframe.south)(vehiclecontroller){Vehicle Controller};
		
		\path (coptframe) edge[-latex'] (vehiclecontroller);

		\node[rectangle,rounded corners,fill=compcolor,draw,yshift=0.7cm]
		at(behavior.north)(envdata){Environmental Data};
		\path (envdata) edge[-latex'] (behavior);

		\path (Decision Making) edge[-latex'] (LTM);
		\path (Decision Making) edge[-latex'] (Lane Change);
		\path (Decision Making) edge[-latex'] (Emergency);

		\end{tikzpicture}
		\caption{Motion planning architecture.}
		\label{fig:mparchitecture}
	\end{figure}
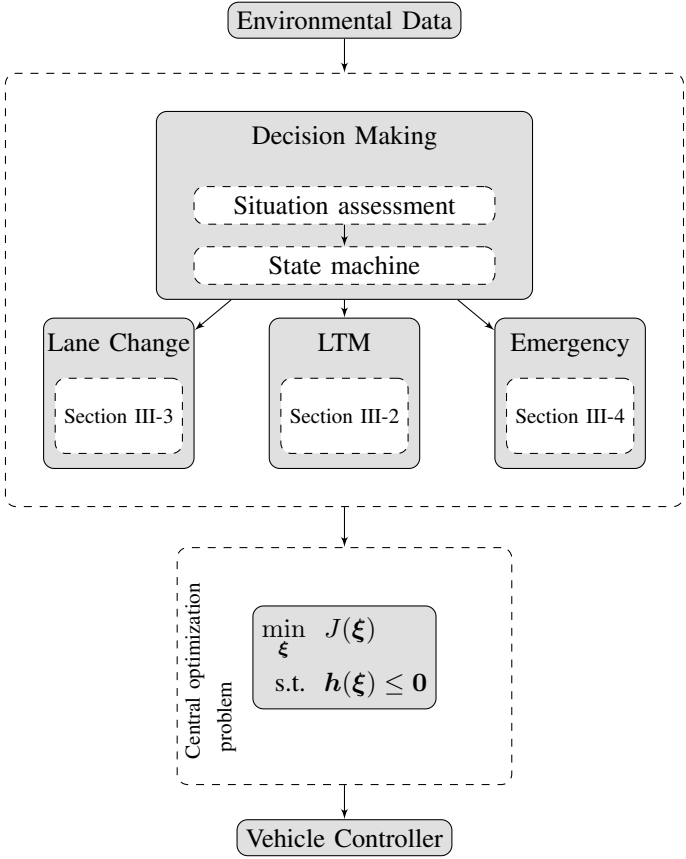

    To beneficially connect the described operational modes and to obtain an
    overall motion planning system, we implemented a decision making module
    which can be  divided into a situation assessment and a finite state
    machine. The main task of the situation assessment consists in determining
    boolean state transition variables for the state machine. Therefore, the
    model "MOBIL", which is based on the IDM, is used to decide whether lane
    changes are desirable \cite{Kesting2007}.
    By this model, the current traffic situation is compared with the hypothetical situation that the ego vehicle is located on a candidate lane. In both situations, the current and the hypothetical one, the accelerations of all relevant surrounding entities as well as the one of the ego vehicle itself is computed using the EIDM/IDM. If in the hypothetical situation the overall acceleration is increased compared  to the current situation, a lane change is assessed as incentive.
     To obtain high consistency
    between decision making and the computation of the behavior trajectory, we
    use the EIDM within MOBIL. However, lane changes may only be performed, if
    there are no roundabouts, intersections or road sections of high curvature
    in reach. To identify safety critical situations and to determine whether
    the emergency mode has to be activated, the distance and the relative velocity between the ego vehicle and its leader is checked. If braking with a constant   acceleration threshold is not sufficient to avoid a rear-end collision, the situation is assessed as critical and the emergency functionality gets active. As mentioned already, the longitudinal traffic mode is the default
    operational mode and always active, provided no lane change or emergency
    maneuver needs to be performed.
    The overall structure of the motion planning system is also depicted in
    Figure \ref{fig:mparchitecture}. 

	\section{Experiments}

    Experiments are carried out on our experimental vehicle which is shown in
    figure \ref{fig:susi}. The luggage space contains two usual consumer
    computers and a MicroAutoBox by the company dSpace, which serves as a
    real-time capable control system. The algorithmic processing chain is
    implemented on the consumer PCs using ROS\footnote{Robot Operating System
    \cite{ros}.}, which is already proven to work as a middleware for autonomous driving (see~e.g.~\cite{ROSHellmund}). Figure \ref{fig:vehiclebuilt} illustrates the hardware
    arrangement and shows which software components are implemented on the
    devices. The sensor setup consists of LiDARs, RaDARs, Cameras and an
    ADMA\footnote{Automotive Data Motion Analyzer by the company GeneSys.}.
    The ADMA contains a DGPS system and allows the determination of the
    vehicle position accurately with errors only up to a few centimeters.
    Sensor data is first passed to the perception PC and object information is
    extracted by the object detection modules \cite{Danzer2019}. This information is then passed
    to the tracking algorithm where the detections are fused and filtered using a Labeled Multi-Bernoulli Filter to
    estimate the states of all surrounding dynamic obstacles \cite{Reuter2014}. The ego motion
    module fuses data from the ADMA together with information from other
    sensors (e.g. odometry) to determine the ego vehicles state as accurately
    as possible. This information is passed to the motion planner which is running on
    the application PC. Information of the static environment is retrieved from
    an HD digital map stored on the application PC. The controller on the
    MicroAutoBox requires a trajectory represented by $N=60$ support points
    over a temporal horizon of $T=\SI{2.95}{\second}$
    (cmp.~\cite{autonomousdrivingatulmuniversity}). The target state is then
    extracted from the trajectory and used together with the ego state for the
    computation of controls by the trajectory controller.
	\begin{figure}
		\begin{tikzpicture}[node distance=0.5cm]
		\def\dstbelow{0.1cm};
			\tikzstyle{maincomp} = [draw,rectangle,rounded corners];	
			\tikzstyle{subcomp} = [draw,rectangle,rounded corners,dashed,fill=white,minimum width=2cm];	
		 \tikzstyle{subcomppc} = [draw,rectangle,rounded corners,dashed,fill=white,minimum width=2.5cm,yshift=-0.5cm,minimum height=0.9cm];			
			\definecolor{compcolor}{RGB}{224,224,224};	
			\newcommand{\sensors}[7]{
				\node[align=left] (#3) at (#1,#2) {#3};	
				\node[align=left,subcomp,below of=#3] (#4) at (#1,#2) {#4};
				\node[align=left,subcomp,right of=#4,anchor=west,xshift=0.6cm] (#5) {#5};
				\node[align=left,subcomp,right of=#5,anchor=west,xshift=0.6cm] (#6) {#6};
				\node[align=left,subcomp,right of=#6,anchor=west,xshift=0.6cm] (#7) {#7};
				
				\begin{pgfonlayer}{background}			
				\node[maincomp,fit=(#3)(#4)(#5)(#6)(#7),fill=compcolor](sensorsfit){};
				\end{pgfonlayer}
			}
		\def\xoff{1cm};
		\def\xofftwo{5cm};
				\def\heightsmall{1cm};
				\def\heightlarge{2.3cm};
				
		\node[](PercPC)at(\xoff,-1.2){Perception PC};
		\node[](ApplPC)at(\xofftwo,-1.2){Application PC};
				
				\sensors{0}{0.8}{Sensors}{LiDAR}{RaDAR}{Camera}{ADMA};				
		\node[subcomp,align=left,minimum width=3cm,minimum height=\heightsmall](egomotion)at(\xoff,-5){Ego motion};				
		\node[subcomp,align=left,minimum width=3cm,minimum height=\heightsmall](vehicletracking)at(\xoff,-3.5){Vehicle tracking};
		\node[subcomp,align=left,minimum width=3cm,minimum height=\heightsmall](sensorpreprocessing)at(\xoff,-2){Preprocessing and \\object detection};
		\node[subcomp,align=left,minimum width=3cm,,minimum height=\heightlarge](motionplanning)at(\xofftwo,-4.3){Motion planning};				
		\node[subcomp,align=left,minimum width=3cm,minimum height=\heightsmall](dm)at(\xofftwo,-2){Digital map};	
				\draw[-latex'] (egomotion)--(motionplanning.west|-egomotion);
								\draw[-latex'] (vehicletracking)--(vehicletracking-|motionplanning.west);
								\draw[-latex'] (sensorpreprocessing)--(vehicletracking);
							\draw[-latex'] (dm)--(motionplanning);
		\node[rectangle,dashed,rounded corners,draw,fit=(PercPC)(egomotion)(vehicletracking)(sensorpreprocessing)]{};
		\node[rectangle,rounded corners,draw,dashed,fit=(ApplPC)(dm)(motionplanning)]{};
		
			\begin{pgfonlayer}{background}			
		\node[inner sep=10pt,fill=compcolor,maincomp,fit=(PercPC)(ApplPC)(egomotion)(vehicletracking)(sensorpreprocessing)
		(motionplanning)(dm)](wrapper){};	
		\end{pgfonlayer}

			\node[align=left] (MAB) at (0,-6.5cm) {MicroAutoBox};	
			\node[align=left,subcomp,below of=MAB,yshift=-0.4cm] (sysfun) at (0,-6.5cm) {
			\begin{tabular}{l}
						System\\
						monitoring
						\end{tabular}
		};
			\node[align=left,subcomp,right of=sysfun,anchor=west,xshift=0.8cm] (sysmon) {
				\begin{tabular}{l}
			Safety\\
			functions
			\end{tabular}
		};
			\node[align=left,subcomp,right of=sysmon,anchor=west,xshift=0.8cm] (control) {
			\begin{tabular}{l}
		Trajectory\\
		controller
		\end{tabular}	
		};
			\node[align=left,subcomp,right of=control,anchor=west,xshift=0.8cm,fill=compcolor,draw=compcolor,minimum width=1cm ] (test) {$\dots$};
			\begin{pgfonlayer}{background}			
			\node[maincomp,fit=(MAB)(sysfun)(sysmon)(control)(test),fill=compcolor](mabfit){};
			\end{pgfonlayer}
			\draw[-latex'](wrapper)--(wrapper|-mabfit.north);
				\draw[-latex'](sensorsfit)--(wrapper);
		\end{tikzpicture}
		\caption{Simplified architecture of the vehicles algorithmic processing chain.}
		\label{fig:vehiclebuilt}
	\end{figure}
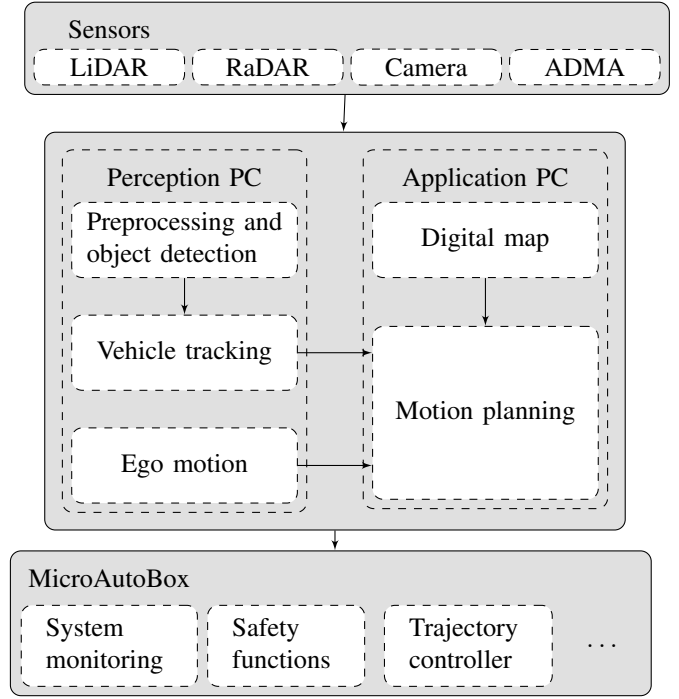
	\begin{figure}[t!]
	\begin{tabular}[r]{r}
		\input{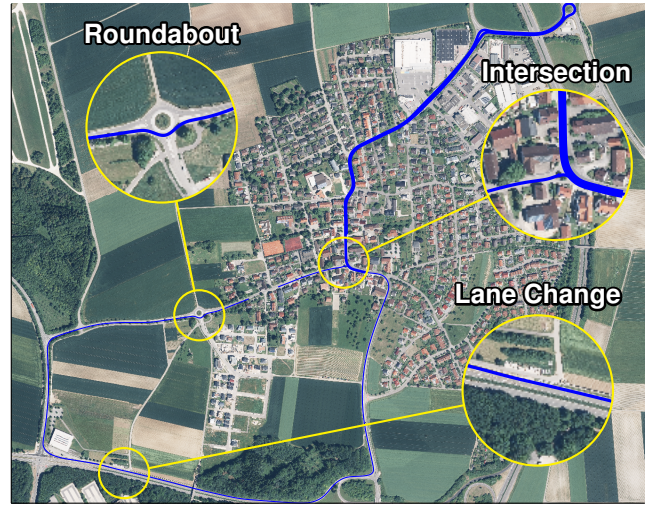} 
	\end{tabular}	
	\caption{Bird's-eye-view of the test track through the district "Lehr" in Ulm, where the positions of the evaluation scenarios are highlighted.}
	\label{fig:lehr}
    \end{figure}
	To test our algorithms, we have chosen a test track close to the university which leads through the district of Ulm called "Lehr".
	The test track is $\approx\SI{4}{\kilo\metre}$ long and contains an intersection, an acceleration strip, a pedestrian crossway, a roundabout, zones of various speed limits, and thus provides a challenging   opportunity to test our algorithms in public road  traffic. 
	Figure \ref{fig:lehr} shows a bird's-eye-view of the test track. The blue line represents the vehicles path which was recorded during an automated drive. 
 	Even though  our vehicle is capable of driving long routes autonomously, it does not reveal much about the driving behavior and the capability of the vehicle  to behave socially.
 	Therefore,  we show some representative scenarios in more detail below. 
 	\subsection{Lane Change}
\begin{figure*}
 		\setlength\columnsep{5pt}
 		\begin{multicols}{3}
 			\begin{tikzpicture}
 			\draw (0, 0) node[inner sep=0] {\includegraphics[clip,width=\linewidth]{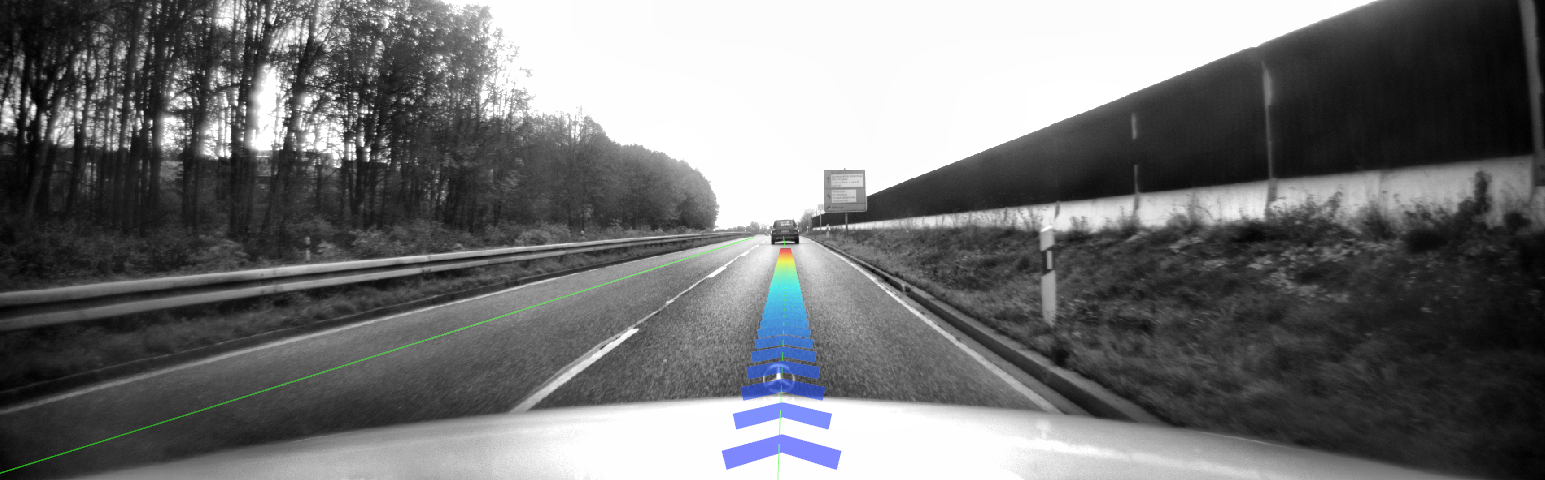}};
 			\draw (-2.7, 0.65) node[circle,fill=white,draw=black,thick,inner sep=1pt]{1};
 			\end{tikzpicture}\,\columnbreak
 			\begin{tikzpicture}
 			\draw (0, 0) node[inner sep=0] {\includegraphics[clip,width=\linewidth]{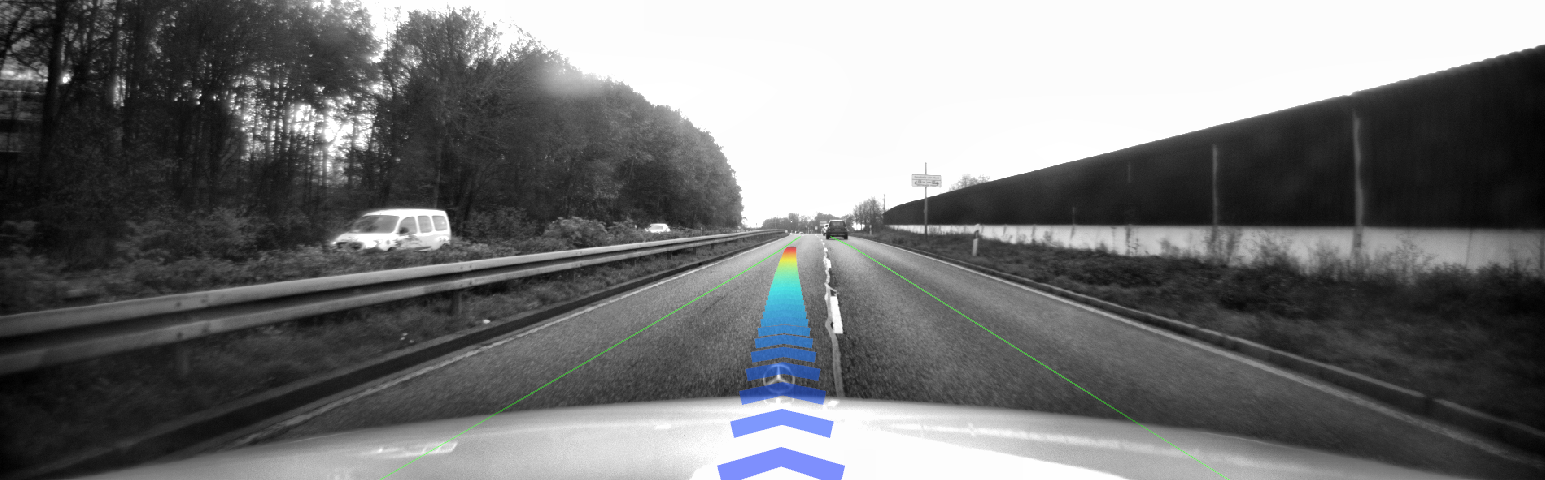}};
 			\draw (-2.7, 0.65) node[circle,fill=white,draw=black,thick,inner sep=1pt]{2};
 			\end{tikzpicture}\,\columnbreak
 			\begin{tikzpicture}
 			\draw (0, 0) node[inner sep=0] {\includegraphics[clip,width=\linewidth]{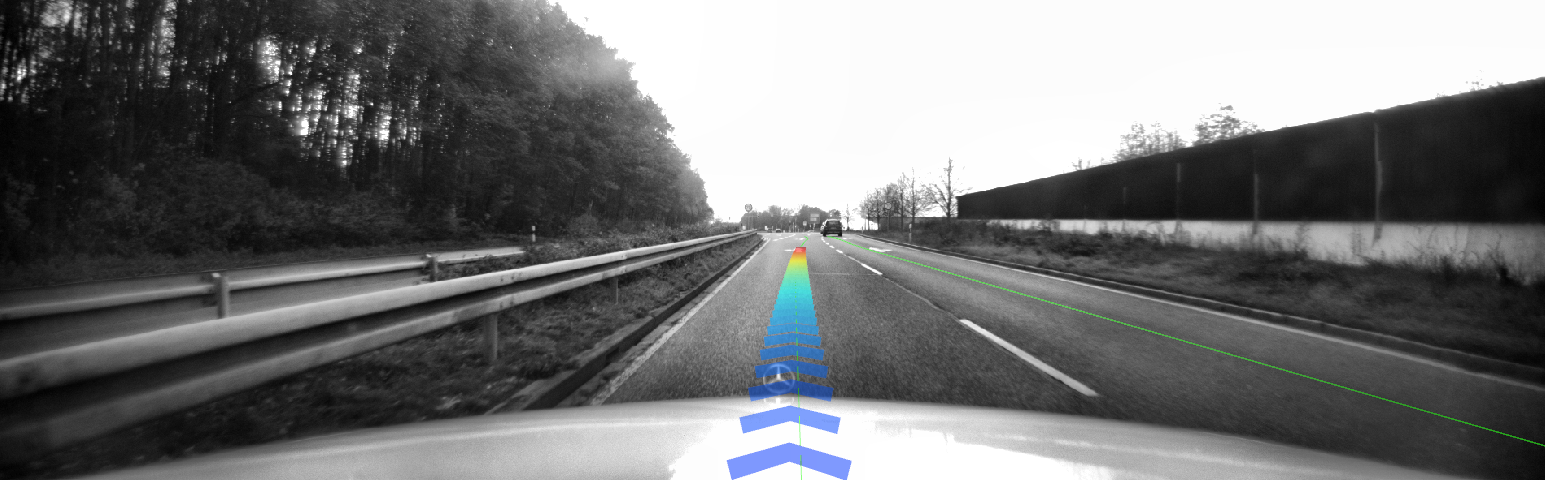}};
 			\draw (-2.7, 0.65) node[circle,fill=white,draw=black,thick,inner sep=1pt]{3};
 			\end{tikzpicture}
 		\end{multicols}
 		\vspace{-0.65cm}
 		\begin{multicols}{3}
 			\begin{tikzpicture}
 			\draw (0, 0) node[inner sep=0] {\includegraphics[trim={0cm 7cm 0cm 6cm}, clip,width=\linewidth]{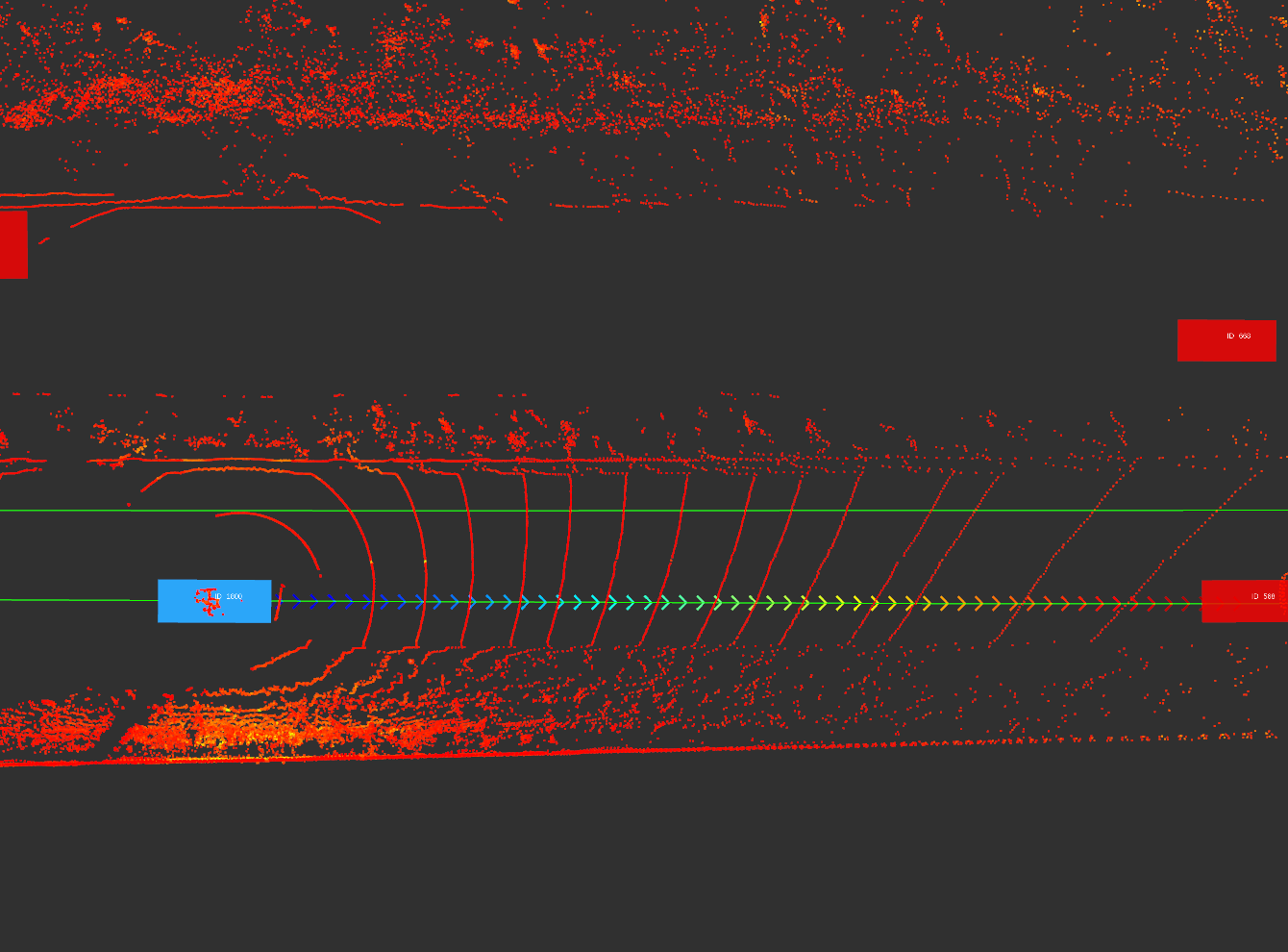}};
 			\draw (-2.7, 1.1) node[circle,fill=white,draw=black,thick,inner sep=1pt]{1};
 			\end{tikzpicture}\,\columnbreak
 			\begin{tikzpicture}
            \draw (0, 0) node[inner sep=0] {\includegraphics[trim={0cm 7cm 0cm 6cm}, clip,width=\linewidth]{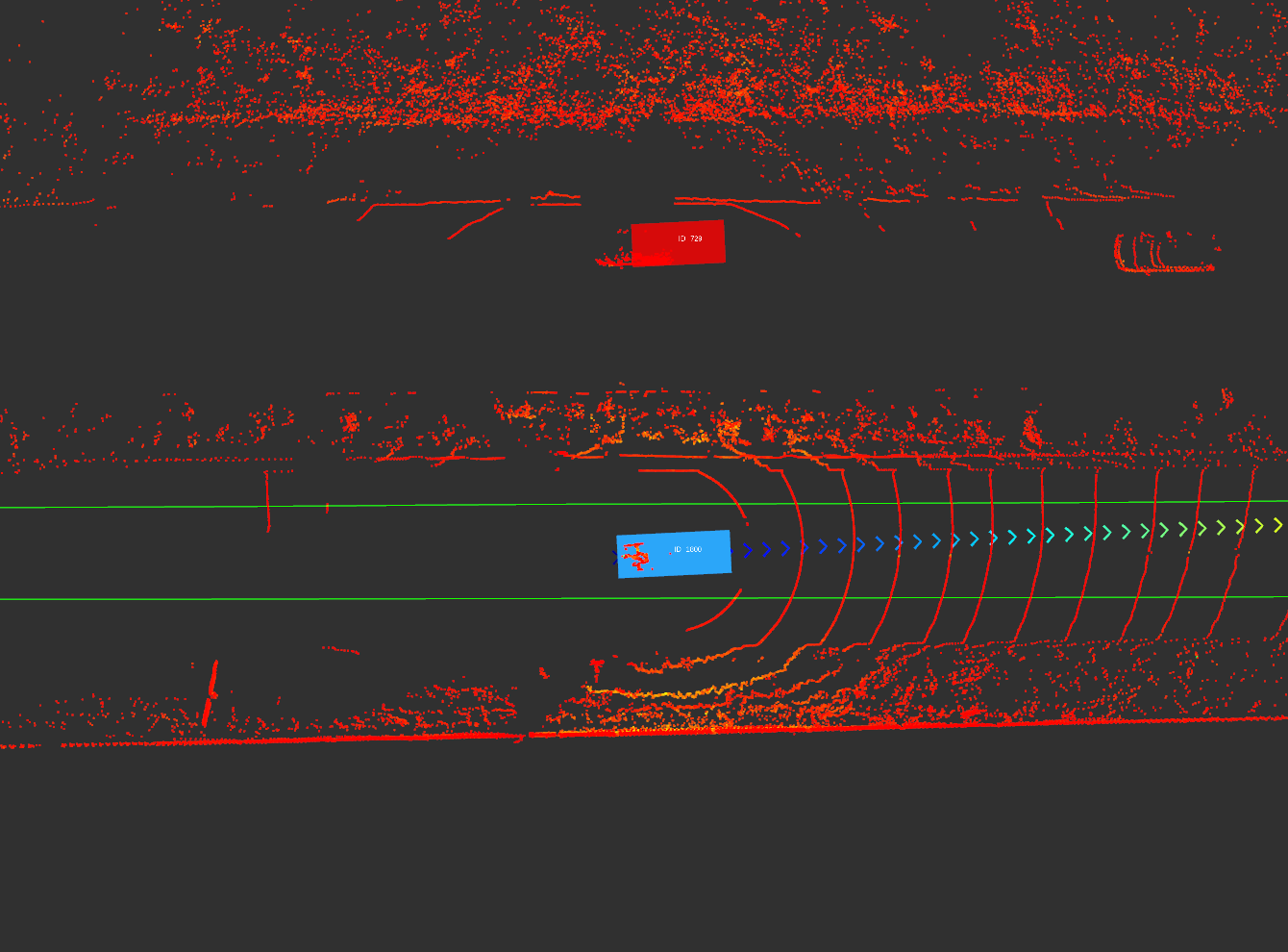}};
 			\draw (-2.7, 1.1) node[circle,fill=white,draw=black,thick,inner sep=1pt]{2};
 			\end{tikzpicture}\,\columnbreak
 			\begin{tikzpicture}
            \draw (0, 0) node[inner sep=0] {\includegraphics[trim={0cm 7cm 0cm 6cm}, clip,width=\linewidth]{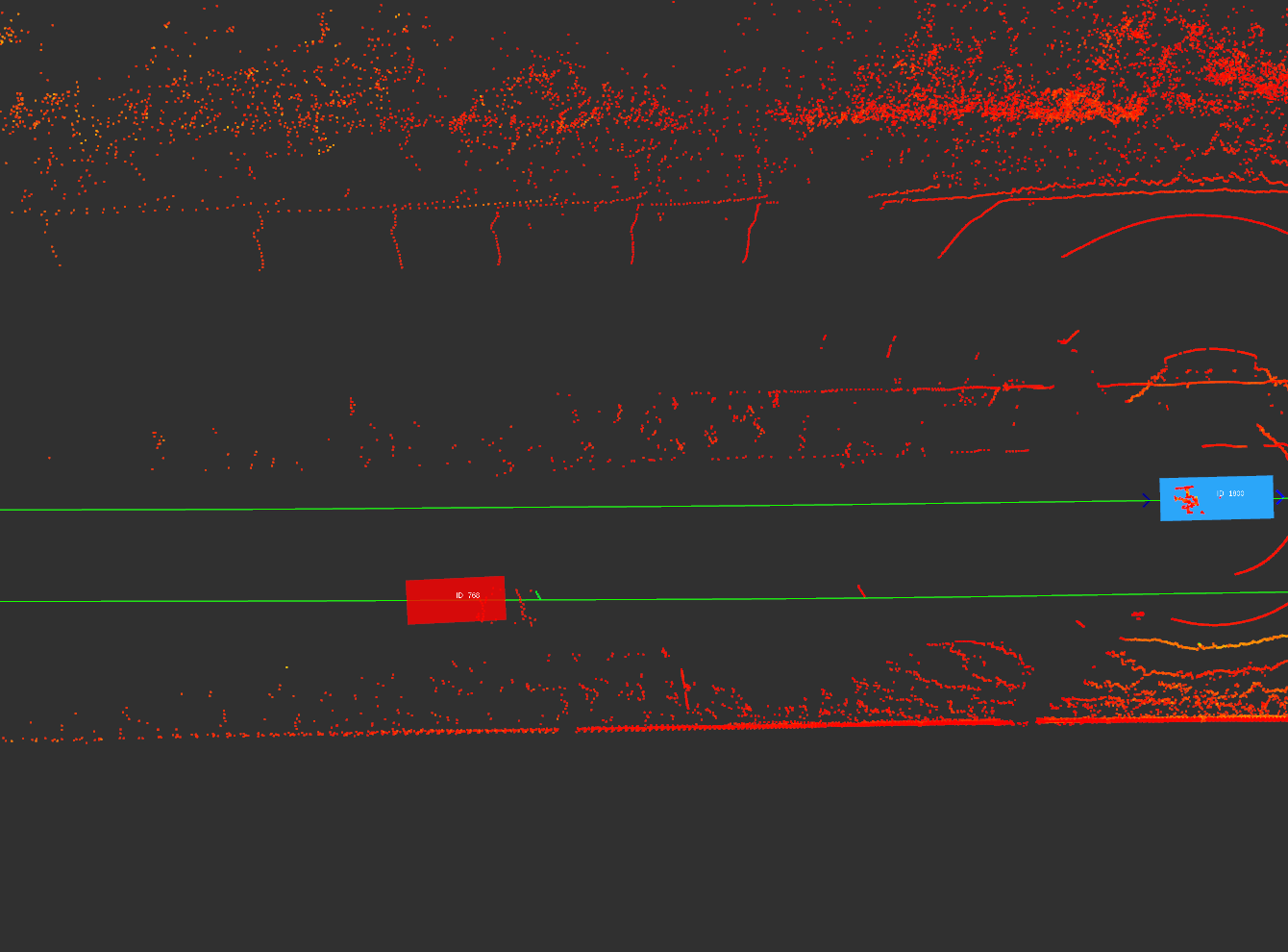}};
 			\draw (-2.7, 1.1) node[circle,fill=white,draw=black,thick,inner sep=1pt]{3};
 			\end{tikzpicture}
 		\end{multicols}
 		\begin{multicols}{3}
 			\input{Images/Tikz/fig9-3-1.tikz}\;\;\;\;\;\;\,\columnbreak
 			\input{Images/Tikz/fig9-3-2.tikz}\;\;\;\;\;\;\,\columnbreak
 			\input{Images/Tikz/fig9-3-3.tikz}
 		\end{multicols}
 		\caption{
            Lane Change scenario for three dedicated time instants:
            The first row shows the drivers perspective by
            camera images. The second row shows recorded as well as processed
            data in top view. The third row depicts measured values during the
            maneuver, where the velocity, the longitudinal (blue) together with
            the lateral (red) acceleration, and the steering wheel angle are shown over time.
            Note that the peak in the lateral acceleration at $t \approx\SI{3.8}{s}$ is due to measurement noise.            
        }
 			\label{fig:lanechange}
 	\end{figure*}
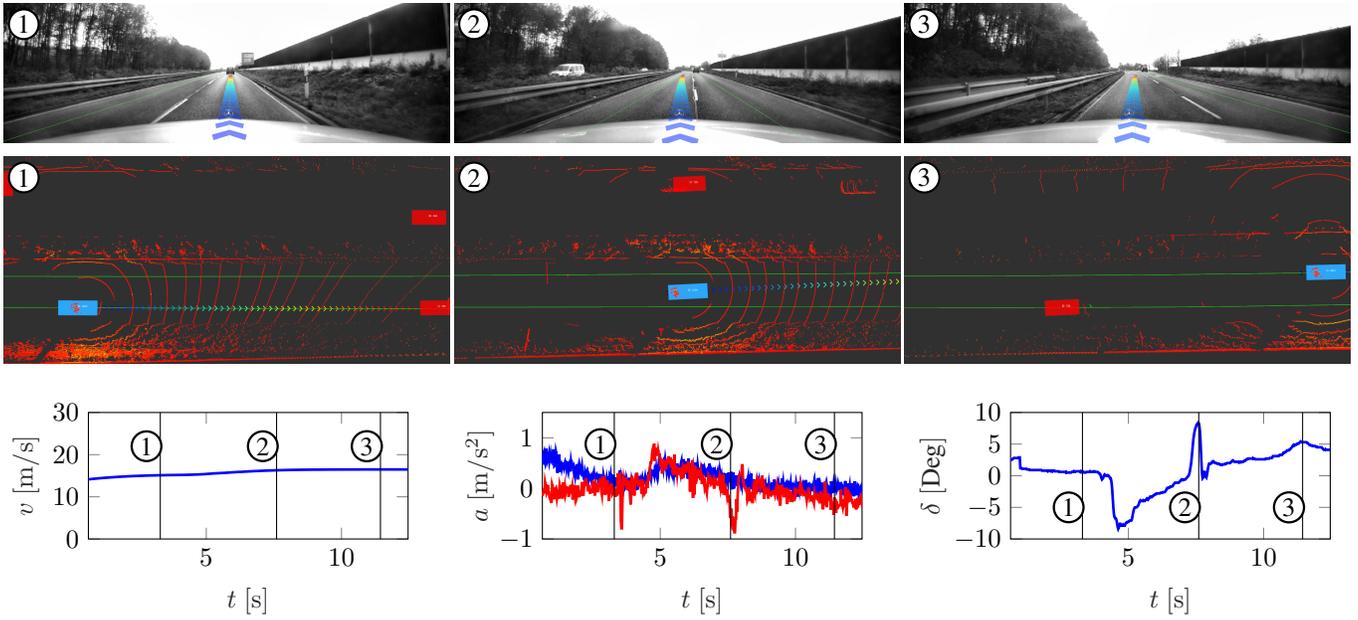

 	The first demonstration scenario corresponds to a lane change maneuver. The recorded data is shown in figure \ref{fig:lanechange}.
 	The first row shows the drivers perspective by camera images for three dedicated time instants and the second row shows the according ROS-RViz display and thus, corresponds to what the vehicle perceives. In order to relate these two perspectives to each other, the trajectory itself is mapped from the RViz-display into the camera images. Beside the trajectory, the center line for each lane is visualized and the blue boxes correspond to the ego vehicle, whereas red boxes represent other vehicles in the scene. Thereby, the trajectory originates in the gravity center of the ego vehicle. The third row corresponds to the measured velocity, acceleration and steering angle during the maneuver.
 	\newline
 	The three images show a transition from the right to the left lane. Whereby the lane change gets triggered shortly after \tikz\node[circle,fill=white,draw=black,thick,inner sep=1pt]{\footnotesize 1};  and executed at \tikz\node[circle,fill=white,draw=black,thick,inner sep=1pt]{\footnotesize 2};.   
 	At \tikz\node[circle,fill=white,draw=black,thick,inner sep=1pt]{\footnotesize 3};, the vehicle has already arrived at the target lane. Even though the reader has not the possibility to take a test drive himself, a view on the measured values shows that the acceleration as well as the steering wheel angle remain small and don't change rapidly during the scenario. This indicates comfortable driving behavior, which was experienced when the data was recorded.
	\subsection{Roundabout}
	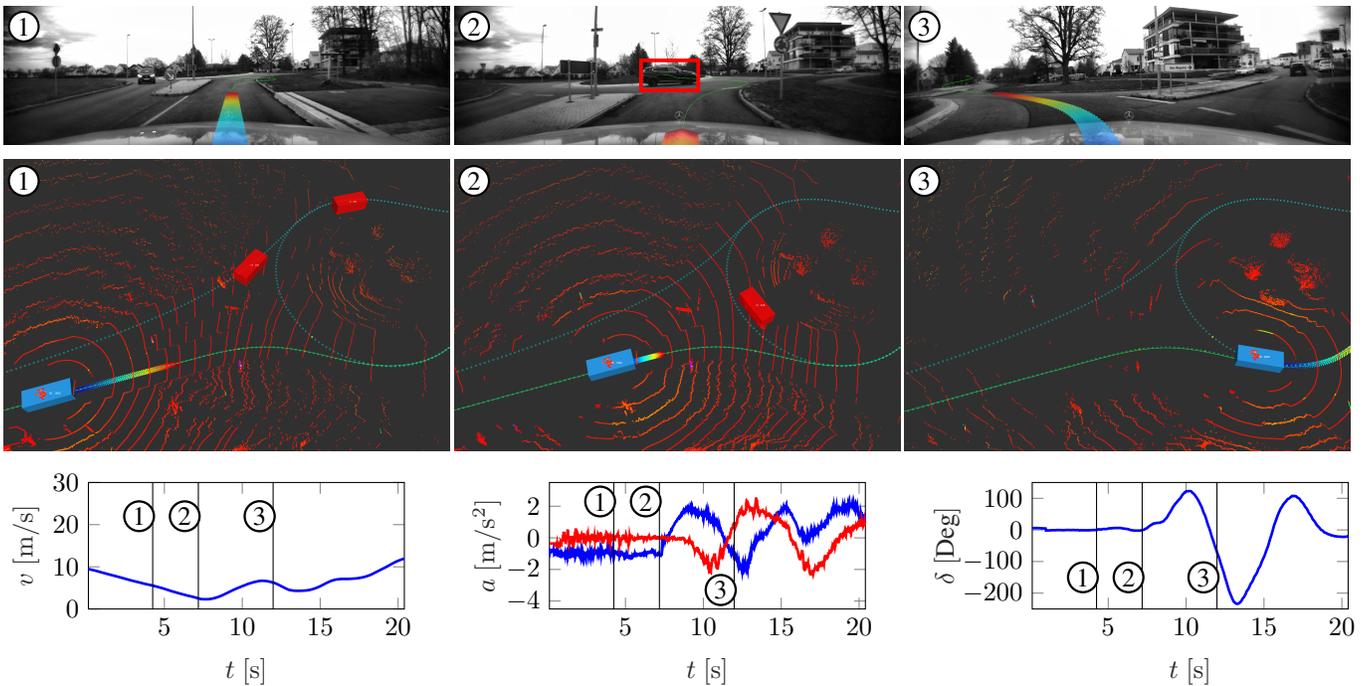
\begin{figure*}
		\setlength\columnsep{5pt}
		\begin{multicols}{3}
			\begin{tikzpicture}
            \draw (0, 0) node[inner sep=0] {\includegraphics[clip,width=\linewidth]{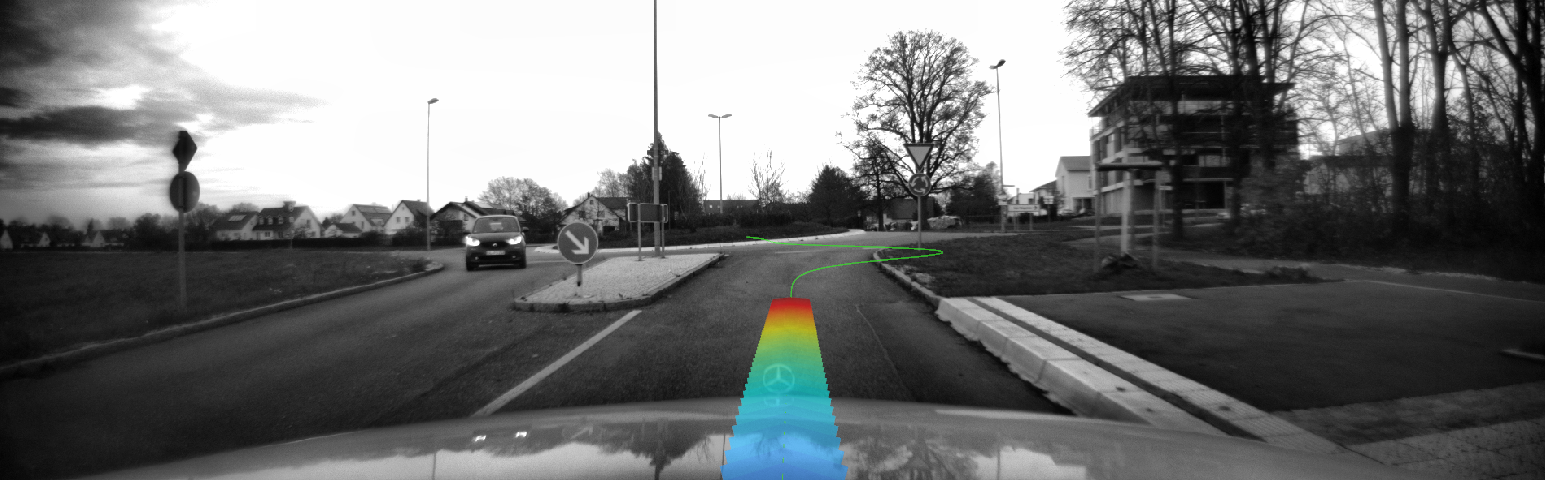}};
			\draw (-2.7, 0.65) node[circle,fill=white,draw=black,thick,inner sep=1pt]{1};
			\end{tikzpicture}\,\columnbreak
			\begin{tikzpicture}
            \draw (0, 0) node[inner sep=0] {\includegraphics[clip,width=\linewidth]{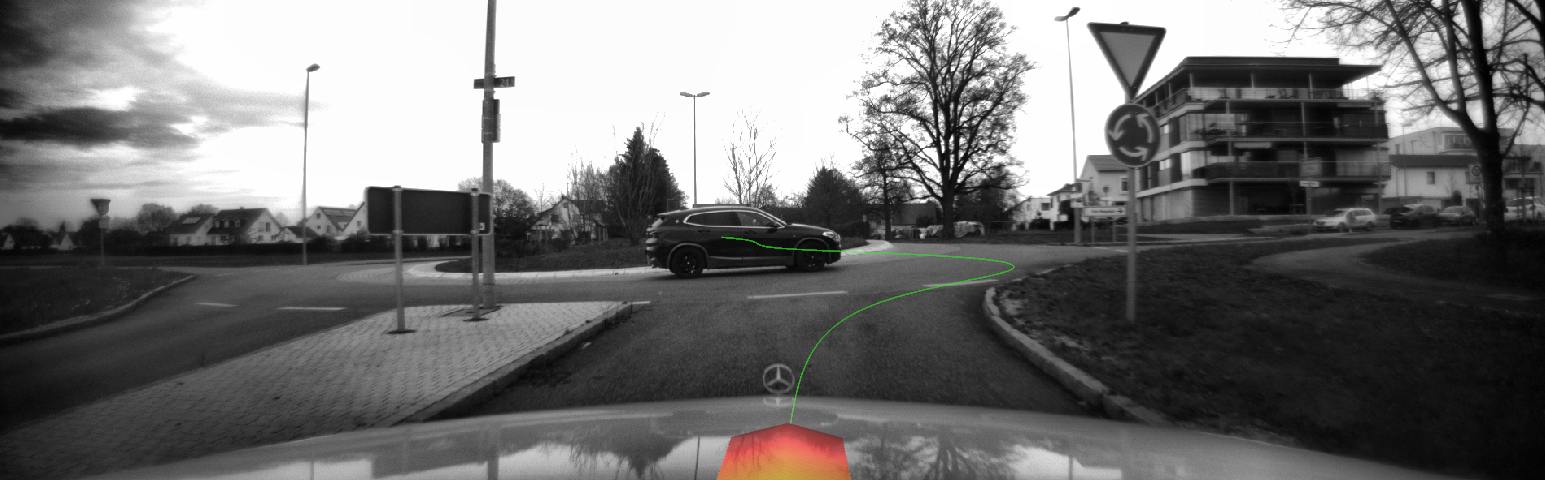}};
            \node[rectangle, draw=red, xscale=1.9, yscale=1.0, line width=0.5mm, inner sep=0pt, minimum size=0.4cm] (a) at (-0.1, 0.0) {};
			\draw (-2.7, 0.65) node[circle,fill=white,draw=black,thick,inner sep=1pt]{2};
			\end{tikzpicture}\,\columnbreak
			\begin{tikzpicture}
            \draw (0, 0) node[inner sep=0] {\includegraphics[clip,width=\linewidth]{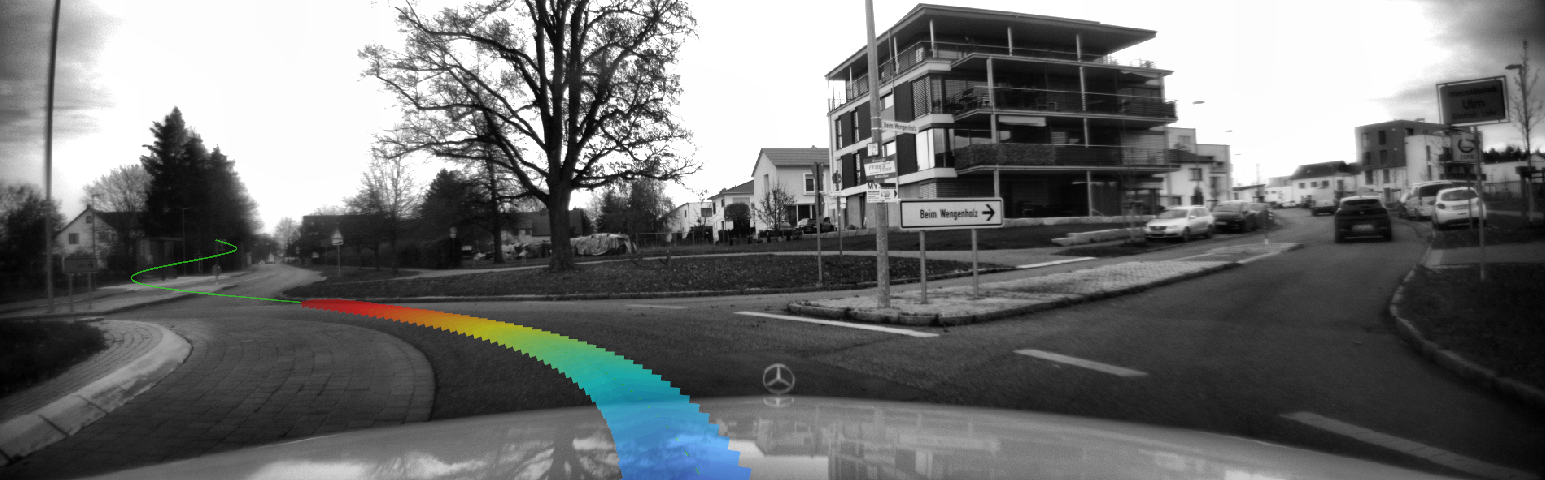}};
			\draw (-2.7, 0.65) node[circle,fill=white,draw=black,thick,inner sep=1pt]{3};
			\end{tikzpicture}
		\end{multicols}
		\vspace{-0.65cm}
		\begin{multicols}{3}
			\begin{tikzpicture}
			\draw (0, 0) node[inner sep=0] {\includegraphics[trim={0cm 2cm 0cm 2cm}, clip,width=\linewidth]{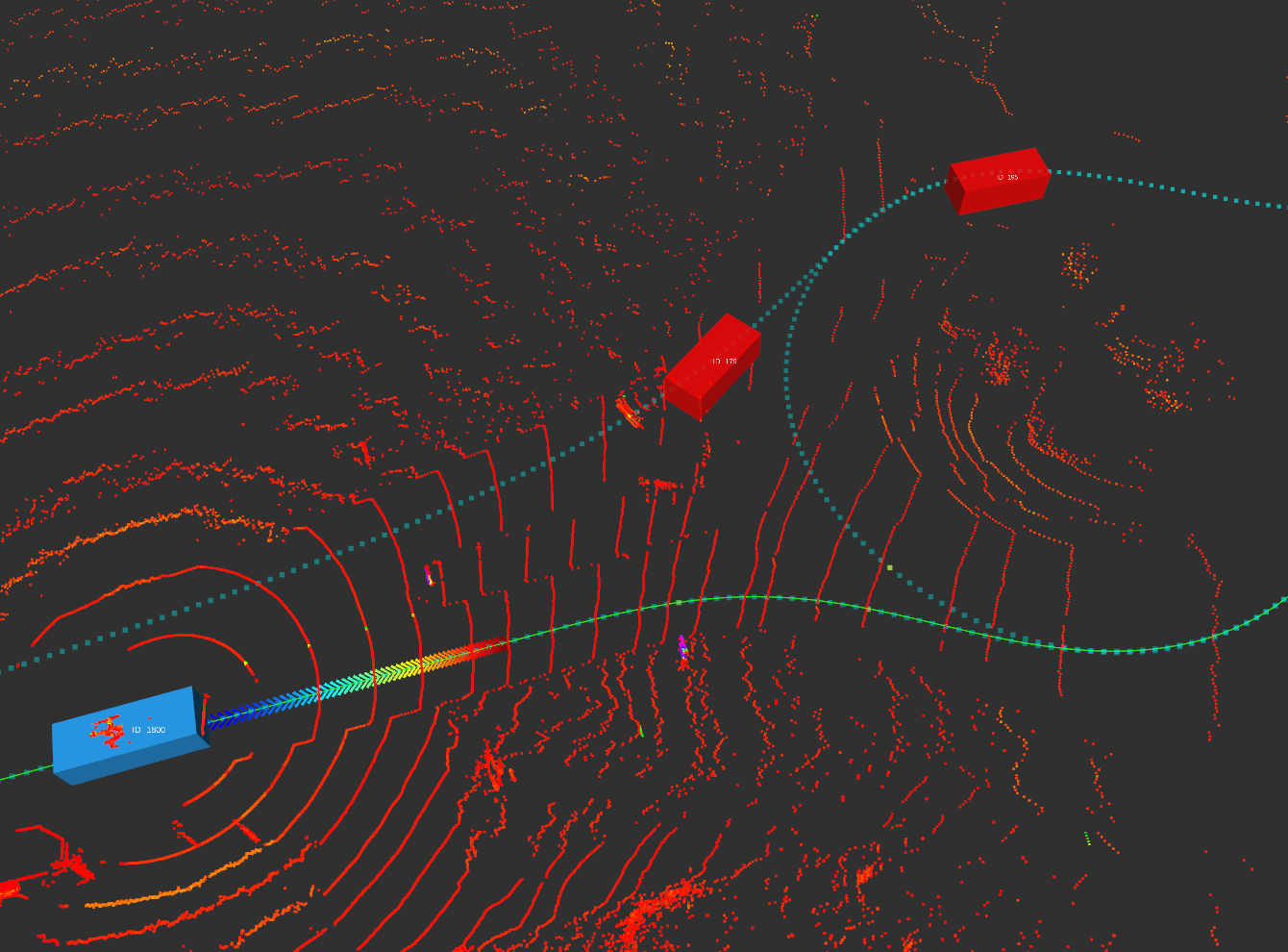}};
			\draw (-2.7, 1.65) node[circle,fill=white,draw=black,thick,inner sep=1pt]{1};
			\end{tikzpicture}\,\columnbreak
			\begin{tikzpicture}
			\draw (0, 0) node[inner sep=0] {\includegraphics[trim={0cm 2cm 0cm 2cm}, clip,width=\linewidth]{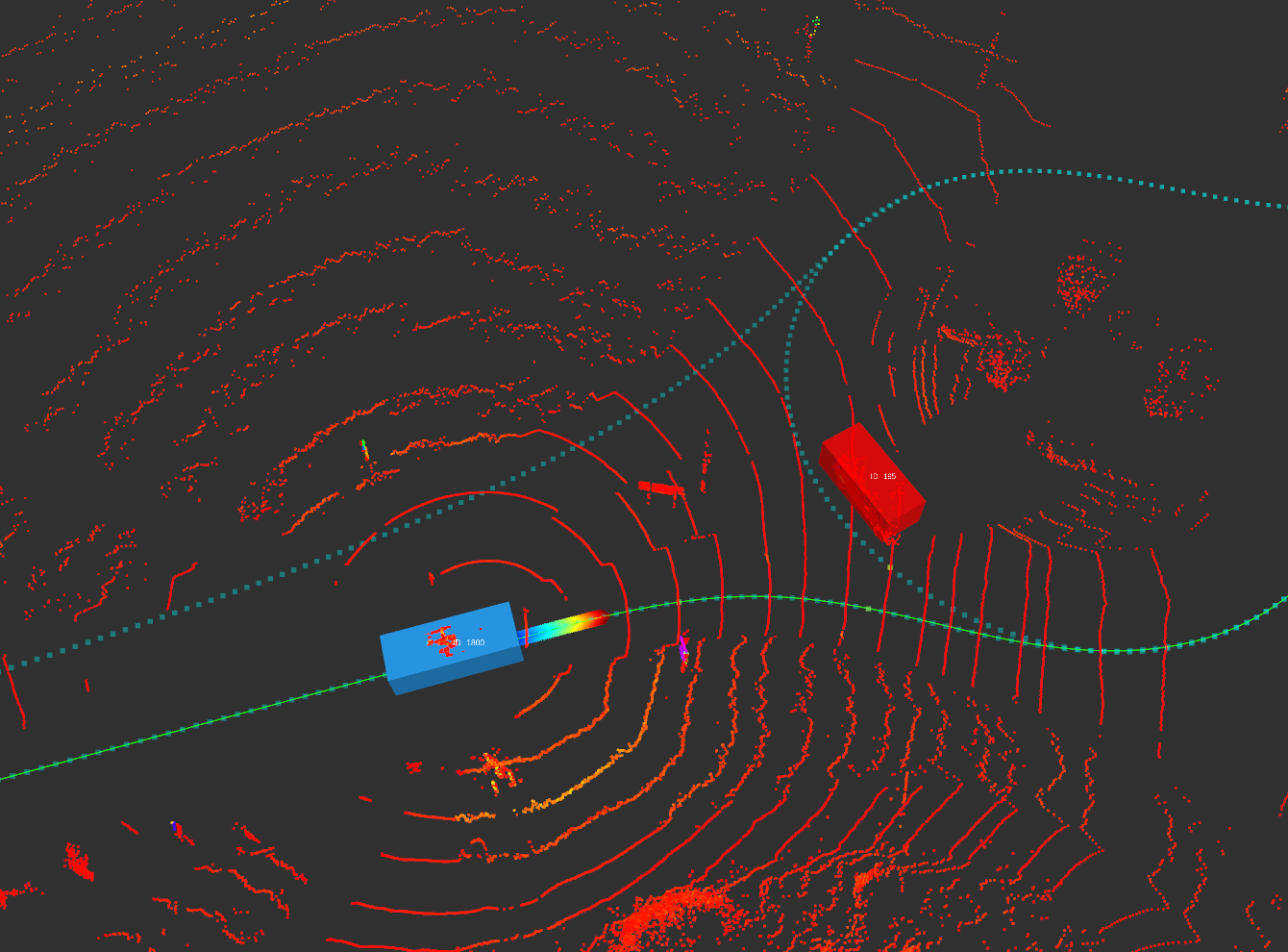}};
			\draw (-2.7, 1.65) node[circle,fill=white,draw=black,thick,inner sep=1pt]{2};
			\end{tikzpicture}\,\columnbreak
			\begin{tikzpicture}
			\draw (0, 0) node[inner sep=0] {\includegraphics[trim={0cm 2cm 0cm 2cm}, clip,width=\linewidth]{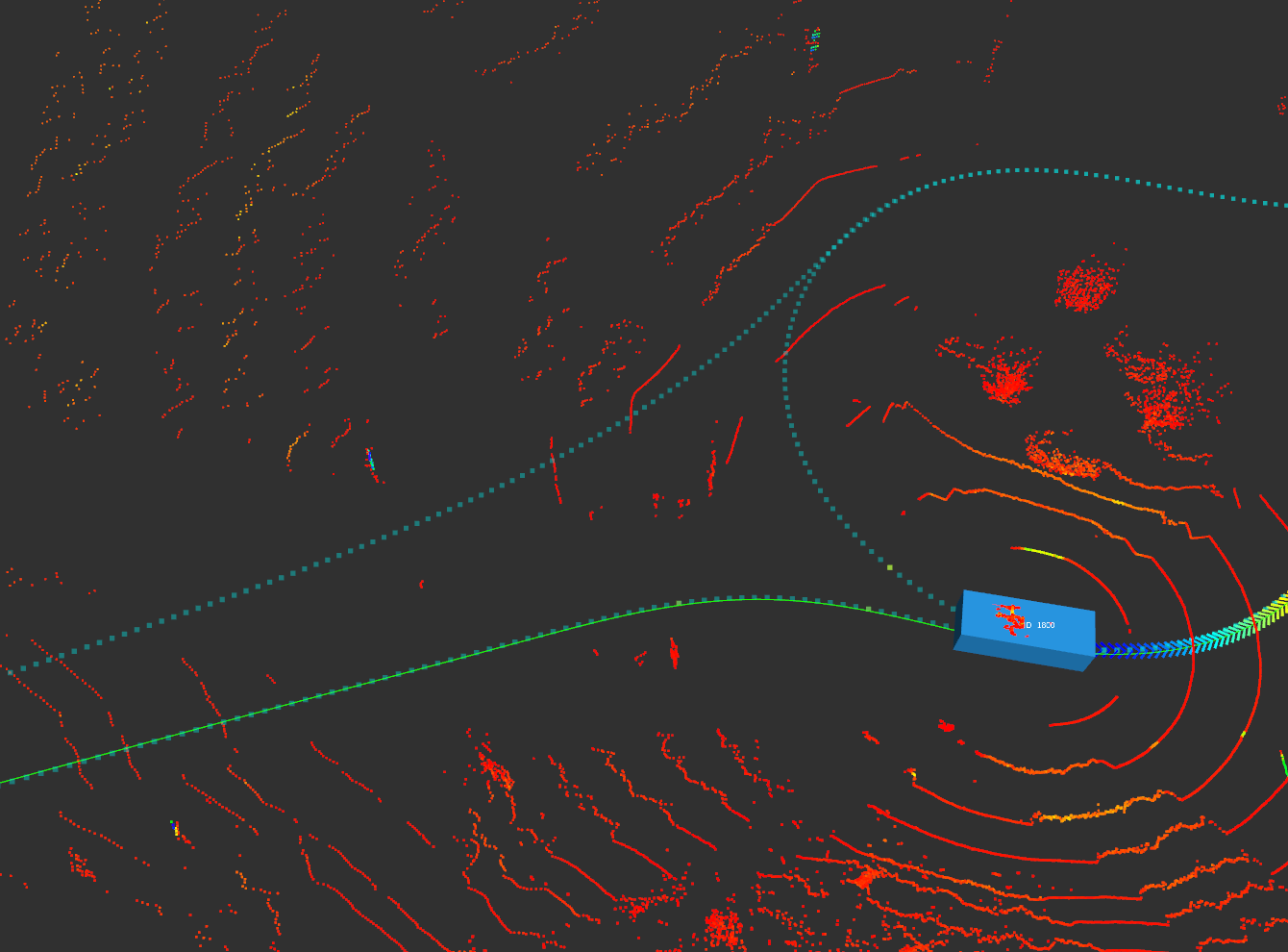}};
			\draw (-2.7, 1.65) node[circle,fill=white,draw=black,thick,inner sep=1pt]{3};
			\end{tikzpicture}
		\end{multicols}
	\vspace{-0.65cm}
		\begin{multicols}{3}
			\input{Images/Tikz/fig10-3-1.tikz}\;\;\;\;\;\;\columnbreak
			\input{Images/Tikz/fig10-3-2.tikz}\;\;\;\;\;\;\columnbreak
			\input{Images/Tikz/fig10-3-3.tikz}
		\end{multicols}
		\caption{
            Roundabout scenario, for three dedicated time instants:
	        The first row shows the drivers perspective by camera images, with vehicles emphasized by red bounding boxes. The second row shows recorded as well as processed data in top view. The third row depicts measured values during the maneuver, where the velocity, the longitudinal (blue) together with the lateral (red) acceleration, and the steering wheel angle are shown over time.
        }
		\label{fig:evalRoundabout}
	\end{figure*}

The second experiment corresponds to a roundabout scenario. According evaluations and pictures are shown in Figure \ref{fig:evalRoundabout}. Besides the center line of the ego vehicle, possible center lines for other vehicles in the roundabout are visualized in the ROS-RViz display. \newline 
Initially, there are two other vehicles involved in the scenario. Therefore, the ego vehicle first has to reduce the velocity while approaching the roundabout. 
At \tikz\node[circle,fill=white,draw=black,thick,inner sep=1pt]{\footnotesize 2};, the velocity is adapted for smooth merging behavior behind the other vehicle into the roundabout using the approach presented in \ref{sec:long_driving}. At \tikz\node[circle,fill=white,draw=black,thick,inner sep=1pt]{\footnotesize 3}; the vehicle merged successfully.
Qualitatively, the driving experience during this scenario can be described as natural and human-like.

	\subsection{Intersection}
	\begin{figure*}
		\setlength\columnsep{5pt}
	\begin{multicols}{3}
		\begin{tikzpicture}
		\draw (0, 0) node[inner sep=0] {\includegraphics[clip,width=\linewidth]{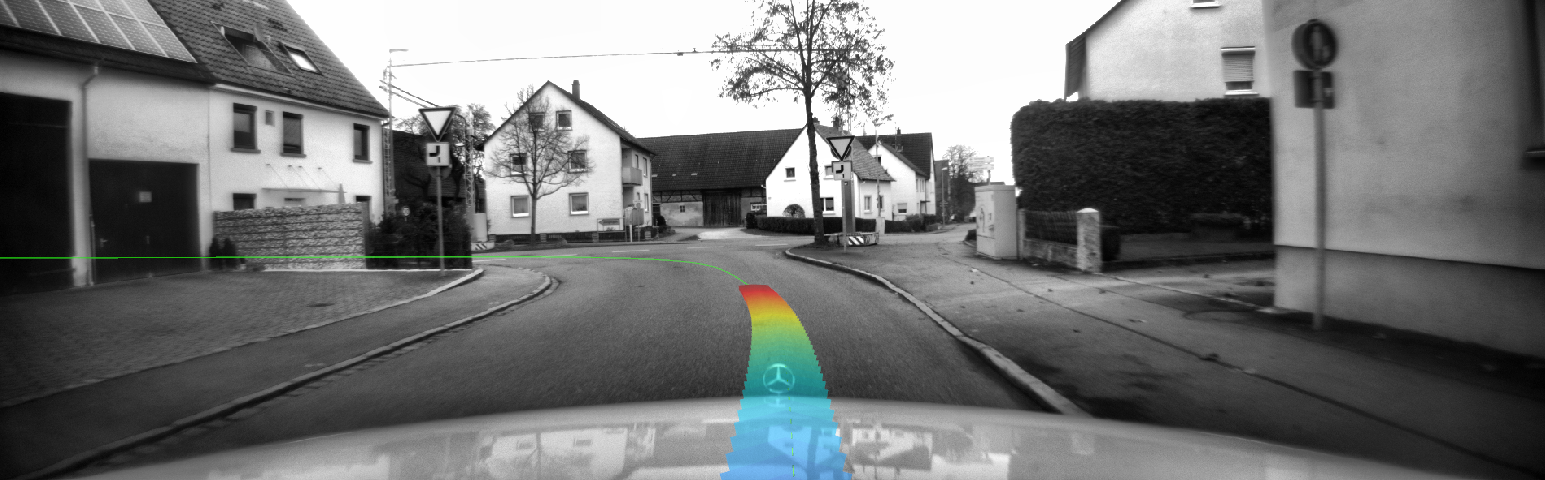}};
		\draw (-2.7, 0.65) node[circle,fill=white,draw=black,thick,inner sep=1pt]{1};
		\end{tikzpicture}\,\columnbreak
		\begin{tikzpicture}
		\draw (0, 0) node[inner sep=0] {\includegraphics[clip,width=\linewidth]{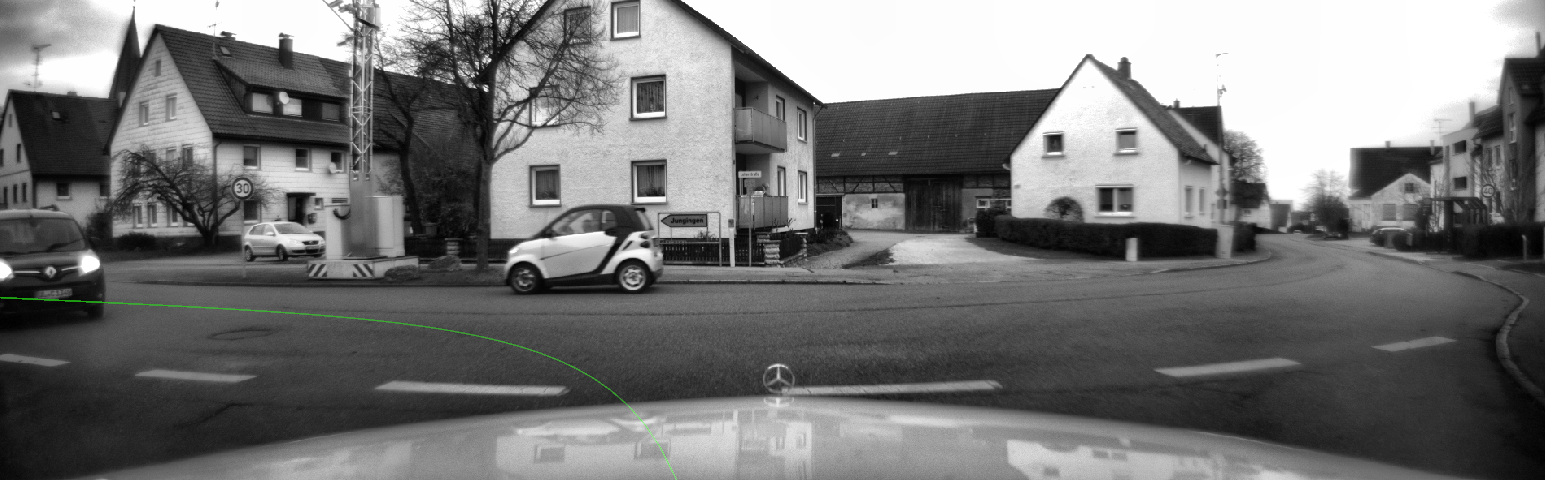}};
        \node[rectangle, draw=red, xscale=1.65, line width=0.5mm, inner sep=0pt, minimum size=0.4cm] (a) at (-0.75, -0.03) {};
        \node[rectangle, draw=red, xscale=1.05, yscale=1.2, line width=0.5mm, inner sep=0pt, minimum size=0.4cm] (a) at (-2.73, -0.1) {};
		\draw (-2.7, 0.65) node[circle,fill=white,draw=black,thick,inner sep=1pt]{2};
		\end{tikzpicture}\,\columnbreak
		\begin{tikzpicture}
		\draw (0, 0) node[inner sep=0] {\includegraphics[clip,width=\linewidth]{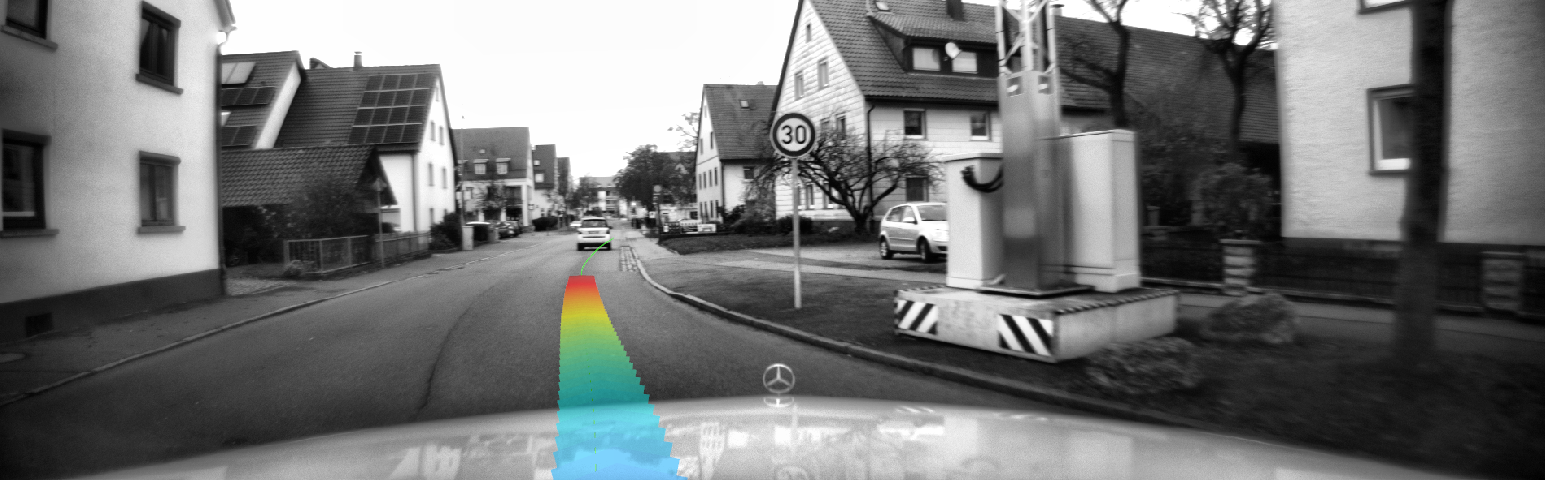}};
		\draw (-2.7, 0.65) node[circle,fill=white,draw=black,thick,inner sep=1pt]{3};
		\end{tikzpicture}
	\end{multicols}
	\vspace{-0.65cm}
	\begin{multicols}{3}
		\begin{tikzpicture}
		\draw (0, 0) node[inner sep=0] {\includegraphics[clip,width=\linewidth]{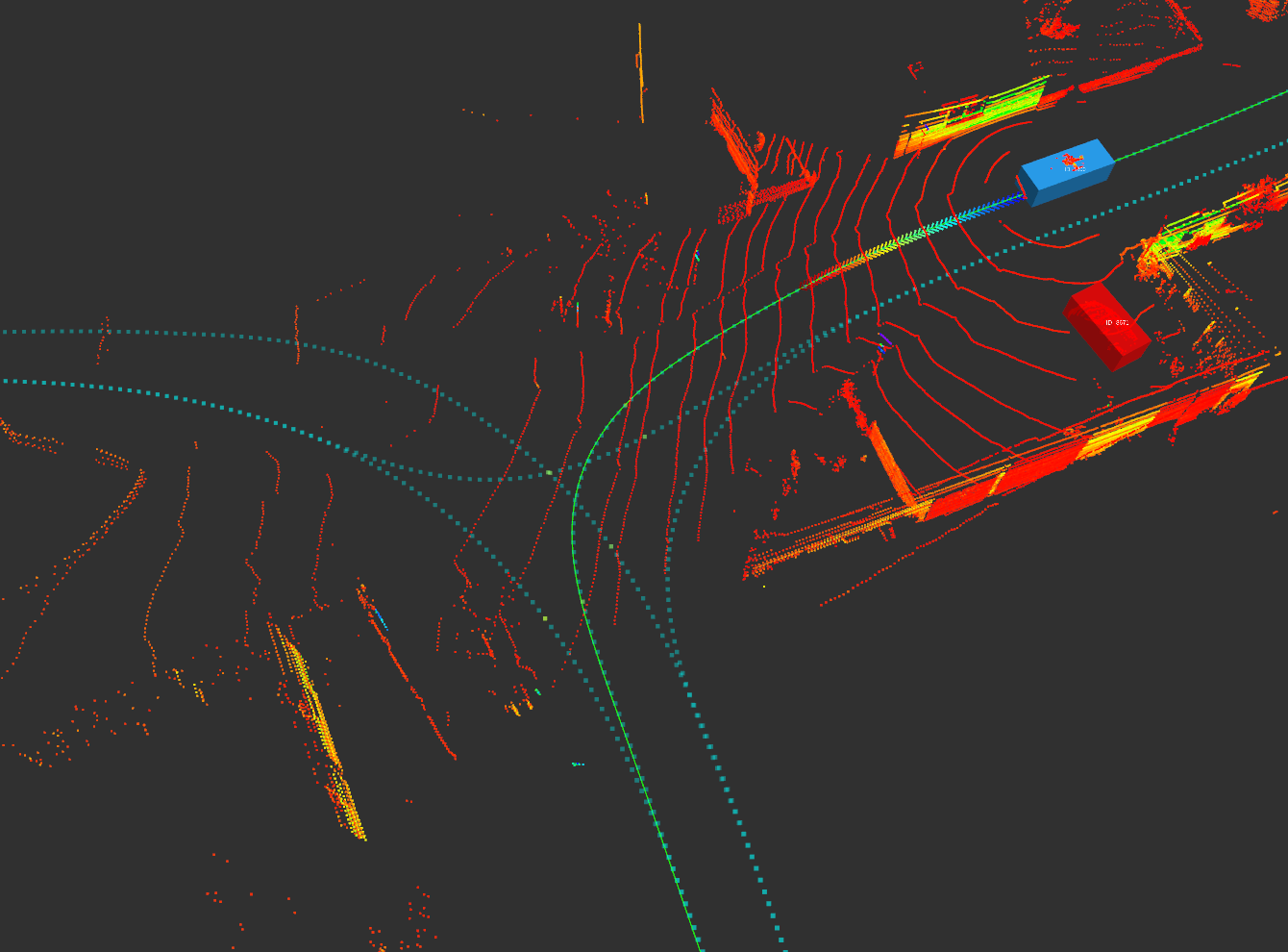}};
		\draw (-2.7, 1.9) node[circle,fill=white,draw=black,thick,inner sep=1pt]{1};
		\end{tikzpicture}\,\columnbreak
		\begin{tikzpicture}
		\draw (0, 0) node[inner sep=0] {\includegraphics[clip,width=\linewidth]{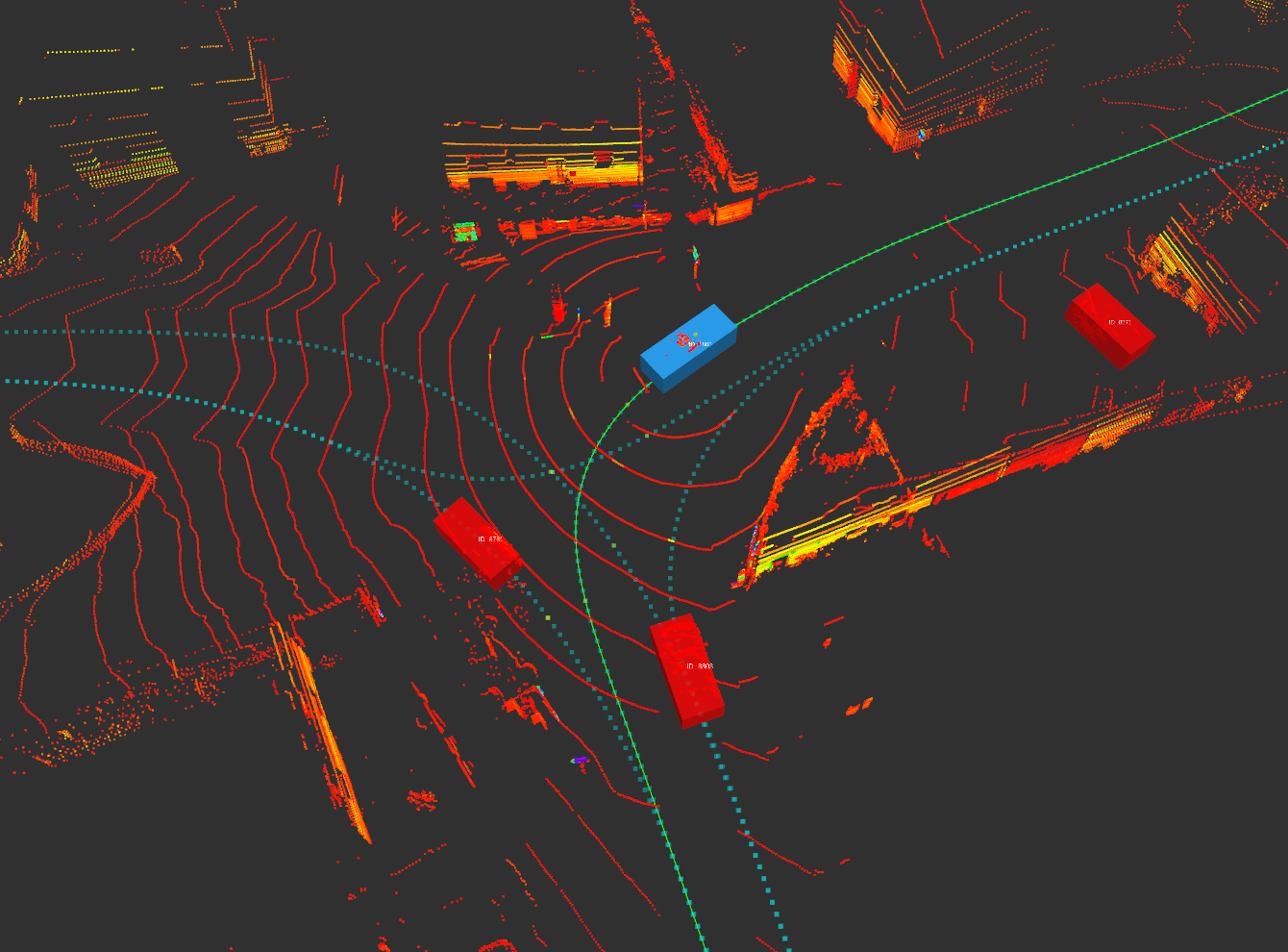}};
		\draw (-2.7, 1.9) node[circle,fill=white,draw=black,thick,inner sep=1pt]{2};
		\end{tikzpicture}\,\columnbreak
		\begin{tikzpicture}
		\draw (0, 0) node[inner sep=0] {\includegraphics[clip,width=\linewidth]{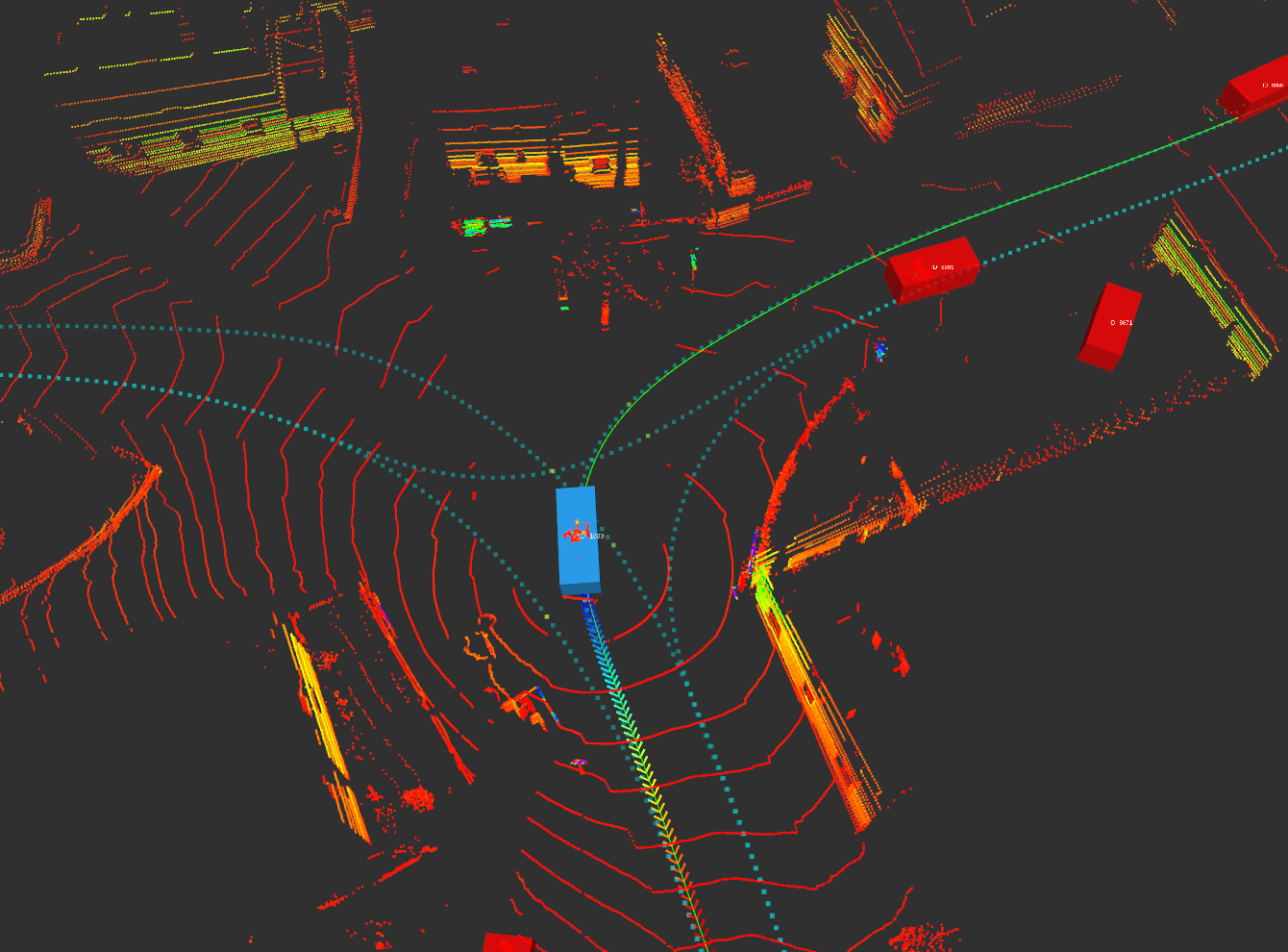}};
		\draw (-2.7, 1.9) node[circle,fill=white,draw=black,thick,inner sep=1pt]{3};
		\end{tikzpicture}
	\end{multicols}
	\begin{multicols}{3}
		\vspace{-0.65cm}
		\input{Images/Tikz/fig11-3-1.tikz}\;\;\;\;\;\;\columnbreak
		\input{Images/Tikz/fig11-3-2.tikz}\;\;\;\;\;\;\columnbreak
		\input{Images/Tikz/fig11-3-3.tikz}
	\end{multicols}
	\caption{
        Left turn scenario, for three dedicated time instants:
            The first row shows the drivers perspective by camera images, with vehicles emphasized by red bounding boxes.
            The second row shows recorded as well as processed data in top view.
            The third row depicts measured values during the maneuver, where the velocity, the longitudinal (blue) together with the lateral (red) acceleration,
            and the steering wheel angle are shown over time.
            Note that the peak in the lateral acceleration at $t \approx\SI{4.5}{s}$ is due to measurement noise.
    }
    \label{fig:evalLeftTurn}
	\end{figure*}  
	
	The third experiment demonstrates a left turn scenario. The corresponding scene is visualized in Figure \ref{fig:evalLeftTurn}. First, the vehicle approaches the intersection and reduces the velocity. In general, there are two other vehicles involved. One is approaching from the left and one approaching from the right, both are prioritized over the ego vehicle. Using the approach presented in \ref{sec:long_driving}, the vehicle remains in standstill up to \tikz\node[circle,fill=white,draw=black,thick,inner sep=1pt]{\footnotesize 2}; as driving in front of the other vehicles is rated as too aggressive and, consequently, not feasible in this scenario. Driving off shortly after \tikz\node[circle,fill=white,draw=black,thick,inner sep=1pt]{\footnotesize 2}; implements courteous behavior as the vehicle approaching from the left is already predicted to take a right turn and therefore, it does not interact with the ego vehicle. Thus, the ego vehicle is able to smoothly merge behind the vehicle approaching from the right. 

\section{Conclusion and Future Work}

We presented a motion planning system that is capable of computing comfortable
trajectories in real-world scenarios by using model knowledge in combination
with continuous optimization.  Thereby, the main task of the behavior
generation consists in computing a maneuver specific  behavior trajectory for
the central optimization problem. Therefore, the EIDM equations are used to
compute the longitudinal motion along the lane's reference line. If lane
changes are to be performed, a single track model is additionally used to also
describe the lateral motion. The resulting behavior trajectory mainly
influences the pose of the global optimum of the central optimization problem
and therefore, the optimal solution and the corresponding final maneuver. Due
to the problem constraints, it is ensured that collisions with the static free
space boundaries, as well as with dynamic obstacles, are avoided.
\newline
This paper corresponds to a continuation of our previous works and shows beside
some theoretical extensions also the practical deployment. We have described
how a level 4 motion planning system can be built and demonstrated the
functionality by real-world results in public road traffic.
However, even though the vehicle is able to cope with a variety of daily traffic situations, there is no possiblity to obtain a driving behavior that differs substantially from something than can be generated using the EIDM/IDM. This becomes an issue, for instance, when driving in urban environments, with cyclists and pedestrians crossing the road. In such situations, more advanced approaches, like e.~g. machine learning techniques could be utilized to compute the behavior trajectory.  

% or


%\appendices

%\section{}

%\section*{Acknowledgment}
%%
%%
%The authors would like to thank all employees of  the Institute of Measurement, Control and Microtechnology of the University of Ulm, who participated in the work on the experimental vehicle.

\ifCLASSOPTIONcaptionsoff
  \newpage
\fi



\bibliographystyle{IEEEtran}
%\begin{thebibliography}{1}

\begin{IEEEbiography}[{\includegraphics[width=1in,height=1.25in,clip,keepaspectratio]{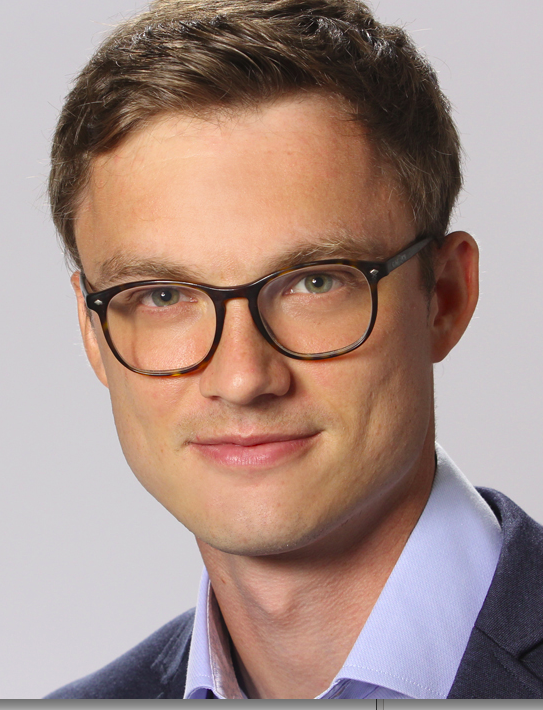}}]{Maximilian J. Graf}
	received  his B. Sc. degree in 2014 and his M. Sc. degree in 2016, both from the University of Ulm/Germany in the field of electrical engineering. From 2016 until 2020 he was a research assistant and PhD student at the institute of Measurement, Control and Microtechnology at the University of Ulm and since 2021 he is with the ZF~Group Friedrichshafen. His main research  interests include motion planning for autonomous driving and vehicle motion control.
\end{IEEEbiography}

\begin{IEEEbiography}[{\includegraphics[width=1in,height=1.25in,clip,keepaspectratio]{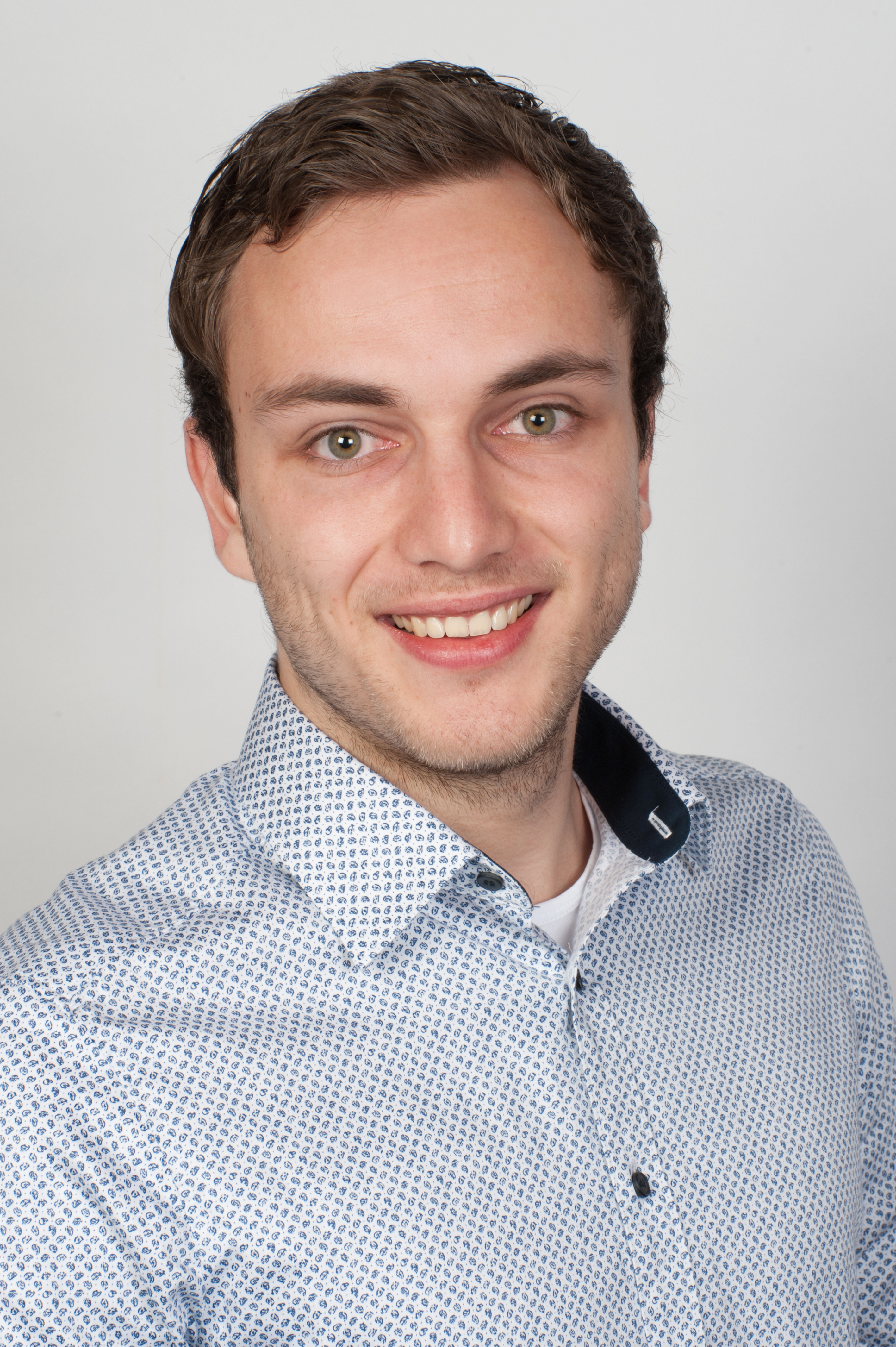}}]{Oliver M. Speidel}
	received the Master’s degree in computer science with distinction from Ulm University, Ulm, Germany in 2017. Currently, he is a research assistent at the institute of Measurement, Control and Microtechnology of Ulm University, where he is working towards the
	Doctoral degree. His main research interests include the areas of decision-making and motion planning with a focus on
	applications for autonomous vehicles.
	% His research interests include motion planning, trajectory planning, graph-based planning as well as automated driving and autonomous vehicles.
\end{IEEEbiography}

\begin{IEEEbiography}[{\includegraphics[width=1in,height=1.25in,clip,keepaspectratio]{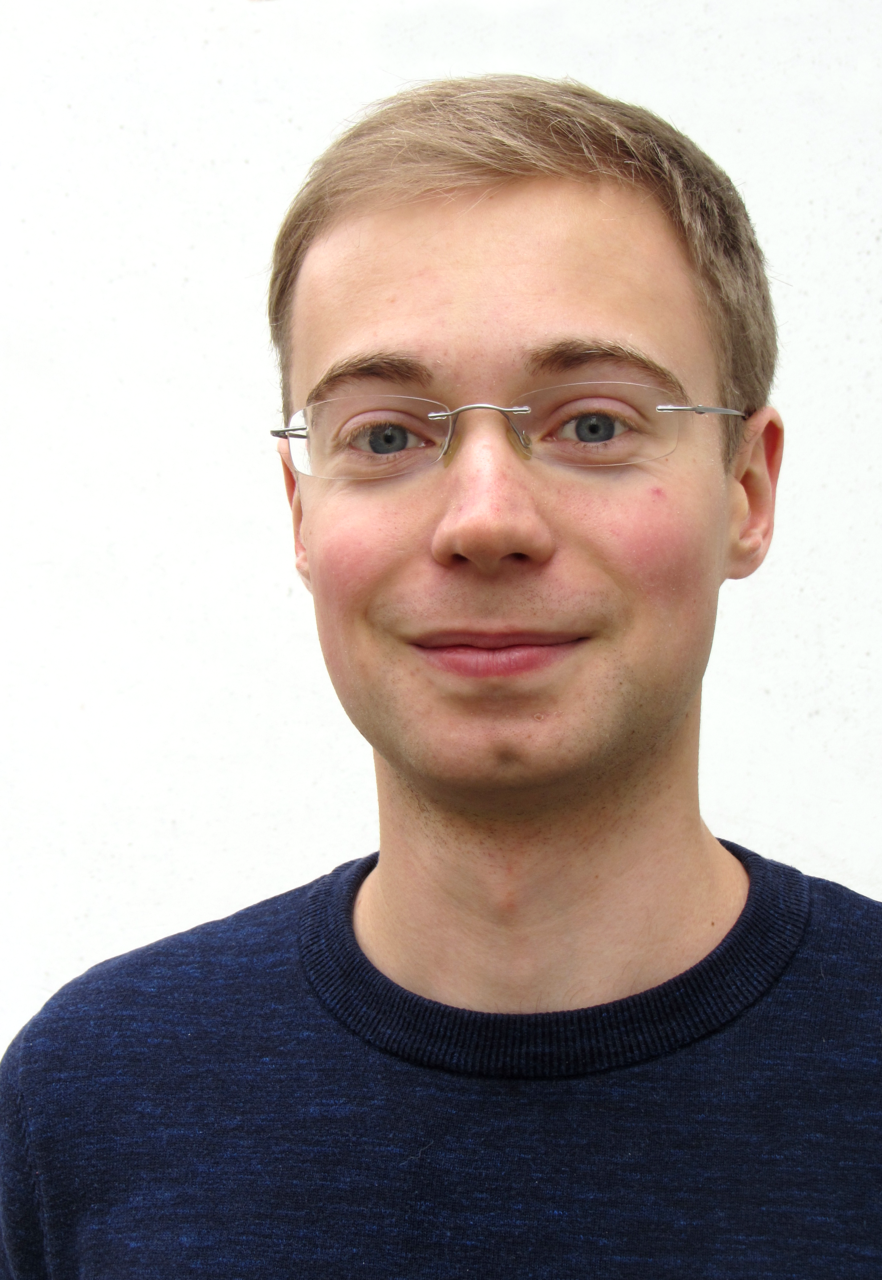}}]{Jona O. Ruof}
	received his M. Sc. degree in media informatics with distinction in 2019 from Ulm University, Ulm, Germany.
    Since 2019, he is a research assistent at the institute of Measurement, Control and Microtechnology of Ulm University, working towards the
	Doctoral degree. His current research interests encompass, decision-making, motion planning, and learning control for automated vehicles.
\end{IEEEbiography}

\begin{IEEEbiography}[{\includegraphics[width=1in,height=1.25in,clip,keepaspectratio]{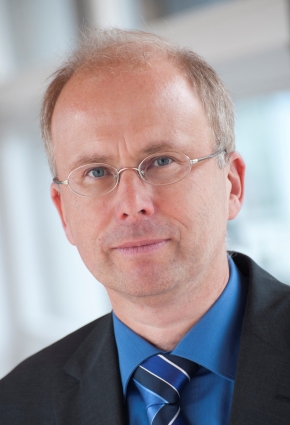}}]{Klaus Dietmayer}
	(M’05) was born in Celle,
	Germany, in 1962. He received the Diploma degree
	(equivalent to M.Sc. degree) in electrical engineering
	from the Technical University of Braunschweig,
	Germany, in 1989, and the Dr.-Ing. degree (equivalent to Ph.D. degree) from the University of Armed
	Forces, Hamburg, Germany, in 1994. In 1994,
	he joined Philips Semiconductors Systems Laboratory, Hamburg, as a Research Engineer. In 1996,
	he became a manager in the field of networks and
	sensors for automotive applications. In 2000, he was
	appointed to a Professorship at Ulm University in the field of measurement
	and control. He is currently a Full Professor and the Director of the Institute
	of Measurement, Control and Microtechnology, School of Engineering and
	Computer Science, Ulm University. His research interests include information
	fusion, multi-object tracking, environment perception for advanced automotive
	driver assistance, and E-mobility. He is a member of the German Society of
	Engineers VDI/VDE.
\end{IEEEbiography}

%\end{thebibliography}



%\newpage


%\vfill


\end{document}

%% file: Images/svgs/fig2.pdf_tex
%% Creator: Inkscape inkscape 0.92.5, www.inkscape.org
%% Accompanies image file 'env2.pdf' (pdf, eps, ps)
\begingroup%
  \makeatletter%
  \providecommand\color[2][]{%
    \errmessage{(Inkscape) Color is used for the text in Inkscape, but the package 'color.sty' is not loaded}%
    \renewcommand\color[2][]{}%
  }%
  \providecommand\transparent[1]{%
    \errmessage{(Inkscape) Transparency is used (non-zero) for the text in Inkscape, but the package 'transparent.sty' is not loaded}%
    \renewcommand\transparent[1]{}%
  }%
  \providecommand\rotatebox[2]{#2}%
  \newcommand*\fsize{\dimexpr\f@size pt\relax}%
  \newcommand*\lineheight[1]{\fontsize{\fsize}{#1\fsize}\selectfont}%
  \ifx\svgwidth\undefined%
    \setlength{\unitlength}{399.08394445bp}%
    \ifx\svgscale\undefined%
      \relax%
    \else%
      \setlength{\unitlength}{\unitlength * \real{\svgscale}}%
    \fi%
  \else%
    \setlength{\unitlength}{\svgwidth}%
  \fi%
  \global\let\svgwidth\undefined%
  \global\let\svgscale\undefined%
  \makeatother%
  \begin{picture}(1,0.54798331)%
    \lineheight{1}%
    \setlength\tabcolsep{0pt}%
    \put(0,0){\includegraphics[width=\unitlength,page=1]{fig2.pdf}}%
    \put(-0.00089561,0.47358451){\color[rgb]{0,0,0}\makebox(0,0)[lt]{\lineheight{1.25}\smash{\begin{tabular}[t]{l}y\end{tabular}}}}%
    \put(0.92560122,0){\color[rgb]{0,0,0}\makebox(0,0)[lt]{\lineheight{1.25}\smash{\begin{tabular}[t]{l}x\end{tabular}}}}%
  \end{picture}%
\endgroup%

%% file: Images/svgs/fig3a.pdf_tex
%% Accompanies image file 'onecircle.pdf' (pdf, eps, ps)
\begingroup%
  \makeatletter%
  \providecommand\color[2][]{%
    \errmessage{(Inkscape) Color is used for the text in Inkscape, but the package 'color.sty' is not loaded}%
    \renewcommand\color[2][]{}%
  }%
  \providecommand\transparent[1]{%
    \errmessage{(Inkscape) Transparency is used (non-zero) for the text in Inkscape, but the package 'transparent.sty' is not loaded}%
    \renewcommand\transparent[1]{}%
  }%
  \providecommand\rotatebox[2]{#2}%
  \newcommand*\fsize{\dimexpr\f@size pt\relax}%
  \newcommand*\lineheight[1]{\fontsize{\fsize}{#1\fsize}\selectfont}%
  \ifx\svgwidth\undefined%
    \setlength{\unitlength}{112.2143678bp}%
    \ifx\svgscale\undefined%
      \relax%
    \else%
      \setlength{\unitlength}{\unitlength * \real{\svgscale}}%
    \fi%
  \else%
    \setlength{\unitlength}{\svgwidth}%
  \fi%
  \global\let\svgwidth\undefined%
  \global\let\svgscale\undefined%
  \makeatother%
  \begin{picture}(1,0.61821692)%
    \lineheight{1}%
    \setlength\tabcolsep{0pt}%
    \put(0,0){\includegraphics[width=\unitlength,page=1]{fig3a.pdf}}%
    \put(0.7590766,0.60763444){\color[rgb]{0,0,0}\makebox(0,0)[lt]{\lineheight{1.25}\smash{\begin{tabular}[t]{l}$r$\end{tabular}}}}%
    \put(0.68034358,0.58510862){\color[rgb]{0,0,0}\makebox(0,0)[lt]{\lineheight{1.25}\smash{\begin{tabular}[t]{l}$\rotatebox{-68}{$\bigg \{$}$\end{tabular}}}}%
    \put(0,0){\includegraphics[width=\unitlength,page=2]{fig3a.pdf}}%
  \end{picture}%
\endgroup%

%% file: Images/svgs/fig3b.pdf_tex
%% Accompanies image file 'dynobstacle.pdf' (pdf, eps, ps)
\begingroup%
  \makeatletter%
  \providecommand\color[2][]{%
    \errmessage{(Inkscape) Color is used for the text in Inkscape, but the package 'color.sty' is not loaded}%
    \renewcommand\color[2][]{}%
  }%
  \providecommand\transparent[1]{%
    \errmessage{(Inkscape) Transparency is used (non-zero) for the text in Inkscape, but the package 'transparent.sty' is not loaded}%
    \renewcommand\transparent[1]{}%
  }%
  \providecommand\rotatebox[2]{#2}%
  \newcommand*\fsize{\dimexpr\f@size pt\relax}%
  \newcommand*\lineheight[1]{\fontsize{\fsize}{#1\fsize}\selectfont}%
  \ifx\svgwidth\undefined%
    \setlength{\unitlength}{180.24078527bp}%
    \ifx\svgscale\undefined%
      \relax%
    \else%
      \setlength{\unitlength}{\unitlength * \real{\svgscale}}%
    \fi%
  \else%
    \setlength{\unitlength}{\svgwidth}%
  \fi%
  \global\let\svgwidth\undefined%
  \global\let\svgscale\undefined%
  \makeatother%
  \begin{picture}(1,0.31837546)%
    \lineheight{1}%
    \setlength\tabcolsep{0pt}%
    \put(0,0){\includegraphics[width=\unitlength,page=1]{fig3b.pdf}}%
    \put(0.27449101,0.31178702){\color[rgb]{0,0,0}\makebox(0,0)[lt]{\lineheight{1.25}\smash{\begin{tabular}[t]{l}$r$\end{tabular}}}}%
    \put(0.21647756,0.29684162){\color[rgb]{0,0,0}\makebox(0,0)[lt]{\lineheight{1.25}\smash{\begin{tabular}[t]{l}$\rotatebox{-90}{$\bigg \{$}$\end{tabular}}}}%
    \put(0,0){\includegraphics[width=\unitlength,page=2]{fig3b.pdf}}%
  \end{picture}%
\endgroup%

%% file: Images/svgs/fig4.pdf_tex
%% Accompanies image file 'staticboundaries.pdf' (pdf, eps, ps)
\begingroup%
  \makeatletter%
  \providecommand\color[2][]{%
    \errmessage{(Inkscape) Color is used for the text in Inkscape, but the package 'color.sty' is not loaded}%
    \renewcommand\color[2][]{}%
  }%
  \providecommand\transparent[1]{%
    \errmessage{(Inkscape) Transparency is used (non-zero) for the text in Inkscape, but the package 'transparent.sty' is not loaded}%
    \renewcommand\transparent[1]{}%
  }%
  \providecommand\rotatebox[2]{#2}%
  \newcommand*\fsize{\dimexpr\f@size pt\relax}%
  \newcommand*\lineheight[1]{\fontsize{\fsize}{#1\fsize}\selectfont}%
  \ifx\svgwidth\undefined%
    \setlength{\unitlength}{427.44710642bp}%
    \ifx\svgscale\undefined%
      \relax%
    \else%
      \setlength{\unitlength}{\unitlength * \real{\svgscale}}%
    \fi%
  \else%
    \setlength{\unitlength}{\svgwidth}%
  \fi%
  \global\let\svgwidth\undefined%
  \global\let\svgscale\undefined%
  \makeatother%
  \begin{picture}(1,0.20979725)%
    \lineheight{1}%
    \setlength\tabcolsep{0pt}%
    \put(0,0){\includegraphics[width=\unitlength,page=1]{fig4.pdf}}%
  \end{picture}%
\endgroup%

%% file: Images/Tikz/fig9-3-1.tikz
\begin{tikzpicture}

\begin{axis}[%
width=4.25cm,
height=1.683cm,
at={(0cm,0cm)},
scale only axis,
xmin=0.642011881,
xmax=12.46199894,
xlabel style={font=\color{white!15!black}},
xlabel={$t\;[\SI[per-mode=repeated-symbol]{}{\second}]$},
ymin=0,
ymax=30,
ylabel style={font=\color{white!15!black}},
ylabel={$v\;[\SI[per-mode=repeated-symbol]{}{\meter\per\second}]$},
axis background/.style={fill=white},
ylabel style={yshift=-0.4cm}
]
\addplot [color=blue, line width=1.0pt, forget plot]
  table[row sep=crcr]{%
0.642011881	14.1500035335685\\
0.662031174	14.160007944913\\
0.682033062	14.1750079365057\\
0.702054262	14.1900079281162\\
0.722006321	14.200007922533\\
0.742020607	14.2100079169577\\
0.762029409	14.2250079086094\\
0.782009363	14.2400079002787\\
0.802028418	14.2550078919655\\
0.822096586	14.2650078864332\\
0.842060328	14.2750078809085\\
0.862061977	14.2850078753916\\
0.882035017	14.3000034965031\\
0.902044773	14.3100034940597\\
0.922039747	14.325007853401\\
0.941986799	14.3350034879661\\
0.962060928	14.3500034843201\\
0.982016802	14.3600008704735\\
1.001998186	14.3700008698678\\
1.0225811	14.3850008689607\\
1.041987896	14.4000008680555\\
1.062039852	14.4100008674531\\
1.082021952	14.420003467406\\
1.102125168	14.430003465003\\
1.122168303	14.4400034626035\\
1.142053843	14.4550034590103\\
1.162023783	14.4700034554246\\
1.182157993	14.475003454231\\
1.202011108	14.4850077666531\\
1.222189903	14.4950077612949\\
1.24209857	14.5050077559442\\
1.262012243	14.5200077479318\\
1.282166004	14.5250077452647\\
1.30210948	14.535007739936\\
1.322170496	14.5450077346147\\
1.342099428	14.5550077293006\\
1.362005711	14.5600137362573\\
1.382188082	14.5700077213432\\
1.402091265	14.5800034293549\\
1.422305346	14.5900034270044\\
1.442149401	14.6000008561644\\
1.462014198	14.6050008558713\\
1.482145309	14.6200008549931\\
1.502088785	14.6300008544087\\
1.522160292	14.6350008541168\\
1.542102575	14.645\\
1.562010288	14.655\\
1.582163095	14.665\\
1.602073908	14.67\\
1.622177124	14.68\\
1.642151356	14.69\\
1.662014008	14.695\\
1.682134151	14.705000850051\\
1.70217967	14.7150008494733\\
1.722165823	14.7250008488964\\
1.742109299	14.73\\
1.762005329	14.74\\
1.782206297	14.75\\
1.802084208	14.7550008471704\\
1.822207689	14.765\\
1.842102051	14.7700008463101\\
1.862024307	14.7750008460237\\
1.882204294	14.7850033818055\\
1.902106047	14.7900033806622\\
1.922158003	14.8000076013494\\
1.942089081	14.8050075987822\\
1.962026119	14.8100075962168\\
1.982143164	14.8200075910912\\
2.00207901	14.8300075859724\\
2.022226334	14.8350134816252\\
2.042124987	14.845021050844\\
2.062014103	14.8500210437561\\
2.082222939	14.8550210366731\\
2.102130413	14.8650134544171\\
2.122152328	14.870013449893\\
2.142102718	14.8750210083885\\
2.162024736	14.8800210013293\\
2.182160616	14.8850209942747\\
2.20205164	14.890020987225\\
2.222170115	14.9000134228127\\
2.242100239	14.9050075478008\\
2.262027979	14.9100075452697\\
2.282144308	14.9150075427403\\
2.302100658	14.9200075402126\\
2.322163582	14.9250075376865\\
2.342063904	14.9300075351622\\
2.361994982	14.9350075326395\\
2.382143974	14.9400075301186\\
2.40207243	14.9450075275993\\
2.42218256	14.9500075250817\\
2.442131042	14.9550133734477\\
2.46202755	14.9550133734477\\
2.48217392	14.9600075200516\\
2.502112865	14.965007517539\\
2.522167683	14.9700075150282\\
2.542099714	14.9800033377833\\
2.562043667	14.9800075100115\\
2.58217597	14.9850075075056\\
2.602068663	14.995007502499\\
2.622182608	14.995007502499\\
2.6420753	15.000003333333\\
2.662009001	15.0050033322222\\
2.682165384	15.0100074950015\\
2.702100515	15.0150074925056\\
2.722188473	15.0200074900114\\
2.742105484	15.0250074875189\\
2.76200366	15.0300033266796\\
2.782128811	15.0350008313934\\
2.80207634	15.040000831117\\
2.822188854	15.0450033233629\\
2.842112541	15.0450033233629\\
2.862010002	15.0500033222588\\
2.882200003	15.0500033222588\\
2.902073622	15.0550008302889\\
2.922177553	15.06\\
2.942096472	15.06\\
2.962039709	15.0650008297378\\
2.982153654	15.0700008294625\\
3.002112389	15.0700008294625\\
3.022202015	15.075\\
3.042122364	15.075\\
3.062047958	15.08\\
3.082171917	15.085\\
3.102081537	15.0850008286377\\
3.122168303	15.0900008283631\\
3.142112732	15.0900008283631\\
3.162014723	15.0950008280887\\
3.182135105	15.1000008278145\\
3.20213604	15.1000008278145\\
3.222180128	15.1050033101618\\
3.242101908	15.1050033101618\\
3.262006998	15.1100033090665\\
3.282162428	15.1100033090665\\
3.302073717	15.1150033079719\\
3.322144032	15.1150033079719\\
3.342067957	15.1200074404744\\
3.362023592	15.1200132275074\\
3.382169962	15.1250132231347\\
3.402055264	15.1250206611429\\
3.422219515	15.1300206543151\\
3.44207406	15.1350132143979\\
3.462005854	15.1300132187649\\
3.482167482	15.1350132143979\\
3.502092838	15.1400074306455\\
3.522217512	15.1450297127473\\
3.542165518	15.1450033014192\\
3.56202507	15.1450297127473\\
3.582141399	15.1500074257408\\
3.602095366	15.155013196959\\
3.622175932	15.1600131926064\\
3.642080307	15.1600074208425\\
3.662027359	15.1600074208425\\
3.682184696	15.1600032981527\\
3.702087164	15.1600032981527\\
3.722172022	15.1650008242664\\
3.742079735	15.1650008242664\\
3.7620399	15.1650008242664\\
3.782176971	15.1700008239947\\
3.802106619	15.1650008242664\\
3.822199345	15.17\\
3.842080116	15.175\\
3.861995697	15.1750008237232\\
3.882146597	15.175\\
3.902102232	15.175\\
3.922180891	15.1800008234519\\
3.942057133	15.18\\
3.962022543	15.185\\
3.982133865	15.185\\
4.002083778	15.185\\
4.022172689	15.19\\
4.042096615	15.1900008229098\\
4.06201911	15.1900008229098\\
4.082199574	15.1900008229098\\
4.102113008	15.19\\
4.122160435	15.195000822639\\
4.142089844	15.195\\
4.162051678	15.195000822639\\
4.182180882	15.2000008223684\\
4.202055454	15.205000822098\\
4.222214699	15.2100032873106\\
4.242144346	15.2100032873106\\
4.262007713	15.2150032862303\\
4.282158852	15.2150032862303\\
4.302107573	15.2200032851508\\
4.322161913	15.2250032840719\\
4.342093706	15.2300008207485\\
4.362052917	15.2350032819163\\
4.38219142	15.2350008204791\\
4.402065992	15.235\\
4.422214985	15.24\\
4.442141771	15.2450008199409\\
4.462005138	15.2500032786882\\
4.48217988	15.2500032786882\\
4.502103329	15.25501311045\\
4.522183895	15.2600131061543\\
4.542001963	15.2650294791723\\
4.562021494	15.2700401112767\\
4.582141638	15.2750662846352\\
4.601983547	15.2800818060637\\
4.622296572	15.2800989852815\\
4.642003298	15.2850989529018\\
4.662042141	15.2900989205433\\
4.68221283	15.2951176850654\\
4.702038527	15.3051176081728\\
4.722152233	15.3100987913207\\
4.742001772	15.3150987590678\\
4.762083292	15.320081592472\\
4.782161951	15.3250660683731\\
4.802028894	15.3350660252899\\
4.822190285	15.3400660037693\\
4.841997147	15.3450659822628\\
4.862009764	15.3500521171754\\
4.882171869	15.3550521002047\\
4.902021646	15.365039863274\\
4.922164917	15.3750398373468\\
4.942040443	15.3800398243958\\
4.962023258	15.390039798519\\
4.982138634	15.4000292207515\\
5.002011061	15.4100397468663\\
5.022208452	15.4200397210902\\
5.042009592	15.4300396953475\\
5.061990499	15.4400518133846\\
5.082176208	15.4500655338416\\
5.102002144	15.4600654914525\\
5.122169971	15.4700654491182\\
5.142003059	15.4800654068386\\
5.162042618	15.4900516461373\\
5.182177782	15.5000516128173\\
5.20200038	15.5100515795403\\
5.222169161	15.5200515463062\\
5.242000103	15.5300515131148\\
5.262005568	15.5350514965352\\
5.282134056	15.5500514468602\\
5.302009821	15.5600393637034\\
5.322175264	15.5600393637034\\
5.342041731	15.5650393510585\\
5.362050533	15.575039325793\\
5.382157087	15.5800393131725\\
5.402051449	15.5850288738905\\
5.422196627	15.5900288646301\\
5.442041874	15.6000288461272\\
5.462004662	15.6050288368846\\
5.482170343	15.6150200127954\\
5.502006054	15.6200200063892\\
5.522168398	15.6300127959001\\
5.542014837	15.635012791808\\
5.562004805	15.6450127836317\\
5.58215785	15.6450127836317\\
5.601986408	15.6500127795475\\
5.622159719	15.6600071839064\\
5.642006874	15.6700071793219\\
5.661997795	15.6750031897923\\
5.682132959	15.6800031887752\\
5.701995373	15.6900031867428\\
5.72217679	15.7000031847131\\
5.742019892	15.7050031836991\\
5.762010574	15.7150031816732\\
5.782159567	15.7300007946599\\
5.801992416	15.7350007944074\\
5.822404146	15.745003175611\\
5.842020988	15.7500031746029\\
5.862004519	15.7600031725885\\
5.882167339	15.7700031705767\\
5.902027845	15.7800007921419\\
5.92217803	15.7900007916403\\
5.942015648	15.8000007911392\\
5.96203351	15.8100007906388\\
5.982154131	15.8150007903889\\
6.00202179	15.8250007898894\\
6.022155046	15.8250007898894\\
6.042008638	15.84\\
6.062045097	15.845\\
6.082176447	15.8500007886435\\
6.102030754	15.86\\
6.122168303	15.8650007878979\\
6.142015219	15.8750007874016\\
6.162034512	15.88\\
6.182168484	15.8900031466328\\
6.202008486	15.9000007861635\\
6.22219348	15.9000007861635\\
6.242008209	15.9050007859164\\
6.261991501	15.9100007856694\\
6.2821877	15.915\\
6.302003145	15.925\\
6.322138309	15.9350007844368\\
6.341986656	15.94\\
6.362041235	15.95\\
6.382148266	15.960000783208\\
6.402024269	15.9650007829627\\
6.42219305	15.9700031308701\\
6.442016125	15.9800031289108\\
6.462002277	15.9850070378464\\
6.482141256	15.9950070334464\\
6.501999855	16.0000070312485\\
6.522139788	16.0050070290519\\
6.542096376	16.0100070268567\\
6.562011242	16.0200124843896\\
6.582169533	16.0300124766015\\
6.602061272	16.0350124727111\\
6.622207403	16.0400070137142\\
6.642106295	16.0500070093443\\
6.662022114	16.0550070071613\\
6.682183027	16.0600031133247\\
6.702072859	16.0650031123558\\
6.722159863	16.0700007778469\\
6.742056131	16.0800007773632\\
6.762032509	16.09000077688\\
6.782183886	16.09\\
6.802063704	16.0950007766387\\
6.822175264	16.1100007759156\\
6.842110157	16.1150007756748\\
6.861999989	16.1200007754342\\
6.882146597	16.1200007754342\\
6.9020679	16.1200007754342\\
6.922154188	16.1300007749535\\
6.942060471	16.1350007747133\\
6.962014437	16.1400007744733\\
6.98218441	16.1500007739938\\
7.002091169	16.1550007737542\\
7.022185087	16.1600007735148\\
7.042077065	16.1650030931021\\
7.062018156	16.1700069573269\\
7.082176685	16.1750030911898\\
7.102084875	16.1800030902346\\
7.122146368	16.1900030883258\\
7.142120361	16.1950030873724\\
7.162061453	16.195006946587\\
7.182176352	16.200006944443\\
7.202130079	16.2100069401589\\
7.222165823	16.2150069380189\\
7.242084503	16.2200069358801\\
7.262024403	16.2250123266517\\
7.282145262	16.2300069316067\\
7.302088261	16.2300123228542\\
7.322201729	16.2350123190591\\
7.342108727	16.2400123152663\\
7.362012625	16.2450123114758\\
7.382150888	16.2500123076876\\
7.402046919	16.2500192307579\\
7.42216754	16.2550192248425\\
7.442087412	16.2600276752532\\
7.462014914	16.2650376575033\\
7.482140779	16.2650376575033\\
7.502044916	16.2700491701777\\
7.522149324	16.2750768047343\\
7.542014122	16.2800929051403\\
7.562052011	16.2851105307885\\
7.582183599	16.2901104968628\\
7.60199666	16.2901104968628\\
7.622199535	16.3001104290738\\
7.6420784	16.295092819619\\
7.66203475	16.3050927626923\\
7.682171345	16.3050766634199\\
7.702017784	16.310049049589\\
7.722157955	16.3100375536048\\
7.74199152	16.3150191541414\\
7.762046337	16.3200122548974\\
7.78216362	16.3200030637252\\
7.802079916	16.3250007656968\\
7.822324038	16.3300007654623\\
7.842059135	16.3300030618491\\
7.862033606	16.3350068870509\\
7.882151604	16.3400122398975\\
7.902106762	16.3400122398975\\
7.922181845	16.3450068828373\\
7.942075253	16.3500030581037\\
7.96202302	16.3550030571688\\
7.982138634	16.3550007642922\\
8.002056837	16.36\\
8.022184849	16.3600030562344\\
8.042066336	16.3650068744257\\
8.061996698	16.3650122211992\\
8.082165003	16.3700122174664\\
8.10209918	16.3700122174664\\
8.12220192	16.3750122137359\\
8.142128468	16.3750122137359\\
8.162009001	16.3750190839583\\
8.182178974	16.380019078133\\
8.202082872	16.3850122062817\\
8.2221632	16.3850122062817\\
8.242114782	16.39000686394\\
8.261997461	16.390012202558\\
8.282184362	16.39500304971\\
8.302082777	16.39500304971\\
8.322158575	16.4000030487802\\
8.342079163	16.4000007621951\\
8.362035036	16.4050007619628\\
8.382153511	16.4050007619628\\
8.402086258	16.41\\
8.422154903	16.41\\
8.442137957	16.4100007617306\\
8.462030649	16.4150030459942\\
8.482141018	16.4150030459942\\
8.50206995	16.4150030459942\\
8.522186756	16.4150007614986\\
8.542097092	16.4150030459942\\
8.562069654	16.4200007612667\\
8.582162857	16.4200007612667\\
8.602098942	16.425000761035\\
8.622184515	16.42\\
8.642097235	16.425000761035\\
8.662009954	16.4250030441397\\
8.682184458	16.4300030432134\\
8.70208025	16.4300030432134\\
8.722199678	16.4300068472293\\
8.742061377	16.4250121765556\\
8.76201725	16.4300190200742\\
8.782147169	16.4350273805674\\
8.802074194	16.4350273805674\\
8.822186708	16.4350273805674\\
8.842088938	16.4350273805674\\
8.86203146	16.4350372679833\\
8.882168055	16.4400372566488\\
8.902080774	16.4400372566488\\
8.92218399	16.4400372566488\\
8.942073584	16.4400372566488\\
8.962017059	16.4450372453212\\
8.982139111	16.4450372453212\\
9.002076387	16.4400372566488\\
9.022206306	16.4450273639176\\
9.042067289	16.4450273639176\\
9.062023401	16.4450121617468\\
9.082197666	16.4450121617468\\
9.102116108	16.4450121617468\\
9.122168541	16.4450121617468\\
9.142102718	16.4450121617468\\
9.16201067	16.4450068409837\\
9.182157755	16.4450030404375\\
9.202145576	16.4450030404375\\
9.222205639	16.4450030404375\\
9.242083549	16.4450030404375\\
9.262011051	16.4500030395134\\
9.282196045	16.4500030395134\\
9.302088022	16.4500030395134\\
9.322224855	16.4500030395134\\
9.342083454	16.4500030395134\\
9.362039328	16.4500068389044\\
9.382145166	16.4550068368263\\
9.402059555	16.4550121543559\\
9.422160625	16.4550068368263\\
9.442074537	16.4550068368263\\
9.462022066	16.4600068347495\\
9.482147932	16.4550068368263\\
9.50248909	16.4600121506638\\
9.522160769	16.4600121506638\\
9.542102814	16.465012146974\\
9.56207037	16.465012146974\\
9.582131863	16.465012146974\\
9.602108955	16.4600189854083\\
9.622174025	16.4650189796429\\
9.64208889	16.4650189796429\\
9.662001371	16.4650273306788\\
9.682150602	16.4650189796429\\
9.702144861	16.4650189796429\\
9.722158909	16.4650189796429\\
9.742101669	16.4650273306788\\
9.762069225	16.4700273223817\\
9.782200336	16.470018973881\\
9.802125454	16.470018973881\\
9.822184563	16.470018973881\\
9.84209919	16.470018973881\\
9.862032652	16.470018973881\\
9.882185698	16.4700121432864\\
9.902075529	16.475012139601\\
9.922151327	16.4750068285267\\
9.942129374	16.4750030349011\\
9.96201849	16.4750030349011\\
9.982164145	16.4750030349011\\
10.002075434	16.4750030349011\\
10.022153854	16.4750030349011\\
10.04221487	16.4750030349011\\
10.062048197	16.4750030349011\\
10.082187891	16.4750030349011\\
10.102066994	16.4750068285267\\
10.122208834	16.4750068285267\\
10.142107248	16.4800121359179\\
10.162016392	16.4800121359179\\
10.182190657	16.4800121359179\\
10.202092886	16.4800121359179\\
10.222184658	16.4800189623677\\
10.242083073	16.4800189623677\\
10.262024164	16.4800189623677\\
10.282155752	16.4800189623677\\
10.302184582	16.4800189623677\\
10.322179794	16.4800189623677\\
10.342079401	16.4800273058026\\
10.362017393	16.4750273140897\\
10.382154226	16.4800273058026\\
10.402059078	16.4800189623677\\
10.42215991	16.4800189623677\\
10.442088127	16.4800189623677\\
10.462037086	16.4750273140897\\
10.482158422	16.4800273058026\\
10.502085447	16.4800273058026\\
10.522191048	16.4800273058026\\
10.542074203	16.4800273058026\\
10.562060118	16.4800189623677\\
10.582148314	16.4800273058026\\
10.60208869	16.4800189623677\\
10.622238636	16.4800189623677\\
10.642085791	16.4800121359179\\
10.662041664	16.4800189623677\\
10.682162285	16.485012132237\\
10.70212698	16.485012132237\\
10.722177505	16.485012132237\\
10.742103338	16.485012132237\\
10.76202035	16.4900121285583\\
10.782142401	16.4900121285583\\
10.802069664	16.4900121285583\\
10.822172403	16.4900121285583\\
10.842101336	16.4900121285583\\
10.862004042	16.4950121248819\\
10.882152796	16.4900068223151\\
10.902067661	16.4900068223151\\
10.922169685	16.4900121285583\\
10.942084789	16.4900068223151\\
10.962040186	16.4950121248819\\
10.982153654	16.4950121248819\\
11.002078533	16.4950121248819\\
11.022383213	16.4900121285583\\
11.042106867	16.4900121285583\\
11.062020779	16.4900121285583\\
11.08216548	16.4900189508684\\
11.102080345	16.4900189508684\\
11.122196198	16.4900272892436\\
11.142080545	16.4850189566163\\
11.162013292	16.4900272892436\\
11.182190418	16.4900272892436\\
11.202085495	16.4900371436816\\
11.222196817	16.4900371436816\\
11.242108583	16.4900371436816\\
11.261999369	16.4850371549475\\
11.282183409	16.4850371549475\\
11.302100897	16.4850371549475\\
11.32214427	16.4850371549475\\
11.342061758	16.4900272892436\\
11.362005711	16.4900272892436\\
11.382149696	16.4850272975206\\
11.402094126	16.4850189566163\\
11.422182083	16.4900272892436\\
11.442009449	16.4900189508684\\
11.462024927	16.4900189508684\\
11.482139349	16.4900272892436\\
11.502108097	16.4900189508684\\
11.522141933	16.4900189508684\\
11.541990042	16.4900121285583\\
11.56203866	16.4900189508684\\
11.582138538	16.4900189508684\\
11.602074623	16.4900189508684\\
11.622191191	16.4900189508684\\
11.642078638	16.4900189508684\\
11.662010193	16.4900189508684\\
11.682131052	16.4950272809717\\
11.702092648	16.4950272809717\\
11.722218513	16.4950272809717\\
11.742126942	16.4900189508684\\
11.762035608	16.4900189508684\\
11.782173157	16.4900189508684\\
11.802094936	16.4900189508684\\
11.822187424	16.4900189508684\\
11.842082024	16.4900272892436\\
11.86200428	16.4900189508684\\
11.882132292	16.4850272975206\\
11.902098179	16.4850189566163\\
11.922159433	16.4850272975206\\
11.942083597	16.4900272892436\\
11.962034941	16.4900371436816\\
11.982154131	16.4900371436816\\
12.002068758	16.4900371436816\\
12.022183418	16.4850371549475\\
12.042089939	16.4900371436816\\
12.062016249	16.4850485288943\\
12.082199812	16.4900485141797\\
12.102077246	16.4900485141797\\
12.12219286	16.4900371436816\\
12.142095089	16.4850371549475\\
12.162034035	16.4850485288943\\
12.182173967	16.4850371549475\\
12.202055454	16.4900371436816\\
12.222174883	16.4850272975206\\
12.242062807	16.4850272975206\\
12.262057781	16.4850189566163\\
12.282171249	16.4850189566163\\
12.302083969	16.485012132237\\
12.322149754	16.485012132237\\
12.342099667	16.4850068243844\\
12.362051249	16.4850068243844\\
12.382143021	16.4850068243844\\
12.402063608	16.4850068243844\\
12.422169447	16.4850068243844\\
12.442089319	16.4850068243844\\
12.46199894	16.4850068243844\\
};
\addplot [color=black, forget plot]
  table[row sep=crcr]{%
3.29999995231628	0\\
3.29999995231628	30\\
};
\node[right, align=left, circle,fill=white,draw=black,thick,line width=0.3mm,inner sep=1pt]
at (axis cs:2.2,22) {1};
\addplot [color=black, forget plot]
  table[row sep=crcr]{%
7.59999990463257	0\\
7.59999990463257	30\\
};
\node[right, align=left, circle,fill=white,draw=black,thick,line width=0.3mm,inner sep=1pt]
at (axis cs:6.5,22) {2};
\addplot [color=black, forget plot]
  table[row sep=crcr]{%
11.4400000572205	0\\
11.4400000572205	30\\
};
\node[right, align=left, circle,fill=white,draw=black,thick,line width=0.3mm,inner sep=1pt]
at (axis cs:10.34,22) {3};
\end{axis}
\end{tikzpicture}%

%% file: Images/Tikz/fig9-3-2.tikz
\begin{tikzpicture}

\begin{axis}[%
width=4.25cm,
height=1.683cm,
at={(0cm,0cm)},
scale only axis,
xmin=0.642011881,
xmax=12.46199894,
xlabel style={font=\color{white!15!black}},
xlabel={$t\;[\SI[per-mode=repeated-symbol]{}{\second}]$},
ymin=-1,
ymax=1.5,
ylabel style={font=\color{white!15!black}},
ylabel={$a\;[\SI[per-mode=repeated-symbol]{}{\metre\per\second\squared}]$},
axis background/.style={fill=white},
ylabel style={yshift=-0.4cm}
]
\addplot [color=blue, line width=1.0pt, forget plot]
  table[row sep=crcr]{%
0.642011881	0.541512\\
0.662031174	0.525816\\
0.682033062	0.702396\\
0.702054262	0.729864\\
0.722006321	0.600372\\
0.742020607	0.537588\\
0.762029409	0.635688\\
0.782009363	0.788724\\
0.802028418	0.66708\\
0.822096586	0.533664\\
0.842060328	0.56898\\
0.862061977	0.604296\\
0.882035017	0.671004\\
0.902044773	0.541512\\
0.922039747	0.604296\\
0.941986799	0.671004\\
0.962060928	0.557208\\
0.982016802	0.533664\\
1.001998186	0.5886\\
1.0225811	0.635688\\
1.041987896	0.773028\\
1.062039852	0.580752\\
1.082021952	0.439488\\
1.102125168	0.45126\\
1.122168303	0.576828\\
1.142053843	0.72594\\
1.162023783	0.671004\\
1.182157993	0.466956\\
1.202011108	0.306072\\
1.222189903	0.514044\\
1.24209857	0.592524\\
1.262012243	0.663156\\
1.282166004	0.329616\\
1.30210948	0.498348\\
1.322170496	0.443412\\
1.342099428	0.466956\\
1.362005711	0.388476\\
1.382188082	0.443412\\
1.402091265	0.486576\\
1.422305346	0.486576\\
1.442149401	0.423792\\
1.462014198	0.4905\\
1.482145309	0.525816\\
1.502088785	0.482652\\
1.522160292	0.408096\\
1.542102575	0.537588\\
1.562010288	0.439488\\
1.582163095	0.384552\\
1.602073908	0.33354\\
1.622177124	0.435564\\
1.642151356	0.478728\\
1.662014008	0.43164\\
1.682134151	0.33354\\
1.70217967	0.525816\\
1.722165823	0.455184\\
1.742109299	0.482652\\
1.762005329	0.43164\\
1.782206297	0.419868\\
1.802084208	0.345312\\
1.822207689	0.415944\\
1.842102051	0.376704\\
1.862024307	0.251136\\
1.882204294	0.317844\\
1.902106047	0.317844\\
1.922158003	0.478728\\
1.942089081	0.364932\\
1.962026119	0.17658\\
1.982143164	0.435564\\
2.00207901	0.4905\\
2.022226334	0.435564\\
2.042124987	0.317844\\
2.062014103	0.298224\\
2.082222939	0.376704\\
2.102130413	0.439488\\
2.122152328	0.298224\\
2.142102718	0.286452\\
2.162024736	0.23544\\
2.182160616	0.258984\\
2.20205164	0.204048\\
2.222170115	0.341388\\
2.242100239	0.302148\\
2.262027979	0.278604\\
2.282144308	0.31392\\
2.302100658	0.204048\\
2.322163582	0.262908\\
2.342063904	0.172656\\
2.361994982	0.278604\\
2.382143974	0.258984\\
2.40207243	0.239364\\
2.42218256	0.200124\\
2.442131042	0.231516\\
2.46202755	0.160884\\
2.48217392	0.204048\\
2.502112865	0.25506\\
2.522167683	0.290376\\
2.542099714	0.325692\\
2.562043667	0.17658\\
2.58217597	0.286452\\
2.602068663	0.27468\\
2.622182608	0.204048\\
2.6420753	0.219744\\
2.662009001	0.262908\\
2.682165384	0.23544\\
2.702100515	0.180504\\
2.722188473	0.21582\\
2.742105484	0.262908\\
2.76200366	0.258984\\
2.782128811	0.258984\\
2.80207634	0.168732\\
2.822188854	0.1962\\
2.842112541	0.145188\\
2.862010002	0.188352\\
2.882200003	0.070632\\
2.902073622	0.258984\\
2.922177553	0.204048\\
2.942096472	0.168732\\
2.962039709	0.17658\\
2.982153654	0.113796\\
3.002112389	0.094176\\
3.022202015	0.192276\\
3.042122364	0.121644\\
3.062047958	0.15696\\
3.082171917	0.125568\\
3.102081537	0.090252\\
3.122168303	0.223668\\
3.142112732	0.184428\\
3.162014723	0.251136\\
3.182135105	0.086328\\
3.20213604	0.149112\\
3.222180128	0.07848\\
3.242101908	0.113796\\
3.262006998	0.266832\\
3.282162428	0.003924\\
3.302073717	0.172656\\
3.322144032	0.090252\\
3.342067957	0.21582\\
3.362023592	0.102024\\
3.382169962	0.086328\\
3.402055264	0.153036\\
3.422219515	0.192276\\
3.44207406	0.113796\\
3.462005854	-0.05886\\
3.482167482	0.168732\\
3.502092838	0.2943\\
3.522217512	0.357084\\
3.542165518	-0.1962\\
3.56202507	0.207972\\
3.582141399	0.15696\\
3.602095366	0.188352\\
3.622175932	0.270756\\
3.642080307	-0.047088\\
3.662027359	-0.003924\\
3.682184696	0.090252\\
3.702087164	0.153036\\
3.722172022	-0.011772\\
3.742079735	0.074556\\
3.7620399	0.145188\\
3.782176971	0.070632\\
3.802106619	-0.03924\\
3.822199345	0.11772\\
3.842080116	0.125568\\
3.861995697	0.207972\\
3.882146597	-0.043164\\
3.902102232	0.003924\\
3.922180891	0.054936\\
3.942057133	0.172656\\
3.962022543	0.121644\\
3.982133865	0.015696\\
4.002083778	0.051012\\
4.022172689	0.168732\\
4.042096615	-0.011772\\
4.06201911	0.113796\\
4.082199574	-0.031392\\
4.102113008	0.149112\\
4.122160435	0.102024\\
4.142089844	0.21582\\
4.162051678	-0.023544\\
4.182180882	0.243288\\
4.202055454	0.160884\\
4.222214699	0.153036\\
4.242144346	0.13734\\
4.262007713	0.180504\\
4.282158852	0.125568\\
4.302107573	0.239364\\
4.322161913	0.133416\\
4.342093706	0.204048\\
4.362052917	0.145188\\
4.38219142	0.149112\\
4.402065992	0.133416\\
4.422214985	0.15696\\
4.442141771	0.223668\\
4.462005138	0.145188\\
4.48217988	0.192276\\
4.502103329	0.17658\\
4.522183895	0.211896\\
4.542001963	0.247212\\
4.562021494	0.1962\\
4.582141638	0.262908\\
4.601983547	0.1962\\
4.622296572	0.207972\\
4.642003298	0.243288\\
4.662042141	0.184428\\
4.68221283	0.27468\\
4.702038527	0.345312\\
4.722152233	0.337464\\
4.742001772	0.286452\\
4.762083292	0.290376\\
4.782161951	0.247212\\
4.802028894	0.376704\\
4.822190285	0.345312\\
4.841997147	0.404172\\
4.862009764	0.270756\\
4.882171869	0.223668\\
4.902021646	0.43164\\
4.922164917	0.345312\\
4.942040443	0.47088\\
4.962023258	0.400248\\
4.982138634	0.423792\\
5.002011061	0.415944\\
5.022208452	0.52974\\
5.042009592	0.47088\\
5.061990499	0.521892\\
5.082176208	0.502272\\
5.102002144	0.506196\\
5.122169971	0.408096\\
5.142003059	0.51012\\
5.162042618	0.474804\\
5.182177782	0.572904\\
5.20200038	0.486576\\
5.222169161	0.463032\\
5.242000103	0.341388\\
5.262005568	0.427716\\
5.282134056	0.557208\\
5.302009821	0.349236\\
5.322175264	0.278604\\
5.342041731	0.364932\\
5.362050533	0.357084\\
5.382157087	0.258984\\
5.402051449	0.482652\\
5.422196627	0.364932\\
5.442041874	0.415944\\
5.462004662	0.302148\\
5.482170343	0.357084\\
5.502006054	0.27468\\
5.522168398	0.419868\\
5.542014837	0.443412\\
5.562004805	0.345312\\
5.58215785	0.306072\\
5.601986408	0.227592\\
5.622159719	0.494424\\
5.642006874	0.376704\\
5.661997795	0.384552\\
5.682132959	0.361008\\
5.701995373	0.3924\\
5.72217679	0.364932\\
5.742019892	0.309996\\
5.762010574	0.494424\\
5.782159567	0.537588\\
5.801992416	0.325692\\
5.822404146	0.439488\\
5.842020988	0.278604\\
5.862004519	0.576828\\
5.882167339	0.349236\\
5.902027845	0.486576\\
5.92217803	0.349236\\
5.942015648	0.459108\\
5.96203351	0.537588\\
5.982154131	0.278604\\
6.00202179	0.455184\\
6.022155046	0.192276\\
6.042008638	0.5886\\
6.062045097	0.47088\\
6.082176447	0.380628\\
6.102030754	0.282528\\
6.122168303	0.494424\\
6.142015219	0.380628\\
6.162034512	0.380628\\
6.182168484	0.364932\\
6.202008486	0.423792\\
6.22219348	0.278604\\
6.242008209	0.302148\\
6.261991501	0.290376\\
6.2821877	0.251136\\
6.302003145	0.514044\\
6.322138309	0.427716\\
6.341986656	0.302148\\
6.362041235	0.357084\\
6.382148266	0.47088\\
6.402024269	0.380628\\
6.42219305	0.243288\\
6.442016125	0.25506\\
6.462002277	0.364932\\
6.482141256	0.361008\\
6.501999855	0.3924\\
6.522139788	0.192276\\
6.542096376	0.364932\\
6.562011242	0.35316\\
6.582169533	0.478728\\
6.602061272	0.27468\\
6.622207403	0.306072\\
6.642106295	0.380628\\
6.662022114	0.329616\\
6.682183027	0.321768\\
6.702072859	0.219744\\
6.722159863	0.349236\\
6.742056131	0.380628\\
6.762032509	0.396324\\
6.782183886	0.15696\\
6.802063704	0.27468\\
6.822175264	0.486576\\
6.842110157	0.376704\\
6.861999989	0.25506\\
6.882146597	0.298224\\
6.9020679	0.13734\\
6.922154188	0.400248\\
6.942060471	0.223668\\
6.962014437	0.23544\\
6.98218441	0.153036\\
7.002091169	0.415944\\
7.022185087	0.207972\\
7.042077065	0.286452\\
7.062018156	0.223668\\
7.082176685	0.266832\\
7.102084875	0.2943\\
7.122146368	0.337464\\
7.142120361	0.247212\\
7.162061453	0.207972\\
7.182176352	0.204048\\
7.202130079	0.341388\\
7.222165823	0.262908\\
7.242084503	0.239364\\
7.262024403	0.278604\\
7.282145262	0.17658\\
7.302088261	0.180504\\
7.322201729	0.282528\\
7.342108727	0.282528\\
7.362012625	0.133416\\
7.382150888	0.168732\\
7.402046919	0.027468\\
7.42216754	0.290376\\
7.442087412	0.17658\\
7.462014914	0.282528\\
7.482140779	0.172656\\
7.502044916	0.109872\\
7.522149324	0.309996\\
7.542014122	0.168732\\
7.562052011	0.270756\\
7.582183599	0.262908\\
7.60199666	0.129492\\
7.622199535	0.325692\\
7.6420784	-0.01962\\
7.66203475	0.2943\\
7.682171345	0.027468\\
7.702017784	0.31392\\
7.722157955	-0.003924\\
7.74199152	0.219744\\
7.762046337	0.219744\\
7.78216362	0.17658\\
7.802079916	0.15696\\
7.822324038	0.25506\\
7.842059135	0.051012\\
7.862033606	0.200124\\
7.882151604	0.219744\\
7.902106762	0.121644\\
7.922181845	0.109872\\
7.942075253	0.239364\\
7.96202302	0.204048\\
7.982138634	0.0981\\
8.002056837	0.15696\\
8.022184849	0.074556\\
8.042066336	0.188352\\
8.061996698	0.180504\\
8.082165003	0.113796\\
8.10209918	0.102024\\
8.12220192	0.074556\\
8.142128468	0.062784\\
8.162009001	0.141264\\
8.182178974	0.133416\\
8.202082872	0.15696\\
8.2221632	0.145188\\
8.242114782	0.204048\\
8.261997461	-0.043164\\
8.282184362	0.266832\\
8.302082777	0.102024\\
8.322158575	0.160884\\
8.342079163	0.035316\\
8.362035036	0.2943\\
8.382153511	-0.007848\\
8.402086258	0.184428\\
8.422154903	-0.01962\\
8.442137957	0.03924\\
8.462030649	0.17658\\
8.482141018	0.074556\\
8.50206995	0.054936\\
8.522186756	0.03924\\
8.542097092	0.027468\\
8.562069654	0.051012\\
8.582162857	0.164808\\
8.602098942	0.109872\\
8.622184515	-0.062784\\
8.642097235	0.051012\\
8.662009954	0.13734\\
8.682184458	0.062784\\
8.70208025	0.023544\\
8.722199678	0.066708\\
8.742061377	-0.07848\\
8.76201725	0.149112\\
8.782147169	0.15696\\
8.802074194	0.007848\\
8.822186708	0.011772\\
8.842088938	0.062784\\
8.86203146	0.094176\\
8.882168055	0.121644\\
8.902080774	0.035316\\
8.92218399	0.051012\\
8.942073584	0.003924\\
8.962017059	0.054936\\
8.982139111	0.047088\\
9.002076387	-0.051012\\
9.022206306	0.086328\\
9.042067289	0.011772\\
9.062023401	-0.027468\\
9.082197666	-0.035316\\
9.102116108	0.023544\\
9.122168541	-0.015696\\
9.142102718	0.113796\\
9.16201067	0.01962\\
9.182157755	-0.047088\\
9.202145576	-0.027468\\
9.222205639	0.17658\\
9.242083549	0.027468\\
9.262011051	0.05886\\
9.282196045	0.027468\\
9.302088022	0.05886\\
9.322224855	-0.035316\\
9.342083454	0.05886\\
9.362039328	-0.007848\\
9.382145166	0.082404\\
9.402059555	0.007848\\
9.422160625	0.074556\\
9.442074537	0.070632\\
9.462022066	0.070632\\
9.482147932	-0.031392\\
9.50248909	0.1962\\
9.522160769	0.031392\\
9.542102814	0.113796\\
9.56207037	-0.023544\\
9.582131863	0.074556\\
9.602108955	-0.102024\\
9.622174025	0.082404\\
9.64208889	0.102024\\
9.662001371	-0.051012\\
9.682150602	0\\
9.702144861	0.043164\\
9.722158909	-0.07848\\
9.742101669	0.145188\\
9.762069225	0.054936\\
9.782200336	0.035316\\
9.802125454	0.031392\\
9.822184563	0.102024\\
9.84209919	-0.035316\\
9.862032652	0.109872\\
9.882185698	-0.023544\\
9.902075529	0.070632\\
9.922151327	-0.062784\\
9.942129374	0.015696\\
9.96201849	-0.031392\\
9.982164145	0.011772\\
10.002075434	0.007848\\
10.022153854	0.0981\\
10.04221487	0.023544\\
10.062048197	0.094176\\
10.082187891	-0.160884\\
10.102066994	0.105948\\
10.122208834	-0.01962\\
10.142107248	0.133416\\
10.162016392	-0.003924\\
10.182190657	0.07848\\
10.202092886	-0.011772\\
10.222184658	-0.031392\\
10.242083073	-0.01962\\
10.262024164	0.035316\\
10.282155752	-0.054936\\
10.302184582	0.145188\\
10.322179794	-0.125568\\
10.342079401	-0.007848\\
10.362017393	-0.066708\\
10.382154226	0.164808\\
10.402059078	0.007848\\
10.42215991	-0.07848\\
10.442088127	0.035316\\
10.462037086	-0.090252\\
10.482158422	0.090252\\
10.502085447	0.094176\\
10.522191048	-0.051012\\
10.542074203	-0.090252\\
10.562060118	0.090252\\
10.582148314	-0.003924\\
10.60208869	0.023544\\
10.622238636	0\\
10.642085791	0.047088\\
10.662041664	0.074556\\
10.682162285	0.121644\\
10.70212698	-0.051012\\
10.722177505	0.11772\\
10.742103338	0.051012\\
10.76202035	0.13734\\
10.782142401	-0.113796\\
10.802069664	0.15696\\
10.822172403	-0.105948\\
10.842101336	0.204048\\
10.862004042	0.027468\\
10.882152796	-0.011772\\
10.902067661	-0.047088\\
10.922169685	0.054936\\
10.942084789	-0.003924\\
10.962040186	0.113796\\
10.982153654	0.003924\\
11.002078533	-0.054936\\
11.022383213	-0.141264\\
11.042106867	-0.047088\\
11.062020779	-0.05886\\
11.08216548	0.105948\\
11.102080345	-0.0981\\
11.122196198	0.015696\\
11.142080545	-0.05886\\
11.162013292	0.03924\\
11.182190418	-0.047088\\
11.202085495	-0.007848\\
11.222196817	0.035316\\
11.242108583	0.023544\\
11.261999369	-0.066708\\
11.282183409	-0.043164\\
11.302100897	-0.102024\\
11.32214427	0.141264\\
11.342061758	0.03924\\
11.362005711	0\\
11.382149696	-0.094176\\
11.402094126	0.035316\\
11.422182083	0.129492\\
11.442009449	0.074556\\
11.462024927	0.043164\\
11.482139349	-0.074556\\
11.502108097	0.047088\\
11.522141933	-0.047088\\
11.541990042	-0.015696\\
11.56203866	-0.011772\\
11.582138538	0.086328\\
11.602074623	0.03924\\
11.622191191	0.035316\\
11.642078638	-0.007848\\
11.662010193	-0.07848\\
11.682131052	0.125568\\
11.702092648	-0.003924\\
11.722218513	0.086328\\
11.742126942	-0.086328\\
11.762035608	-0.102024\\
11.782173157	-0.070632\\
11.802094936	-0.01962\\
11.822187424	0.007848\\
11.842082024	0.145188\\
11.86200428	-0.074556\\
11.882132292	-0.11772\\
11.902098179	-0.047088\\
11.922159433	0.007848\\
11.942083597	0.074556\\
11.962034941	0.074556\\
11.982154131	-0.01962\\
12.002068758	0.043164\\
12.022183418	-0.133416\\
12.042089939	0.070632\\
12.062016249	-0.0981\\
12.082199812	0.102024\\
12.102077246	0.027468\\
12.12219286	-0.047088\\
12.142095089	-0.01962\\
12.162034035	-0.15696\\
12.182173967	0.141264\\
12.202055454	0.051012\\
12.222174883	-0.05886\\
12.242062807	-0.13734\\
12.262057781	0.066708\\
12.282171249	-0.051012\\
12.302083969	0.074556\\
12.322149754	0.062784\\
12.342099667	-0.113796\\
12.362051249	-0.043164\\
12.382143021	-0.03924\\
12.402063608	0.003924\\
12.422169447	0.031392\\
12.442089319	-0.051012\\
12.46199894	0.05886\\
};
\addplot [color=red, line width=1.0pt, forget plot]
  table[row sep=crcr]{%
0.642011881	-0.149112\\
0.662031174	-0.121644\\
0.682033062	-0.054936\\
0.702054262	-0.066708\\
0.722006321	-0.070632\\
0.742020607	0.11772\\
0.762029409	0.05886\\
0.782009363	0.05886\\
0.802028418	-0.090252\\
0.822096586	-0.062784\\
0.842060328	0.066708\\
0.862061977	-0.066708\\
0.882035017	-0.066708\\
0.902044773	-0.168732\\
0.922039747	-0.207972\\
0.941986799	-0.219744\\
0.962060928	-0.266832\\
0.982016802	-0.266832\\
1.001998186	0.074556\\
1.0225811	0.035316\\
1.041987896	-0.121644\\
1.062039852	-0.121644\\
1.082021952	-0.21582\\
1.102125168	-0.015696\\
1.122168303	-0.015696\\
1.142053843	0.003924\\
1.162023783	-0.074556\\
1.182157993	-0.192276\\
1.202011108	-0.125568\\
1.222189903	-0.180504\\
1.24209857	-0.180504\\
1.262012243	-0.070632\\
1.282166004	0.066708\\
1.30210948	0.015696\\
1.322170496	-0.17658\\
1.342099428	-0.17658\\
1.362005711	-0.180504\\
1.382188082	-0.180504\\
1.402091265	0.121644\\
1.422305346	-0.03924\\
1.442149401	-0.03924\\
1.462014198	-0.102024\\
1.482145309	0.01962\\
1.502088785	-0.066708\\
1.522160292	0.003924\\
1.542102575	0.01962\\
1.562010288	0.01962\\
1.582163095	0.007848\\
1.602073908	-0.145188\\
1.622177124	-0.145188\\
1.642151356	-0.07848\\
1.662014008	-0.07848\\
1.682134151	0.066708\\
1.70217967	-0.031392\\
1.722165823	-0.031392\\
1.742109299	0.031392\\
1.762005329	-0.219744\\
1.782206297	-0.05886\\
1.802084208	-0.011772\\
1.822207689	0.043164\\
1.842102051	0.043164\\
1.862024307	-0.168732\\
1.882204294	-0.113796\\
1.902106047	-0.192276\\
1.922158003	-0.141264\\
1.942089081	0.035316\\
1.962026119	0.035316\\
1.982143164	-0.102024\\
2.00207901	-0.153036\\
2.022226334	-0.153036\\
2.042124987	-0.113796\\
2.062014103	-0.113796\\
2.082222939	-0.023544\\
2.102130413	0.102024\\
2.122152328	0.102024\\
2.142102718	-0.125568\\
2.162024736	-0.125568\\
2.182160616	0.047088\\
2.20205164	0.121644\\
2.222170115	0.121644\\
2.242100239	0.17658\\
2.262027979	0.17658\\
2.282144308	0.074556\\
2.302100658	0.074556\\
2.322163582	-0.05886\\
2.342063904	0.043164\\
2.361994982	0.043164\\
2.382143974	0.043164\\
2.40207243	-0.086328\\
2.42218256	-0.086328\\
2.442131042	-0.05886\\
2.46202755	-0.066708\\
2.48217392	-0.066708\\
2.502112865	0.054936\\
2.522167683	0.149112\\
2.542099714	0.011772\\
2.562043667	0.011772\\
2.58217597	-0.141264\\
2.602068663	-0.035316\\
2.622182608	0.0981\\
2.6420753	0.066708\\
2.662009001	0.066708\\
2.682165384	-0.07848\\
2.702100515	0.015696\\
2.722188473	0.054936\\
2.742105484	0.070632\\
2.76200366	0.070632\\
2.782128811	0.192276\\
2.80207634	0.043164\\
2.822188854	-0.05886\\
2.842112541	-0.121644\\
2.862010002	0.133416\\
2.882200003	0.094176\\
2.902073622	0.094176\\
2.922177553	0.211896\\
2.942096472	0.211896\\
2.962039709	-0.125568\\
2.982153654	-0.011772\\
3.002112389	-0.011772\\
3.022202015	-0.03924\\
3.042122364	0.164808\\
3.062047958	-0.027468\\
3.082171917	0.074556\\
3.102081537	-0.251136\\
3.122168303	0.031392\\
3.142112732	-0.035316\\
3.162014723	-0.035316\\
3.182135105	0.074556\\
3.20213604	0.074556\\
3.222180128	-0.121644\\
3.242101908	-0.102024\\
3.262006998	-0.051012\\
3.282162428	0\\
3.302073717	0\\
3.322144032	-0.01962\\
3.342067957	-0.145188\\
3.362023592	-0.145188\\
3.382169962	-0.121644\\
3.402055264	-0.121644\\
3.422219515	-0.043164\\
3.44207406	0.298224\\
3.462005854	-0.023544\\
3.482167482	-0.023544\\
3.502092838	-0.07848\\
3.522217512	0.306072\\
3.542165518	-0.623916\\
3.56202507	-0.788724\\
3.582141399	-0.788724\\
3.602095366	-0.270756\\
3.622175932	0.05886\\
3.642080307	0.180504\\
3.662027359	0.102024\\
3.682184696	0.102024\\
3.702087164	0.07848\\
3.722172022	0.07848\\
3.742079735	0.211896\\
3.7620399	-0.007848\\
3.782176971	-0.007848\\
3.802106619	-0.003924\\
3.822199345	0.129492\\
3.842080116	0.129492\\
3.861995697	-0.007848\\
3.882146597	0.102024\\
3.902102232	-0.317844\\
3.922180891	-0.21582\\
3.942057133	0.223668\\
3.962022543	0.223668\\
3.982133865	0.051012\\
4.002083778	-0.007848\\
4.022172689	0.047088\\
4.042096615	-0.141264\\
4.06201911	0.090252\\
4.082199574	0.172656\\
4.102113008	0.125568\\
4.122160435	0.125568\\
4.142089844	0.306072\\
4.162051678	0.153036\\
4.182180882	0.062784\\
4.202055454	0.054936\\
4.222214699	0.054936\\
4.242144346	0.0981\\
4.262007713	0.149112\\
4.282158852	0.149112\\
4.302107573	-0.082404\\
4.322161913	-0.082404\\
4.342093706	0.168732\\
4.362052917	-0.054936\\
4.38219142	-0.003924\\
4.402065992	-0.003924\\
4.422214985	-0.0981\\
4.442141771	-0.074556\\
4.462005138	-0.074556\\
4.48217988	0.121644\\
4.502103329	0.121644\\
4.522183895	0.003924\\
4.542001963	0.125568\\
4.562021494	0.266832\\
4.582141638	0.180504\\
4.601983547	0.180504\\
4.622296572	0.43164\\
4.642003298	0.506196\\
4.662042141	0.506196\\
4.68221283	0.561132\\
4.702038527	0.702396\\
4.722152233	0.780876\\
4.742001772	0.792648\\
4.762083292	0.86328\\
4.782161951	0.86328\\
4.802028894	0.541512\\
4.822190285	0.773028\\
4.841997147	0.773028\\
4.862009764	0.776952\\
4.882171869	0.66708\\
4.902021646	0.66708\\
4.922164917	0.753408\\
4.942040443	0.678852\\
4.962023258	0.655308\\
4.982138634	0.655308\\
5.002011061	0.463032\\
5.022208452	0.463032\\
5.042009592	0.584676\\
5.061990499	0.31392\\
5.082176208	0.31392\\
5.102002144	0.545436\\
5.122169971	0.545436\\
5.142003059	0.52974\\
5.162042618	0.52974\\
5.182177782	0.533664\\
5.20200038	0.37278\\
5.222169161	0.463032\\
5.242000103	0.498348\\
5.262005568	0.498348\\
5.282134056	0.56898\\
5.302009821	0.612144\\
5.322175264	0.612144\\
5.342041731	0.502272\\
5.362050533	0.502272\\
5.382157087	0.60822\\
5.402051449	0.41202\\
5.422196627	0.54936\\
5.442041874	0.54936\\
5.462004662	0.325692\\
5.482170343	0.337464\\
5.502006054	0.266832\\
5.522168398	0.66708\\
5.542014837	0.290376\\
5.562004805	0.427716\\
5.58215785	0.427716\\
5.601986408	0.361008\\
5.622159719	0.266832\\
5.642006874	0.388476\\
5.661997795	0.388476\\
5.682132959	0.388476\\
5.701995373	0.309996\\
5.72217679	0.317844\\
5.742019892	0.317844\\
5.762010574	0.41202\\
5.782159567	0.207972\\
5.801992416	0.207972\\
5.822404146	0.321768\\
5.842020988	0.223668\\
5.862004519	0.223668\\
5.882167339	0.357084\\
5.902027845	0.455184\\
5.92217803	0.455184\\
5.942015648	0.325692\\
5.96203351	0.325692\\
5.982154131	0.380628\\
6.00202179	0.380628\\
6.022155046	0.506196\\
6.042008638	0.506196\\
6.062045097	0.404172\\
6.082176447	0.286452\\
6.102030754	-0.062784\\
6.122168303	0.309996\\
6.142015219	0.309996\\
6.162034512	0.145188\\
6.182168484	0.557208\\
6.202008486	-0.164808\\
6.22219348	-0.164808\\
6.242008209	0.188352\\
6.261991501	-0.121644\\
6.2821877	0.443412\\
6.302003145	0.109872\\
6.322138309	0.109872\\
6.341986656	0.160884\\
6.362041235	0.270756\\
6.382148266	0.270756\\
6.402024269	0.35316\\
6.42219305	0.341388\\
6.442016125	0.33354\\
6.462002277	0.33354\\
6.482141256	0.168732\\
6.501999855	0.302148\\
6.522139788	0.21582\\
6.542096376	0.086328\\
6.562011242	0.247212\\
6.582169533	0.247212\\
6.602061272	0.172656\\
6.622207403	0.172656\\
6.642106295	0.035316\\
6.662022114	0.086328\\
6.682183027	-0.027468\\
6.702072859	-0.027468\\
6.722159863	-0.086328\\
6.742056131	0.047088\\
6.762032509	0.047088\\
6.782183886	0.251136\\
6.802063704	0.223668\\
6.822175264	0.227592\\
6.842110157	0.160884\\
6.861999989	0.180504\\
6.882146597	0.086328\\
6.9020679	0.102024\\
6.922154188	0.15696\\
6.942060471	0.15696\\
6.962014437	0.184428\\
6.98218441	0.015696\\
7.002091169	0.015696\\
7.022185087	0.192276\\
7.042077065	0.192276\\
7.062018156	0.15696\\
7.082176685	0.07848\\
7.102084875	0.07848\\
7.122146368	0.074556\\
7.142120361	0.074556\\
7.162061453	0.109872\\
7.182176352	0.1962\\
7.202130079	0.105948\\
7.222165823	0.160884\\
7.242084503	0.007848\\
7.262024403	0.066708\\
7.282145262	0.223668\\
7.302088261	-0.082404\\
7.322201729	0.266832\\
7.342108727	-0.102024\\
7.362012625	-0.102024\\
7.382150888	0.062784\\
7.402046919	0.066708\\
7.42216754	-0.113796\\
7.442087412	-0.113796\\
7.462014914	-0.011772\\
7.482140779	-0.172656\\
7.502044916	-0.129492\\
7.522149324	-0.15696\\
7.542014122	-0.251136\\
7.562052011	-0.251136\\
7.582183599	-0.561132\\
7.60199666	-0.561132\\
7.622199535	-0.400248\\
7.6420784	-0.623916\\
7.66203475	-0.56898\\
7.682171345	-0.56898\\
7.702017784	-0.788724\\
7.722157955	-0.780876\\
7.74199152	-0.867204\\
7.762046337	-0.867204\\
7.78216362	-0.596448\\
7.802079916	-0.596448\\
7.822324038	-0.211896\\
7.842059135	-0.459108\\
7.862033606	-0.133416\\
7.882151604	0.070632\\
7.902106762	0.231516\\
7.922181845	0.266832\\
7.942075253	0.266832\\
7.96202302	0.266832\\
7.982138634	0.278604\\
8.002056837	0.278604\\
8.022184849	0.455184\\
8.042066336	0.455184\\
8.061996698	0.13734\\
8.082165003	0.13734\\
8.10209918	0.03924\\
8.12220192	0.011772\\
8.142128468	0.011772\\
8.162009001	-0.031392\\
8.182178974	-0.031392\\
8.202082872	-0.086328\\
8.2221632	-0.243288\\
8.242114782	-0.211896\\
8.261997461	-0.003924\\
8.282184362	-0.003924\\
8.302082777	-0.051012\\
8.322158575	-0.231516\\
8.342079163	-0.231516\\
8.362035036	-0.0981\\
8.382153511	-0.0981\\
8.402086258	-0.270756\\
8.422154903	-0.270756\\
8.442137957	-0.031392\\
8.462030649	-0.090252\\
8.482141018	-0.090252\\
8.50206995	-0.129492\\
8.522186756	0.168732\\
8.542097092	-0.086328\\
8.562069654	-0.086328\\
8.582162857	0.066708\\
8.602098942	0.211896\\
8.622184515	0.1962\\
8.642097235	0.17658\\
8.662009954	0.145188\\
8.682184458	0.145188\\
8.70208025	0\\
8.722199678	0.03924\\
8.742061377	0.03924\\
8.76201725	0.251136\\
8.782147169	0.168732\\
8.802074194	0.141264\\
8.822186708	-0.007848\\
8.842088938	-0.204048\\
8.86203146	0.011772\\
8.882168055	-0.03924\\
8.902080774	-0.090252\\
8.92218399	-0.090252\\
8.942073584	-0.051012\\
8.962017059	-0.094176\\
8.982139111	-0.180504\\
9.002076387	-0.180504\\
9.022206306	-0.070632\\
9.042067289	-0.129492\\
9.062023401	-0.129492\\
9.082197666	-0.086328\\
9.102116108	-0.086328\\
9.122168541	-0.129492\\
9.142102718	-0.11772\\
9.16201067	-0.05886\\
9.182157755	-0.05886\\
9.202145576	-0.086328\\
9.222205639	-0.086328\\
9.242083549	-0.129492\\
9.262011051	-0.047088\\
9.282196045	0.003924\\
9.302088022	-0.105948\\
9.322224855	0.011772\\
9.342083454	-0.070632\\
9.362039328	0.015696\\
9.382145166	0.043164\\
9.402059555	0.043164\\
9.422160625	-0.317844\\
9.442074537	-0.317844\\
9.462022066	-0.01962\\
9.482147932	0.051012\\
9.50248909	0.051012\\
9.522160769	-0.023544\\
9.542102814	-0.03924\\
9.56207037	-0.153036\\
9.582131863	-0.153036\\
9.602108955	-0.01962\\
9.622174025	-0.01962\\
9.64208889	-0.007848\\
9.662001371	-0.007848\\
9.682150602	-0.286452\\
9.702144861	-0.286452\\
9.722158909	-0.074556\\
9.742101669	-0.031392\\
9.762069225	-0.031392\\
9.782200336	-0.133416\\
9.802125454	-0.13734\\
9.822184563	-0.270756\\
9.84209919	-0.062784\\
9.862032652	-0.149112\\
9.882185698	-0.23544\\
9.902075529	-0.23544\\
9.922151327	-0.227592\\
9.942129374	-0.231516\\
9.96201849	-0.231516\\
9.982164145	-0.011772\\
10.002075434	-0.011772\\
10.022153854	-0.090252\\
10.04221487	-0.047088\\
10.062048197	-0.047088\\
10.082187891	-0.074556\\
10.102066994	-0.074556\\
10.122208834	-0.051012\\
10.142107248	-0.051012\\
10.162016392	0.023544\\
10.182190657	-0.149112\\
10.202092886	-0.05886\\
10.222184658	0.031392\\
10.242083073	-0.023544\\
10.262024164	-0.047088\\
10.282155752	-0.047088\\
10.302184582	-0.125568\\
10.322179794	-0.11772\\
10.342079401	-0.11772\\
10.362017393	-0.0981\\
10.382154226	-0.11772\\
10.402059078	-0.11772\\
10.42215991	-0.231516\\
10.442088127	-0.164808\\
10.462037086	-0.164808\\
10.482158422	0.011772\\
10.502085447	-0.074556\\
10.522191048	-0.074556\\
10.542074203	-0.227592\\
10.562060118	-0.21582\\
10.582148314	-0.160884\\
10.60208869	-0.160884\\
10.622238636	-0.239364\\
10.642085791	-0.239364\\
10.662041664	-0.247212\\
10.682162285	-0.290376\\
10.70212698	-0.109872\\
10.722177505	-0.180504\\
10.742103338	-0.180504\\
10.76202035	-0.17658\\
10.782142401	-0.149112\\
10.802069664	-0.211896\\
10.822172403	-0.211896\\
10.842101336	-0.17658\\
10.862004042	-0.223668\\
10.882152796	-0.27468\\
10.902067661	-0.27468\\
10.922169685	-0.086328\\
10.942084789	-0.086328\\
10.962040186	-0.1962\\
10.982153654	-0.17658\\
11.002078533	-0.3924\\
11.022383213	-0.3924\\
11.042106867	-0.102024\\
11.062020779	-0.188352\\
11.08216548	0.113796\\
11.102080345	0.113796\\
11.122196198	-0.302148\\
11.142080545	-0.349236\\
11.162013292	-0.349236\\
11.182190418	-0.133416\\
11.202085495	-0.153036\\
11.222196817	-0.13734\\
11.242108583	-0.13734\\
11.261999369	-0.211896\\
11.282183409	-0.211896\\
11.302100897	-0.400248\\
11.32214427	-0.400248\\
11.342061758	-0.290376\\
11.362005711	-0.357084\\
11.382149696	-0.357084\\
11.402094126	-0.282528\\
11.422182083	-0.521892\\
11.442009449	-0.180504\\
11.462024927	-0.368856\\
11.482139349	-0.219744\\
11.502108097	-0.486576\\
11.522141933	-0.258984\\
11.541990042	-0.388476\\
11.56203866	-0.388476\\
11.582138538	-0.345312\\
11.602074623	-0.345312\\
11.622191191	-0.345312\\
11.642078638	-0.325692\\
11.662010193	-0.321768\\
11.682131052	-0.027468\\
11.702092648	-0.317844\\
11.722218513	-0.317844\\
11.742126942	-0.364932\\
11.762035608	-0.239364\\
11.782173157	-0.239364\\
11.802094936	-0.31392\\
11.822187424	-0.282528\\
11.842082024	0\\
11.86200428	0\\
11.882132292	-0.541512\\
11.902098179	-0.376704\\
11.922159433	-0.047088\\
11.942083597	-0.047088\\
11.962034941	-0.090252\\
11.982154131	-0.090252\\
12.002068758	-0.231516\\
12.022183418	-0.219744\\
12.042089939	-0.11772\\
12.062016249	-0.231516\\
12.082199812	-0.231516\\
12.102077246	-0.282528\\
12.12219286	-0.290376\\
12.142095089	-0.357084\\
12.162034035	-0.357084\\
12.182173967	-0.357084\\
12.202055454	-0.231516\\
12.222174883	-0.231516\\
12.242062807	-0.415944\\
12.262057781	-0.258984\\
12.282171249	-0.380628\\
12.302083969	-0.188352\\
12.322149754	-0.325692\\
12.342099667	-0.306072\\
12.362051249	-0.306072\\
12.382143021	-0.207972\\
12.402063608	-0.207972\\
12.422169447	-0.172656\\
12.442089319	-0.239364\\
12.46199894	-0.133416\\
};
\addplot [color=black, forget plot]
  table[row sep=crcr]{%
3.29999995231628	-1\\
3.29999995231628	1.5\\
};
\node[right, align=left, circle,fill=white,draw=black,thick,line width=0.3mm,inner sep=1pt]
at (axis cs:2.2,0.875) {1};
\addplot [color=black, forget plot]
  table[row sep=crcr]{%
7.59999990463257	-1\\
7.59999990463257	1.5\\
};
\node[right, align=left, circle,fill=white,draw=black,thick,line width=0.3mm,inner sep=1pt]
at (axis cs:6.5,0.875) {2};
\addplot [color=black, forget plot]
  table[row sep=crcr]{%
11.4400000572205	-1\\
11.4400000572205	1.5\\
};
\node[right, align=left, circle,fill=white,draw=black,thick,line width=0.3mm,inner sep=1pt]
at (axis cs:10.34,0.875) {3};
\end{axis}
\end{tikzpicture}%

%% file: Images/Tikz/fig9-3-3.tikz
\begin{tikzpicture}

\begin{axis}[%
width=4.25cm,
height=1.683cm,
at={(0cm,0cm)},
scale only axis,
xmin=0.642011881,
xmax=12.46199894,
xlabel style={font=\color{white!15!black}},
xlabel={$t\;[\SI[per-mode=repeated-symbol]{}{\second}]$},
ymin=-10,
ymax=10,
ylabel style={font=\color{white!15!black}},
ylabel={$\delta\;[\text{Deg}]$},
axis background/.style={fill=white},
ylabel style={yshift=-0.2cm}
]
\addplot [color=blue, line width=1.0pt, forget plot]
  table[row sep=crcr]{%
0.642011881	2.5\\
0.662031174	2.5\\
0.682033062	2.5\\
0.702054262	2.60000000000002\\
0.722006321	2.70000000000005\\
0.742020607	2.70000000000005\\
0.762029409	2.79999999999995\\
0.782009363	2.79999999999995\\
0.802028418	2.79999999999995\\
0.822096586	2.79999999999995\\
0.842060328	2.79999999999995\\
0.862061977	2.79999999999995\\
0.882035017	2.79999999999995\\
0.902044773	2.79999999999995\\
0.922039747	2.79999999999995\\
0.941986799	2.89999999999998\\
0.962060928	2.89999999999998\\
0.982016802	2.89999999999998\\
1.001998186	1.20000000000005\\
1.0225811	1.20000000000005\\
1.041987896	1.20000000000005\\
1.062039852	1.10000000000002\\
1.082021952	1.10000000000002\\
1.102125168	1.10000000000002\\
1.122168303	1.10000000000002\\
1.142053843	1.10000000000002\\
1.162023783	1.10000000000002\\
1.182157993	1.10000000000002\\
1.202011108	1.10000000000002\\
1.222189903	1.10000000000002\\
1.24209857	1\\
1.262012243	1\\
1.282166004	1\\
1.30210948	1\\
1.322170496	1\\
1.342099428	0.899999999999977\\
1.362005711	0.899999999999977\\
1.382188082	0.899999999999977\\
1.402091265	0.899999999999977\\
1.422305346	0.899999999999977\\
1.442149401	0.899999999999977\\
1.462014198	0.899999999999977\\
1.482145309	0.899999999999977\\
1.502088785	0.899999999999977\\
1.522160292	0.899999999999977\\
1.542102575	0.899999999999977\\
1.562010288	0.899999999999977\\
1.582163095	0.899999999999977\\
1.602073908	0.899999999999977\\
1.622177124	0.899999999999977\\
1.642151356	0.899999999999977\\
1.662014008	0.899999999999977\\
1.682134151	0.899999999999977\\
1.70217967	0.899999999999977\\
1.722165823	0.899999999999977\\
1.742109299	0.899999999999977\\
1.762005329	0.899999999999977\\
1.782206297	0.899999999999977\\
1.802084208	0.799999999999955\\
1.822207689	0.799999999999955\\
1.842102051	0.799999999999955\\
1.862024307	0.799999999999955\\
1.882204294	0.799999999999955\\
1.902106047	0.799999999999955\\
1.922158003	0.799999999999955\\
1.942089081	0.799999999999955\\
1.962026119	0.799999999999955\\
1.982143164	0.799999999999955\\
2.00207901	0.799999999999955\\
2.022226334	0.700000000000045\\
2.042124987	0.799999999999955\\
2.062014103	0.799999999999955\\
2.082222939	0.799999999999955\\
2.102130413	0.799999999999955\\
2.122152328	0.799999999999955\\
2.142102718	0.700000000000045\\
2.162024736	0.799999999999955\\
2.182160616	0.799999999999955\\
2.20205164	0.700000000000045\\
2.222170115	0.799999999999955\\
2.242100239	0.799999999999955\\
2.262027979	0.799999999999955\\
2.282144308	0.700000000000045\\
2.302100658	0.700000000000045\\
2.322163582	0.700000000000045\\
2.342063904	0.700000000000045\\
2.361994982	0.700000000000045\\
2.382143974	0.700000000000045\\
2.40207243	0.700000000000045\\
2.42218256	0.700000000000045\\
2.442131042	0.600000000000023\\
2.46202755	0.600000000000023\\
2.48217392	0.600000000000023\\
2.502112865	0.600000000000023\\
2.522167683	0.600000000000023\\
2.542099714	0.600000000000023\\
2.562043667	0.600000000000023\\
2.58217597	0.600000000000023\\
2.602068663	0.600000000000023\\
2.622182608	0.600000000000023\\
2.6420753	0.600000000000023\\
2.662009001	0.600000000000023\\
2.682165384	0.5\\
2.702100515	0.600000000000023\\
2.722188473	0.5\\
2.742105484	0.5\\
2.76200366	0.5\\
2.782128811	0.5\\
2.80207634	0.5\\
2.822188854	0.5\\
2.842112541	0.5\\
2.862010002	0.5\\
2.882200003	0.5\\
2.902073622	0.600000000000023\\
2.922177553	0.600000000000023\\
2.942096472	0.600000000000023\\
2.962039709	0.600000000000023\\
2.982153654	0.700000000000045\\
3.002112389	0.600000000000023\\
3.022202015	0.600000000000023\\
3.042122364	0.700000000000045\\
3.062047958	0.700000000000045\\
3.082171917	0.700000000000045\\
3.102081537	0.700000000000045\\
3.122168303	0.700000000000045\\
3.142112732	0.700000000000045\\
3.162014723	0.700000000000045\\
3.182135105	0.700000000000045\\
3.20213604	0.700000000000045\\
3.222180128	0.600000000000023\\
3.242101908	0.700000000000045\\
3.262006998	0.600000000000023\\
3.282162428	0.700000000000045\\
3.302073717	0.700000000000045\\
3.322144032	0.600000000000023\\
3.342067957	0.600000000000023\\
3.362023592	0.5\\
3.382169962	0.5\\
3.402055264	0.5\\
3.422219515	0.5\\
3.44207406	0.600000000000023\\
3.462005854	0.600000000000023\\
3.482167482	0.600000000000023\\
3.502092838	0.600000000000023\\
3.522217512	0.600000000000023\\
3.542165518	0.600000000000023\\
3.56202507	0.600000000000023\\
3.582141399	0.700000000000045\\
3.602095366	0.600000000000023\\
3.622175932	0.600000000000023\\
3.642080307	0.600000000000023\\
3.662027359	0.600000000000023\\
3.682184696	0.600000000000023\\
3.702087164	0.700000000000045\\
3.722172022	0.700000000000045\\
3.742079735	0.700000000000045\\
3.7620399	0.600000000000023\\
3.782176971	0.600000000000023\\
3.802106619	0.600000000000023\\
3.822199345	0.600000000000023\\
3.842080116	0.600000000000023\\
3.861995697	0.600000000000023\\
3.882146597	0.5\\
3.902102232	0.399999999999977\\
3.922180891	0.399999999999977\\
3.942057133	0.299999999999955\\
3.962022543	0.200000000000045\\
3.982133865	-0.100000000000023\\
4.002083778	-0.399999999999977\\
4.022172689	-0.600000000000023\\
4.042096615	-0.5\\
4.06201911	-0.399999999999977\\
4.082199574	-0.100000000000023\\
4.102113008	0.100000000000023\\
4.122160435	0.200000000000045\\
4.142089844	0.100000000000023\\
4.162051678	0\\
4.182180882	0\\
4.202055454	0\\
4.222214699	0.100000000000023\\
4.242144346	0.100000000000023\\
4.262007713	0\\
4.282158852	-0.100000000000023\\
4.302107573	-0.200000000000045\\
4.322161913	-0.399999999999977\\
4.342093706	-0.399999999999977\\
4.362052917	-0.700000000000045\\
4.38219142	-1.29999999999995\\
4.402065992	-2.29999999999995\\
4.422214985	-3.70000000000005\\
4.442141771	-5\\
4.462005138	-6\\
4.48217988	-6.5\\
4.502103329	-6.70000000000005\\
4.522183895	-6.70000000000005\\
4.542001963	-6.79999999999995\\
4.562021494	-7.10000000000002\\
4.582141638	-7.60000000000002\\
4.601983547	-8.10000000000002\\
4.622296572	-8.39999999999998\\
4.642003298	-8.20000000000005\\
4.662042141	-7.89999999999998\\
4.68221283	-7.60000000000002\\
4.702038527	-7.60000000000002\\
4.722152233	-7.79999999999995\\
4.742001772	-7.89999999999998\\
4.762083292	-7.89999999999998\\
4.782161951	-7.79999999999995\\
4.802028894	-7.70000000000005\\
4.822190285	-7.79999999999995\\
4.841997147	-7.89999999999998\\
4.862009764	-7.89999999999998\\
4.882171869	-7.70000000000005\\
4.902021646	-7.5\\
4.922164917	-7.39999999999998\\
4.942040443	-7.39999999999998\\
4.962023258	-7.29999999999995\\
4.982138634	-7.29999999999995\\
5.002011061	-7.29999999999995\\
5.022208452	-7.20000000000005\\
5.042009592	-7.10000000000002\\
5.061990499	-7.10000000000002\\
5.082176208	-7.10000000000002\\
5.102002144	-7.10000000000002\\
5.122169971	-7\\
5.142003059	-6.60000000000002\\
5.162042618	-6.10000000000002\\
5.182177782	-5.70000000000005\\
5.20200038	-5.29999999999995\\
5.222169161	-5.10000000000002\\
5.242000103	-5\\
5.262005568	-5\\
5.282134056	-5\\
5.302009821	-5\\
5.322175264	-4.89999999999998\\
5.342041731	-4.79999999999995\\
5.362050533	-4.70000000000005\\
5.382157087	-4.5\\
5.402051449	-4.20000000000005\\
5.422196627	-4\\
5.442041874	-3.89999999999998\\
5.462004662	-4\\
5.482170343	-4\\
5.502006054	-4\\
5.522168398	-4\\
5.542014837	-3.89999999999998\\
5.562004805	-4\\
5.58215785	-4\\
5.601986408	-4\\
5.622159719	-3.89999999999998\\
5.642006874	-3.79999999999995\\
5.661997795	-3.70000000000005\\
5.682132959	-3.60000000000002\\
5.701995373	-3.70000000000005\\
5.72217679	-3.79999999999995\\
5.742019892	-3.70000000000005\\
5.762010574	-3.60000000000002\\
5.782159567	-3.39999999999998\\
5.801992416	-3.29999999999995\\
5.822404146	-3.29999999999995\\
5.842020988	-3.39999999999998\\
5.862004519	-3.39999999999998\\
5.882167339	-3.29999999999995\\
5.902027845	-3.20000000000005\\
5.92217803	-3.10000000000002\\
5.942015648	-3.10000000000002\\
5.96203351	-3\\
5.982154131	-3\\
6.00202179	-2.89999999999998\\
6.022155046	-2.89999999999998\\
6.042008638	-2.79999999999995\\
6.062045097	-2.79999999999995\\
6.082176447	-2.79999999999995\\
6.102030754	-2.79999999999995\\
6.122168303	-2.79999999999995\\
6.142015219	-2.79999999999995\\
6.162034512	-2.79999999999995\\
6.182168484	-2.70000000000005\\
6.202008486	-2.70000000000005\\
6.22219348	-2.70000000000005\\
6.242008209	-2.70000000000005\\
6.261991501	-2.60000000000002\\
6.2821877	-2.60000000000002\\
6.302003145	-2.60000000000002\\
6.322138309	-2.60000000000002\\
6.341986656	-2.60000000000002\\
6.362041235	-2.5\\
6.382148266	-2.39999999999998\\
6.402024269	-2.20000000000005\\
6.42219305	-2\\
6.442016125	-2\\
6.462002277	-1.89999999999998\\
6.482141256	-1.89999999999998\\
6.501999855	-1.89999999999998\\
6.522139788	-1.79999999999995\\
6.542096376	-1.79999999999995\\
6.562011242	-1.70000000000005\\
6.582169533	-1.70000000000005\\
6.602061272	-1.70000000000005\\
6.622207403	-1.70000000000005\\
6.642106295	-1.60000000000002\\
6.662022114	-1.39999999999998\\
6.682183027	-1.29999999999995\\
6.702072859	-1.20000000000005\\
6.722159863	-1.10000000000002\\
6.742056131	-1.10000000000002\\
6.762032509	-1.10000000000002\\
6.782183886	-1.10000000000002\\
6.802063704	-1.10000000000002\\
6.822175264	-1.10000000000002\\
6.842110157	-1.10000000000002\\
6.861999989	-1.10000000000002\\
6.882146597	-1\\
6.9020679	-0.899999999999977\\
6.922154188	-0.799999999999955\\
6.942060471	-0.700000000000045\\
6.962014437	-0.600000000000023\\
6.98218441	-0.700000000000045\\
7.002091169	-0.700000000000045\\
7.022185087	-0.700000000000045\\
7.042077065	-0.600000000000023\\
7.062018156	-0.600000000000023\\
7.082176685	-0.600000000000023\\
7.102084875	-0.600000000000023\\
7.122146368	-0.5\\
7.142120361	-0.399999999999977\\
7.162061453	-0.299999999999955\\
7.182176352	-0.200000000000045\\
7.202130079	-0.100000000000023\\
7.222165823	0\\
7.242084503	0.200000000000045\\
7.262024403	0.5\\
7.282145262	1\\
7.302088261	1.5\\
7.322201729	2.10000000000002\\
7.342108727	3\\
7.362012625	3.89999999999998\\
7.382150888	4.89999999999998\\
7.402046919	5.79999999999995\\
7.42216754	6.39999999999998\\
7.442087412	6.89999999999998\\
7.462014914	7.29999999999995\\
7.482140779	7.60000000000002\\
7.502044916	7.79999999999995\\
7.522149324	8.10000000000002\\
7.542014122	8.20000000000005\\
7.562052011	8.29999999999995\\
7.582183599	8.39999999999998\\
7.60199666	8.29999999999995\\
7.622199535	7.89999999999998\\
7.6420784	7.10000000000002\\
7.66203475	5.70000000000005\\
7.682171345	3.70000000000005\\
7.702017784	1.79999999999995\\
7.722157955	0.299999999999955\\
7.74199152	-0.399999999999977\\
7.762046337	-0.299999999999955\\
7.78216362	0.200000000000045\\
7.802079916	0.600000000000023\\
7.822324038	0.600000000000023\\
7.842059135	0.200000000000045\\
7.862033606	-0.200000000000045\\
7.882151604	-0.299999999999955\\
7.902106762	0.100000000000023\\
7.922181845	0.600000000000023\\
7.942075253	1.20000000000005\\
7.96202302	1.60000000000002\\
7.982138634	1.89999999999998\\
8.002056837	1.89999999999998\\
8.022184849	1.89999999999998\\
8.042066336	1.89999999999998\\
8.061996698	2\\
8.082165003	2\\
8.10209918	2.20000000000005\\
8.12220192	2.29999999999995\\
8.142128468	2.29999999999995\\
8.162009001	2.20000000000005\\
8.182178974	2.10000000000002\\
8.202082872	2\\
8.2221632	2\\
8.242114782	2\\
8.261997461	2\\
8.282184362	1.89999999999998\\
8.302082777	1.89999999999998\\
8.322158575	1.89999999999998\\
8.342079163	1.89999999999998\\
8.362035036	1.79999999999995\\
8.382153511	1.79999999999995\\
8.402086258	1.79999999999995\\
8.422154903	1.70000000000005\\
8.442137957	1.70000000000005\\
8.462030649	1.70000000000005\\
8.482141018	1.79999999999995\\
8.50206995	1.79999999999995\\
8.522186756	1.79999999999995\\
8.542097092	1.79999999999995\\
8.562069654	1.79999999999995\\
8.582162857	1.89999999999998\\
8.602098942	1.89999999999998\\
8.622184515	1.89999999999998\\
8.642097235	2\\
8.662009954	2\\
8.682184458	2\\
8.70208025	2\\
8.722199678	2\\
8.742061377	2\\
8.76201725	2\\
8.782147169	2.10000000000002\\
8.802074194	2.20000000000005\\
8.822186708	2.29999999999995\\
8.842088938	2.29999999999995\\
8.86203146	2.39999999999998\\
8.882168055	2.29999999999995\\
8.902080774	2.29999999999995\\
8.92218399	2.20000000000005\\
8.942073584	2.20000000000005\\
8.962017059	2.20000000000005\\
8.982139111	2.20000000000005\\
9.002076387	2.20000000000005\\
9.022206306	2.20000000000005\\
9.042067289	2.20000000000005\\
9.062023401	2.20000000000005\\
9.082197666	2.10000000000002\\
9.102116108	2.20000000000005\\
9.122168541	2.20000000000005\\
9.142102718	2.10000000000002\\
9.16201067	2.10000000000002\\
9.182157755	2.20000000000005\\
9.202145576	2.20000000000005\\
9.222205639	2.20000000000005\\
9.242083549	2.20000000000005\\
9.262011051	2.29999999999995\\
9.282196045	2.29999999999995\\
9.302088022	2.39999999999998\\
9.322224855	2.39999999999998\\
9.342083454	2.5\\
9.362039328	2.60000000000002\\
9.382145166	2.60000000000002\\
9.402059555	2.60000000000002\\
9.422160625	2.70000000000005\\
9.442074537	2.70000000000005\\
9.462022066	2.70000000000005\\
9.482147932	2.70000000000005\\
9.50248909	2.70000000000005\\
9.522160769	2.70000000000005\\
9.542102814	2.70000000000005\\
9.56207037	2.70000000000005\\
9.582131863	2.70000000000005\\
9.602108955	2.79999999999995\\
9.622174025	2.70000000000005\\
9.64208889	2.70000000000005\\
9.662001371	2.70000000000005\\
9.682150602	2.60000000000002\\
9.702144861	2.60000000000002\\
9.722158909	2.60000000000002\\
9.742101669	2.60000000000002\\
9.762069225	2.60000000000002\\
9.782200336	2.60000000000002\\
9.802125454	2.60000000000002\\
9.822184563	2.60000000000002\\
9.84209919	2.60000000000002\\
9.862032652	2.60000000000002\\
9.882185698	2.60000000000002\\
9.902075529	2.60000000000002\\
9.922151327	2.60000000000002\\
9.942129374	2.60000000000002\\
9.96201849	2.60000000000002\\
9.982164145	2.60000000000002\\
10.002075434	2.60000000000002\\
10.022153854	2.60000000000002\\
10.04221487	2.60000000000002\\
10.062048197	2.60000000000002\\
10.082187891	2.70000000000005\\
10.102066994	2.70000000000005\\
10.122208834	2.70000000000005\\
10.142107248	2.70000000000005\\
10.162016392	2.70000000000005\\
10.182190657	2.70000000000005\\
10.202092886	2.70000000000005\\
10.222184658	2.79999999999995\\
10.242083073	2.79999999999995\\
10.262024164	2.79999999999995\\
10.282155752	2.79999999999995\\
10.302184582	2.79999999999995\\
10.322179794	2.89999999999998\\
10.342079401	3.10000000000002\\
10.362017393	3.10000000000002\\
10.382154226	3.10000000000002\\
10.402059078	3.10000000000002\\
10.42215991	3.10000000000002\\
10.442088127	3.10000000000002\\
10.462037086	3.10000000000002\\
10.482158422	3.10000000000002\\
10.502085447	3.10000000000002\\
10.522191048	3.10000000000002\\
10.542074203	3.10000000000002\\
10.562060118	3.20000000000005\\
10.582148314	3.20000000000005\\
10.60208869	3.29999999999995\\
10.622238636	3.29999999999995\\
10.642085791	3.29999999999995\\
10.662041664	3.29999999999995\\
10.682162285	3.29999999999995\\
10.70212698	3.39999999999998\\
10.722177505	3.60000000000002\\
10.742103338	3.60000000000002\\
10.76202035	3.70000000000005\\
10.782142401	3.70000000000005\\
10.802069664	3.70000000000005\\
10.822172403	3.79999999999995\\
10.842101336	3.89999999999998\\
10.862004042	4\\
10.882152796	4\\
10.902067661	4\\
10.922169685	4\\
10.942084789	4.10000000000002\\
10.962040186	4.10000000000002\\
10.982153654	4.20000000000005\\
11.002078533	4.29999999999995\\
11.022383213	4.39999999999998\\
11.042106867	4.5\\
11.062020779	4.60000000000002\\
11.08216548	4.79999999999995\\
11.102080345	4.79999999999995\\
11.122196198	4.89999999999998\\
11.142080545	4.79999999999995\\
11.162013292	4.89999999999998\\
11.182190418	4.89999999999998\\
11.202085495	4.89999999999998\\
11.222196817	5\\
11.242108583	5\\
11.261999369	5.10000000000002\\
11.282183409	5.10000000000002\\
11.302100897	5.10000000000002\\
11.32214427	5.20000000000005\\
11.342061758	5.29999999999995\\
11.362005711	5.39999999999998\\
11.382149696	5.29999999999995\\
11.402094126	5.29999999999995\\
11.422182083	5.29999999999995\\
11.442009449	5.29999999999995\\
11.462024927	5.29999999999995\\
11.482139349	5.29999999999995\\
11.502108097	5.29999999999995\\
11.522141933	5.29999999999995\\
11.541990042	5.29999999999995\\
11.56203866	5.29999999999995\\
11.582138538	5.29999999999995\\
11.602074623	5.20000000000005\\
11.622191191	5.20000000000005\\
11.642078638	5.10000000000002\\
11.662010193	5\\
11.682131052	4.89999999999998\\
11.702092648	4.89999999999998\\
11.722218513	4.79999999999995\\
11.742126942	4.79999999999995\\
11.762035608	4.79999999999995\\
11.782173157	4.79999999999995\\
11.802094936	4.70000000000005\\
11.822187424	4.70000000000005\\
11.842082024	4.70000000000005\\
11.86200428	4.70000000000005\\
11.882132292	4.70000000000005\\
11.902098179	4.70000000000005\\
11.922159433	4.60000000000002\\
11.942083597	4.60000000000002\\
11.962034941	4.60000000000002\\
11.982154131	4.5\\
12.002068758	4.5\\
12.022183418	4.5\\
12.042089939	4.5\\
12.062016249	4.5\\
12.082199812	4.5\\
12.102077246	4.5\\
12.12219286	4.5\\
12.142095089	4.5\\
12.162034035	4.39999999999998\\
12.182173967	4.29999999999995\\
12.202055454	4.20000000000005\\
12.222174883	4.10000000000002\\
12.242062807	4.10000000000002\\
12.262057781	4.10000000000002\\
12.282171249	4.20000000000005\\
12.302083969	4.10000000000002\\
12.322149754	4.10000000000002\\
12.342099667	4.10000000000002\\
12.362051249	4.10000000000002\\
12.382143021	4.10000000000002\\
12.402063608	4.10000000000002\\
12.422169447	4.10000000000002\\
12.442089319	4.10000000000002\\
12.46199894	4.10000000000002\\
};
\addplot [color=black, forget plot]
  table[row sep=crcr]{%
3.29999995231628	-10\\
3.29999995231628	10\\
};
\node[right, align=left, circle,fill=white,draw=black,thick,line width=0.3mm,inner sep=1pt]
at (axis cs:2.2,-5) {1};
\addplot [color=black, forget plot]
  table[row sep=crcr]{%
7.59999990463257	-10\\
7.59999990463257	10\\
};
\node[right, align=left, circle,fill=white,draw=black,thick,line width=0.3mm,inner sep=1pt]
at (axis cs:6.5,-5) {2};
\addplot [color=black, forget plot]
  table[row sep=crcr]{%
11.4400000572205	-10\\
11.4400000572205	10\\
};
\node[right, align=left, circle,fill=white,draw=black,thick,line width=0.3mm,inner sep=1pt]
at (axis cs:10.34,-5) {3};
\end{axis}
\end{tikzpicture}%

%% file: Images/Tikz/fig10-3-1.tikz
\begin{tikzpicture}

\begin{axis}[%
width=4.2cm,
height=1.683cm,
at={(0cm,0cm)},
scale only axis,
xmin=0.122153044,
xmax=20.401994467,
xlabel style={font=\color{white!15!black}},
xlabel={$t\;[\SI[per-mode=repeated-symbol]{}{\second}]$},
ymin=0,
ymax=30,
ylabel style={font=\color{white!15!black}},
ylabel={$v\;[\SI[per-mode=repeated-symbol]{}{\meter\per\second}]$},
axis background/.style={fill=white},
ylabel style={yshift=-0.4cm}
]
\addplot [color=blue, line width=1.0pt, forget plot]
  table[row sep=crcr]{%
0.122153044	9.47\\
0.142704725	9.45\\
0.162019968	9.43500132485417\\
0.182151318	9.42000132696381\\
0.202072382	9.40000132978714\\
0.222170353	9.385\\
0.242097378	9.37\\
0.262028217	9.3500013368983\\
0.282156229	9.33500535618486\\
0.302108049	9.32000536480532\\
0.322176456	9.30500537345358\\
0.342110395	9.29001210978759\\
0.362020254	9.27001213591438\\
0.382161379	9.25501215558359\\
0.402102709	9.24001217531665\\
0.422178984	9.22001220172728\\
0.442130327	9.20001222825274\\
0.46200633	9.18000544662148\\
0.482154846	9.16500545553575\\
0.502096176	9.15\\
0.522157907	9.13000547644962\\
0.542132139	9.11000548847255\\
0.562018394	9.09002200217359\\
0.582155943	9.07503443519638\\
0.602100611	9.05503451125395\\
0.622170925	9.03502213611013\\
0.642130375	9.01501247919269\\
0.661995411	8.99500555864197\\
0.682149649	8.98\\
0.702078104	8.96000139508918\\
0.722169399	8.94000139821018\\
0.742068052	8.915\\
0.76201272	8.89500140528376\\
0.782187939	8.87500563380103\\
0.802058935	8.8500225988412\\
0.822188854	8.83003539064255\\
0.84211874	8.80505110717706\\
0.862021446	8.79005119439017\\
0.882141829	8.76503565309349\\
0.902095318	8.74502287018164\\
0.922189951	8.7200129013666\\
0.942067385	8.70500574382349\\
0.962010384	8.68500575705048\\
0.982141018	8.66500144258499\\
1.00208354	8.64500578368806\\
1.022191286	8.6200130510342\\
1.04208827	8.6000130813854\\
1.062000513	8.5850131042416\\
1.082139254	8.56001314251328\\
1.10210681	8.54000585479893\\
1.122172117	8.52000146713602\\
1.142074108	8.5\\
1.162016153	8.48000147405648\\
1.182153702	8.46501329000729\\
1.202116489	8.44001332937336\\
1.222205639	8.42001336103453\\
1.242153645	8.40000595237884\\
1.26202631	8.37500149253718\\
1.282181978	8.36\\
1.302097559	8.34000149880082\\
1.322151184	8.32000600961321\\
1.342127562	8.29500602772536\\
1.362022638	8.28000603864514\\
1.382176876	8.26000151331705\\
1.402080297	8.24000151699015\\
1.422161579	8.22000152068112\\
1.442111731	8.2000015243901\\
1.462035418	8.18000611246715\\
1.482175827	8.1600015318626\\
1.502425671	8.14000614250383\\
1.522183895	8.13000615005918\\
1.542054653	8.10500616902912\\
1.562038183	8.08500618428953\\
1.582180977	8.06501394915099\\
1.602072954	8.04000621890307\\
1.622181654	8.02502492207968\\
1.64208889	8.01502495317388\\
1.662036657	7.99502501559564\\
1.682142735	7.97003920943931\\
1.702091694	7.95005660357208\\
1.722146988	7.93003940721608\\
1.742088318	7.91002528440965\\
1.762031794	7.89502533244828\\
1.782210588	7.87500634920379\\
1.802090406	7.85\\
1.822175741	7.835\\
1.842198849	7.81500639795004\\
1.862034321	7.79500641436555\\
1.882190228	7.77500643086551\\
1.902107477	7.75500161186315\\
1.922172308	7.735\\
1.942081451	7.72000161917081\\
1.962022543	7.70000649350376\\
1.982192516	7.67500651465522\\
2.002118826	7.66000652741236\\
2.02216363	7.64000654449982\\
2.04212141	7.62000656167697\\
2.062012196	7.60000657894452\\
2.082183361	7.58001484167412\\
2.102087498	7.56001488093774\\
2.122143269	7.54001492040964\\
2.142081261	7.5200149600915\\
2.162018299	7.49500667111111\\
2.182207823	7.47500167224062\\
2.202086687	7.46000167560303\\
2.222201586	7.44\\
2.242088079	7.42\\
2.262051105	7.395\\
2.282233238	7.37000169606493\\
2.302112341	7.35500679809339\\
2.322165251	7.33000682127923\\
2.342108011	7.31001538986068\\
2.362074614	7.29001543208243\\
2.382153273	7.27001547453649\\
2.402099133	7.25000689654844\\
2.422190428	7.23000172890712\\
2.44206953	7.21\\
2.462044001	7.19000173852552\\
2.482162237	7.16500697836366\\
2.50207901	7.1450069978972\\
2.522173405	7.1250070175404\\
2.542060137	7.10000704225003\\
2.562032938	7.08500176429053\\
2.582175016	7.06\\
2.602091312	7.04\\
2.622170687	7.02\\
2.642087221	6.99500178699048\\
2.661997318	6.97500179211447\\
2.682151079	6.95500179726792\\
2.702081203	6.93500720980159\\
2.722151279	6.92000722542975\\
2.742081404	6.89501631615183\\
2.762029409	6.87501636361689\\
2.782158852	6.85501641135891\\
2.802078962	6.83500731528504\\
2.822157145	6.82000733137436\\
2.842126608	6.80000183823505\\
2.862025261	6.775\\
2.882153511	6.75500185048087\\
2.902089834	6.74000185459915\\
2.922161102	6.72000186011879\\
2.9420681	6.7\\
2.962011576	6.68000187125722\\
2.982153654	6.66000750750328\\
3.002092123	6.64500752445022\\
3.022182465	6.62500754716551\\
3.042096138	6.61000756429219\\
3.062021494	6.59000189681308\\
3.082163334	6.57500190114041\\
3.102095366	6.55500190694099\\
3.1221807	6.53500191277707\\
3.142077208	6.52000766870715\\
3.16202116	6.50000769230314\\
3.182166815	6.48501734770232\\
3.20211935	6.46501740136869\\
3.222192287	6.44501745536814\\
3.242128134	6.42500778209646\\
3.262024879	6.40500780639649\\
3.282130003	6.39000782472134\\
3.302080154	6.37500784313243\\
3.322163582	6.35500786781574\\
3.342069626	6.33500789265491\\
3.361999273	6.31500791765141\\
3.382153988	6.30000198412667\\
3.402083397	6.28\\
3.422180891	6.265\\
3.442085266	6.24500200160096\\
3.462021589	6.23000802567701\\
3.482152939	6.21500804504709\\
3.502100468	6.19000201938578\\
3.522212744	6.175\\
3.542113781	6.15500203086888\\
3.562012196	6.13500814995384\\
3.582176685	6.11500817660942\\
3.602067471	6.09501845772431\\
3.622155666	6.07503292172149\\
3.642103434	6.05503303046317\\
3.662031174	6.03503313992558\\
3.682188272	6.01503325011591\\
3.702080488	6.00003333324074\\
3.722182035	5.98501879696296\\
3.742081165	5.96500838222378\\
3.762041092	5.94500210260686\\
3.782184362	5.93\\
3.802086115	5.91500211327097\\
3.822208643	5.89500848175811\\
3.842411518	5.88001913262193\\
3.862011194	5.86501918155431\\
3.882148266	5.85003418793429\\
3.902097464	5.83503427582049\\
3.922157049	5.82001932986481\\
3.942069292	5.80001939651929\\
3.962007523	5.7850086430359\\
3.982177258	5.77000216637741\\
4.002087593	5.75500217202392\\
4.022173405	5.735\\
4.042154551	5.72000218531427\\
4.06201911	5.70500219106005\\
4.082152367	5.69500877962449\\
4.102093697	5.68000880281008\\
4.122160673	5.66000883391537\\
4.142062426	5.64500885738898\\
4.162030935	5.62500222222178\\
4.182180405	5.61000891264889\\
4.202072144	5.59000223613551\\
4.222162008	5.57\\
4.242100716	5.555\\
4.262014389	5.53500225835546\\
4.28219533	5.51500226654532\\
4.302089214	5.49500909917354\\
4.322158575	5.47500913241247\\
4.342076063	5.45500916589514\\
4.362028837	5.43500919962423\\
4.382173061	5.4200092250844\\
4.402078867	5.40000925925132\\
4.4221766	5.38000232341957\\
4.442105055	5.36\\
4.46200633	5.34\\
4.482172966	5.32000939848794\\
4.502092361	5.30002122637259\\
4.522158146	5.28003787865201\\
4.542140245	5.26003802267626\\
4.562050581	5.24005963706521\\
4.582169533	5.22003831403564\\
4.60207653	5.20002163457038\\
4.622186661	5.18000965250066\\
4.642095804	5.16\\
4.662044764	5.14000243190604\\
4.682162046	5.12000976561569\\
4.702105522	5.10000980391215\\
4.722176552	5.08002214562102\\
4.742126703	5.06002223315274\\
4.762061596	5.04003968238346\\
4.782148838	5.02003984047936\\
4.802083969	5.00003999984\\
4.82220006	4.98002259031021\\
4.842048645	4.96001008063492\\
4.862048626	4.94001012144712\\
4.882157326	4.92000254064975\\
4.902056932	4.90000255101975\\
4.92215395	4.87500256410189\\
4.942088842	4.8500025773189\\
4.962026834	4.83000258799103\\
4.982174397	4.805\\
5.002012014	4.78\\
5.022206068	4.75500262881105\\
5.042008877	4.73000264270539\\
5.062007904	4.70501062698056\\
5.082171679	4.68001068374849\\
5.10203433	4.66000268240266\\
5.122157097	4.63500269687084\\
5.141997099	4.61\\
5.161998987	4.59000272331074\\
5.182173491	4.56501095288938\\
5.202002525	4.54002477966806\\
5.222178936	4.51502491687477\\
5.242048025	4.49002505560938\\
5.261997223	4.46502519589755\\
5.282162666	4.44001126124698\\
5.302005768	4.41501132501379\\
5.322175503	4.39000284737949\\
5.342036486	4.37\\
5.362068892	4.345\\
5.382175207	4.32\\
5.402026415	4.2950029103599\\
5.422189951	4.27000292739947\\
5.442039967	4.25000294117545\\
5.462042093	4.22500295857885\\
5.482180834	4.20500297265055\\
5.502022743	4.18000299042955\\
5.522157431	4.1550030084225\\
5.542028427	4.135\\
5.562045097	4.115\\
5.582159758	4.09500305250191\\
5.602006435	4.07500306748351\\
5.622204542	4.05000308641858\\
5.642046213	4.03000310173578\\
5.662028074	4.01001246880854\\
5.682136536	3.99001253130864\\
5.702010393	3.97001259443846\\
5.722163677	3.94501267425087\\
5.741996527	3.92501273883283\\
5.762047768	3.90501280407632\\
5.782159328	3.88500321750189\\
5.802027225	3.8650032341513\\
5.822189093	3.84000325520695\\
5.842361212	3.82000327224991\\
5.862008333	3.8\\
5.882154942	3.78000330687686\\
5.902008057	3.76000332446662\\
5.922199488	3.7400033422445\\
5.941994667	3.71500336473603\\
5.962021589	3.69500338294838\\
5.98218298	3.67500340135897\\
6.002015829	3.655\\
6.022216558	3.63\\
6.042048216	3.61\\
6.062056303	3.59\\
6.08216095	3.57000350139884\\
6.102020502	3.55000352112501\\
6.122161865	3.53000354107471\\
6.142045021	3.51000356125176\\
6.16202569	3.49\\
6.182163954	3.47\\
6.202054739	3.4500036231865\\
6.222176075	3.43000364431293\\
6.242068768	3.41500366032015\\
6.262020826	3.395\\
6.282161474	3.375\\
6.302022219	3.355\\
6.322189093	3.335\\
6.342045546	3.315\\
6.362016678	3.295\\
6.38218379	3.275\\
6.402023554	3.255\\
6.42219162	3.24000385802239\\
6.442023993	3.22\\
6.462028742	3.2\\
6.482155561	3.18\\
6.502026796	3.16500394944461\\
6.522170305	3.14500397456029\\
6.542007208	3.12500399999744\\
6.562035084	3.11000401929001\\
6.582168818	3.0900040453048\\
6.60201931	3.075\\
6.62219882	3.055\\
6.642049789	3.04\\
6.662009001	3.02\\
6.68214035	3.0050041597309\\
6.702040672	2.985\\
6.722191811	2.97\\
6.742039204	2.95500423011542\\
6.762017965	2.94000425169761\\
6.782172441	2.925\\
6.802040339	2.91\\
6.822195053	2.89\\
6.842060566	2.87\\
6.862022638	2.85000438596154\\
6.882184267	2.83000441695768\\
6.902058125	2.81000444839506\\
6.922169924	2.79000448028314\\
6.942036152	2.7700045126317\\
6.962060928	2.75000454545079\\
6.982192516	2.725\\
7.001996994	2.705\\
7.022202253	2.685\\
7.042022705	2.665\\
7.062073708	2.645\\
7.082153797	2.625\\
7.102023363	2.6\\
7.122164726	2.58\\
7.141997576	2.565\\
7.162062645	2.54\\
7.182226181	2.52\\
7.202002287	2.5\\
7.222168207	2.48\\
7.242021084	2.455\\
7.262044907	2.435\\
7.282160997	2.41500517597789\\
7.302033901	2.3900052301198\\
7.322174311	2.37\\
7.342018366	2.355\\
7.362043142	2.35\\
7.382188559	2.34\\
7.402025938	2.34\\
7.422181129	2.33500535331292\\
7.442045927	2.33000536480069\\
7.461986542	2.32500537633787\\
7.482157707	2.32500537633787\\
7.502041578	2.32000538792478\\
7.52215457	2.32000538792478\\
7.541994095	2.31500539956174\\
7.562044144	2.31000541124907\\
7.582208633	2.31000541124907\\
7.601991415	2.31000541124907\\
7.622181654	2.31000541124907\\
7.642034292	2.31\\
7.662019968	2.31000541124907\\
7.682176352	2.31\\
7.702009916	2.31\\
7.722177982	2.315\\
7.742027044	2.315\\
7.76205492	2.32\\
7.78215456	2.32\\
7.802002907	2.325\\
7.82217288	2.33\\
7.842028141	2.335\\
7.862007141	2.34\\
7.882144928	2.35\\
7.902017117	2.355\\
7.922207117	2.365\\
7.941999197	2.375\\
7.962014675	2.38500524108439\\
7.982160568	2.39500521920099\\
8.00202775	2.40500519749958\\
8.022207737	2.42000516528374\\
8.042006493	2.43500513346481\\
8.062014818	2.4500051020355\\
8.082175493	2.4650202838922\\
8.102050066	2.48002016120837\\
8.12214613	2.49502003999968\\
8.142054081	2.51001992023968\\
8.162019491	2.53001976276866\\
8.182166815	2.55001960776775\\
8.202024937	2.57001945517928\\
8.222177029	2.59001930494736\\
8.242496252	2.61000478926764\\
8.262017727	2.63000475284742\\
8.282183409	2.65000471697693\\
8.302018642	2.67000468164384\\
8.322143316	2.69500463821493\\
8.341991186	2.72000459558435\\
8.362090826	2.745\\
8.382203579	2.76500452079196\\
8.402018785	2.79000448028314\\
8.422168493	2.81500444049383\\
8.442065001	2.84500439366972\\
8.462030411	2.87000435539739\\
8.482177019	2.89500431778607\\
8.502048016	2.92000428081878\\
8.522155046	2.95000423728509\\
8.541997671	2.97500420167771\\
8.562011242	3.0050041597309\\
8.58215785	3.03500411861335\\
8.602002859	3.06000408496459\\
8.622174025	3.09501615504669\\
8.642019033	3.12501599995904\\
8.662014723	3.15501584782074\\
8.682170153	3.19003526626274\\
8.702004671	3.22003493769866\\
8.72217226	3.25506144335249\\
8.742035627	3.28506088223643\\
8.762006521	3.32006024041733\\
8.782174349	3.35505961198903\\
8.802015543	3.39005899653679\\
8.822201967	3.4250583936628\\
8.84207201	3.4600325142981\\
8.862038374	3.49503218869297\\
8.882151842	3.53003186954452\\
8.90201664	3.56503155666258\\
8.9221735	3.6000138888621\\
8.942082882	3.63503094897416\\
8.962004662	3.6750306121174\\
8.982151985	3.71003032332621\\
9.002033949	3.75002999988\\
9.022185564	3.78502972247247\\
9.042088509	3.82005235566216\\
9.062009811	3.85505188032535\\
9.082170963	3.89505134754344\\
9.102047443	3.93007951573502\\
9.122205019	3.96507881384469\\
9.142018557	4.00011249841801\\
9.162004709	4.04011138460315\\
9.182167768	4.07511042795162\\
9.202019215	4.11514884299462\\
9.222215652	4.15014758773709\\
9.2420578	4.18519115453524\\
9.262014627	4.22523963817438\\
9.282175064	4.26029341712516\\
9.302053452	4.3002906878489\\
9.322166443	4.33034929307094\\
9.342011452	4.36534649254787\\
9.362027884	4.40028408173836\\
9.382169485	4.43528183997365\\
9.40199542	4.47027963331155\\
9.422189236	4.50527746093401\\
9.442024231	4.54022301214379\\
9.462031126	4.5752213061228\\
9.482210636	4.61017353252565\\
9.50201416	4.64017241058993\\
9.522164106	4.67017130306801\\
9.541999578	4.70517002880874\\
9.562057734	4.73516895157924\\
9.58217001	4.7702122594283\\
9.602027416	4.80021093286535\\
9.62221837	4.83025879223878\\
9.642043591	4.86031120402799\\
9.662041426	4.89530897901246\\
9.682175636	4.92536546867337\\
9.702047348	4.95536325610948\\
9.72215867	4.98536107017335\\
9.742034674	5.01542121860168\\
9.762027502	5.04541871404148\\
9.782195807	5.07541623908818\\
9.802001715	5.1054137932199\\
9.822241306	5.13547709565528\\
9.842035294	5.16547432478374\\
9.861996412	5.19547158591018\\
9.882143974	5.22546887848354\\
9.902062178	5.25540198272216\\
9.92215991	5.29039932330254\\
9.941990614	5.32046050638476\\
9.962040186	5.3504579243276\\
9.982160091	5.37545579462802\\
10.002008915	5.40545326499083\\
10.022162676	5.43045117830922\\
10.042040348	5.45538724564994\\
10.062033176	5.48532815791362\\
10.082159519	5.51532637293569\\
10.102014542	5.55032431484864\\
10.122166157	5.58027105076447\\
10.14205265	5.61026960136498\\
10.162023067	5.64026816738353\\
10.182164669	5.67031745143074\\
10.20201087	5.70037060198721\\
10.222209692	5.7354903887985\\
10.242053509	5.7605555287663\\
10.262000322	5.79062388693999\\
10.282187939	5.82062067137174\\
10.301998377	5.84561801694226\\
10.32216692	5.87561486144216\\
10.342014313	5.90561173799971\\
10.362031221	5.93553914990037\\
10.382137775	5.96053688856969\\
10.402019739	5.98546990636491\\
10.422216654	6.00535177987102\\
10.442014933	6.03025082397076\\
10.462009192	6.05516721486698\\
10.482165575	6.08010073929701\\
10.502165794	6.10007377004574\\
10.522184372	6.11505110362947\\
10.542012691	6.14005089555453\\
10.562042475	6.16507299226862\\
10.582167625	6.18509902911829\\
10.602003336	6.21012882314047\\
10.62219882	6.23016251794446\\
10.642024279	6.2501999968001\\
10.662052393	6.27024122342993\\
10.682150364	6.29533557485223\\
10.702017546	6.31038826063817\\
10.722202063	6.33557021900949\\
10.742022038	6.35563726151831\\
10.762051105	6.37070835935848\\
10.782175779	6.39063377138762\\
10.802012682	6.40563228729218\\
10.822185516	6.42556223220972\\
10.842025518	6.44543637002182\\
10.862024546	6.46532675121683\\
10.882167339	6.48023340629024\\
10.901996136	6.49512317050262\\
10.922170639	6.51506907100761\\
10.942038536	6.53004785587365\\
10.962023258	6.55001717555\\
10.982150793	6.56501713630665\\
11.002026796	6.58000759877981\\
11.022168398	6.59500758149678\\
11.042107582	6.60501703252914\\
11.062018871	6.62003021141143\\
11.082143784	6.63004713407077\\
11.102014303	6.64006777073849\\
11.122162104	6.65009210462532\\
11.142023325	6.660091966332\\
11.162003279	6.66512002892671\\
11.182168961	6.67009182845334\\
11.202023745	6.67504681631522\\
11.222198009	6.68002994005266\\
11.242036581	6.68501682869984\\
11.262020588	6.68500186985763\\
11.282158136	6.69\\
11.302039862	6.69000186846013\\
11.322206497	6.68500186985763\\
11.341994762	6.68500186985763\\
11.362016439	6.68000187125722\\
11.382194281	6.6750074906325\\
11.402049065	6.67000749624766\\
11.42216301	6.66501687919843\\
11.44204402	6.66003002996233\\
11.462008238	6.65003007511996\\
11.482185125	6.64504702767407\\
11.502063513	6.63504709855175\\
11.522179365	6.6250471696434\\
11.542031527	6.62004720526976\\
11.562044382	6.61004727668418\\
11.582202435	6.60004734831501\\
11.602005959	6.59006828492695\\
11.622162104	6.58006838870235\\
11.642018795	6.56509329712838\\
11.662035704	6.55012213626586\\
11.682209015	6.53512241660399\\
11.701995373	6.52015528956175\\
11.722155094	6.50519215703887\\
11.742041111	6.48519274964129\\
11.76204586	6.4701931964973\\
11.782194138	6.45019379553824\\
11.802010298	6.43519424726247\\
11.822166204	6.41519485284742\\
11.842052221	6.39519546222006\\
11.862076521	6.38019592175664\\
11.882140398	6.35523799711702\\
11.902008533	6.33523874846087\\
11.922194719	6.31523950456354\\
11.942014456	6.29524026547041\\
11.962010145	6.27519920002545\\
11.982151985	6.25524180188104\\
12.002012491	6.23024277215583\\
12.022212505	6.21024355400012\\
12.042027712	6.18529101983084\\
12.062018633	6.16034292876622\\
12.082137346	6.14039900983641\\
12.102040052	6.1105237091431\\
12.122159719	6.08059413215518\\
12.142022371	6.05566883176417\\
12.162007332	6.02567216167624\\
12.182183743	5.9956755249096\\
12.202021599	5.96067949146739\\
12.22219348	5.93068292863478\\
12.242077351	5.89568698287146\\
12.262014866	5.86069108552908\\
12.282147408	5.82562013866335\\
12.302027941	5.79062388693999\\
12.322198868	5.75562768080077\\
12.342032194	5.72055941320427\\
12.362026215	5.68056335234455\\
12.382187128	5.64549820653589\\
12.402066231	5.60543709268064\\
12.422172785	5.56537959172598\\
12.442011118	5.52538233971189\\
12.462028265	5.48538512777362\\
12.482155561	5.45038760089592\\
12.50212431	5.41045284611187\\
12.522170067	5.37052371747858\\
12.542024374	5.33559977884399\\
12.562016249	5.30068155995057\\
12.582173347	5.26076990563168\\
12.60203433	5.22577506213193\\
12.62221384	5.18587022591194\\
12.642044067	5.15097078228949\\
12.662013531	5.11107865327858\\
12.682140589	5.08119080531326\\
12.702059507	5.04131183324341\\
12.722218513	5.00132232514562\\
12.742001295	4.96644993934299\\
12.762009859	4.92634245663048\\
12.782172441	4.89123706233914\\
12.802034855	4.85602975691048\\
12.822225571	4.82093611241634\\
12.842023134	4.78075569340246\\
12.862039804	4.75067363644357\\
12.882176399	4.71559646280298\\
12.902061462	4.68560028171418\\
12.922202587	4.65060479937825\\
12.942026854	4.62053027259859\\
12.962030411	4.59061270420409\\
12.982169151	4.56061673460948\\
13.002050877	4.54061945113219\\
13.022166729	4.52062219168999\\
13.042048693	4.50062495660325\\
13.062007904	4.48071422878094\\
13.082170963	4.47080809250408\\
13.102045774	4.45081172371962\\
13.122222662	4.44091206848323\\
13.142009258	4.42601965653114\\
13.162029028	4.41602196552508\\
13.182162762	4.40625124113458\\
13.202048779	4.39637634876724\\
13.222162247	4.38664165393071\\
13.24207449	4.37178739190277\\
13.262017488	4.36193764283718\\
13.282202244	4.35209432342637\\
13.302010059	4.34209914672615\\
13.322175026	4.33694881224116\\
13.342008591	4.33180389676172\\
13.362028837	4.33152686705277\\
13.382196903	4.32627437872357\\
13.402005672	4.32093739829681\\
13.422161102	4.32074067724505\\
13.442029476	4.31556774943923\\
13.462056637	4.31049011134465\\
13.48216176	4.31049011134465\\
13.502006292	4.31049011134465\\
13.522183418	4.30549068051482\\
13.542004347	4.30556906807915\\
13.56208086	4.30565326054015\\
13.582190752	4.30574325755728\\
13.602019787	4.3108380855699\\
13.622189045	4.3108380855699\\
13.642015696	4.31093957276137\\
13.662033081	4.31093957276137\\
13.682160139	4.31093957276137\\
13.702014208	4.31593848426967\\
13.722155094	4.31593848426967\\
13.741992235	4.31593848426967\\
13.76205492	4.31593848426967\\
13.782170057	4.32093739829681\\
13.801992893	4.32583517947691\\
13.822155237	4.32583517947691\\
13.842068911	4.3308342152523\\
13.86199522	4.3308342152523\\
13.882196903	4.33583325325133\\
13.902000666	4.34083229346631\\
13.922158957	4.3458313358896\\
13.942016602	4.35073556999273\\
13.96201849	4.35573472562322\\
13.982173681	4.36073388318985\\
14.002015829	4.36573304268596\\
14.022208691	4.37073220410494\\
14.042024612	4.37573136744019\\
14.062039852	4.38572969983331\\
14.082187176	4.39072886887815\\
14.102016687	4.40072721263202\\
14.122161388	4.4057263873282\\
14.142022133	4.41572474232713\\
14.162013292	4.42572310475927\\
14.182175159	4.43563411475744\\
14.201994419	4.45063197759599\\
14.222209215	4.4606305608064\\
14.242069721	4.470548064835\\
14.262035131	4.48054684162547\\
14.282157898	4.49554501701407\\
14.302041531	4.50546889901595\\
14.322193146	4.52554140407532\\
14.342046499	4.54053961550827\\
14.362018108	4.55553783871894\\
14.382190943	4.57053607359137\\
14.402009964	4.59053373803091\\
14.422169924	4.60545871765235\\
14.442035675	4.62545673420474\\
14.462081909	4.6454547678349\\
14.48216176	4.66545281832321\\
14.502019167	4.68545088545382\\
14.52216506	4.71052014113091\\
14.542030334	4.73051794204398\\
14.562053204	4.75559144586664\\
14.582164764	4.77558896891263\\
14.602011204	4.8006666203768\\
14.62218523	4.82566316686111\\
14.64201045	4.85574402125977\\
14.66202426	4.880740210255\\
14.682141304	4.90565235213422\\
14.702003956	4.93564838699031\\
14.722154617	4.96056700388171\\
14.742044926	4.99049095781167\\
14.762069941	5.02042079909643\\
14.782160044	5.05041829950748\\
14.802019119	5.07541623908818\\
14.822211266	5.10535258331881\\
14.842029095	5.13535052357675\\
14.862046957	5.16534848775956\\
14.882148504	5.20034614232553\\
14.902021885	5.23034415693652\\
14.922165394	5.26534186924268\\
14.942038774	5.29533993243116\\
14.962046146	5.33033770037134\\
14.982152224	5.36533549743164\\
15.002025843	5.4003333230459\\
15.022171974	5.43533117666256\\
15.042080164	5.47027650123831\\
15.062054157	5.50527474337112\\
15.082161188	5.54027300771361\\
15.102033854	5.57532286060637\\
15.122178078	5.61532056075163\\
15.142013073	5.65031857508937\\
15.162030697	5.69031633567063\\
15.182167053	5.72531440184729\\
15.202033281	5.7653122205133\\
15.222221851	5.80531006923833\\
15.2420187	5.84530794740534\\
15.262054443	5.88530585441402\\
15.282158136	5.92525526876269\\
15.302040577	5.96525355705858\\
15.322177887	6.00025207803806\\
15.342006445	6.04020695009699\\
15.362061739	6.07516666438049\\
15.382174253	6.11513082443867\\
15.401994705	6.15009959268954\\
15.422181129	6.19007269747295\\
15.442026138	6.23003210264602\\
15.462063313	6.2650179568777\\
15.482153177	6.30000793650294\\
15.502030611	6.33500197316465\\
15.522151232	6.37\\
15.542061329	6.4\\
15.562007427	6.435\\
15.582166672	6.465\\
15.602015734	6.495\\
15.622195959	6.53000191424168\\
15.642052889	6.56000190548753\\
15.662012339	6.59000189681308\\
15.682186842	6.62000188821725\\
15.702052832	6.64500188111335\\
15.722197056	6.67500187265891\\
15.74203229	6.70000186567138\\
15.762060881	6.7250018587358\\
15.782194376	6.7500018518516\\
15.802023888	6.77000184638084\\
15.822170258	6.79500183958768\\
15.842051506	6.815\\
15.862038612	6.84000182748514\\
15.88216114	6.86000728862587\\
15.901991606	6.88001635172475\\
15.922177792	6.90002898544637\\
15.942011356	6.92004515881219\\
15.962031841	6.93506488794445\\
15.982192993	6.95006474789984\\
16.002021551	6.96508793914334\\
16.022186518	6.98011461223954\\
16.042070389	6.99514474474975\\
16.062039375	7.00517844169583\\
16.082175016	7.01517818733067\\
16.102033615	7.02521529919191\\
16.122169256	7.03521499316119\\
16.142048597	7.04021484047184\\
16.162024975	7.04521468799922\\
16.182143211	7.05517717708067\\
16.202010155	7.0601770516043\\
16.222177744	7.06517692630553\\
16.242027044	7.07017680118397\\
16.262030125	7.0802136267206\\
16.282165766	7.08534579254958\\
16.302013397	7.09545100751178\\
16.322194338	7.10057039962284\\
16.342005014	7.10570369492002\\
16.362016916	7.11085086329337\\
16.382160902	7.11601187463877\\
16.40202713	7.12101116415359\\
16.422223806	7.12609640686961\\
16.442056656	7.12609640686961\\
16.462023735	7.131009746172\\
16.482177973	7.12592800974021\\
16.502076387	7.13084847686445\\
16.522184372	7.13070122778959\\
16.542040825	7.13070122778959\\
16.562006235	7.12563330518769\\
16.582221985	7.12563330518769\\
16.602005959	7.12563330518769\\
16.622155905	7.12070221256303\\
16.642079353	7.12070221256303\\
16.66200304	7.12077418543798\\
16.682188034	7.12084966840334\\
16.702009439	7.12601045466536\\
16.722154617	7.12118669885855\\
16.742036343	7.12627883260261\\
16.76201272	7.12637530586202\\
16.782194853	7.13127793596632\\
16.802005053	7.13127793596632\\
16.822202444	7.13618420446109\\
16.842010736	7.1360948704456\\
16.861999512	7.13600903867141\\
16.882180214	7.14092606039301\\
16.902031898	7.14584669580869\\
16.922189713	7.1508461037838\\
16.942047596	7.15577039877608\\
16.962030411	7.15569877789724\\
16.982206821	7.16069828997145\\
17.002003193	7.16562976995044\\
17.022186279	7.17556443772892\\
17.042018175	7.17550346665654\\
17.062042475	7.18044566861974\\
17.082158804	7.18544535850075\\
17.102040529	7.19044504881304\\
17.122158766	7.19544473955571\\
17.142002344	7.20044443072787\\
17.162016392	7.20550137048075\\
17.182179689	7.21050102281388\\
17.201996088	7.22050032892458\\
17.222167015	7.22549998270016\\
17.2420187	7.23056014427651\\
17.262016535	7.23549929168679\\
17.282165527	7.24055937065639\\
17.302018881	7.25049825874057\\
17.322158813	7.25544106171362\\
17.342066288	7.26544045464554\\
17.36203742	7.27538658766667\\
17.382156134	7.28043954717021\\
17.401993513	7.29038579226093\\
17.422166586	7.30043834300379\\
17.442014456	7.31543744419977\\
17.462014198	7.32543684704196\\
17.482141733	7.33543625151224\\
17.502054214	7.34543565760398\\
17.522158623	7.36043476976734\\
17.54201293	7.37043417988384\\
17.562024117	7.38543329805368\\
17.582157373	7.40048815957434\\
17.602024794	7.41543154509567\\
17.622184992	7.43043067392463\\
17.642039776	7.44537776073182\\
17.662021875	7.46037700119773\\
17.6821661	7.47532775201195\\
17.702030182	7.49028203741354\\
17.722173929	7.50523983627439\\
17.742020369	7.52523919885607\\
17.762018919	7.54023872301136\\
17.782183647	7.55520019589157\\
17.802010298	7.57523762003543\\
17.82222414	7.5902371504453\\
17.842059374	7.61023652720466\\
17.862042665	7.62523606191966\\
17.882157326	7.65023529049924\\
17.902052641	7.67019719433601\\
17.922207832	7.68519680945127\\
17.942019939	7.71016212540307\\
17.962049961	7.73013098207268\\
17.982177734	7.75510315856598\\
18.002047062	7.7800787271081\\
18.022161961	7.80507847494181\\
18.042015553	7.83007822438576\\
18.062008619	7.85505728814246\\
18.082187176	7.88005710639206\\
18.102060556	7.91005688980806\\
18.12223959	7.9400566748607\\
18.142025948	7.96503923405277\\
18.162018299	7.99503908683378\\
18.18220377	8.02002493761709\\
18.20204711	8.05002484468216\\
18.222159147	8.08001392325533\\
18.242021084	8.11501386320442\\
18.262038469	8.14501381214298\\
18.282178402	8.18001375304467\\
18.302013874	8.21001370279003\\
18.322201729	8.24501364462182\\
18.342014313	8.28001358694537\\
18.362061024	8.31502405288163\\
18.382189989	8.35501346498017\\
18.402012348	8.39500595592403\\
18.422197104	8.43000593119602\\
18.44211483	8.4700014757968\\
18.46204257	8.505\\
18.482144117	8.545\\
18.502032518	8.585\\
18.522205114	8.625\\
18.542003393	8.66000144341789\\
18.56207037	8.70000143678149\\
18.58215332	8.73500572409658\\
18.601989031	8.77000570125242\\
18.622177839	8.81000567536707\\
18.642030239	8.8450014132277\\
18.662031651	8.885\\
18.682155609	8.92000140134518\\
18.702023029	8.96000558035541\\
18.722157001	8.99501250693961\\
18.742028713	9.03001245846317\\
18.762127161	9.07001240351963\\
18.782155275	9.10500549148654\\
18.802005768	9.14000547045788\\
18.822184801	9.17500136239772\\
18.842027903	9.21500135648389\\
18.862023354	9.25000135135125\\
18.88215661	9.29000134553274\\
18.902007818	9.33000133976411\\
18.922172785	9.36500533902677\\
18.942003012	9.40501196171488\\
18.962057114	9.44501191105654\\
18.982169151	9.48001186708118\\
19.002014637	9.52001181721956\\
19.022171259	9.55501177393309\\
19.042019367	9.59000521376292\\
19.062022209	9.63001168223591\\
19.082147837	9.66001164595571\\
19.102004051	9.70002061853479\\
19.122156858	9.74003208413607\\
19.142049789	9.77503196925718\\
19.162013531	9.81504584808446\\
19.182175159	9.85504566199467\\
19.202008724	9.89003159752283\\
19.222201824	9.93503145440416\\
19.242077351	9.97502005010516\\
19.262018919	10.01001998\\
19.282177448	10.0500199004778\\
19.302042007	10.0900198215861\\
19.32219243	10.1300308489165\\
19.342000008	10.1750307124844\\
19.36203742	10.2100306072019\\
19.3821702	10.2500304877595\\
19.402042627	10.2900303692458\\
19.422222853	10.3300435623476\\
19.442041636	10.370030134961\\
19.462012053	10.410043227576\\
19.482167006	10.4550430415183\\
19.502070189	10.4950428774732\\
19.522199869	10.5350581393745\\
19.542028427	10.5750579194631\\
19.562001705	10.620057674043\\
19.582165003	10.6600574576313\\
19.602050543	10.7000572428375\\
19.622229338	10.7400570296437\\
19.642035484	10.7750417632601\\
19.662012577	10.8150566341559\\
19.682184458	10.8550564254637\\
19.702040434	10.8950562183038\\
19.722159863	10.9350731593346\\
19.742070913	10.9700729259199\\
19.76202178	11.0100919614688\\
19.782196522	11.0500916285794\\
19.802010298	11.0900912980913\\
19.822179794	11.1250910108637\\
19.84202981	11.1601120066064\\
19.862037182	11.2001350438287\\
19.88216424	11.240160141208\\
19.902036905	11.2751596441026\\
19.922178745	11.3101867800669\\
19.942070246	11.3401587290478\\
19.962021589	11.3701583102435\\
19.982163429	11.3951579629244\\
20.002016068	11.4251323843534\\
20.022198677	11.450157204161\\
20.042007446	11.4801567933543\\
20.062027216	11.505156452652\\
20.082159042	11.5301832162373\\
20.102087975	11.5651826617654\\
20.122168541	11.5951821891681\\
20.142026424	11.6251548376785\\
20.162014246	11.6501545054132\\
20.182190418	11.6801541085724\\
20.202024937	11.7051537794255\\
20.222181797	11.7351533863005\\
20.242038488	11.7601796329818\\
20.261988878	11.7851792519249\\
20.282178164	11.8101788724811\\
20.302005768	11.8301785700808\\
20.322232485	11.8551781935153\\
20.342035294	11.8751778934044\\
20.36203599	11.9052057941054\\
20.382168293	11.9302053628594\\
20.401994467	11.9552352548998\\
};
\addplot [color=black, forget plot]
  table[row sep=crcr]{%
4.25	0\\
4.25	30\\
};
\node[right, align=left, circle,fill=white,draw=black,thick,line width=0.3mm,inner sep=1pt]
at (axis cs:2.35,22) {1};
\addplot [color=black, forget plot]
  table[row sep=crcr]{%
7.1800000667572	0\\
7.1800000667572	30\\
};
\node[right, align=left, circle,fill=white,draw=black,thick,line width=0.3mm,inner sep=1pt]
at (axis cs:5.28,22) {2};
\addplot [color=black, forget plot]
  table[row sep=crcr]{%
11.9800000190735	0\\
11.9800000190735	30\\
};
\node[right, align=left, circle,fill=white,draw=black,thick,line width=0.3mm,inner sep=1pt]
at (axis cs:10.08,22) {3};
\end{axis}
\end{tikzpicture}%

%% file: Images/Tikz/fig10-3-2.tikz
\begin{tikzpicture}

\begin{axis}[%
width=4.2cm,
height=1.683cm,
at={(0cm,0cm)},
scale only axis,
xmin=0.122153044,
xmax=20.401994467,
xlabel style={font=\color{white!15!black}},
xlabel={$t\;[\SI[per-mode=repeated-symbol]{}{\second}]$},
ymin=-4.5,
ymax=3.5,
ylabel style={font=\color{white!15!black}},
ylabel={$a\;[\SI[per-mode=repeated-symbol]{}{\metre\per\second\squared}]$},
axis background/.style={fill=white},
ylabel style={yshift=-0.4cm}
]
\addplot [color=blue, line width=1.0pt, forget plot]
  table[row sep=crcr]{%
0.122153044	-0.827964\\
0.142704725	-0.914292\\
0.162019968	-0.812268\\
0.182151318	-0.847584\\
0.202072382	-0.871128\\
0.222170353	-0.82404\\
0.242097378	-0.82404\\
0.262028217	-0.82404\\
0.282156229	-0.7848\\
0.302108049	-0.792648\\
0.322176456	-0.831888\\
0.342110395	-0.80442\\
0.362020254	-0.820116\\
0.382161379	-0.722016\\
0.402102709	-0.84366\\
0.422178984	-1.024164\\
0.442130327	-0.992772\\
0.46200633	-0.875052\\
0.482154846	-0.855432\\
0.502096176	-0.682776\\
0.522157907	-1.145808\\
0.542132139	-0.937836\\
0.562018394	-0.878976\\
0.582155943	-0.929988\\
0.602100611	-0.769104\\
0.622170925	-1.016316\\
0.642130375	-1.028088\\
0.661995411	-1.051632\\
0.682149649	-0.84366\\
0.702078104	-0.926064\\
0.722169399	-1.071252\\
0.742068052	-1.086948\\
0.76201272	-1.051632\\
0.782187939	-1.106568\\
0.802058935	-1.083024\\
0.822188854	-1.016316\\
0.84211874	-1.232136\\
0.862021446	-0.776952\\
0.882141829	-1.271376\\
0.902095318	-0.977076\\
0.922189951	-1.208592\\
0.942067385	-0.937836\\
0.962010384	-0.945684\\
0.982141018	-0.984924\\
1.00208354	-1.075176\\
1.022191286	-1.067328\\
1.04208827	-0.981\\
1.062000513	-0.973152\\
1.082139254	-1.083024\\
1.10210681	-1.024164\\
1.122172117	-0.94176\\
1.142074108	-1.083024\\
1.162016153	-0.937836\\
1.182153702	-0.800496\\
1.202116489	-1.361628\\
1.222205639	-0.992772\\
1.242153645	-0.96138\\
1.26202631	-1.102644\\
1.282181978	-0.851508\\
1.302097559	-0.894672\\
1.322151184	-1.161504\\
1.342127562	-1.094796\\
1.362022638	-0.871128\\
1.382176876	-0.996696\\
1.402080297	-1.047708\\
1.422161579	-0.871128\\
1.442111731	-1.043784\\
1.462035418	-1.004544\\
1.482175827	-1.051632\\
1.502425671	-0.933912\\
1.522183895	-0.604296\\
1.542054653	-1.25568\\
1.562038183	-0.953532\\
1.582180977	-0.96138\\
1.602072954	-1.1772\\
1.622181654	-0.86328\\
1.64208889	-0.514044\\
1.662036657	-1.00062\\
1.682142735	-1.19682\\
1.702091694	-0.906444\\
1.722146988	-0.988848\\
1.742088318	-1.05948\\
1.762031794	-0.710244\\
1.782210588	-1.15758\\
1.802090406	-1.090872\\
1.822175741	-0.835812\\
1.842198849	-1.032012\\
1.862034321	-0.992772\\
1.882190228	-0.996696\\
1.902107477	-0.894672\\
1.922172308	-1.145808\\
1.942081451	-0.733788\\
1.962022543	-1.055556\\
1.982192516	-1.086948\\
2.002118826	-0.769104\\
2.02216363	-1.09872\\
2.04212141	-0.918216\\
2.062012196	-0.988848\\
2.082183361	-1.067328\\
2.102087498	-0.855432\\
2.122143269	-1.106568\\
2.142081261	-0.94176\\
2.162018299	-1.149732\\
2.182207823	-1.028088\\
2.202086687	-0.886824\\
2.222201586	-1.071252\\
2.242088079	-0.965304\\
2.262051105	-1.122264\\
2.282233238	-1.21644\\
2.302112341	-0.898596\\
2.322165251	-1.134036\\
2.342108011	-0.992772\\
2.362074614	-1.028088\\
2.382153273	-1.024164\\
2.402099133	-1.051632\\
2.422190428	-1.15758\\
2.44206953	-0.973152\\
2.462044001	-0.981\\
2.482162237	-1.126188\\
2.50207901	-0.867204\\
2.522173405	-1.114416\\
2.542060137	-1.106568\\
2.562032938	-0.867204\\
2.582175016	-1.200744\\
2.602091312	-0.929988\\
2.622170687	-0.988848\\
2.642087221	-1.181124\\
2.661997318	-1.075176\\
2.682151079	-1.00062\\
2.702081203	-1.016316\\
2.722151279	-0.878976\\
2.742081404	-1.071252\\
2.762029409	-1.204668\\
2.782158852	-1.02024\\
2.802078962	-0.906444\\
2.822157145	-0.835812\\
2.842126608	-0.996696\\
2.862025261	-1.102644\\
2.882153511	-0.969228\\
2.902089834	-0.906444\\
2.922161102	-0.926064\\
2.9420681	-0.875052\\
2.962011576	-1.094796\\
2.982153654	-0.984924\\
3.002092123	-0.886824\\
3.022182465	-0.871128\\
3.042096138	-0.80442\\
3.062021494	-0.8829\\
3.082163334	-0.886824\\
3.102095366	-0.906444\\
3.1221807	-0.906444\\
3.142077208	-0.945684\\
3.16202116	-0.878976\\
3.182166815	-0.90252\\
3.20211935	-1.02024\\
3.222192287	-0.906444\\
3.242128134	-0.953532\\
3.262024879	-1.024164\\
3.282130003	-0.788724\\
3.302080154	-0.827964\\
3.322163582	-0.890748\\
3.342069626	-0.965304\\
3.361999273	-0.988848\\
3.382153988	-0.7848\\
3.402083397	-0.926064\\
3.422180891	-0.816192\\
3.442085266	-1.012392\\
3.462021589	-0.820116\\
3.482152939	-0.84366\\
3.502100468	-1.086948\\
3.522212744	-0.906444\\
3.542113781	-1.012392\\
3.562012196	-1.012392\\
3.582176685	-0.937836\\
3.602067471	-0.914292\\
3.622155666	-1.012392\\
3.642103434	-1.067328\\
3.662031174	-0.965304\\
3.682188272	-0.957456\\
3.702080488	-0.80442\\
3.722182035	-0.800496\\
3.742081165	-0.894672\\
3.762041092	-0.906444\\
3.782184362	-0.757332\\
3.802086115	-0.74556\\
3.822208643	-0.996696\\
3.842411518	-0.710244\\
3.862011194	-0.812268\\
3.882148266	-0.76518\\
3.902097464	-0.827964\\
3.922157049	-0.702396\\
3.942069292	-0.957456\\
3.962007523	-0.6867\\
3.982177258	-0.796572\\
4.002087593	-0.898596\\
4.022173405	-1.008468\\
4.042154551	-0.596448\\
4.06201911	-0.733788\\
4.082152367	-0.690624\\
4.102093697	-0.749484\\
4.122160673	-0.820116\\
4.142062426	-0.871128\\
4.162030935	-0.969228\\
4.182180405	-0.827964\\
4.202072144	-0.992772\\
4.222162008	-0.831888\\
4.242100716	-0.965304\\
4.262014389	-0.94176\\
4.28219533	-1.043784\\
4.302089214	-0.90252\\
4.322158575	-0.886824\\
4.342076063	-0.988848\\
4.362028837	-0.988848\\
4.382173061	-0.933912\\
4.402078867	-0.918216\\
4.4221766	-0.984924\\
4.442105055	-0.996696\\
4.46200633	-0.992772\\
4.482172966	-1.043784\\
4.502092361	-1.032012\\
4.522158146	-1.028088\\
4.542140245	-1.055556\\
4.562050581	-0.929988\\
4.582169533	-1.024164\\
4.60207653	-0.851508\\
4.622186661	-1.09872\\
4.642095804	-1.035936\\
4.662044764	-1.012392\\
4.682162046	-1.043784\\
4.702105522	-0.96138\\
4.722176552	-1.004544\\
4.742126703	-0.984924\\
4.762061596	-0.965304\\
4.782148838	-0.996696\\
4.802083969	-0.94176\\
4.82220006	-0.965304\\
4.842048645	-0.988848\\
4.862048626	-1.024164\\
4.882157326	-1.067328\\
4.902056932	-1.047708\\
4.92215395	-1.204668\\
4.942088842	-1.169352\\
4.962026834	-1.153656\\
4.982174397	-1.251756\\
5.002012014	-1.220364\\
5.022206068	-1.212516\\
5.042008877	-1.114416\\
5.062007904	-1.232136\\
5.082171679	-1.224288\\
5.10203433	-1.21644\\
5.122157097	-1.09872\\
5.141997099	-1.188972\\
5.161998987	-1.208592\\
5.182173491	-1.220364\\
5.202002525	-1.208592\\
5.222178936	-1.185048\\
5.242048025	-1.239984\\
5.261997223	-1.220364\\
5.282162666	-1.239984\\
5.302005768	-1.232136\\
5.322175503	-1.25568\\
5.342036486	-1.181124\\
5.362068892	-1.19682\\
5.382175207	-1.181124\\
5.402026415	-1.243908\\
5.422189951	-1.220364\\
5.442039967	-1.130112\\
5.462042093	-1.1772\\
5.482180834	-1.126188\\
5.502022743	-1.145808\\
5.522157431	-1.106568\\
5.542028427	-1.067328\\
5.562045097	-1.05948\\
5.582159758	-1.083024\\
5.602006435	-0.984924\\
5.622204542	-1.055556\\
5.642046213	-1.05948\\
5.662028074	-1.043784\\
5.682136536	-1.083024\\
5.702010393	-1.024164\\
5.722163677	-1.043784\\
5.741996527	-1.067328\\
5.762047768	-1.086948\\
5.782159328	-1.03986\\
5.802027225	-1.051632\\
5.822189093	-1.067328\\
5.842361212	-1.055556\\
5.862008333	-1.028088\\
5.882154942	-1.028088\\
5.902008057	-1.055556\\
5.922199488	-1.028088\\
5.941994667	-1.055556\\
5.962021589	-1.094796\\
5.98218298	-1.063404\\
6.002015829	-1.032012\\
6.022216558	-1.05948\\
6.042048216	-1.012392\\
6.062056303	-1.004544\\
6.08216095	-1.03986\\
6.102020502	-0.992772\\
6.122161865	-0.992772\\
6.142045021	-0.973152\\
6.16202569	-0.973152\\
6.182163954	-1.028088\\
6.202054739	-0.957456\\
6.222176075	-0.981\\
6.242068768	-0.918216\\
6.262020826	-1.004544\\
6.282161474	-0.981\\
6.302022219	-1.028088\\
6.322189093	-0.996696\\
6.342045546	-1.043784\\
6.362016678	-0.949608\\
6.38218379	-0.933912\\
6.402023554	-0.926064\\
6.42219162	-0.945684\\
6.442023993	-0.92214\\
6.462028742	-0.929988\\
6.482155561	-0.949608\\
6.502026796	-0.949608\\
6.522170305	-0.90252\\
6.542007208	-0.910368\\
6.562035084	-0.8829\\
6.582168818	-0.878976\\
6.60201931	-0.867204\\
6.62219882	-0.871128\\
6.642049789	-0.898596\\
6.662009001	-0.855432\\
6.68214035	-0.86328\\
6.702040672	-0.800496\\
6.722191811	-0.827964\\
6.742039204	-0.780876\\
6.762017965	-0.80442\\
6.782172441	-0.780876\\
6.802040339	-0.792648\\
6.822195053	-0.831888\\
6.842060566	-1.00062\\
6.862022638	-1.032012\\
6.882184267	-1.024164\\
6.902058125	-0.969228\\
6.922169924	-1.071252\\
6.942036152	-1.024164\\
6.962060928	-1.035936\\
6.982192516	-1.043784\\
7.001996994	-1.067328\\
7.022202253	-1.083024\\
7.042022705	-1.024164\\
7.062073708	-0.992772\\
7.082153797	-1.016316\\
7.102023363	-1.024164\\
7.122164726	-1.028088\\
7.141997576	-0.988848\\
7.162062645	-1.05948\\
7.182226181	-1.05948\\
7.202002287	-1.09872\\
7.222168207	-1.051632\\
7.242021084	-1.067328\\
7.262044907	-1.067328\\
7.282160997	-1.090872\\
7.302033901	-1.032012\\
7.322174311	-1.035936\\
7.342018366	-0.733788\\
7.362043142	-0.325692\\
7.382188559	-0.427716\\
7.402025938	-0.188352\\
7.422181129	-0.239364\\
7.442045927	-0.192276\\
7.461986542	-0.172656\\
7.482157707	-0.149112\\
7.502041578	-0.184428\\
7.52215457	-0.070632\\
7.541994095	-0.184428\\
7.562044144	-0.074556\\
7.582208633	-0.074556\\
7.601991415	-0.011772\\
7.622181654	-0.03924\\
7.642034292	0.015696\\
7.662019968	-0.035316\\
7.682176352	0.023544\\
7.702009916	0.074556\\
7.722177982	0.047088\\
7.742027044	0.105948\\
7.76205492	0.160884\\
7.78215456	0.15696\\
7.802002907	0.231516\\
7.82217288	0.207972\\
7.842028141	0.23544\\
7.862007141	0.345312\\
7.882144928	0.364932\\
7.902017117	0.396324\\
7.922207117	0.47088\\
7.941999197	0.443412\\
7.962014675	0.45126\\
7.982160568	0.533664\\
8.00202775	0.592524\\
8.022207737	0.655308\\
8.042006493	0.6867\\
8.062014818	0.729864\\
8.082175493	0.761256\\
8.102050066	0.7848\\
8.12214613	0.808344\\
8.142054081	0.847584\\
8.162019491	0.92214\\
8.182166815	0.937836\\
8.202024937	0.965304\\
8.222177029	1.004544\\
8.242496252	1.032012\\
8.262017727	1.055556\\
8.282183409	1.051632\\
8.302018642	1.075176\\
8.322143316	1.126188\\
8.341991186	1.185048\\
8.362090826	1.19682\\
8.382203579	1.243908\\
8.402018785	1.259604\\
8.422168493	1.25568\\
8.442065001	1.290996\\
8.462030411	1.267452\\
8.482177019	1.330236\\
8.502048016	1.322388\\
8.522155046	1.377324\\
8.541997671	1.35378\\
8.562011242	1.39302\\
8.58215785	1.4715\\
8.602002859	1.455804\\
8.622174025	1.5696\\
8.642019033	1.553904\\
8.662014723	1.561752\\
8.682170153	1.60884\\
8.702004671	1.60884\\
8.72217226	1.68732\\
8.742035627	1.644156\\
8.762006521	1.738332\\
8.782174349	1.703016\\
8.802015543	1.738332\\
8.822201967	1.742256\\
8.84207201	1.742256\\
8.862038374	1.801116\\
8.882151842	1.730484\\
8.90201664	1.789344\\
8.9221735	1.801116\\
8.942082882	1.801116\\
8.962004662	1.871748\\
8.982151985	1.812888\\
9.002033949	1.84428\\
9.022185564	1.7658\\
9.042088509	1.867824\\
9.062009811	1.801116\\
9.082170963	1.848204\\
9.102047443	1.828584\\
9.122205019	1.836432\\
9.142018557	1.781496\\
9.162004709	1.918836\\
9.182167768	1.840356\\
9.202019215	1.907064\\
9.222215652	1.797192\\
9.2420578	1.90314\\
9.262014627	1.769724\\
9.282175064	1.907064\\
9.302053452	1.816812\\
9.322166443	1.781496\\
9.342011452	1.710864\\
9.362027884	1.7658\\
9.382169485	1.718712\\
9.40199542	1.773648\\
9.422189236	1.80504\\
9.442024231	1.750104\\
9.462031126	1.679472\\
9.482210636	1.730484\\
9.50201416	1.600992\\
9.522164106	1.691244\\
9.541999578	1.640232\\
9.562057734	1.600992\\
9.58217001	1.561752\\
9.602027416	1.612764\\
9.62221837	1.514664\\
9.642043591	1.593144\\
9.662041426	1.561752\\
9.682175636	1.636308\\
9.702047348	1.49112\\
9.72215867	1.54998\\
9.742034674	1.51074\\
9.762027502	1.561752\\
9.782195807	1.573524\\
9.802001715	1.565676\\
9.822241306	1.542132\\
9.842035294	1.522512\\
9.861996412	1.502892\\
9.882143974	1.597068\\
9.902062178	1.526436\\
9.92215991	1.5696\\
9.941990614	1.538208\\
9.962040186	1.534284\\
9.982160091	1.3734\\
10.002008915	1.447956\\
10.022162676	1.271376\\
10.042040348	1.440108\\
10.062033176	1.444032\\
10.082159519	1.593144\\
10.102014542	1.620612\\
10.122166157	1.542132\\
10.14205265	1.573524\\
10.162023067	1.506816\\
10.182164669	1.64808\\
10.20201087	1.546056\\
10.222209692	1.6677\\
10.242053509	1.440108\\
10.262000322	1.498968\\
10.282187939	1.396944\\
10.301998377	1.455804\\
10.32216692	1.577448\\
10.342014313	1.39302\\
10.362031221	1.5696\\
10.382137775	1.23606\\
10.402019739	1.263528\\
10.422216654	1.188972\\
10.442014933	1.298844\\
10.462009192	1.232136\\
10.482165575	1.169352\\
10.502165794	1.130112\\
10.522184372	0.886824\\
10.542012691	1.15758\\
10.562042475	1.169352\\
10.582167625	1.106568\\
10.602003336	1.220364\\
10.62219882	1.05948\\
10.642024279	1.13796\\
10.662052393	1.047708\\
10.682150364	1.130112\\
10.702017546	0.918216\\
10.722202063	1.094796\\
10.742022038	1.00062\\
10.762051105	0.910368\\
10.782175779	0.926064\\
10.802012682	0.890748\\
10.822185516	0.973152\\
10.842025518	1.035936\\
10.862024546	0.949608\\
10.882167339	0.839736\\
10.901996136	0.859356\\
10.922170639	0.949608\\
10.942038536	0.80442\\
10.962023258	0.871128\\
10.982150793	0.835812\\
11.002026796	0.70632\\
11.022168398	0.796572\\
11.042107582	0.655308\\
11.062018871	0.572904\\
11.082143784	0.52974\\
11.102014303	0.553284\\
11.122162104	0.447336\\
11.142023325	0.443412\\
11.162003279	0.35316\\
11.182168961	0.337464\\
11.202023745	0.239364\\
11.222198009	0.25506\\
11.242036581	0.188352\\
11.262020588	0.0981\\
11.282158136	0.082404\\
11.302039862	0\\
11.322206497	-0.0981\\
11.341994762	-0.066708\\
11.362016439	-0.309996\\
11.382194281	-0.211896\\
11.402049065	-0.258984\\
11.42216301	-0.321768\\
11.44204402	-0.25506\\
11.462008238	-0.361008\\
11.482185125	-0.400248\\
11.502063513	-0.376704\\
11.522179365	-0.419868\\
11.542031527	-0.443412\\
11.562044382	-0.502272\\
11.582202435	-0.463032\\
11.602005959	-0.463032\\
11.622162104	-0.56898\\
11.642018795	-0.659232\\
11.662035704	-0.808344\\
11.682209015	-0.7848\\
11.701995373	-0.761256\\
11.722155094	-0.76518\\
11.742041111	-0.871128\\
11.76204586	-0.94176\\
11.782194138	-0.831888\\
11.802010298	-0.929988\\
11.822166204	-0.839736\\
11.842052221	-0.953532\\
11.862076521	-0.981\\
11.882140398	-0.988848\\
11.902008533	-1.012392\\
11.922194719	-1.083024\\
11.942014456	-0.90252\\
11.962010145	-1.016316\\
11.982151985	-1.15758\\
12.002012491	-1.153656\\
12.022212505	-1.086948\\
12.042027712	-1.169352\\
12.062018633	-1.13796\\
12.082137346	-1.21644\\
12.102040052	-1.39302\\
12.122159719	-1.408716\\
12.142022371	-1.306692\\
12.162007332	-1.495044\\
12.182183743	-1.506816\\
12.202021599	-1.5696\\
12.22219348	-1.632384\\
12.242077351	-1.74618\\
12.262014866	-1.58922\\
12.282147408	-1.80504\\
12.302027941	-1.848204\\
12.322198868	-1.699092\\
12.342032194	-1.812888\\
12.362026215	-1.859976\\
12.382187128	-1.914912\\
12.402066231	-1.938456\\
12.422172785	-1.930608\\
12.442011118	-1.875672\\
12.462028265	-1.965924\\
12.482155561	-1.848204\\
12.50212431	-2.052252\\
12.522170067	-1.90314\\
12.542024374	-1.895292\\
12.562016249	-1.722636\\
12.582173347	-1.887444\\
12.60203433	-1.793268\\
12.62221384	-1.946304\\
12.642044067	-1.871748\\
12.662013531	-1.74618\\
12.682140589	-1.624536\\
12.702059507	-1.8639\\
12.722218513	-1.958076\\
12.742001295	-1.836432\\
12.762009859	-1.90314\\
12.782172441	-1.604916\\
12.802034855	-1.769724\\
12.822225571	-1.80504\\
12.842023134	-1.92276\\
12.862039804	-1.5696\\
12.882176399	-1.600992\\
12.902061462	-1.49112\\
12.922202587	-1.624536\\
12.942026854	-1.538208\\
12.962030411	-1.495044\\
12.982169151	-1.51074\\
13.002050877	-1.067328\\
13.022166729	-0.945684\\
13.042048693	-1.016316\\
13.062007904	-0.796572\\
13.082170963	-0.64746\\
13.102045774	-0.788724\\
13.122222662	-0.60822\\
13.142009258	-0.561132\\
13.162029028	-0.541512\\
13.182162762	-0.51012\\
13.202048779	-0.541512\\
13.222162247	-0.478728\\
13.24207449	-0.553284\\
13.262017488	-0.423792\\
13.282202244	-0.498348\\
13.302010059	-0.384552\\
13.322175026	-0.408096\\
13.342008591	-0.070632\\
13.362028837	-0.121644\\
13.382196903	-0.164808\\
13.402005672	-0.200124\\
13.422161102	-0.03924\\
13.442029476	-0.121644\\
13.462056637	-0.15696\\
13.48216176	-0.0981\\
13.502006292	-0.11772\\
13.522183418	-0.011772\\
13.542004347	0.023544\\
13.56208086	0.003924\\
13.582190752	-0.023544\\
13.602019787	0.105948\\
13.622189045	0.015696\\
13.642015696	0.031392\\
13.662033081	0.094176\\
13.682160139	0.070632\\
13.702014208	0.192276\\
13.722155094	0.082404\\
13.741992235	0.023544\\
13.76205492	0.062784\\
13.782170057	0.204048\\
13.801992893	0.17658\\
13.822155237	0.0981\\
13.842068911	0.200124\\
13.86199522	0.13734\\
13.882196903	0.227592\\
13.902000666	0.258984\\
13.922158957	0.282528\\
13.942016602	0.258984\\
13.96201849	0.25506\\
13.982173681	0.2943\\
14.002015829	0.282528\\
14.022208691	0.341388\\
14.042024612	0.2943\\
14.062039852	0.408096\\
14.082187176	0.364932\\
14.102016687	0.41202\\
14.122161388	0.376704\\
14.142022133	0.478728\\
14.162013292	0.514044\\
14.182175159	0.502272\\
14.201994419	0.600372\\
14.222209215	0.54936\\
14.242069721	0.596448\\
14.262035131	0.604296\\
14.282157898	0.631764\\
14.302041531	0.694548\\
14.322193146	0.7848\\
14.342046499	0.749484\\
14.362018108	0.796572\\
14.382190943	0.831888\\
14.402009964	0.914292\\
14.422169924	0.90252\\
14.442035675	1.008468\\
14.462081909	0.988848\\
14.48216176	0.969228\\
14.502019167	1.055556\\
14.52216506	1.09872\\
14.542030334	1.083024\\
14.562053204	1.204668\\
14.582164764	1.145808\\
14.602011204	1.279224\\
14.62218523	1.247832\\
14.64201045	1.322388\\
14.66202426	1.29492\\
14.682141304	1.420488\\
14.702003956	1.377324\\
14.722154617	1.424412\\
14.742044926	1.385172\\
14.762069941	1.4715\\
14.782160044	1.436184\\
14.802019119	1.459728\\
14.822211266	1.534284\\
14.842029095	1.49112\\
14.862046957	1.557828\\
14.882148504	1.54998\\
14.902021885	1.616688\\
14.922165394	1.652004\\
14.942038774	1.6677\\
14.962046146	1.699092\\
14.982152224	1.68732\\
15.002025843	1.773648\\
15.022171974	1.695168\\
15.042080164	1.801116\\
15.062054157	1.78542\\
15.082161188	1.769724\\
15.102033854	1.832508\\
15.122178078	1.859976\\
15.142013073	1.938456\\
15.162030697	1.828584\\
15.182167053	1.938456\\
15.202033281	1.930608\\
15.222221851	1.914912\\
15.2420187	1.969848\\
15.262054443	1.973772\\
15.282158136	1.989468\\
15.302040577	2.036556\\
15.322177887	1.891368\\
15.342006445	1.98162\\
15.362061739	1.832508\\
15.382174253	1.926684\\
15.401994705	1.92276\\
15.422181129	1.90314\\
15.442026138	1.836432\\
15.462063313	1.808964\\
15.482153177	1.703016\\
15.502030611	1.730484\\
15.522151232	1.734408\\
15.542061329	1.679472\\
15.562007427	1.640232\\
15.582166672	1.597068\\
15.602015734	1.577448\\
15.622195959	1.498968\\
15.642052889	1.51074\\
15.662012339	1.514664\\
15.682186842	1.487196\\
15.702052832	1.416564\\
15.722197056	1.338084\\
15.74203229	1.283148\\
15.762060881	1.271376\\
15.782194376	1.25568\\
15.802023888	1.185048\\
15.822170258	1.161504\\
15.842051506	1.106568\\
15.862038612	1.083024\\
15.88216114	1.043784\\
15.901991606	1.024164\\
15.922177792	0.973152\\
15.942011356	0.957456\\
15.962031841	0.86328\\
15.982192993	0.835812\\
16.002021551	0.773028\\
16.022186518	0.702396\\
16.042070389	0.659232\\
16.062039375	0.619992\\
16.082175016	0.541512\\
16.102033615	0.415944\\
16.122169256	0.45126\\
16.142048597	0.435564\\
16.162024975	0.309996\\
16.182143211	0.35316\\
16.202010155	0.35316\\
16.222177744	0.266832\\
16.242027044	0.337464\\
16.262030125	0.345312\\
16.282165766	0.388476\\
16.302013397	0.466956\\
16.322194338	0.396324\\
16.342005014	0.211896\\
16.362016916	0.31392\\
16.382160902	0.168732\\
16.40202713	0.3924\\
16.422223806	0.341388\\
16.442056656	0.066708\\
16.462023735	0.121644\\
16.482177973	0.01962\\
16.502076387	0.133416\\
16.522184372	0.082404\\
16.542040825	0.03924\\
16.562006235	-0.153036\\
16.582221985	0.070632\\
16.602005959	-0.11772\\
16.622155905	-0.027468\\
16.642079353	0.043164\\
16.66200304	-0.047088\\
16.682188034	0.17658\\
16.702009439	0.070632\\
16.722154617	0.01962\\
16.742036343	0.180504\\
16.76201272	0.090252\\
16.782194853	0.207972\\
16.802005053	0.223668\\
16.822202444	0.168732\\
16.842010736	0.03924\\
16.861999512	0.145188\\
16.882180214	0.219744\\
16.902031898	0.384552\\
16.922189713	0.258984\\
16.942047596	0.223668\\
16.962030411	0.200124\\
16.982206821	0.384552\\
17.002003193	0.282528\\
17.022186279	0.2943\\
17.042018175	0.172656\\
17.062042475	0.321768\\
17.082158804	0.27468\\
17.102040529	0.35316\\
17.122158766	0.168732\\
17.142002344	0.415944\\
17.162016392	0.302148\\
17.182179689	0.282528\\
17.201996088	0.364932\\
17.222167015	0.25506\\
17.2420187	0.2943\\
17.262016535	0.341388\\
17.282165527	0.361008\\
17.302018881	0.380628\\
17.322158813	0.459108\\
17.342066288	0.415944\\
17.36203742	0.525816\\
17.382156134	0.41202\\
17.401993513	0.545436\\
17.422166586	0.56898\\
17.442014456	0.62784\\
17.462014198	0.584676\\
17.482141733	0.592524\\
17.502054214	0.580752\\
17.522158623	0.796572\\
17.54201293	0.584676\\
17.562024117	0.773028\\
17.582157373	0.718092\\
17.602024794	0.812268\\
17.622184992	0.698472\\
17.642039776	0.702396\\
17.662021875	0.812268\\
17.6821661	0.827964\\
17.702030182	0.84366\\
17.722173929	0.82404\\
17.742020369	0.780876\\
17.762018919	0.86328\\
17.782183647	0.918216\\
17.802010298	0.84366\\
17.82222414	0.906444\\
17.842059374	0.855432\\
17.862042665	0.945684\\
17.882157326	1.016316\\
17.902052641	0.996696\\
17.922207832	1.055556\\
17.942019939	1.141884\\
17.962049961	1.094796\\
17.982177734	1.204668\\
18.002047062	1.13796\\
18.022161961	1.306692\\
18.042015553	1.251756\\
18.062008619	1.31454\\
18.082187176	1.318464\\
18.102060556	1.361628\\
18.12223959	1.444032\\
18.142025948	1.436184\\
18.162018299	1.428336\\
18.18220377	1.365552\\
18.20204711	1.479348\\
18.222159147	1.487196\\
18.242021084	1.640232\\
18.262038469	1.561752\\
18.282178402	1.644156\\
18.302013874	1.62846\\
18.322201729	1.659852\\
18.342014313	1.867824\\
18.362061024	1.797192\\
18.382189989	1.965924\\
18.402012348	1.801116\\
18.422197104	1.938456\\
18.44211483	1.92276\\
18.46204257	1.88352\\
18.482144117	1.926684\\
18.502032518	1.84428\\
18.522205114	2.028708\\
18.542003393	1.781496\\
18.56207037	1.946304\\
18.58215332	1.773648\\
18.601989031	1.887444\\
18.622177839	1.899216\\
18.642030239	1.848204\\
18.662031651	1.8639\\
18.682155609	1.840356\\
18.702023029	1.84428\\
18.722157001	1.7658\\
18.742028713	1.895292\\
18.762127161	1.859976\\
18.782155275	1.78542\\
18.802005768	1.777572\\
18.822184801	1.828584\\
18.842027903	1.891368\\
18.862023354	1.90314\\
18.88215661	1.812888\\
18.902007818	1.92276\\
18.922172785	1.934532\\
18.942003012	2.028708\\
18.962057114	1.808964\\
18.982169151	1.879596\\
19.002014637	1.94238\\
19.022171259	1.836432\\
19.042019367	1.757952\\
19.062022209	1.88352\\
19.082147837	1.600992\\
19.102004051	1.997316\\
19.122156858	1.828584\\
19.142049789	1.910988\\
19.162013531	1.891368\\
19.182175159	1.94238\\
19.202008724	1.84428\\
19.222201824	2.169972\\
19.242077351	1.92276\\
19.262018919	1.973772\\
19.282177448	1.781496\\
19.302042007	2.091492\\
19.32219243	2.013012\\
19.342000008	2.126808\\
19.36203742	1.859976\\
19.3821702	1.958076\\
19.402042627	2.107188\\
19.422222853	1.962\\
19.442041636	1.969848\\
19.462012053	2.048328\\
19.482167006	2.107188\\
19.502070189	2.09934\\
19.522199869	1.989468\\
19.542028427	1.993392\\
19.562001705	2.130732\\
19.582165003	1.977696\\
19.602050543	2.016936\\
19.622229338	2.028708\\
19.642035484	1.836432\\
19.662012577	2.052252\\
19.682184458	1.907064\\
19.702040434	1.950228\\
19.722159863	1.962\\
19.742070913	1.8639\\
19.76202178	1.98162\\
19.782196522	1.848204\\
19.802010298	1.973772\\
19.822179794	1.867824\\
19.84202981	1.8639\\
19.862037182	1.852128\\
19.88216424	1.969848\\
19.902036905	1.797192\\
19.922178745	1.652004\\
19.942070246	1.498968\\
19.962021589	1.440108\\
19.982163429	1.381248\\
20.002016068	1.444032\\
20.022198677	1.396944\\
20.042007446	1.322388\\
20.062027216	1.35378\\
20.082159042	1.302768\\
20.102087975	1.597068\\
20.122168541	1.49112\\
20.142026424	1.640232\\
20.162014246	1.279224\\
20.182190418	1.506816\\
20.202024937	1.287072\\
20.222181797	1.483272\\
20.242038488	1.243908\\
20.261988878	1.13796\\
20.282178164	1.169352\\
20.302005768	1.220364\\
20.322232485	1.267452\\
20.342035294	0.816192\\
20.36203599	1.483272\\
20.382168293	1.149732\\
20.401994467	1.487196\\
};
\addplot [color=red, line width=1.0pt, forget plot]
  table[row sep=crcr]{%
0.122153044	-0.357084\\
0.142704725	-0.357084\\
0.162019968	-0.357084\\
0.182151318	-0.357084\\
0.202072382	-0.357084\\
0.222170353	-0.357084\\
0.242097378	-0.357084\\
0.262028217	-0.357084\\
0.282156229	-0.357084\\
0.302108049	-0.357084\\
0.322176456	-0.357084\\
0.342110395	-0.357084\\
0.362020254	-0.357084\\
0.382161379	-0.357084\\
0.402102709	-0.357084\\
0.422178984	-0.357084\\
0.442130327	-0.357084\\
0.46200633	-0.357084\\
0.482154846	-0.439488\\
0.502096176	-0.419868\\
0.522157907	-0.419868\\
0.542132139	-0.400248\\
0.562018394	-0.400248\\
0.582155943	-0.364932\\
0.602100611	-0.600372\\
0.622170925	-0.423792\\
0.642130375	-0.561132\\
0.661995411	-0.561132\\
0.682149649	-0.466956\\
0.702078104	-0.427716\\
0.722169399	-0.427716\\
0.742068052	-0.572904\\
0.76201272	-0.572904\\
0.782187939	-0.329616\\
0.802058935	-0.309996\\
0.822188854	-0.239364\\
0.84211874	-0.286452\\
0.862021446	-0.54936\\
0.882141829	-0.364932\\
0.902095318	-0.341388\\
0.922189951	-0.160884\\
0.942067385	-0.160884\\
0.962010384	-0.15696\\
0.982141018	0.054936\\
1.00208354	0.047088\\
1.022191286	0.17658\\
1.04208827	0.388476\\
1.062000513	0.062784\\
1.082139254	0.062784\\
1.10210681	-0.102024\\
1.122172117	-0.102024\\
1.142074108	-0.23544\\
1.162016153	-0.211896\\
1.182153702	-0.239364\\
1.202116489	-0.047088\\
1.222205639	0.105948\\
1.242153645	0.105948\\
1.26202631	0.219744\\
1.282181978	0.459108\\
1.302097559	0.459108\\
1.322151184	0.200124\\
1.342127562	0.062784\\
1.362022638	-0.01962\\
1.382176876	-0.01962\\
1.402080297	-0.286452\\
1.422161579	-0.184428\\
1.442111731	-0.172656\\
1.462035418	-0.1962\\
1.482175827	0.109872\\
1.502425671	-0.05886\\
1.522183895	-0.121644\\
1.542054653	0.129492\\
1.562038183	-0.082404\\
1.582180977	-0.082404\\
1.602072954	-0.262908\\
1.622181654	-0.525816\\
1.64208889	-0.525816\\
1.662036657	0.027468\\
1.682142735	-0.121644\\
1.702091694	-0.207972\\
1.722146988	-0.047088\\
1.742088318	0.129492\\
1.762031794	0.361008\\
1.782210588	0.0981\\
1.802090406	0.521892\\
1.822175741	0.521892\\
1.842198849	0.054936\\
1.862034321	0.47088\\
1.882190228	0.035316\\
1.902107477	-0.1962\\
1.922172308	-0.1962\\
1.942081451	-0.251136\\
1.962022543	-0.251136\\
1.982192516	-0.007848\\
2.002118826	0.066708\\
2.02216363	0.003924\\
2.04212141	0.003924\\
2.062012196	-0.070632\\
2.082183361	-0.051012\\
2.102087498	-0.11772\\
2.122143269	-0.007848\\
2.142081261	0.141264\\
2.162018299	0.141264\\
2.182207823	0.207972\\
2.202086687	0.129492\\
2.222201586	0.129492\\
2.242088079	0.074556\\
2.262051105	0\\
2.282233238	0.184428\\
2.302112341	0.184428\\
2.322165251	0.219744\\
2.342108011	0.219744\\
2.362074614	0.207972\\
2.382153273	-0.023544\\
2.402099133	-0.298224\\
2.422190428	-0.298224\\
2.44206953	-0.33354\\
2.462044001	-0.247212\\
2.482162237	-0.023544\\
2.50207901	-0.023544\\
2.522173405	-0.043164\\
2.542060137	0.11772\\
2.562032938	0.043164\\
2.582175016	0.043164\\
2.602091312	0.160884\\
2.622170687	-0.062784\\
2.642087221	-0.133416\\
2.661997318	-0.133416\\
2.682151079	-0.066708\\
2.702081203	-0.066708\\
2.722151279	-0.160884\\
2.742081404	-0.070632\\
2.762029409	-0.109872\\
2.782158852	-0.035316\\
2.802078962	0.113796\\
2.822157145	0.113796\\
2.842126608	0.192276\\
2.862025261	0.192276\\
2.882153511	0.129492\\
2.902089834	0.01962\\
2.922161102	-0.105948\\
2.9420681	-0.11772\\
2.962011576	-0.11772\\
2.982153654	-0.125568\\
3.002092123	-0.125568\\
3.022182465	0.066708\\
3.042096138	-0.102024\\
3.062021494	0.102024\\
3.082163334	0.11772\\
3.102095366	0.11772\\
3.1221807	-0.11772\\
3.142077208	-0.1962\\
3.16202116	-0.1962\\
3.182166815	-0.168732\\
3.20211935	-0.168732\\
3.222192287	-0.082404\\
3.242128134	0.184428\\
3.262024879	0.133416\\
3.282130003	0.133416\\
3.302080154	0.13734\\
3.322163582	0.003924\\
3.342069626	0.047088\\
3.361999273	0.047088\\
3.382153988	0.27468\\
3.402083397	0.27468\\
3.422180891	0.129492\\
3.442085266	0.188352\\
3.462021589	0.145188\\
3.482152939	-0.023544\\
3.502100468	-0.164808\\
3.522212744	-0.164808\\
3.542113781	-0.286452\\
3.562012196	-0.286452\\
3.582176685	-0.172656\\
3.602067471	-0.168732\\
3.622155666	-0.168732\\
3.642103434	0.027468\\
3.662031174	0.027468\\
3.682188272	-0.035316\\
3.702080488	-0.007848\\
3.722182035	0.164808\\
3.742081165	0.200124\\
3.762041092	0.286452\\
3.782184362	0.239364\\
3.802086115	0.27468\\
3.822208643	0.239364\\
3.842411518	0.239364\\
3.862011194	0.035316\\
3.882148266	0.125568\\
3.902097464	0.125568\\
3.922157049	-0.125568\\
3.942069292	-0.05886\\
3.962007523	-0.25506\\
3.982177258	-0.25506\\
4.002087593	0.011772\\
4.022173405	0.011772\\
4.042154551	-0.207972\\
4.06201911	-0.035316\\
4.082152367	-0.133416\\
4.102093697	-0.133416\\
4.122160673	-0.05886\\
4.142062426	-0.01962\\
4.162030935	-0.027468\\
4.182180405	-0.051012\\
4.202072144	0.184428\\
4.222162008	0.184428\\
4.242100716	0.043164\\
4.262014389	-0.090252\\
4.28219533	-0.090252\\
4.302089214	-0.054936\\
4.322158575	-0.054936\\
4.342076063	-0.07848\\
4.362028837	0.023544\\
4.382173061	0.051012\\
4.402078867	0.094176\\
4.4221766	0.094176\\
4.442105055	0.113796\\
4.46200633	0.164808\\
4.482172966	0.200124\\
4.502092361	0.200124\\
4.522158146	0.25506\\
4.542140245	0.113796\\
4.562050581	0.113796\\
4.582169533	0.105948\\
4.60207653	-0.21582\\
4.622186661	-0.439488\\
4.642095804	-0.33354\\
4.662044764	-0.33354\\
4.682162046	-0.35316\\
4.702105522	-0.102024\\
4.722176552	-0.105948\\
4.742126703	-0.109872\\
4.762061596	-0.109872\\
4.782148838	-0.011772\\
4.802083969	-0.011772\\
4.82220006	0.090252\\
4.842048645	0.090252\\
4.862048626	0.141264\\
4.882157326	0.141264\\
4.902056932	0.113796\\
4.92215395	0.066708\\
4.942088842	-0.01962\\
4.962026834	-0.01962\\
4.982174397	-0.070632\\
5.002012014	0.070632\\
5.022206068	0.074556\\
5.042008877	0.207972\\
5.062007904	0.145188\\
5.082171679	-0.05886\\
5.10203433	-0.05886\\
5.122157097	-0.207972\\
5.141997099	-0.207972\\
5.161998987	-0.192276\\
5.182173491	-0.337464\\
5.202002525	-0.129492\\
5.222178936	-0.149112\\
5.242048025	-0.149112\\
5.261997223	0.0981\\
5.282162666	-0.031392\\
5.302005768	0.102024\\
5.322175503	0.102024\\
5.342036486	0.13734\\
5.362068892	0.109872\\
5.382175207	0.109872\\
5.402026415	0.047088\\
5.422189951	0.047088\\
5.442039967	0.054936\\
5.462042093	-0.031392\\
5.482180834	-0.094176\\
5.502022743	-0.094176\\
5.522157431	-0.11772\\
5.542028427	-0.149112\\
5.562045097	-0.149112\\
5.582159758	-0.0981\\
5.602006435	-0.051012\\
5.622204542	-0.094176\\
5.642046213	-0.05886\\
5.662028074	-0.074556\\
5.682136536	-0.074556\\
5.702010393	-0.066708\\
5.722163677	-0.066708\\
5.741996527	-0.070632\\
5.762047768	-0.070632\\
5.782159328	0.011772\\
5.802027225	0.086328\\
5.822189093	0.031392\\
5.842361212	0.003924\\
5.862008333	0.015696\\
5.882154942	0.015696\\
5.902008057	-0.047088\\
5.922199488	-0.047088\\
5.941994667	-0.03924\\
5.962021589	-0.015696\\
5.98218298	-0.015696\\
6.002015829	0.054936\\
6.022216558	0.070632\\
6.042048216	0.070632\\
6.062056303	0.070632\\
6.08216095	0.070632\\
6.102020502	0.054936\\
6.122161865	0.035316\\
6.142045021	-0.03924\\
6.16202569	-0.03924\\
6.182163954	-0.153036\\
6.202054739	-0.113796\\
6.222176075	-0.105948\\
6.242068768	0\\
6.262020826	0.043164\\
6.282161474	0.043164\\
6.302022219	0.027468\\
6.322189093	-0.015696\\
6.342045546	-0.015696\\
6.362016678	-0.015696\\
6.38218379	-0.07848\\
6.402023554	-0.01962\\
6.42219162	-0.003924\\
6.442023993	0.047088\\
6.462028742	0.051012\\
6.482155561	0.07848\\
6.502026796	0.07848\\
6.522170305	0.125568\\
6.542007208	0.05886\\
6.562035084	-0.003924\\
6.582168818	-0.082404\\
6.60201931	-0.082404\\
6.62219882	-0.074556\\
6.642049789	-0.070632\\
6.662009001	-0.027468\\
6.68214035	-0.027468\\
6.702040672	-0.031392\\
6.722191811	0.015696\\
6.742039204	0.011772\\
6.762017965	-0.011772\\
6.782172441	-0.015696\\
6.802040339	-0.015696\\
6.822195053	0.066708\\
6.842060566	0.066708\\
6.862022638	0.070632\\
6.882184267	0.102024\\
6.902058125	0.102024\\
6.922169924	0.003924\\
6.942036152	-0.01962\\
6.962060928	-0.01962\\
6.982192516	-0.047088\\
7.001996994	-0.047088\\
7.022202253	-0.011772\\
7.042022705	0.043164\\
7.062073708	0.043164\\
7.082153797	0.043164\\
7.102023363	0.047088\\
7.122164726	0.01962\\
7.141997576	-0.027468\\
7.162062645	-0.031392\\
7.182226181	-0.03924\\
7.202002287	-0.03924\\
7.222168207	-0.047088\\
7.242021084	-0.003924\\
7.262044907	-0.011772\\
7.282160997	-0.011772\\
7.302033901	-0.01962\\
7.322174311	0.023544\\
7.342018366	0.023544\\
7.362043142	0.086328\\
7.382188559	0.086328\\
7.402025938	0.035316\\
7.422181129	0.03924\\
7.442045927	0.047088\\
7.461986542	0.047088\\
7.482157707	0.043164\\
7.502041578	-0.015696\\
7.52215457	0\\
7.541994095	0\\
7.562044144	-0.05886\\
7.582208633	-0.07848\\
7.601991415	-0.027468\\
7.622181654	-0.027468\\
7.642034292	-0.051012\\
7.662019968	0.023544\\
7.682176352	0.023544\\
7.702009916	0.031392\\
7.722177982	-0.082404\\
7.742027044	-0.01962\\
7.76205492	-0.043164\\
7.78215456	-0.070632\\
7.802002907	-0.003924\\
7.82217288	-0.003924\\
7.842028141	-0.090252\\
7.862007141	-0.062784\\
7.882144928	-0.007848\\
7.902017117	-0.007848\\
7.922207117	-0.007848\\
7.941999197	0.011772\\
7.962014675	-0.01962\\
7.982160568	-0.01962\\
8.00202775	0\\
8.022207737	0.01962\\
8.042006493	-0.03924\\
8.062014818	0.011772\\
8.082175493	-0.003924\\
8.102050066	-0.003924\\
8.12214613	0.027468\\
8.142054081	-0.015696\\
8.162019491	-0.011772\\
8.182166815	-0.054936\\
8.202024937	-0.054936\\
8.222177029	-0.102024\\
8.242496252	-0.11772\\
8.262017727	-0.105948\\
8.282183409	-0.105948\\
8.302018642	-0.090252\\
8.322143316	-0.090252\\
8.341991186	-0.066708\\
8.362090826	-0.066708\\
8.382203579	-0.031392\\
8.402018785	-0.031392\\
8.422168493	-0.03924\\
8.442065001	-0.03924\\
8.462030411	-0.035316\\
8.482177019	-0.05886\\
8.502048016	-0.051012\\
8.522155046	-0.086328\\
8.541997671	-0.086328\\
8.562011242	-0.03924\\
8.58215785	-0.03924\\
8.602002859	-0.062784\\
8.622174025	-0.031392\\
8.642019033	-0.031392\\
8.662014723	-0.023544\\
8.682170153	-0.023544\\
8.702004671	0.003924\\
8.72217226	-0.003924\\
8.742035627	-0.023544\\
8.762006521	-0.082404\\
8.782174349	-0.082404\\
8.802015543	-0.168732\\
8.822201967	-0.180504\\
8.84207201	-0.282528\\
8.862038374	-0.258984\\
8.882151842	-0.258984\\
8.90201664	-0.223668\\
8.9221735	-0.223668\\
8.942082882	-0.219744\\
8.962004662	-0.200124\\
8.982151985	-0.21582\\
9.002033949	-0.1962\\
9.022185564	-0.1962\\
9.042088509	-0.251136\\
9.062009811	-0.184428\\
9.082170963	-0.266832\\
9.102047443	-0.266832\\
9.122205019	-0.172656\\
9.142018557	-0.290376\\
9.162004709	-0.317844\\
9.182167768	-0.317844\\
9.202019215	-0.27468\\
9.222215652	-0.27468\\
9.2420578	-0.258984\\
9.262014627	-0.321768\\
9.282175064	-0.298224\\
9.302053452	-0.298224\\
9.322166443	-0.455184\\
9.342011452	-0.466956\\
9.362027884	-0.592524\\
9.382169485	-0.655308\\
9.40199542	-0.671004\\
9.422189236	-0.698472\\
9.442024231	-0.702396\\
9.462031126	-0.867204\\
9.482210636	-0.800496\\
9.50201416	-0.800496\\
9.522164106	-0.773028\\
9.541999578	-0.773028\\
9.562057734	-0.682776\\
9.58217001	-0.682776\\
9.602027416	-0.639612\\
9.62221837	-0.639612\\
9.642043591	-0.64746\\
9.662041426	-0.808344\\
9.682175636	-0.788724\\
9.702047348	-0.788724\\
9.72215867	-0.906444\\
9.742034674	-0.94176\\
9.762027502	-1.004544\\
9.782195807	-1.004544\\
9.802001715	-1.008468\\
9.822241306	-1.035936\\
9.842035294	-1.035936\\
9.861996412	-1.212516\\
9.882143974	-1.220364\\
9.902062178	-1.306692\\
9.92215991	-1.306692\\
9.941990614	-1.267452\\
9.962040186	-1.149732\\
9.982160091	-1.287072\\
10.002008915	-1.220364\\
10.022162676	-1.389096\\
10.042040348	-1.459728\\
10.062033176	-1.58922\\
10.082159519	-1.553904\\
10.102014542	-1.616688\\
10.122166157	-1.616688\\
10.14205265	-1.479348\\
10.162023067	-1.4715\\
10.182164669	-1.365552\\
10.20201087	-1.365552\\
10.222209692	-1.228212\\
10.242053509	-1.263528\\
10.262000322	-1.408716\\
10.282187939	-1.585296\\
10.301998377	-1.49112\\
10.32216692	-1.70694\\
10.342014313	-1.70694\\
10.362031221	-1.722636\\
10.382137775	-1.958076\\
10.402019739	-1.90314\\
10.422216654	-2.17782\\
10.442014933	-2.17782\\
10.462009192	-2.134656\\
10.482165575	-2.162124\\
10.502165794	-1.895292\\
10.522184372	-1.789344\\
10.542012691	-1.612764\\
10.562042475	-1.400868\\
10.582167625	-1.400868\\
10.602003336	-1.290996\\
10.62219882	-1.290996\\
10.642024279	-1.263528\\
10.662052393	-1.263528\\
10.682150364	-1.130112\\
10.702017546	-1.130112\\
10.722202063	-1.055556\\
10.742022038	-1.055556\\
10.762051105	-1.2753\\
10.782175779	-1.546056\\
10.802012682	-1.546056\\
10.822185516	-1.722636\\
10.842025518	-1.836432\\
10.862024546	-1.836432\\
10.882167339	-1.92276\\
10.901996136	-1.930608\\
10.922170639	-1.930608\\
10.942038536	-1.777572\\
10.962023258	-1.703016\\
10.982150793	-1.561752\\
11.002026796	-1.279224\\
11.022168398	-1.279224\\
11.042107582	-1.114416\\
11.062018871	-0.906444\\
11.082143784	-0.906444\\
11.102014303	-0.749484\\
11.122162104	-0.839736\\
11.142023325	-0.62784\\
11.162003279	-0.894672\\
11.182168961	-1.055556\\
11.202023745	-1.134036\\
11.222198009	-1.055556\\
11.242036581	-1.055556\\
11.262020588	-1.102644\\
11.282158136	-0.977076\\
11.302039862	-0.733788\\
11.322206497	-0.733788\\
11.341994762	-0.561132\\
11.362016439	-0.498348\\
11.382194281	-0.498348\\
11.402049065	-0.47088\\
11.42216301	-0.494424\\
11.44204402	-0.423792\\
11.462008238	-0.423792\\
11.482185125	-0.341388\\
11.502063513	-0.219744\\
11.522179365	-0.219744\\
11.542031527	-0.070632\\
11.562044382	-0.031392\\
11.582202435	0.05886\\
11.602005959	0.05886\\
11.622162104	-0.027468\\
11.642018795	-0.01962\\
11.662035704	0.031392\\
11.682209015	0.031392\\
11.701995373	0.054936\\
11.722155094	0.1962\\
11.742041111	0.298224\\
11.76204586	0.298224\\
11.782194138	0.494424\\
11.802010298	0.576828\\
11.822166204	0.576828\\
11.842052221	0.604296\\
11.862076521	0.604296\\
11.882140398	0.616068\\
11.902008533	0.741636\\
11.922194719	0.722016\\
11.942014456	0.722016\\
11.962010145	0.910368\\
11.982151985	0.910368\\
12.002012491	0.867204\\
12.022212505	0.831888\\
12.042027712	0.831888\\
12.062018633	0.729864\\
12.082137346	0.808344\\
12.102040052	0.808344\\
12.122159719	0.957456\\
12.142022371	0.965304\\
12.162007332	0.965304\\
12.182183743	1.21644\\
12.202021599	1.326312\\
12.22219348	1.357704\\
12.242077351	1.357704\\
12.262014866	1.408716\\
12.282147408	1.45188\\
12.302027941	1.553904\\
12.322198868	1.557828\\
12.342032194	1.675548\\
12.362026215	1.655928\\
12.382187128	1.699092\\
12.402066231	1.699092\\
12.422172785	1.797192\\
12.442011118	1.836432\\
12.462028265	1.6677\\
12.482155561	1.573524\\
12.50212431	1.495044\\
12.522170067	1.483272\\
12.542024374	1.463652\\
12.562016249	1.495044\\
12.582173347	1.495044\\
12.60203433	1.577448\\
12.62221384	1.542132\\
12.642044067	1.553904\\
12.662013531	1.553904\\
12.682140589	1.553904\\
12.702059507	1.624536\\
12.722218513	1.777572\\
12.742001295	1.777572\\
12.762009859	2.115036\\
12.782172441	2.067948\\
12.802034855	2.299464\\
12.822225571	2.268072\\
12.842023134	2.268072\\
12.862039804	2.181744\\
12.882176399	2.115036\\
12.902061462	1.989468\\
12.922202587	1.989468\\
12.942026854	1.946304\\
12.962030411	1.859976\\
12.982169151	1.918836\\
13.002050877	1.918836\\
13.022166729	1.867824\\
13.042048693	1.867824\\
13.062007904	1.757952\\
13.082170963	1.72656\\
13.102045774	1.72656\\
13.122222662	1.769724\\
13.142009258	1.722636\\
13.162029028	1.754028\\
13.182162762	1.60884\\
13.202048779	1.616688\\
13.222162247	1.463652\\
13.24207449	1.542132\\
13.262017488	1.636308\\
13.282202244	1.636308\\
13.302010059	1.7658\\
13.322175026	1.895292\\
13.342008591	2.13858\\
13.362028837	2.307312\\
13.382196903	2.476044\\
13.402005672	2.476044\\
13.422161102	2.303388\\
13.442029476	2.303388\\
13.462056637	2.071872\\
13.48216176	1.926684\\
13.502006292	1.926684\\
13.522183418	1.644156\\
13.542004347	1.644156\\
13.56208086	1.600992\\
13.582190752	1.546056\\
13.602019787	1.640232\\
13.622189045	1.640232\\
13.642015696	1.58922\\
13.662033081	1.679472\\
13.682160139	1.659852\\
13.702014208	1.659852\\
13.722155094	1.691244\\
13.741992235	1.691244\\
13.76205492	1.718712\\
13.782170057	1.718712\\
13.801992893	1.738332\\
13.822155237	1.675548\\
13.842068911	1.683396\\
13.86199522	1.683396\\
13.882196903	1.624536\\
13.902000666	1.624536\\
13.922158957	1.581372\\
13.942016602	1.538208\\
13.96201849	1.538208\\
13.982173681	1.561752\\
14.002015829	1.4715\\
14.022208691	1.573524\\
14.042024612	1.440108\\
14.062039852	1.440108\\
14.082187176	1.400868\\
14.102016687	1.400868\\
14.122161388	1.404792\\
14.142022133	1.41264\\
14.162013292	1.475424\\
14.182175159	1.487196\\
14.201994419	1.534284\\
14.222209215	1.534284\\
14.242069721	1.495044\\
14.262035131	1.495044\\
14.282157898	1.416564\\
14.302041531	1.416564\\
14.322193146	1.369476\\
14.342046499	1.2753\\
14.362018108	1.283148\\
14.382190943	1.271376\\
14.402009964	1.330236\\
14.422169924	1.310616\\
14.442035675	1.322388\\
14.462081909	1.322388\\
14.48216176	1.23606\\
14.502019167	1.21644\\
14.52216506	1.15758\\
14.542030334	1.110492\\
14.562053204	1.090872\\
14.582164764	1.090872\\
14.602011204	1.047708\\
14.62218523	1.130112\\
14.64201045	1.075176\\
14.66202426	1.075176\\
14.682141304	1.29492\\
14.702003956	1.298844\\
14.722154617	1.298844\\
14.742044926	1.385172\\
14.762069941	1.310616\\
14.782160044	1.208592\\
14.802019119	1.165428\\
14.822211266	1.161504\\
14.842029095	1.169352\\
14.862046957	1.169352\\
14.882148504	1.1772\\
14.902021885	1.024164\\
14.922165394	1.024164\\
14.942038774	1.008468\\
14.962046146	1.083024\\
14.982152224	1.02024\\
15.002025843	1.032012\\
15.022171974	1.032012\\
15.042080164	1.02024\\
15.062054157	1.02024\\
15.082161188	0.929988\\
15.102033854	0.84366\\
15.122178078	0.769104\\
15.142013073	0.769104\\
15.162030697	0.820116\\
15.182167053	0.761256\\
15.202033281	0.761256\\
15.222221851	0.82404\\
15.2420187	0.733788\\
15.262054443	0.698472\\
15.282158136	0.698472\\
15.302040577	0.729864\\
15.322177887	0.788724\\
15.342006445	0.639612\\
15.362061739	0.651384\\
15.382174253	0.651384\\
15.401994705	0.776952\\
15.422181129	0.776952\\
15.442026138	0.773028\\
15.462063313	0.631764\\
15.482153177	0.592524\\
15.502030611	0.592524\\
15.522151232	0.33354\\
15.542061329	0.27468\\
15.562007427	0.145188\\
15.582166672	0.145188\\
15.602015734	-0.051012\\
15.622195959	-0.051012\\
15.642052889	-0.125568\\
15.662012339	-0.125568\\
15.682186842	-0.188352\\
15.702052832	-0.188352\\
15.722197056	-0.188352\\
15.74203229	-0.266832\\
15.762060881	-0.37278\\
15.782194376	-0.361008\\
15.802023888	-0.361008\\
15.822170258	-0.415944\\
15.842051506	-0.415944\\
15.862038612	-0.321768\\
15.88216114	-0.341388\\
15.901991606	-0.33354\\
15.922177792	-0.384552\\
15.942011356	-0.506196\\
15.962031841	-0.62784\\
15.982192993	-0.62784\\
16.002021551	-0.678852\\
16.022186518	-0.698472\\
16.042070389	-0.780876\\
16.062039375	-0.827964\\
16.082175016	-0.94176\\
16.102033615	-1.043784\\
16.122169256	-1.134036\\
16.142048597	-1.220364\\
16.162024975	-1.263528\\
16.182143211	-1.263528\\
16.202010155	-1.345932\\
16.222177744	-1.345932\\
16.242027044	-1.15758\\
16.262030125	-1.15758\\
16.282165766	-1.094796\\
16.302013397	-0.92214\\
16.322194338	-0.851508\\
16.342005014	-0.86328\\
16.362016916	-0.86328\\
16.382160902	-1.106568\\
16.40202713	-1.106568\\
16.422223806	-1.581372\\
16.442056656	-1.581372\\
16.462023735	-1.950228\\
16.482177973	-1.958076\\
16.502076387	-2.032632\\
16.522184372	-2.032632\\
16.542040825	-2.013012\\
16.562006235	-1.950228\\
16.582221985	-1.859976\\
16.602005959	-1.714788\\
16.622155905	-1.750104\\
16.642079353	-1.750104\\
16.66200304	-1.644156\\
16.682188034	-1.644156\\
16.702009439	-1.593144\\
16.722154617	-1.53036\\
16.742036343	-1.714788\\
16.76201272	-1.812888\\
16.782194853	-1.812888\\
16.802005053	-2.00124\\
16.822202444	-2.17782\\
16.842010736	-2.220984\\
16.861999512	-2.287692\\
16.882180214	-2.283768\\
16.902031898	-2.307312\\
16.922189713	-2.307312\\
16.942047596	-2.252376\\
16.962030411	-2.181744\\
16.982206821	-2.220984\\
17.002003193	-2.228832\\
17.022186279	-2.228832\\
17.042018175	-2.303388\\
17.062042475	-2.205288\\
17.082158804	-2.11896\\
17.102040529	-2.11896\\
17.122158766	-2.024784\\
17.142002344	-1.993392\\
17.162016392	-1.993392\\
17.182179689	-1.867824\\
17.201996088	-1.910988\\
17.222167015	-1.910988\\
17.2420187	-1.879596\\
17.262016535	-1.918836\\
17.282165527	-1.965924\\
17.302018881	-1.973772\\
17.322158813	-2.052252\\
17.342066288	-2.052252\\
17.36203742	-1.934532\\
17.382156134	-1.714788\\
17.401993513	-1.891368\\
17.422166586	-1.604916\\
17.442014456	-1.604916\\
17.462014198	-1.695168\\
17.482141733	-1.679472\\
17.502054214	-1.70694\\
17.522158623	-1.70694\\
17.54201293	-1.64808\\
17.562024117	-1.561752\\
17.582157373	-1.561752\\
17.602024794	-1.679472\\
17.622184992	-1.644156\\
17.642039776	-1.683396\\
17.662021875	-1.640232\\
17.6821661	-1.640232\\
17.702030182	-1.561752\\
17.722173929	-1.561752\\
17.742020369	-1.644156\\
17.762018919	-1.396944\\
17.782183647	-1.302768\\
17.802010298	-1.1772\\
17.82222414	-1.1772\\
17.842059374	-1.243908\\
17.862042665	-1.1772\\
17.882157326	-1.1772\\
17.902052641	-1.298844\\
17.922207832	-1.287072\\
17.942019939	-1.29492\\
17.962049961	-1.21644\\
17.982177734	-1.21644\\
18.002047062	-1.188972\\
18.022161961	-1.004544\\
18.042015553	-1.004544\\
18.062008619	-0.977076\\
18.082187176	-0.977076\\
18.102060556	-0.8829\\
18.12223959	-0.914292\\
18.142025948	-0.926064\\
18.162018299	-0.926064\\
18.18220377	-0.84366\\
18.20204711	-0.84366\\
18.222159147	-0.780876\\
18.242021084	-0.780876\\
18.262038469	-0.600372\\
18.282178402	-0.600372\\
18.302013874	-0.459108\\
18.322201729	-0.498348\\
18.342014313	-0.51012\\
18.362061024	-0.51012\\
18.382189989	-0.435564\\
18.402012348	-0.557208\\
18.422197104	-0.517968\\
18.44211483	-0.60822\\
18.46204257	-0.478728\\
18.482144117	-0.478728\\
18.502032518	-0.21582\\
18.522205114	-0.113796\\
18.542003393	-0.07848\\
18.56207037	-0.07848\\
18.58215332	-0.066708\\
18.601989031	-0.066708\\
18.622177839	-0.153036\\
18.642030239	-0.172656\\
18.662031651	-0.270756\\
18.682155609	-0.270756\\
18.702023029	-0.262908\\
18.722157001	-0.054936\\
18.742028713	-0.054936\\
18.762127161	0.11772\\
18.782155275	0.192276\\
18.802005768	0.192276\\
18.822184801	0.164808\\
18.842027903	0.227592\\
18.862023354	0.227592\\
18.88215661	0.105948\\
18.902007818	0.090252\\
18.922172785	0.090252\\
18.942003012	0.051012\\
18.962057114	0.0981\\
18.982169151	0.298224\\
19.002014637	0.329616\\
19.022171259	0.329616\\
19.042019367	0.286452\\
19.062022209	0.368856\\
19.082147837	0.211896\\
19.102004051	0.180504\\
19.122156858	0.149112\\
19.142049789	0.341388\\
19.162013531	0.258984\\
19.182175159	0.455184\\
19.202008724	0.455184\\
19.222201824	0.443412\\
19.242077351	0.561132\\
19.262018919	0.592524\\
19.282177448	0.404172\\
19.302042007	0.404172\\
19.32219243	0.502272\\
19.342000008	0.459108\\
19.36203742	0.459108\\
19.3821702	0.43164\\
19.402042627	0.584676\\
19.422222853	0.584676\\
19.442041636	0.545436\\
19.462012053	0.447336\\
19.482167006	0.639612\\
19.502070189	0.639612\\
19.522199869	0.439488\\
19.542028427	0.514044\\
19.562001705	0.663156\\
19.582165003	0.663156\\
19.602050543	0.694548\\
19.622229338	0.773028\\
19.642035484	0.773028\\
19.662012577	0.761256\\
19.682184458	0.612144\\
19.702040434	0.64746\\
19.722159863	0.698472\\
19.742070913	0.722016\\
19.76202178	0.698472\\
19.782196522	0.698472\\
19.802010298	0.808344\\
19.822179794	0.757332\\
19.84202981	0.722016\\
19.862037182	0.635688\\
19.88216424	0.671004\\
19.902036905	0.682776\\
19.922178745	0.847584\\
19.942070246	0.96138\\
19.962021589	0.96138\\
19.982163429	0.926064\\
20.002016068	1.114416\\
20.022198677	1.114416\\
20.042007446	0.894672\\
20.062027216	0.776952\\
20.082159042	0.776952\\
20.102087975	0.773028\\
20.122168541	0.90252\\
20.142026424	0.90252\\
20.162014246	1.090872\\
20.182190418	0.94176\\
20.202024937	0.86328\\
20.222181797	0.86328\\
20.242038488	0.8829\\
20.261988878	0.76518\\
20.282178164	0.831888\\
20.302005768	0.894672\\
20.322232485	0.894672\\
20.342035294	1.024164\\
20.36203599	0.820116\\
20.382168293	0.808344\\
20.401994467	0.753408\\
};
\addplot [color=black, forget plot]
  table[row sep=crcr]{%
4.25	-4.5\\
4.25	3.5\\
};
\node[right, align=left, circle,fill=white,draw=black,thick,line width=0.3mm,inner sep=1pt]
at (axis cs:2.35,2.3) {1};
\addplot [color=black, forget plot]
  table[row sep=crcr]{%
7.1800000667572	-4.5\\
7.1800000667572	3.5\\
};
\node[right, align=left, circle,fill=white,draw=black,thick,line width=0.3mm,inner sep=1pt]
at (axis cs:5.28,2.3) {2};
\addplot [color=black, forget plot]
  table[row sep=crcr]{%
11.9800000190735	-4.5\\
11.9800000190735	3.5\\
};
\node[right, align=left, circle,fill=white,draw=black,thick,line width=0.3mm,inner sep=1pt]
at (axis cs:10.08,-3.3) {3};
\end{axis}
\end{tikzpicture}%

%% file: Images/Tikz/fig10-3-3.tikz
\begin{tikzpicture}

\begin{axis}[%
width=4.2cm,
height=1.683cm,
at={(0cm,0cm)},
scale only axis,
xmin=0.122153044,
xmax=20.401994467,
xlabel style={font=\color{white!15!black}},
xlabel={$t\;[\SI[per-mode=repeated-symbol]{}{\second}]$},
ymin=-250,
ymax=150,
ylabel style={font=\color{white!15!black}},
ylabel={$\delta\;[\text{Deg}]$},
axis background/.style={fill=white},
ylabel style={yshift=-0.1cm}
]
\addplot [color=blue, line width=1.0pt, forget plot]
  table[row sep=crcr]{%
0.122153044	5.89999999999998\\
0.142704725	5.89999999999998\\
0.162019968	5.89999999999998\\
0.182151318	5.89999999999998\\
0.202072382	5.89999999999998\\
0.222170353	5.89999999999998\\
0.242097378	5.89999999999998\\
0.262028217	5.89999999999998\\
0.282156229	5.89999999999998\\
0.302108049	5.89999999999998\\
0.322176456	5.89999999999998\\
0.342110395	5.89999999999998\\
0.362020254	5.89999999999998\\
0.382161379	5.89999999999998\\
0.402102709	5.89999999999998\\
0.422178984	5.89999999999998\\
0.442130327	5.89999999999998\\
0.46200633	5.89999999999998\\
0.482154846	5.89999999999998\\
0.502096176	5.89999999999998\\
0.522157907	5.89999999999998\\
0.542132139	5.89999999999998\\
0.562018394	5.79999999999995\\
0.582155943	5.70000000000005\\
0.602100611	5.60000000000002\\
0.622170925	5.60000000000002\\
0.642130375	5.60000000000002\\
0.661995411	5.5\\
0.682149649	5.39999999999998\\
0.702078104	5.20000000000005\\
0.722169399	5\\
0.742068052	5\\
0.76201272	5\\
0.782187939	5.10000000000002\\
0.802058935	5.10000000000002\\
0.822188854	5\\
0.84211874	4.89999999999998\\
0.862021446	4.79999999999995\\
0.882141829	4.70000000000005\\
0.902095318	4.79999999999995\\
0.922189951	4.79999999999995\\
0.942067385	4.70000000000005\\
0.962010384	4.5\\
0.982141018	4.5\\
1.00208354	-1.60000000000002\\
1.022191286	-1.60000000000002\\
1.04208827	-1.60000000000002\\
1.062000513	-1.60000000000002\\
1.082139254	-1.60000000000002\\
1.10210681	-1.60000000000002\\
1.122172117	-1.60000000000002\\
1.142074108	-1.60000000000002\\
1.162016153	-1.60000000000002\\
1.182153702	-1.60000000000002\\
1.202116489	-1.5\\
1.222205639	-1.39999999999998\\
1.242153645	-1.29999999999995\\
1.26202631	-1.29999999999995\\
1.282181978	-1.29999999999995\\
1.302097559	-1.20000000000005\\
1.322151184	-1.10000000000002\\
1.342127562	-1\\
1.362022638	-1\\
1.382176876	-1\\
1.402080297	-1\\
1.422161579	-0.899999999999977\\
1.442111731	-0.799999999999955\\
1.462035418	-0.600000000000023\\
1.482175827	-0.600000000000023\\
1.502425671	-0.700000000000045\\
1.522183895	-0.799999999999955\\
1.542054653	-1\\
1.562038183	-1\\
1.582180977	-0.899999999999977\\
1.602072954	-0.899999999999977\\
1.622181654	-1\\
1.64208889	-0.899999999999977\\
1.662036657	-0.799999999999955\\
1.682142735	-0.600000000000023\\
1.702091694	-0.299999999999955\\
1.722146988	-0.200000000000045\\
1.742088318	-0.299999999999955\\
1.762031794	-0.399999999999977\\
1.782210588	-0.5\\
1.802090406	-0.5\\
1.822175741	-0.399999999999977\\
1.842198849	-0.399999999999977\\
1.862034321	-0.5\\
1.882190228	-0.5\\
1.902107477	-0.5\\
1.922172308	-0.399999999999977\\
1.942081451	-0.399999999999977\\
1.962022543	-0.399999999999977\\
1.982192516	-0.399999999999977\\
2.002118826	-0.600000000000023\\
2.02216363	-0.700000000000045\\
2.04212141	-0.899999999999977\\
2.062012196	-0.899999999999977\\
2.082183361	-1\\
2.102087498	-0.899999999999977\\
2.122143269	-0.899999999999977\\
2.142081261	-0.799999999999955\\
2.162018299	-0.799999999999955\\
2.182207823	-0.799999999999955\\
2.202086687	-0.700000000000045\\
2.222201586	-0.5\\
2.242088079	-0.299999999999955\\
2.262051105	-0.200000000000045\\
2.282233238	-0.200000000000045\\
2.302112341	-0.299999999999955\\
2.322165251	-0.299999999999955\\
2.342108011	-0.299999999999955\\
2.362074614	-0.200000000000045\\
2.382153273	-0.200000000000045\\
2.402099133	-0.299999999999955\\
2.422190428	-0.299999999999955\\
2.44206953	-0.200000000000045\\
2.462044001	-0.100000000000023\\
2.482162237	-0.100000000000023\\
2.50207901	-0.200000000000045\\
2.522173405	-0.299999999999955\\
2.542060137	-0.399999999999977\\
2.562032938	-0.399999999999977\\
2.582175016	-0.399999999999977\\
2.602091312	-0.399999999999977\\
2.622170687	-0.399999999999977\\
2.642087221	-0.399999999999977\\
2.661997318	-0.399999999999977\\
2.682151079	-0.399999999999977\\
2.702081203	-0.299999999999955\\
2.722151279	-0.200000000000045\\
2.742081404	0\\
2.762029409	0\\
2.782158852	-0.100000000000023\\
2.802078962	-0.200000000000045\\
2.822157145	-0.200000000000045\\
2.842126608	-0.299999999999955\\
2.862025261	-0.399999999999977\\
2.882153511	-0.5\\
2.902089834	-0.5\\
2.922161102	-0.600000000000023\\
2.9420681	-0.5\\
2.962011576	-0.5\\
2.982153654	-0.399999999999977\\
3.002092123	-0.399999999999977\\
3.022182465	-0.299999999999955\\
3.042096138	-0.299999999999955\\
3.062021494	-0.399999999999977\\
3.082163334	-0.5\\
3.102095366	-0.700000000000045\\
3.1221807	-1\\
3.142077208	-1.39999999999998\\
3.16202116	-1.60000000000002\\
3.182166815	-1.5\\
3.20211935	-1.29999999999995\\
3.222192287	-0.899999999999977\\
3.242128134	-0.700000000000045\\
3.262024879	-0.700000000000045\\
3.282130003	-0.899999999999977\\
3.302080154	-1.10000000000002\\
3.322163582	-1.10000000000002\\
3.342069626	-1\\
3.361999273	-0.899999999999977\\
3.382153988	-1\\
3.402083397	-1.10000000000002\\
3.422180891	-1.10000000000002\\
3.442085266	-1.20000000000005\\
3.462021589	-1.20000000000005\\
3.482152939	-1.20000000000005\\
3.502100468	-1.20000000000005\\
3.522212744	-1.20000000000005\\
3.542113781	-1.20000000000005\\
3.562012196	-1.10000000000002\\
3.582176685	-1\\
3.602067471	-1.10000000000002\\
3.622155666	-1.20000000000005\\
3.642103434	-1.20000000000005\\
3.662031174	-1.20000000000005\\
3.682188272	-1.10000000000002\\
3.702080488	-1.10000000000002\\
3.722182035	-1.10000000000002\\
3.742081165	-1.10000000000002\\
3.762041092	-1.10000000000002\\
3.782184362	-1.10000000000002\\
3.802086115	-1\\
3.822208643	-0.899999999999977\\
3.842411518	-0.799999999999955\\
3.862011194	-0.700000000000045\\
3.882148266	-0.600000000000023\\
3.902097464	-0.600000000000023\\
3.922157049	-0.5\\
3.942069292	-0.5\\
3.962007523	-0.5\\
3.982177258	-0.5\\
4.002087593	-0.5\\
4.022173405	-0.399999999999977\\
4.042154551	-0.200000000000045\\
4.06201911	0\\
4.082152367	0\\
4.102093697	-0.200000000000045\\
4.122160673	-0.399999999999977\\
4.142062426	-0.399999999999977\\
4.162030935	-0.299999999999955\\
4.182180405	-0.200000000000045\\
4.202072144	-0.200000000000045\\
4.222162008	-0.200000000000045\\
4.242100716	-0.200000000000045\\
4.262014389	-0.100000000000023\\
4.28219533	0.200000000000045\\
4.302089214	0.600000000000023\\
4.322158575	1.20000000000005\\
4.342076063	1.70000000000005\\
4.362028837	2\\
4.382173061	2\\
4.402078867	1.89999999999998\\
4.4221766	1.60000000000002\\
4.442105055	1.39999999999998\\
4.46200633	1.39999999999998\\
4.482172966	1.5\\
4.502092361	1.60000000000002\\
4.522158146	1.60000000000002\\
4.542140245	1.60000000000002\\
4.562050581	1.70000000000005\\
4.582169533	1.79999999999995\\
4.60207653	2\\
4.622186661	2.29999999999995\\
4.642095804	2.60000000000002\\
4.662044764	2.79999999999995\\
4.682162046	2.89999999999998\\
4.702105522	2.89999999999998\\
4.722176552	2.79999999999995\\
4.742126703	2.79999999999995\\
4.762061596	2.89999999999998\\
4.782148838	2.79999999999995\\
4.802083969	2.79999999999995\\
4.82220006	2.79999999999995\\
4.842048645	2.79999999999995\\
4.862048626	2.79999999999995\\
4.882157326	2.89999999999998\\
4.902056932	2.89999999999998\\
4.92215395	3.10000000000002\\
4.942088842	3.29999999999995\\
4.962026834	3.60000000000002\\
4.982174397	4\\
5.002012014	4.39999999999998\\
5.022206068	4.70000000000005\\
5.042008877	4.89999999999998\\
5.062007904	4.89999999999998\\
5.082171679	4.79999999999995\\
5.10203433	4.79999999999995\\
5.122157097	4.89999999999998\\
5.141997099	5.10000000000002\\
5.161998987	5.29999999999995\\
5.182173491	5.29999999999995\\
5.202002525	5.29999999999995\\
5.222178936	5.29999999999995\\
5.242048025	5.29999999999995\\
5.261997223	5.39999999999998\\
5.282162666	5.60000000000002\\
5.302005768	5.79999999999995\\
5.322175503	5.89999999999998\\
5.342036486	5.89999999999998\\
5.362068892	5.89999999999998\\
5.382175207	6\\
5.402026415	6.10000000000002\\
5.422189951	6.10000000000002\\
5.442039967	6.20000000000005\\
5.462042093	6.20000000000005\\
5.482180834	6.20000000000005\\
5.502022743	6.20000000000005\\
5.522157431	6.20000000000005\\
5.542028427	6.20000000000005\\
5.562045097	6.10000000000002\\
5.582159758	6.20000000000005\\
5.602006435	6.20000000000005\\
5.622204542	6.20000000000005\\
5.642046213	6.29999999999995\\
5.662028074	6.29999999999995\\
5.682136536	6.29999999999995\\
5.702010393	6.20000000000005\\
5.722163677	6.29999999999995\\
5.741996527	6.29999999999995\\
5.762047768	6.29999999999995\\
5.782159328	6.20000000000005\\
5.802027225	6.10000000000002\\
5.822189093	5.89999999999998\\
5.842361212	5.60000000000002\\
5.862008333	5.29999999999995\\
5.882154942	5.10000000000002\\
5.902008057	4.89999999999998\\
5.922199488	4.89999999999998\\
5.941994667	4.79999999999995\\
5.962021589	4.79999999999995\\
5.98218298	4.5\\
6.002015829	4.20000000000005\\
6.022216558	3.89999999999998\\
6.042048216	3.70000000000005\\
6.062056303	3.70000000000005\\
6.08216095	3.79999999999995\\
6.102020502	3.79999999999995\\
6.122161865	3.70000000000005\\
6.142045021	3.60000000000002\\
6.16202569	3.39999999999998\\
6.182163954	3.29999999999995\\
6.202054739	3.10000000000002\\
6.222176075	2.89999999999998\\
6.242068768	2.60000000000002\\
6.262020826	2.39999999999998\\
6.282161474	2.20000000000005\\
6.302022219	2\\
6.322189093	1.70000000000005\\
6.342045546	1.20000000000005\\
6.362016678	0.799999999999955\\
6.38218379	0.5\\
6.402023554	0.200000000000045\\
6.42219162	0\\
6.442023993	-0.100000000000023\\
6.462028742	-0.399999999999977\\
6.482155561	-0.600000000000023\\
6.502026796	-0.799999999999955\\
6.522170305	-1\\
6.542007208	-1.10000000000002\\
6.562035084	-1.20000000000005\\
6.582168818	-1.29999999999995\\
6.60201931	-1.60000000000002\\
6.62219882	-1.79999999999995\\
6.642049789	-1.89999999999998\\
6.662009001	-2\\
6.68214035	-2\\
6.702040672	-2\\
6.722191811	-2\\
6.742039204	-2\\
6.762017965	-2\\
6.782172441	-2.10000000000002\\
6.802040339	-2.10000000000002\\
6.822195053	-2.10000000000002\\
6.842060566	-2.10000000000002\\
6.862022638	-2.10000000000002\\
6.882184267	-2.10000000000002\\
6.902058125	-2.10000000000002\\
6.922169924	-2.10000000000002\\
6.942036152	-2.10000000000002\\
6.962060928	-2.10000000000002\\
6.982192516	-2.10000000000002\\
7.001996994	-2.10000000000002\\
7.022202253	-2\\
7.042022705	-1.79999999999995\\
7.062073708	-1.70000000000005\\
7.082153797	-1.60000000000002\\
7.102023363	-1.39999999999998\\
7.122164726	-1.20000000000005\\
7.141997576	-1.10000000000002\\
7.162062645	-0.899999999999977\\
7.182226181	-0.700000000000045\\
7.202002287	-0.399999999999977\\
7.222168207	-0.100000000000023\\
7.242021084	0.200000000000045\\
7.262044907	0.399999999999977\\
7.282160997	0.600000000000023\\
7.302033901	0.899999999999977\\
7.322174311	1.29999999999995\\
7.342018366	1.89999999999998\\
7.362043142	2.60000000000002\\
7.382188559	3.29999999999995\\
7.402025938	3.89999999999998\\
7.422181129	4.29999999999995\\
7.442045927	4.60000000000002\\
7.461986542	4.89999999999998\\
7.482157707	5.39999999999998\\
7.502041578	6\\
7.52215457	7\\
7.541994095	8\\
7.562044144	9.10000000000002\\
7.582208633	9.89999999999998\\
7.601991415	10.6\\
7.622181654	11.3\\
7.642034292	12\\
7.662019968	12.7\\
7.682176352	13.4\\
7.702009916	14.1\\
7.722177982	14.7\\
7.742027044	15.3\\
7.76205492	15.8\\
7.78215456	16.3\\
7.802002907	16.8\\
7.82217288	17.5\\
7.842028141	18.3\\
7.862007141	19.1\\
7.882144928	19.8\\
7.902017117	20.3\\
7.922207117	20.5\\
7.941999197	20.5\\
7.962014675	20.5\\
7.982160568	20.6\\
8.00202775	20.8\\
8.022207737	21\\
8.042006493	21.3\\
8.062014818	21.5\\
8.082175493	21.8\\
8.102050066	21.9\\
8.12214613	22.1\\
8.142054081	22.2\\
8.162019491	22.3\\
8.182166815	22.4\\
8.202024937	22.4\\
8.222177029	22.5\\
8.242496252	22.5\\
8.262017727	22.6\\
8.282183409	22.7\\
8.302018642	22.8\\
8.322143316	23.1\\
8.341991186	23.4\\
8.362090826	23.8\\
8.382203579	24.2\\
8.402018785	24.6\\
8.422168493	24.9\\
8.442065001	25.4\\
8.462030411	25.8\\
8.482177019	26.1\\
8.502048016	26.5\\
8.522155046	26.9\\
8.541997671	27.4\\
8.562011242	28.3\\
8.58215785	29.4\\
8.602002859	30.6\\
8.622174025	31.9\\
8.642019033	33\\
8.662014723	33.9\\
8.682170153	34.8\\
8.702004671	35.7\\
8.72217226	36.8\\
8.742035627	37.9\\
8.762006521	39.2\\
8.782174349	40.6\\
8.802015543	42\\
8.822201967	43.7\\
8.84207201	45.7\\
8.862038374	47.5\\
8.882151842	49.3\\
8.90201664	50.9\\
8.9221735	52.4\\
8.942082882	53.7\\
8.962004662	55.2\\
8.982151985	56.8\\
9.002033949	58.4\\
9.022185564	60.1\\
9.042088509	61.8\\
9.062009811	63.6\\
9.082170963	65.5\\
9.102047443	67.4\\
9.122205019	69.3\\
9.142018557	71.1\\
9.162004709	72.7\\
9.182167768	74.1\\
9.202019215	75.5\\
9.222215652	77\\
9.2420578	78.4\\
9.262014627	79.8\\
9.282175064	81.1\\
9.302053452	82.3\\
9.322166443	83.4\\
9.342011452	84.5\\
9.362027884	85.6\\
9.382169485	86.7\\
9.40199542	87.9\\
9.422189236	89.1\\
9.442024231	90.3\\
9.462031126	91.5\\
9.482210636	92.7\\
9.50201416	93.9\\
9.522164106	95.1\\
9.541999578	96.3\\
9.562057734	97.2\\
9.58217001	97.9\\
9.602027416	98.7\\
9.62221837	99.7\\
9.642043591	101.1\\
9.662041426	102.7\\
9.682175636	104.6\\
9.702047348	106.3\\
9.72215867	108\\
9.742034674	109.4\\
9.762027502	110.5\\
9.782195807	111.8\\
9.802001715	113\\
9.822241306	114.1\\
9.842035294	115.2\\
9.861996412	116.2\\
9.882143974	117.2\\
9.902062178	118\\
9.92215991	119\\
9.941990614	119.9\\
9.962040186	120.7\\
9.982160091	121.3\\
10.002008915	121.8\\
10.022162676	121.9\\
10.042040348	122\\
10.062033176	122.2\\
10.082159519	122.3\\
10.102014542	122.5\\
10.122166157	122.6\\
10.14205265	122.7\\
10.162023067	122.8\\
10.182164669	122.8\\
10.20201087	122.8\\
10.222209692	122.8\\
10.242053509	122.8\\
10.262000322	122.8\\
10.282187939	122.6\\
10.301998377	122.4\\
10.32216692	121.9\\
10.342014313	121.2\\
10.362031221	120\\
10.382137775	118.8\\
10.402019739	117.8\\
10.422216654	116.8\\
10.442014933	116.1\\
10.462009192	115.3\\
10.482165575	114.4\\
10.502165794	113.4\\
10.522184372	112.2\\
10.542012691	111\\
10.562042475	109.9\\
10.582167625	108.5\\
10.602003336	107.2\\
10.62219882	105.8\\
10.642024279	104.4\\
10.662052393	102.9\\
10.682150364	101.6\\
10.702017546	100.2\\
10.722202063	98.6\\
10.742022038	97\\
10.762051105	95.2\\
10.782175779	93.4\\
10.802012682	91.6\\
10.822185516	89.8\\
10.842025518	87.8\\
10.862024546	85.8\\
10.882167339	83.7\\
10.901996136	81.7\\
10.922170639	79.7\\
10.942038536	77.9\\
10.962023258	76\\
10.982150793	73.9\\
11.002026796	71.7\\
11.022168398	69.3\\
11.042107582	66.5\\
11.062018871	63.5\\
11.082143784	60.5\\
11.102014303	57.7\\
11.122162104	55.1\\
11.142023325	52.8\\
11.162003279	50.7\\
11.182168961	48.7\\
11.202023745	46.6\\
11.222198009	44.3\\
11.242036581	41.8\\
11.262020588	38.8\\
11.282158136	35.6\\
11.302039862	32.3\\
11.322206497	29\\
11.341994762	25.6\\
11.362016439	22.3\\
11.382194281	19.2\\
11.402049065	16.3\\
11.42216301	13.6\\
11.44204402	10.9\\
11.462008238	8.39999999999998\\
11.482185125	5.79999999999995\\
11.502063513	3\\
11.522179365	0.100000000000023\\
11.542031527	-3\\
11.562044382	-6.29999999999995\\
11.582202435	-9.89999999999998\\
11.602005959	-13.3\\
11.622162104	-16.8\\
11.642018795	-20\\
11.662035704	-22.9\\
11.682209015	-25.9\\
11.701995373	-28.6\\
11.722155094	-31.4\\
11.742041111	-34.2\\
11.76204586	-37.1\\
11.782194138	-40.1\\
11.802010298	-43.3\\
11.822166204	-46.6\\
11.842052221	-49.9\\
11.862076521	-52.9\\
11.882140398	-55.7\\
11.902008533	-58.4\\
11.922194719	-61.1\\
11.942014456	-63.9\\
11.962010145	-67\\
11.982151985	-70.2\\
12.002012491	-73.4\\
12.022212505	-76.3\\
12.042027712	-79\\
12.062018633	-81.4\\
12.082137346	-83.7\\
12.102040052	-86.1\\
12.122159719	-88.9\\
12.142022371	-92.1\\
12.162007332	-95.5\\
12.182183743	-99.1\\
12.202021599	-102.5\\
12.22219348	-105.8\\
12.242077351	-108.8\\
12.262014866	-111.8\\
12.282147408	-114.7\\
12.302027941	-117.8\\
12.322198868	-120.8\\
12.342032194	-123.9\\
12.362026215	-126.8\\
12.382187128	-129.8\\
12.402066231	-132.8\\
12.422172785	-136.1\\
12.442011118	-139.5\\
12.462028265	-143\\
12.482155561	-146.6\\
12.50212431	-150.1\\
12.522170067	-153.6\\
12.542024374	-156.8\\
12.562016249	-159.8\\
12.582173347	-162.9\\
12.60203433	-166.1\\
12.62221384	-169.8\\
12.642044067	-173.3\\
12.662013531	-176.7\\
12.682140589	-180.1\\
12.702059507	-183.5\\
12.722218513	-186.6\\
12.742001295	-189.7\\
12.762009859	-192.5\\
12.782172441	-195.2\\
12.802034855	-197.9\\
12.822225571	-200.4\\
12.842023134	-202.7\\
12.862039804	-205\\
12.882176399	-207.4\\
12.902061462	-209.9\\
12.922202587	-212.3\\
12.942026854	-214.7\\
12.962030411	-217.1\\
12.982169151	-219.1\\
13.002050877	-221.2\\
13.022166729	-223.2\\
13.042048693	-225.1\\
13.062007904	-226.6\\
13.082170963	-227.8\\
13.102045774	-228.7\\
13.122222662	-229.4\\
13.142009258	-230.1\\
13.162029028	-230.9\\
13.182162762	-231.8\\
13.202048779	-232.6\\
13.222162247	-233.2\\
13.24207449	-233.6\\
13.262017488	-233.7\\
13.282202244	-233.6\\
13.302010059	-233.5\\
13.322175026	-233.4\\
13.342008591	-233.2\\
13.362028837	-232.9\\
13.382196903	-232.5\\
13.402005672	-231.9\\
13.422161102	-231\\
13.442029476	-230\\
13.462056637	-229\\
13.48216176	-228\\
13.502006292	-227.1\\
13.522183418	-226.1\\
13.542004347	-225\\
13.56208086	-223.4\\
13.582190752	-221.5\\
13.602019787	-219.5\\
13.622189045	-217.9\\
13.642015696	-216.5\\
13.662033081	-215.4\\
13.682160139	-214.3\\
13.702014208	-213.1\\
13.722155094	-211.5\\
13.741992235	-209.8\\
13.76205492	-208.1\\
13.782170057	-206.4\\
13.801992893	-204.7\\
13.822155237	-202.5\\
13.842068911	-200.1\\
13.86199522	-197.5\\
13.882196903	-195\\
13.902000666	-193.1\\
13.922158957	-191.7\\
13.942016602	-190.5\\
13.96201849	-189.4\\
13.982173681	-187.9\\
14.002015829	-186.2\\
14.022208691	-184.4\\
14.042024612	-182.5\\
14.062039852	-180.3\\
14.082187176	-177.9\\
14.102016687	-175.3\\
14.122161388	-173\\
14.142022133	-171\\
14.162013292	-169.5\\
14.182175159	-168.2\\
14.201994419	-166.9\\
14.222209215	-165.5\\
14.242069721	-163.8\\
14.262035131	-162\\
14.282157898	-160.1\\
14.302041531	-158\\
14.322193146	-155.5\\
14.342046499	-153\\
14.362018108	-150.5\\
14.382190943	-148.4\\
14.402009964	-146.6\\
14.422169924	-145.2\\
14.442035675	-143.9\\
14.462081909	-142.5\\
14.48216176	-141\\
14.502019167	-139.3\\
14.52216506	-137.7\\
14.542030334	-135.9\\
14.562053204	-133.9\\
14.582164764	-131.3\\
14.602011204	-128.5\\
14.62218523	-125.7\\
14.64201045	-123.2\\
14.66202426	-121.3\\
14.682141304	-119.8\\
14.702003956	-118.5\\
14.722154617	-117\\
14.742044926	-115.2\\
14.762069941	-113.1\\
14.782160044	-110.6\\
14.802019119	-108.1\\
14.822211266	-105.3\\
14.842029095	-102.5\\
14.862046957	-100\\
14.882148504	-97.7\\
14.902021885	-95.5\\
14.922165394	-93.4\\
14.942038774	-91.3\\
14.962046146	-89.2\\
14.982152224	-86.9\\
15.002025843	-84.9\\
15.022171974	-82.8\\
15.042080164	-80.6\\
15.062054157	-78\\
15.082161188	-75\\
15.102033854	-71.9\\
15.122178078	-68.9\\
15.142013073	-66\\
15.162030697	-63.4\\
15.182167053	-61\\
15.202033281	-58.6\\
15.222221851	-56.2\\
15.2420187	-53.5\\
15.262054443	-50.5\\
15.282158136	-47.4\\
15.302040577	-43.9\\
15.322177887	-40.4\\
15.342006445	-36.8\\
15.362061739	-33.4\\
15.382174253	-30.1\\
15.401994705	-26.9\\
15.422181129	-24\\
15.442026138	-21.1\\
15.462063313	-18\\
15.482153177	-14.7\\
15.502030611	-11.5\\
15.522151232	-8.20000000000005\\
15.542061329	-4.70000000000005\\
15.562007427	-1.20000000000005\\
15.582166672	2.5\\
15.602015734	6.29999999999995\\
15.622195959	9.89999999999998\\
15.642052889	13.4\\
15.662012339	16.5\\
15.682186842	19.2\\
15.702052832	21.9\\
15.722197056	24.4\\
15.74203229	27\\
15.762060881	29.8\\
15.782194376	32.6\\
15.802023888	35.4\\
15.822170258	38.1\\
15.842051506	41\\
15.862038612	43.7\\
15.88216114	46.4\\
15.901991606	49\\
15.922177792	51.4\\
15.942011356	53.7\\
15.962031841	55.9\\
15.982192993	58\\
16.002021551	60.1\\
16.022186518	62.3\\
16.042070389	64.5\\
16.062039375	66.5\\
16.082175016	68.4\\
16.102033615	69.9\\
16.122169256	71.2\\
16.142048597	72.5\\
16.162024975	74.1\\
16.182143211	76\\
16.202010155	77.9\\
16.222177744	79.5\\
16.242027044	80.9\\
16.262030125	82.1\\
16.282165766	83.4\\
16.302013397	84.7\\
16.322194338	85.8\\
16.342005014	86.9\\
16.362016916	87.8\\
16.382160902	88.8\\
16.40202713	89.9\\
16.422223806	91\\
16.442056656	92.4\\
16.462023735	93.9\\
16.482177973	95.2\\
16.502076387	96.5\\
16.522184372	97.4\\
16.542040825	98.1\\
16.562006235	98.5\\
16.582221985	98.8\\
16.602005959	99.3\\
16.622155905	100\\
16.642079353	100.8\\
16.66200304	101.7\\
16.682188034	102.4\\
16.702009439	103.2\\
16.722154617	104.1\\
16.742036343	105\\
16.76201272	105.9\\
16.782194853	106.7\\
16.802005053	107.4\\
16.822202444	107.7\\
16.842010736	107.5\\
16.861999512	107\\
16.882180214	106.8\\
16.902031898	106.8\\
16.922189713	107.1\\
16.942047596	107.3\\
16.962030411	107.2\\
16.982206821	107\\
17.002003193	106.9\\
17.022186279	106.8\\
17.042018175	106.7\\
17.062042475	106.2\\
17.082158804	105.5\\
17.102040529	104.3\\
17.122158766	103.1\\
17.142002344	102.1\\
17.162016392	101.7\\
17.182179689	101.3\\
17.201996088	101\\
17.222167015	100.5\\
17.2420187	99.8\\
17.262016535	98.9\\
17.282165527	97.8\\
17.302018881	96.4\\
17.322158813	94.9\\
17.342066288	93.4\\
17.36203742	92\\
17.382156134	90.8\\
17.401993513	89.8\\
17.422166586	88.8\\
17.442014456	87.9\\
17.462014198	86.8\\
17.482141733	85.7\\
17.502054214	84.5\\
17.522158623	83.3\\
17.54201293	82\\
17.562024117	80.4\\
17.582157373	78.5\\
17.602024794	76.6\\
17.622184992	74.7\\
17.642039776	73.1\\
17.662021875	71.8\\
17.6821661	70.8\\
17.702030182	69.8\\
17.722173929	68.7\\
17.742020369	67.3\\
17.762018919	65.7\\
17.782183647	63.9\\
17.802010298	61.9\\
17.82222414	59.8\\
17.842059374	57.7\\
17.862042665	55.6\\
17.882157326	53.8\\
17.902052641	52.2\\
17.922207832	50.9\\
17.942019939	49.9\\
17.962049961	48.7\\
17.982177734	47.4\\
18.002047062	46\\
18.022161961	44.5\\
18.042015553	42.6\\
18.062008619	40.6\\
18.082187176	38.3\\
18.102060556	36.2\\
18.12223959	34.3\\
18.142025948	32.8\\
18.162018299	31.6\\
18.18220377	30.7\\
18.20204711	29.8\\
18.222159147	28.7\\
18.242021084	27.3\\
18.262038469	25.8\\
18.282178402	24.2\\
18.302013874	22.6\\
18.322201729	20.9\\
18.342014313	19.2\\
18.362061024	17.7\\
18.382189989	16.1\\
18.402012348	14.7\\
18.422197104	13.7\\
18.44211483	12.8\\
18.46204257	11.7\\
18.482144117	10.8\\
18.502032518	9.89999999999998\\
18.522205114	8.79999999999995\\
18.542003393	7.79999999999995\\
18.56207037	6.60000000000002\\
18.58215332	5.39999999999998\\
18.601989031	4\\
18.622177839	2.60000000000002\\
18.642030239	1.29999999999995\\
18.662031651	0.299999999999955\\
18.682155609	-0.399999999999977\\
18.702023029	-0.799999999999955\\
18.722157001	-1.20000000000005\\
18.742028713	-1.5\\
18.762127161	-2.10000000000002\\
18.782155275	-3\\
18.802005768	-4.20000000000005\\
18.822184801	-5.39999999999998\\
18.842027903	-6.5\\
18.862023354	-7.5\\
18.88215661	-8.10000000000002\\
18.902007818	-8.39999999999998\\
18.922172785	-8.60000000000002\\
18.942003012	-8.89999999999998\\
18.962057114	-9.20000000000005\\
18.982169151	-9.60000000000002\\
19.002014637	-10.2\\
19.022171259	-10.7\\
19.042019367	-11.3\\
19.062022209	-11.9\\
19.082147837	-12.4\\
19.102004051	-13\\
19.122156858	-13.5\\
19.142049789	-13.9\\
19.162013531	-14.1\\
19.182175159	-14.3\\
19.202008724	-14.5\\
19.222201824	-14.8\\
19.242077351	-15.3\\
19.262018919	-15.9\\
19.282177448	-16.5\\
19.302042007	-17\\
19.32219243	-17.2\\
19.342000008	-17.2\\
19.36203742	-17.2\\
19.3821702	-17.3\\
19.402042627	-17.5\\
19.422222853	-17.9\\
19.442041636	-18.3\\
19.462012053	-18.8\\
19.482167006	-19.2\\
19.502070189	-19.4\\
19.522199869	-19.6\\
19.542028427	-19.8\\
19.562001705	-19.9\\
19.582165003	-20\\
19.602050543	-20\\
19.622229338	-20.1\\
19.642035484	-20.2\\
19.662012577	-20.4\\
19.682184458	-20.7\\
19.702040434	-20.9\\
19.722159863	-21\\
19.742070913	-21\\
19.76202178	-21.1\\
19.782196522	-21.1\\
19.802010298	-21.2\\
19.822179794	-21.2\\
19.84202981	-21.3\\
19.862037182	-21.3\\
19.88216424	-21.4\\
19.902036905	-21.5\\
19.922178745	-21.6\\
19.942070246	-21.6\\
19.962021589	-21.6\\
19.982163429	-21.7\\
20.002016068	-21.8\\
20.022198677	-21.9\\
20.042007446	-21.9\\
20.062027216	-21.9\\
20.082159042	-21.9\\
20.102087975	-21.8\\
20.122168541	-21.7\\
20.142026424	-21.7\\
20.162014246	-21.8\\
20.182190418	-21.8\\
20.202024937	-21.8\\
20.222181797	-21.7\\
20.242038488	-21.6\\
20.261988878	-21.5\\
20.282178164	-21.3\\
20.302005768	-21.1\\
20.322232485	-20.8\\
20.342035294	-20.5\\
20.36203599	-20.3\\
20.382168293	-20\\
20.401994467	-19.9\\
};
\addplot [color=black, forget plot]
  table[row sep=crcr]{%
4.25	-250\\
4.25	150\\
};
\node[right, align=left, circle,fill=white,draw=black,thick,line width=0.3mm,inner sep=1pt]
at (axis cs:2.35,-150) {1};
\addplot [color=black, forget plot]
  table[row sep=crcr]{%
7.1800000667572	-250\\
7.1800000667572	150\\
};
\node[right, align=left, circle,fill=white,draw=black,thick,line width=0.3mm,inner sep=1pt]
at (axis cs:5.28,-150) {2};
\addplot [color=black, forget plot]
  table[row sep=crcr]{%
11.9800000190735	-250\\
11.9800000190735	150\\
};
\node[right, align=left, circle,fill=white,draw=black,thick,line width=0.3mm,inner sep=1pt]
at (axis cs:10.08,-150) {3};
\end{axis}
\end{tikzpicture}%

%% file: Images/Tikz/fig11-3-1.tikz
\begin{tikzpicture}

\begin{axis}[%
width=4.25cm,
height=1.683cm,
at={(0cm,0cm)},
scale only axis,
xmin=0.042082548,
xmax=31.541993618,
xlabel style={font=\color{white!15!black}},
xlabel={$t\;[\SI[per-mode=repeated-symbol]{}{\second}]$},
ymin=0,
ymax=30,
ylabel style={font=\color{white!15!black}},
ylabel={$v\;[\SI[per-mode=repeated-symbol]{}{\meter\per\second}]$},
axis background/.style={fill=white},
ylabel style={yshift=-0.4cm}
]
\addplot [color=blue, line width=1.0pt, forget plot]
  table[row sep=crcr]{%
0.042082548	10.3650108538293\\
0.062074423	10.3700108485961\\
0.08214736	10.3600048262537\\
0.102083445	10.3650108538293\\
0.122158766	10.3650108538293\\
0.142064571	10.3550048285841\\
0.162027836	10.3550048285841\\
0.182141781	10.3500048309167\\
0.202076912	10.355\\
0.222186804	10.3500012077294\\
0.24206996	10.345\\
0.262013435	10.3450012083131\\
0.282138109	10.345\\
0.302066565	10.3400012088974\\
0.322176218	10.3400012088974\\
0.342066765	10.3450012083131\\
0.36198473	10.3450012083131\\
0.38214159	10.34\\
0.402073145	10.3400012088974\\
0.422165871	10.3350012094823\\
0.44208169	10.3350048379282\\
0.462033033	10.3350108853353\\
0.482142448	10.3350193516993\\
0.502058506	10.3350193516993\\
0.522171021	10.3300193610661\\
0.542063236	10.3250193704419\\
0.562014103	10.3250193704419\\
0.582153797	10.3200193798268\\
0.602102041	10.3200193798268\\
0.622177362	10.3200193798268\\
0.642128944	10.3150193892207\\
0.662020683	10.3150193892207\\
0.682162285	10.3100193986238\\
0.702095509	10.3050194080361\\
0.722220182	10.3050194080361\\
0.742111921	10.3000194174574\\
0.762013197	10.3000303397611\\
0.782218933	10.2950303544963\\
0.802009821	10.2950194268879\\
0.822173595	10.2900109329388\\
0.842021942	10.2850109382538\\
0.862068653	10.2850048614476\\
0.882125616	10.2850048614476\\
0.902014017	10.2850012153621\\
0.922170401	10.2800012159532\\
0.94209528	10.2750048661789\\
0.96203804	10.2750048661789\\
0.982150078	10.2700109542298\\
1.001995802	10.2700109542298\\
1.022184372	10.2650109595655\\
1.041991234	10.2600109649064\\
1.062036276	10.2550109702525\\
1.082123518	10.2500109756039\\
1.102082253	10.2500109756039\\
1.122177362	10.2450195216993\\
1.141994715	10.2450195216993\\
1.162002563	10.2450195216993\\
1.182144642	10.2450195216993\\
1.202020168	10.2450305026388\\
1.222158432	10.2400195312314\\
1.241981506	10.2400195312314\\
1.262023687	10.2400195312314\\
1.282156467	10.2400109863222\\
1.302074909	10.2400109863222\\
1.322268009	10.2350109916893\\
1.342069626	10.2300109970615\\
1.362036943	10.2300195503234\\
1.382166862	10.2250195598835\\
1.401998997	10.2200195694529\\
1.422176838	10.2150195790317\\
1.442102194	10.2050195982173\\
1.462082386	10.2050195982173\\
1.482149124	10.2000196078243\\
1.50210166	10.1950306522345\\
1.522209644	10.1900306672748\\
1.542084455	10.1850306823298\\
1.56201601	10.1850196367017\\
1.582152605	10.1850196367017\\
1.602059126	10.1800110510746\\
1.622139215	10.1750049140037\\
1.6420784	10.175\\
1.662032366	10.17\\
1.682159185	10.1700012291051\\
1.702077627	10.165\\
1.722183228	10.165004918838\\
1.742052317	10.165004918838\\
1.761993408	10.1650196753376\\
1.782148838	10.1600196850203\\
1.802083254	10.1600307578275\\
1.822393179	10.160044291242\\
1.842071772	10.1550443130495\\
1.862071514	10.1550443130495\\
1.882152319	10.1500443348785\\
1.902098656	10.1450443567291\\
1.922165871	10.1450443567291\\
1.942107439	10.1400443786011\\
1.96204114	10.1350444004948\\
1.982191563	10.1350444004948\\
2.002092361	10.1300308489165\\
2.022161007	10.1300197433174\\
2.042059898	10.1250197530672\\
2.062010765	10.1200197628266\\
2.082136154	10.1100111275903\\
2.102054596	10.1100111275903\\
2.122177124	10.1000111386077\\
2.142044306	10.0950111441246\\
2.162009954	10.090011149647\\
2.182146549	10.0850198314133\\
2.202073097	10.0850198314133\\
2.222182274	10.0800111607081\\
2.242100477	10.0750111662469\\
2.262007952	10.0700111717912\\
2.282197952	10.0700049652421\\
2.302063227	10.0700049652421\\
2.322173834	10.0650012419274\\
2.342066526	10.0650012419274\\
2.362056971	10.0600012425447\\
2.38217783	10.0550012431625\\
2.402078152	10.0550012431625\\
2.422175407	10.0550012431625\\
2.442069769	10.050001243781\\
2.462165833	10.0450012444001\\
2.482060909	10.045\\
2.502035141	10.04\\
2.522202015	10.04\\
2.542052746	10.035\\
2.561997175	10.03\\
2.582178116	10.03\\
2.60208106	10.025\\
2.622206688	10.0200012475049\\
2.642068863	10.015\\
2.662001371	10.01\\
2.682148933	10\\
2.702095985	9.995\\
2.722168922	9.98500125187774\\
2.742069483	9.97500125313275\\
2.762212515	9.97000125376121\\
2.782103539	9.96500125439029\\
2.801998615	9.96000125502\\
2.822190285	9.955\\
2.842061758	9.95000125628133\\
2.86200285	9.94500125691294\\
2.882155657	9.94000503017981\\
2.902054071	9.93501132359697\\
2.922168732	9.93502013083013\\
2.942016125	9.93503145440416\\
2.962005377	9.93503145440416\\
2.982170582	9.93003147024218\\
3.002002239	9.92503148609615\\
3.022175312	9.92002016126984\\
3.042074442	9.92002016126984\\
3.062019825	9.91502017143687\\
3.082145929	9.91003153375407\\
3.102072716	9.90503154967212\\
3.122160673	9.90004545444111\\
3.142097235	9.89506189975586\\
3.162013769	9.88506196237535\\
3.182151079	9.88008097132812\\
3.202117682	9.87506202512166\\
3.222162008	9.86506208799519\\
3.24206996	9.86004563883961\\
3.262001038	9.85004568517324\\
3.282155991	9.84504570837536\\
3.302068949	9.84006224573808\\
3.322217226	9.8350622773829\\
3.342102528	9.82506234076914\\
3.361999273	9.8300623090599\\
3.382189989	9.83003179038603\\
3.402045965	9.83000508646867\\
3.422164679	9.825\\
3.442015648	9.81002038733865\\
3.462039471	9.8000624998007\\
3.482130527	9.80004591825977\\
3.502067804	9.80002040814202\\
3.522150517	9.79500127616122\\
3.542077541	9.78502043942679\\
3.56203866	9.78018404734798\\
3.582189322	9.77532736024733\\
3.602053165	9.77051175732367\\
3.622231483	9.77051175732367\\
3.64200592	9.76541473773644\\
3.662053108	9.75515504746081\\
3.682157516	9.74503206767428\\
3.702059746	9.74000513346887\\
3.722196341	9.74006288480726\\
3.742063999	9.74002053385926\\
3.762039423	9.73506291710536\\
3.782186747	9.72500128534696\\
3.802117586	9.725\\
3.822149277	9.7100630790948\\
3.842074156	9.7100205973005\\
3.86200738	9.7051043271054\\
3.882135153	9.70512879873317\\
3.902033567	9.69010448860073\\
3.922189236	9.68512906470533\\
3.942101002	9.68010459654233\\
3.962026596	9.67008272973918\\
3.982176304	9.66008281538\\
4.002102137	9.65508285826693\\
4.022188663	9.64506350419737\\
4.042072058	9.63506357010684\\
4.062014341	9.62506363615327\\
4.082127094	9.61504680175817\\
4.10209775	9.60503253508285\\
4.12220645	9.59502084416704\\
4.14205718	9.58500521648267\\
4.161990643	9.58000130480158\\
4.182168007	9.57000130616501\\
4.202083349	9.56001176777519\\
4.222192049	9.55002094238542\\
4.242104053	9.54503273959812\\
4.262029648	9.53503277393424\\
4.282184124	9.5250118110163\\
4.302097559	9.5150052548593\\
4.322163105	9.5100013144058\\
4.342095852	9.505\\
4.36203599	9.49500131648227\\
4.382214546	9.48500527147982\\
4.402095795	9.48001186708118\\
4.422175646	9.47502110815591\\
4.442076445	9.46503301631854\\
4.461998224	9.46003303376896\\
4.48216033	9.45002116399747\\
4.502059221	9.4400211864169\\
4.522181511	9.43501192368086\\
4.542098999	9.42501193633196\\
4.562008858	9.41500531067296\\
4.582146883	9.41000531349478\\
4.602077723	9.40000531914743\\
4.622160435	9.39001198082303\\
4.642110348	9.38001199359574\\
4.662068844	9.37001200639572\\
4.682150602	9.36501201280596\\
4.702089787	9.355012025647\\
4.722190619	9.34001204496011\\
4.742033005	9.33500535618486\\
4.76201272	9.32500536192875\\
4.78217864	9.32000536480532\\
4.802061319	9.31500536768498\\
4.822193384	9.30500537345358\\
4.842078686	9.30000537634253\\
4.862027645	9.29000134553274\\
4.882078886	9.28000134698266\\
4.902020931	9.27000134843571\\
4.922174454	9.26000134989191\\
4.94198966	9.25000135135125\\
4.962014675	9.24000135281375\\
4.982167959	9.23500135354619\\
5.002066851	9.22500135501345\\
5.022088051	9.22000135574827\\
5.042010069	9.21000135722031\\
5.062077045	9.20500543182892\\
5.082146645	9.19500543773629\\
5.101998568	9.1900054406948\\
5.122180223	9.18500544365653\\
5.142085791	9.17501226157219\\
5.162003517	9.16500545553575\\
5.182151556	9.16001228165116\\
5.202069044	9.15002185789739\\
5.222151041	9.1450218698481\\
5.24200058	9.13502189378876\\
5.262019396	9.13003422775621\\
5.28213954	9.12003426528651\\
5.301988125	9.11503428408253\\
5.32215786	9.10503432173652\\
5.341994524	9.10003434059455\\
5.36200428	9.09003437837283\\
5.382158279	9.08003441623434\\
5.401992321	9.07502203854073\\
5.422123909	9.06502206285236\\
5.442040682	9.05502208721768\\
5.462051153	9.05003453032087\\
5.48214221	9.04003456851798\\
5.502030611	9.03004983374954\\
5.522213697	9.02504986135811\\
5.542008877	9.01504991666713\\
5.562020063	9.00504997209899\\
5.58216095	8.99503474145598\\
5.602070093	8.98503478012189\\
5.62218833	8.9800222716873\\
5.642028093	8.97002229651632\\
5.662093639	8.96502230895161\\
5.682007074	8.95502233386383\\
5.702009678	8.95002234634082\\
5.722153664	8.94003495518893\\
5.741987944	8.93503497474968\\
5.761996508	8.92505042002565\\
5.782096148	8.92005044828784\\
5.80200243	8.91505047658172\\
5.822184324	8.90505053326482\\
5.842010736	8.90005056165413\\
5.862154245	8.89003515178652\\
5.882057905	8.88503517156798\\
5.902000189	8.87503521119776\\
5.922207832	8.87003523104615\\
5.942038298	8.86003527081016\\
5.962017298	8.85505081860065\\
5.982093573	8.84505087605493\\
6.002003193	8.84005090483081\\
6.02215147	8.83003539064255\\
6.042025089	8.82502266286042\\
6.062041759	8.81502268856978\\
6.082153082	8.81001276957077\\
6.102032185	8.80001278408162\\
6.122202635	8.7900056882803\\
6.142030954	8.78500569151779\\
6.162008762	8.77500569800385\\
6.182173014	8.7650057045047\\
6.202021122	8.75500571102041\\
6.222186804	8.75000571428385\\
6.241998672	8.74000572082193\\
6.262037039	8.73000572737498\\
6.282183409	8.72000143348612\\
6.302028894	8.71500143430854\\
6.322185278	8.70500143595623\\
6.342019558	8.6950014376077\\
6.362130165	8.68500143926298\\
6.382010698	8.67500144092207\\
6.402008057	8.66500144258499\\
6.422150373	8.66000577367013\\
6.441997766	8.65000578034489\\
6.462204456	8.64001302082352\\
6.482050896	8.63001303591136\\
6.501999855	8.6200130510342\\
6.52216506	8.61501305860879\\
6.541989326	8.60501307378438\\
6.562157154	8.59501308899527\\
6.582008123	8.5850131042416\\
6.602034807	8.57501311952349\\
6.6221416	8.56501313484107\\
6.642022371	8.55502337810949\\
6.662213326	8.54003659242746\\
6.681981802	8.53005275481928\\
6.702016592	8.51507193158108\\
6.722084761	8.50507201615601\\
6.742046356	8.49509417252099\\
6.762132645	8.48009433909789\\
6.782027721	8.47011953870782\\
6.802029848	8.45514784022136\\
6.822382689	8.44514801528073\\
6.842010736	8.43514819075516\\
6.862016201	8.42014845474829\\
6.882183313	8.41012039152829\\
6.901999235	8.39512060663812\\
6.922183514	8.38509540792471\\
6.942009211	8.37509552184332\\
6.962037086	8.36009569323222\\
6.982182264	8.35009580783358\\
7.001996994	8.33509598025122\\
7.022175312	8.32512162073324\\
7.042007923	8.31012184026203\\
7.062049866	8.30009638498253\\
7.082154036	8.28509655948559\\
7.102008104	8.27007406254624\\
7.122209072	8.25505451223673\\
7.142029524	8.24503790167153\\
7.162026167	8.23002430130069\\
7.182145119	8.21001370279003\\
7.201986551	8.19501372787136\\
7.22215271	8.18001375304467\\
7.242026567	8.16501377831048\\
7.262038469	8.14502455490467\\
7.282135963	8.13005535036509\\
7.30200696	8.11507547715978\\
7.322199821	8.09512507624187\\
7.341990471	8.08015470148932\\
7.362025499	8.06018765290238\\
7.382147789	8.04518800277532\\
7.402028084	8.03018835395534\\
7.422196388	8.01012640349701\\
7.442009211	7.99505628498011\\
7.462033749	7.9750141065706\\
7.482156515	7.96\\
7.502078533	7.94501415983634\\
7.522161245	7.92503943207856\\
7.54201889	7.91002528440965\\
7.562041998	7.89001425854225\\
7.582206249	7.87500158730143\\
7.602006674	7.86000636132058\\
7.622170448	7.84502549390376\\
7.641988993	7.8250575077759\\
7.662045956	7.81007842470228\\
7.682159185	7.79510262921535\\
7.702026844	7.78010282708397\\
7.722176552	7.76010309209871\\
7.742031097	7.74507908287578\\
7.762196302	7.73005821452853\\
7.782044888	7.71004053167037\\
7.80202055	7.69002600775836\\
7.822185278	7.67501465796646\\
7.841990948	7.65501469626284\\
7.8620193	7.64000654449982\\
7.882154465	7.62000164041977\\
7.902038813	7.60000164473666\\
7.922205448	7.58000164907634\\
7.941994429	7.56\\
7.962000608	7.54\\
7.982166052	7.52500166112939\\
8.002037287	7.50000166666648\\
8.02215147	7.48500167000649\\
8.042023897	7.46000167560303\\
8.062022448	7.44\\
8.082145929	7.42\\
8.102045536	7.400001689189\\
8.122267246	7.38000169376674\\
8.142007351	7.36000169836937\\
8.162024975	7.34000170299708\\
8.182183027	7.32000170765007\\
8.202043295	7.30000684931186\\
8.222168446	7.28000686812863\\
8.242039204	7.26000688704907\\
8.262290478	7.24500172532761\\
8.282201052	7.22001558170064\\
8.302013636	7.20500693962192\\
8.322190762	7.18500695893887\\
8.342035055	7.16501570130869\\
8.361993313	7.15001573424842\\
8.382153273	7.13001577838366\\
8.402017832	7.11002812933957\\
8.422385931	7.0950281888094\\
8.442017555	7.07502826849476\\
8.462018967	7.06001593482621\\
8.482140779	7.0400159800955\\
8.502005577	7.02500177935921\\
8.522188663	7.01000178316668\\
8.542012691	6.99000715307216\\
8.562009573	6.97000717359746\\
8.582140446	6.95500718906889\\
8.60201478	6.935\\
8.622196436	6.91500723065421\\
8.641987324	6.90001630432856\\
8.662022591	6.88502904859522\\
8.682162523	6.87004548747677\\
8.70200491	6.85004562028604\\
8.722198009	6.83006588548017\\
8.741976738	6.81006607897457\\
8.761998653	6.79504598954268\\
8.782134056	6.77504612530424\\
8.802022219	6.76004622765259\\
8.822170258	6.7450296515286\\
8.841993093	6.72501672860373\\
8.86208415	6.7050074571174\\
8.882139921	6.69000186846013\\
8.902025938	6.67500187265891\\
8.922168255	6.65500187828674\\
8.941991568	6.64000188252985\\
8.96199131	6.62500754716551\\
8.982138872	6.61000189107386\\
9.002015829	6.59000189681308\\
9.022140265	6.57500760455834\\
9.04202199	6.56000190548753\\
9.062007427	6.54\\
9.082135201	6.52500191570853\\
9.102025509	6.51000768048702\\
9.122186422	6.49000770415567\\
9.142015934	6.47500772200312\\
9.161997557	6.46000773993344\\
9.182144642	6.44000776397048\\
9.202023745	6.425001945525\\
9.222172022	6.41\\
9.242002964	6.395\\
9.2621696	6.375\\
9.2820158	6.355\\
9.302052498	6.34\\
9.322240829	6.325\\
9.3420403	6.31000198098226\\
9.362011194	6.29000198728109\\
9.382217407	6.27501792826124\\
9.402034044	6.26001797122021\\
9.42215085	6.24003205119974\\
9.442016363	6.22503212843115\\
9.462054968	6.2050181305134\\
9.482158661	6.18500808406909\\
9.502022028	6.1700020259316\\
9.522222281	6.15500203086888\\
9.542011261	6.13500203748947\\
9.562206268	6.12000816992919\\
9.582033873	6.1050081900027\\
9.601994753	6.08500821692132\\
9.622182846	6.07000205930772\\
9.642109394	6.055\\
9.66202116	6.035\\
9.682155132	6.02\\
9.702076912	6.00000208333297\\
9.72217083	5.985\\
9.7421	5.97\\
9.762033224	5.95500209907604\\
9.782195091	5.94500210260686\\
9.80210948	5.93\\
9.822205305	5.915\\
9.842089415	5.905\\
9.862030268	5.8900021222407\\
9.882162809	5.88500849617059\\
9.902081966	5.87000851788138\\
9.922156811	5.85500853970342\\
9.942095995	5.84501924718816\\
9.962002039	5.8350192801738\\
9.982184649	5.82000859105895\\
10.002010584	5.80500861325804\\
10.023465395	5.79000215889424\\
10.042046547	5.77\\
10.062082767	5.755\\
10.082168818	5.735\\
10.102005005	5.72000218531427\\
10.122171879	5.70000219298203\\
10.141993761	5.68000880281008\\
10.162005424	5.66500882611846\\
10.182159185	5.65001991146934\\
10.20200634	5.63001998220255\\
10.222197533	5.61502003558313\\
10.242011786	5.59502010720248\\
10.262054443	5.57503587432404\\
10.282153845	5.56003597110666\\
10.302008867	5.54003610096541\\
10.322174549	5.52003623176515\\
10.342010975	5.50002045450742\\
10.362035513	5.48502051044479\\
10.382130623	5.46500914912317\\
10.402011156	5.44500229568363\\
10.422178268	5.425\\
10.442026138	5.41\\
10.462051392	5.39000231910896\\
10.482138157	5.37000232774624\\
10.502046824	5.35\\
10.52218771	5.32500234741732\\
10.542002678	5.31000235404844\\
10.562021255	5.2900094517874\\
10.582208157	5.27000948765749\\
10.602012396	5.2500214285277\\
10.622630835	5.23502148992724\\
10.642049551	5.21502157234273\\
10.662068844	5.19502165539279\\
10.682162046	5.17502173908477\\
10.70199728	5.15502182342616\\
10.722176313	5.13500973708911\\
10.74203825	5.1200219726091\\
10.762045622	5.10002205877583\\
10.782156229	5.08502212384568\\
10.802023888	5.06002223315274\\
10.822166443	5.04003968238346\\
10.842008352	5.02003984047936\\
10.86204505	5.00003999984\\
10.882124424	4.98006275060867\\
10.902003527	4.96006300363211\\
10.922164202	4.94006325870429\\
10.942017317	4.92006351585018\\
10.962011099	4.9050407745502\\
10.982141495	4.88504094148657\\
11.001995325	4.87002310056123\\
11.022160769	4.85001030926739\\
11.041996956	4.83001035195578\\
11.062013865	4.8100025987519\\
11.082167149	4.79001043840199\\
11.102060795	4.77001048216878\\
11.122785091	4.75502365924713\\
11.141993284	4.73504223845997\\
11.162014008	4.71506627737087\\
11.182183027	4.6950665596986\\
11.20200181	4.67506684444191\\
11.222160101	4.65506713163194\\
11.242029905	4.63504314974521\\
11.262011051	4.61502437696704\\
11.282140493	4.59501088137993\\
11.302013159	4.5750109289487\\
11.322203159	4.55001098899772\\
11.341996431	4.53001103751415\\
11.361999273	4.51001108646087\\
11.382134199	4.4900027839635\\
11.402023792	4.47000279641971\\
11.422180176	4.45\\
11.442027569	4.43\\
11.462018728	4.40500283768354\\
11.482165337	4.38500285062621\\
11.502015114	4.36500286368749\\
11.522163391	4.345\\
11.541991234	4.33000288683507\\
11.56204319	4.31000290023104\\
11.582142353	4.29001165499582\\
11.601999283	4.27501169589043\\
11.622165918	4.26001173707303\\
11.642030716	4.24001179243643\\
11.662001848	4.22501183430295\\
11.682154179	4.20501189058961\\
11.702039242	4.18501194741425\\
11.722190619	4.16501200478462\\
11.742001534	4.1500271083452\\
11.76201725	4.13504836731083\\
11.782306194	4.12004854340334\\
11.80202961	4.10004878019762\\
11.822197199	4.08504895931493\\
11.842004776	4.06504920019426\\
11.862013817	4.05002777768252\\
11.882138729	4.03001240692879\\
11.902037144	4.01500311332382\\
11.922207832	3.995\\
11.942010164	3.975\\
11.962032795	3.96\\
11.982157469	3.935\\
12.002034426	3.91500319284672\\
12.022168159	3.89000321336628\\
12.041992426	3.87001291987507\\
12.062012911	3.84501300387918\\
12.082139969	3.82501307187309\\
12.102028608	3.80501314058178\\
12.122175932	3.78501321001658\\
12.142016888	3.76001329784883\\
12.162007093	3.73501338685687\\
12.182150602	3.71501345892582\\
12.201984882	3.69003048767893\\
12.222195625	3.67003065382294\\
12.242027998	3.64503086406686\\
12.262045145	3.62505517199394\\
12.282132149	3.60503120652235\\
12.301986933	3.5850313806158\\
12.322165728	3.56503155666258\\
12.342005491	3.5450141043443\\
12.362001896	3.52500354609751\\
12.382138014	3.50500356633199\\
12.402010679	3.485\\
12.422173023	3.47\\
12.441994429	3.4500036231865\\
12.46200633	3.43500363900826\\
12.482149124	3.41501464125705\\
12.502034187	3.40001470585055\\
12.522147179	3.38001479286704\\
12.542021513	3.3650037147082\\
12.562004566	3.34500373691869\\
12.58214283	3.33000375375164\\
12.602001905	3.315\\
12.622149944	3.3\\
12.641993761	3.28500380517283\\
12.662008047	3.27000382262774\\
12.682161331	3.26001533738723\\
12.701981068	3.24503466853592\\
12.722173214	3.2350347757018\\
12.742001534	3.22003493769866\\
12.762007236	3.2100623046913\\
12.782156467	3.19509780757961\\
12.801985979	3.18009826892189\\
12.822182894	3.17009857890886\\
12.842017889	3.15509904757363\\
12.862043858	3.14009952071586\\
12.88212204	3.12509999840005\\
12.902052879	3.10510064249132\\
12.922176838	3.09006472424123\\
12.941979647	3.07003664473244\\
12.962032318	3.05001639339857\\
12.982141495	3.03000412540973\\
13.002001524	3.01500414593413\\
13.022152662	2.9950041736198\\
13.042025328	2.97500420167771\\
13.062027693	2.95500423011542\\
13.082137346	2.94001700675353\\
13.102001429	2.92501709396714\\
13.122209072	2.90501721165297\\
13.141997099	2.88503899453716\\
13.162031651	2.87001742154991\\
13.182130337	2.85001754380565\\
13.202013254	2.83000441695768\\
13.222183228	2.81000444839506\\
13.242014885	2.79\\
13.262023926	2.77\\
13.282182693	2.75\\
13.302033186	2.73\\
13.322187185	2.715\\
13.342011929	2.69500463821493\\
13.362032652	2.67001872652609\\
13.382173061	2.65501883232492\\
13.402025938	2.63504269415127\\
13.422155619	2.61504302067863\\
13.442006588	2.59504335223903\\
13.462030411	2.57501941740252\\
13.482156277	2.55501956939668\\
13.502064466	2.53501972378915\\
13.522164583	2.52000496031258\\
13.542022943	2.500004999995\\
13.561992407	2.485005030176\\
13.582232714	2.4650202838922\\
13.60199976	2.45002040807827\\
13.622158289	2.43502053379433\\
13.642034531	2.415020703845\\
13.6620543	2.40002083324291\\
13.682138205	2.38502096426845\\
13.702019453	2.36500528540636\\
13.722175598	2.34500533048435\\
13.742042065	2.32500537633787\\
13.762019157	2.305\\
13.782135963	2.285\\
13.801988363	2.265\\
13.822141886	2.24500556792183\\
13.841995239	2.22500561797044\\
13.862030745	2.20500566892695\\
13.882129192	2.18500572081631\\
13.902008533	2.16000578702928\\
13.922157764	2.14000584111352\\
13.942016602	2.11500591015723\\
13.96199894	2.09502386621251\\
13.982160807	2.07000603863853\\
14.002003908	2.05000609755191\\
14.022171021	2.0250061728301\\
14.042001486	2.00500623440427\\
14.062019587	1.98500629721923\\
14.082140923	1.96000637754064\\
14.102076769	1.94000644328827\\
14.122169018	1.91500652740402\\
14.141991854	1.89500659629459\\
14.162217855	1.87500666665481\\
14.182018042	1.85500673853224\\
14.202015162	1.83500681197646\\
14.222146988	1.82000686811891\\
14.242033482	1.80000694443105\\
14.262063742	1.78000702245806\\
14.282141447	1.765\\
14.301989079	1.74500716330908\\
14.322176218	1.72500724636159\\
14.341989994	1.7100073099259\\
14.36200428	1.69500737461523\\
14.3821702	1.68002976164114\\
14.402010918	1.66503002975922\\
14.422150373	1.65003030275204\\
14.442054033	1.6400304875215\\
14.462033749	1.62503076893947\\
14.482139349	1.6100310556011\\
14.502010345	1.60003124969483\\
14.522170782	1.58507097632882\\
14.542027712	1.57507142695181\\
14.562043428	1.56007211371782\\
14.58217597	1.5500725789459\\
14.602017879	1.53507328815272\\
14.622169018	1.5250737687076\\
14.642028809	1.51507425560598\\
14.662008762	1.50007499812509\\
14.682173729	1.49003355666911\\
14.702037334	1.47503389791557\\
14.722192049	1.46503412929529\\
14.741998196	1.45003448234861\\
14.76201582	1.44003472180361\\
14.782176256	1.42500877190283\\
14.802025318	1.41500883389469\\
14.822154284	1.40000892854296\\
14.8419981	1.39000899277667\\
14.862028122	1.38\\
14.882154465	1.365\\
14.902001619	1.355\\
14.922166586	1.34500929364819\\
14.942019224	1.33000939846303\\
14.962034941	1.320009469663\\
14.982142687	1.31000954194998\\
15.002000093	1.30000961534906\\
15.022146463	1.29000968988609\\
15.042012453	1.28000976558775\\
15.06201601	1.27000984248155\\
15.082141876	1.26000992059587\\
15.10201335	1.25000999996\\
15.122178078	1.24001008060419\\
15.141991377	1.23001016255964\\
15.162016153	1.22001024585862\\
15.182161808	1.21001033053441\\
15.201999187	1.19501046020527\\
15.222200155	1.18501054847626\\
15.242055893	1.17501063824971\\
15.262008667	1.16501072956432\\
15.282166243	1.15501082246012\\
15.302016258	1.14501091697852\\
15.322191	1.13504405200856\\
15.341993332	1.12501111105624\\
15.362011433	1.11501121070597\\
15.382182598	1.10501131215929\\
15.402018785	1.09501141546561\\
15.422198772	1.08501152067616\\
15.442034245	1.07501162784409\\
15.462046862	1.06501173702453\\
15.482173443	1.0550118482747\\
15.502015114	1.04501196165403\\
15.522414684	1.03501207722422\\
15.542039871	1.03001213585084\\
15.562013388	1.02001225482834\\
15.58214283	1.0100123761618\\
15.602000952	1.00001249992188\\
15.622148275	0.990012626182111\\
15.642017603	0.985\\
15.662034512	0.97\\
15.682162046	0.955\\
15.702001095	0.935013368888381\\
15.722177267	0.915013661100205\\
15.741961718	0.89505586417832\\
15.762009382	0.870057469366248\\
15.782168627	0.850058821494136\\
15.801988125	0.830060238777885\\
15.822143555	0.810061726043146\\
15.84201169	0.785063691683675\\
15.862009764	0.760016447190454\\
15.882174969	0.740016891699102\\
15.901987791	0.715\\
15.92216754	0.69\\
15.941969395	0.670018656456669\\
15.961994886	0.645019379553824\\
15.982156992	0.62501999968001\\
16.00205946	0.600020832971656\\
16.022174835	0.575\\
16.041994572	0.555\\
16.062048912	0.530023584380922\\
16.082163095	0.510024509214998\\
16.102010965	0.485103081829007\\
16.122182608	0.465107514452304\\
16.141976595	0.445112345369121\\
16.162011385	0.425117630779999\\
16.182132483	0.400124980474851\\
16.201993942	0.380131556174964\\
16.222128391	0.360138862107382\\
16.241983175	0.340036762718386\\
16.261989355	0.320039060116105\\
16.282131672	0.305040980853393\\
16.301984072	0.29\\
16.322142363	0.275\\
16.341975689	0.26\\
16.362000465	0.25\\
16.38217926	0.235053185470863\\
16.401986361	0.225055548698538\\
16.422143936	0.220056810846654\\
16.442050934	0.210059515376\\
16.462029934	0.200062490237426\\
16.482128382	0.195064092031312\\
16.501989365	0.190065778087482\\
16.522183657	0.185067555233217\\
16.541991472	0.180069431053691\\
16.561990738	0.175071414000116\\
16.582096577	0.170073513516949\\
16.602063894	0.170073513516949\\
16.622097969	0.165075740192192\\
16.642044306	0.155080624192708\\
16.662005901	0.155080624192708\\
16.682119608	0.155080624192708\\
16.70204854	0.150083310198036\\
16.72213316	0.150083310198036\\
16.742088079	0.145086181285469\\
16.761964321	0.140356688476182\\
16.782146931	0.145344418537486\\
16.802073479	0.140089257261219\\
16.822122812	0.140356688476182\\
16.842083931	0.140356688476182\\
16.861982107	0.140089257261219\\
16.882124901	0.140089257261219\\
16.902066469	0.140089257261219\\
16.92209816	0.145086181285469\\
16.942107677	0.140089257261219\\
16.962004423	0.140089257261219\\
16.982093573	0.145086181285469\\
17.002074718	0.145086181285469\\
17.022132635	0.145086181285469\\
17.042076826	0.145086181285469\\
17.061970949	0.145086181285469\\
17.082075357	0.150083310198036\\
17.102052212	0.155322245670091\\
17.122137308	0.155322245670091\\
17.142031193	0.155322245670091\\
17.16200757	0.160312195418814\\
17.182082891	0.160312195418814\\
17.20207262	0.165302752548165\\
17.222100496	0.170293863659264\\
17.242043972	0.170293863659264\\
17.261957645	0.170293863659264\\
17.282083273	0.175071414000116\\
17.302093744	0.180277563773199\\
17.322118044	0.185270073136489\\
17.34204793	0.180277563773199\\
17.361960173	0.185270073136489\\
17.382108212	0.190262975904404\\
17.402024508	0.195256241897666\\
17.422100306	0.200249843945008\\
17.442066431	0.195256241897666\\
17.462002516	0.200249843945008\\
17.48214221	0.205243757517738\\
17.501986027	0.210237960416286\\
17.522157192	0.215232432500309\\
17.54199338	0.220227155455452\\
17.561994553	0.220227155455452\\
17.582138062	0.220227155455452\\
17.602009773	0.225222112591104\\
17.622126341	0.225222112591104\\
17.642013788	0.225222112591104\\
17.662002325	0.225222112591104\\
17.682146549	0.225222112591104\\
17.701998472	0.225222112591104\\
17.722178459	0.225222112591104\\
17.742015362	0.225222112591104\\
17.762011766	0.220227155455452\\
17.782153606	0.220227155455452\\
17.801993847	0.220227155455452\\
17.822180986	0.215232432500309\\
17.841971636	0.215232432500309\\
17.861990213	0.210237960416286\\
17.882103443	0.210237960416286\\
17.901988268	0.205243757517738\\
17.922172308	0.200249843945008\\
17.941988468	0.195256241897666\\
17.962006807	0.195256241897666\\
17.982125759	0.190262975904404\\
18.00197053	0.190262975904404\\
18.022149801	0.185270073136489\\
18.042065859	0.180277563773199\\
18.062117338	0.170293863659264\\
18.082151651	0.170293863659264\\
18.10199523	0.165302752548165\\
18.122134924	0.160312195418814\\
18.142056704	0.155322245670091\\
18.162050724	0.150332963783729\\
18.18215251	0.145344418537486\\
18.201961994	0.140356688476182\\
18.222989321	0.140356688476182\\
18.242043018	0.135369863706809\\
18.262099743	0.130384048104053\\
18.282104254	0.130384048104053\\
18.301981449	0.125399362039845\\
18.322139263	0.125399362039845\\
18.34203124	0.120415945787923\\
18.362072468	0.115433963806152\\
18.382111073	0.110453610171873\\
18.401993513	0.110453610171873\\
18.422118425	0.106066017177982\\
18.442046881	0.101118742080783\\
18.462067366	0.0912414379544733\\
18.482110023	0.0863133825081604\\
18.501984835	0.0813941029804985\\
18.522121668	0.0764852927038918\\
18.542054415	0.0715891053163818\\
18.562091351	0.0618465843842649\\
18.582144976	0.0618465843842649\\
18.602002382	0.0570087712549569\\
18.622121572	0.0522015325445528\\
18.642043829	0.0522015325445528\\
18.662081718	0.0474341649025257\\
18.68214798	0.0427200187265877\\
18.701971769	0.0380788655293195\\
18.722103596	0.0380788655293195\\
18.742054224	0.0335410196624968\\
18.762120247	0.0335410196624968\\
18.782105923	0.0291547594742265\\
18.801943541	0.0291547594742265\\
18.822140217	0.025\\
18.842042446	0.025\\
18.862076283	0.025\\
18.882124662	0.0212132034355964\\
18.901986361	0.0212132034355964\\
18.922109604	0.0212132034355964\\
18.942097187	0.0212132034355964\\
18.962079048	0.0180277563773199\\
18.98213625	0.0180277563773199\\
19.001996517	0.0180277563773199\\
19.022141933	0.0180277563773199\\
19.042103291	0.0180277563773199\\
19.062106609	0.0180277563773199\\
19.082149506	0.0212132034355964\\
19.10197401	0.0212132034355964\\
19.122107983	0.0212132034355964\\
19.142064571	0.0212132034355964\\
19.16207552	0.0212132034355964\\
19.182133913	0.025\\
19.20197773	0.025\\
19.22210598	0.0291547594742265\\
19.242041588	0.0291547594742265\\
19.262112141	0.0269258240356725\\
19.282094955	0.0291547594742265\\
19.301951408	0.0291547594742265\\
19.322146893	0.0335410196624968\\
19.342025757	0.0335410196624968\\
19.36207962	0.0335410196624968\\
19.382131338	0.0316227766016838\\
19.401996851	0.0335410196624968\\
19.42209363	0.0291547594742265\\
19.442054033	0.0291547594742265\\
19.462082148	0.0269258240356725\\
19.482102394	0.0291547594742265\\
19.501953125	0.0335410196624968\\
19.522129059	0.0364005494464026\\
19.54203248	0.0412310562561766\\
19.562086105	0.0509901951359279\\
19.582078218	0.0559016994374947\\
19.601973534	0.0608276253029822\\
19.62209034	0.0657647321898295\\
19.642031431	0.0707106781186548\\
19.662068129	0.0707106781186548\\
19.682100773	0.0756637297521078\\
19.701996803	0.0756637297521078\\
19.722111464	0.0806225774829855\\
19.742014885	0.0855862138431185\\
19.76205492	0.0905538513813742\\
19.78210783	0.095524865872714\\
19.801952839	0.100498756211209\\
19.822114706	0.105475115548645\\
19.842041016	0.105475115548645\\
19.862072468	0.100498756211209\\
19.882095337	0.100498756211209\\
19.901971579	0.100498756211209\\
19.922175407	0.100498756211209\\
19.942037582	0.100498756211209\\
19.962098598	0.100498756211209\\
19.982095718	0.100498756211209\\
20.001986742	0.100498756211209\\
20.022093058	0.095524865872714\\
20.042049885	0.095524865872714\\
20.062073946	0.095524865872714\\
20.082108974	0.095524865872714\\
20.102021933	0.095524865872714\\
20.12212491	0.0905538513813742\\
20.142061234	0.0905538513813742\\
20.162065506	0.0855862138431185\\
20.182135582	0.0855862138431185\\
20.201997042	0.0855862138431185\\
20.222113132	0.0855862138431185\\
20.242067337	0.0806225774829855\\
20.262081623	0.0806225774829855\\
20.282110453	0.0806225774829855\\
20.301987171	0.0756637297521078\\
20.32218504	0.0764852927038918\\
20.342083931	0.0715891053163818\\
20.362050056	0.0657647321898295\\
20.382139683	0.0657647321898295\\
20.401967287	0.0657647321898295\\
20.422115803	0.0657647321898295\\
20.442018032	0.0608276253029822\\
20.462073326	0.0608276253029822\\
20.482110739	0.0608276253029822\\
20.501979351	0.0559016994374947\\
20.522124767	0.0522015325445528\\
20.542053938	0.0412310562561766\\
20.562147379	0.0364005494464026\\
20.582041979	0.0269258240356725\\
20.601985216	0.0180277563773199\\
20.622135162	0.0111803398874989\\
20.642071724	0.0111803398874989\\
20.662116528	0.014142135623731\\
20.682121754	0.0223606797749979\\
20.702011585	0.0316227766016838\\
20.72214818	0.0364005494464026\\
20.742044687	0.0474341649025257\\
20.762064934	0.0522015325445528\\
20.782114506	0.0618465843842649\\
20.801952362	0.0667083203206317\\
20.822112322	0.0667083203206317\\
20.842067003	0.0657647321898295\\
20.862078905	0.0608276253029822\\
20.882071257	0.0559016994374947\\
20.901990891	0.0509901951359279\\
20.922109604	0.0460977222864644\\
20.942028046	0.0412310562561766\\
20.962088346	0.0364005494464026\\
20.982127428	0.0316227766016838\\
21.001998425	0.0316227766016838\\
21.022111893	0.0316227766016838\\
21.042070627	0.0316227766016838\\
21.0620718	0.0316227766016838\\
21.082098961	0.0316227766016838\\
21.101972342	0.0364005494464026\\
21.122126341	0.0412310562561766\\
21.142024517	0.0412310562561766\\
21.162059784	0.0412310562561766\\
21.182094336	0.0460977222864644\\
21.201954126	0.0460977222864644\\
21.222126961	0.0460977222864644\\
21.242029667	0.0412310562561766\\
21.262105465	0.0412310562561766\\
21.282117128	0.0412310562561766\\
21.301984787	0.0364005494464026\\
21.32212925	0.0364005494464026\\
21.34205842	0.0364005494464026\\
21.362068653	0.0364005494464026\\
21.382144451	0.0316227766016838\\
21.402006865	0.0316227766016838\\
21.422141314	0.0316227766016838\\
21.442042828	0.0316227766016838\\
21.462052345	0.0316227766016838\\
21.48212266	0.0316227766016838\\
21.501994848	0.0316227766016838\\
21.522117138	0.0316227766016838\\
21.542080402	0.0364005494464026\\
21.562087774	0.0364005494464026\\
21.582148314	0.0364005494464026\\
21.601935863	0.0364005494464026\\
21.622135878	0.0364005494464026\\
21.642077923	0.0364005494464026\\
21.662090302	0.0316227766016838\\
21.682119846	0.0316227766016838\\
21.702045679	0.0316227766016838\\
21.722146749	0.0316227766016838\\
21.742062569	0.0316227766016838\\
21.762056112	0.0269258240356725\\
21.782142401	0.0269258240356725\\
21.801938295	0.0269258240356725\\
21.822087765	0.0316227766016838\\
21.842089176	0.0316227766016838\\
21.862075567	0.0316227766016838\\
21.882115126	0.0316227766016838\\
21.901961088	0.0316227766016838\\
21.922130823	0.0316227766016838\\
21.942079544	0.0316227766016838\\
21.962067366	0.0316227766016838\\
21.982094288	0.0316227766016838\\
22.001972437	0.0316227766016838\\
22.022099018	0.0316227766016838\\
22.042035818	0.0316227766016838\\
22.062061071	0.0269258240356725\\
22.082083464	0.0269258240356725\\
22.101983786	0.0269258240356725\\
22.122139454	0.0269258240356725\\
22.142041922	0.0269258240356725\\
22.162076473	0.0269258240356725\\
22.182096958	0.0269258240356725\\
22.201950312	0.0269258240356725\\
22.222105026	0.0269258240356725\\
22.242027521	0.0269258240356725\\
22.262073755	0.0269258240356725\\
22.282116652	0.0206155281280883\\
22.301965952	0.0206155281280883\\
22.322112083	0.0223606797749979\\
22.342087507	0.0223606797749979\\
22.362069368	0.0223606797749979\\
22.382096529	0.0206155281280883\\
22.40201211	0.0206155281280883\\
22.422106504	0.0206155281280883\\
22.442033052	0.0206155281280883\\
22.462079525	0.0206155281280883\\
22.482102156	0.0158113883008419\\
22.502007484	0.0158113883008419\\
22.522216558	0.0158113883008419\\
22.542050838	0.0158113883008419\\
22.56208086	0.0158113883008419\\
22.582110882	0.0158113883008419\\
22.601999998	0.0158113883008419\\
22.62213397	0.0158113883008419\\
22.642079115	0.0158113883008419\\
22.662142754	0.0158113883008419\\
22.682119846	0.0158113883008419\\
22.701997519	0.0158113883008419\\
22.722173452	0.0158113883008419\\
22.742031813	0.0158113883008419\\
22.762136936	0.0158113883008419\\
22.782159567	0.0111803398874989\\
22.80195117	0.0158113883008419\\
22.822102785	0.0158113883008419\\
22.842041254	0.0158113883008419\\
22.862128019	0.0111803398874989\\
22.882138968	0.0111803398874989\\
22.901978731	0.0111803398874989\\
22.922153234	0.0111803398874989\\
22.942016125	0.0111803398874989\\
22.962067842	0.0111803398874989\\
22.982091665	0.0111803398874989\\
23.001998425	0.0111803398874989\\
23.022127628	0.0111803398874989\\
23.042078495	0.0111803398874989\\
23.062091589	0.0111803398874989\\
23.082093477	0.0111803398874989\\
23.1019876	0.0111803398874989\\
23.122159719	0.0111803398874989\\
23.142050505	0.0111803398874989\\
23.162089109	0.0111803398874989\\
23.182124615	0.0111803398874989\\
23.201952696	0.0111803398874989\\
23.222137928	0.0111803398874989\\
23.242037773	0.0111803398874989\\
23.262085438	0.0111803398874989\\
23.282083511	0.0111803398874989\\
23.301974058	0.00707106781186548\\
23.322146416	0.0111803398874989\\
23.34202981	0.0111803398874989\\
23.362091303	0.00707106781186548\\
23.382113695	0.00707106781186548\\
23.401981354	0.00707106781186548\\
23.42235136	0.00707106781186548\\
23.442028999	0.00707106781186548\\
23.4620924	0.00707106781186548\\
23.482110739	0.00707106781186548\\
23.501963854	0.00707106781186548\\
23.522117138	0.00707106781186548\\
23.542038918	0.00707106781186548\\
23.56214571	0.00707106781186548\\
23.582130194	0.00707106781186548\\
23.601999044	0.00707106781186548\\
23.622128725	0.00707106781186548\\
23.642071247	0.00707106781186548\\
23.662063122	0.00707106781186548\\
23.682090282	0.00707106781186548\\
23.701986551	0.00707106781186548\\
23.722147942	0.005\\
23.742023468	0.005\\
23.762069225	0.005\\
23.782132864	0.005\\
23.801974297	0.005\\
23.822155952	0.005\\
23.842038155	0.005\\
23.86207509	0.005\\
23.882091522	0.005\\
23.901991844	0.005\\
23.92219615	0.005\\
23.942068815	0.005\\
23.962062836	0.005\\
23.982153893	0.005\\
24.001996994	0.005\\
24.02212429	0.005\\
24.042058945	0.005\\
24.062047243	0.005\\
24.082148075	0.005\\
24.101991177	0.005\\
24.122166872	0.005\\
24.142047405	0.005\\
24.162062168	0.005\\
24.182142496	0.005\\
24.201944351	0.005\\
24.222126722	0\\
24.242063284	0\\
24.262085915	0\\
24.282108784	0\\
24.302018881	0\\
24.32211256	0\\
24.342081308	0\\
24.362098455	0\\
24.382167339	0\\
24.401962757	0.005\\
24.422138929	0\\
24.442104578	0.005\\
24.462070942	0\\
24.482081175	0.005\\
24.501969337	0\\
24.522120237	0\\
24.542026281	0\\
24.562080383	0\\
24.582099199	0\\
24.60195756	0\\
24.622094393	0.005\\
24.642066956	0.005\\
24.662106752	0.005\\
24.682098866	0.005\\
24.701967239	0.01\\
24.722149849	0.015\\
24.742028475	0.0254950975679639\\
24.762083054	0.0353553390593274\\
24.782258987	0.04\\
24.801982403	0.05\\
24.822084904	0.06\\
24.842042446	0.075\\
24.862068176	0.09\\
24.882104397	0.105\\
24.901977301	0.125\\
24.922142029	0.145\\
24.941977024	0.17\\
24.961987257	0.19\\
24.982146025	0.215\\
25.001965761	0.235\\
25.022131443	0.26\\
25.041999102	0.28\\
25.06201458	0.3\\
25.082130671	0.325\\
25.102022648	0.345\\
25.122161627	0.365\\
25.141970634	0.385\\
25.1620152	0.405\\
25.182200909	0.425\\
25.20199275	0.44\\
25.222173214	0.46\\
25.242016792	0.475\\
25.262026548	0.495\\
25.282164812	0.515\\
25.302000761	0.535\\
25.322451115	0.560022320983727\\
25.341986418	0.58\\
25.362017155	0.61\\
25.382159948	0.635\\
25.402005196	0.665\\
25.422199011	0.69\\
25.442036152	0.72\\
25.462018251	0.75\\
25.482153416	0.78\\
25.501992464	0.810015431951762\\
25.522176027	0.840014880820572\\
25.541989326	0.87501428559767\\
25.562006712	0.905013812049297\\
25.582149982	0.945013227420654\\
25.60199666	0.980012755019035\\
25.622157097	1.02001225482834\\
25.642018318	1.06001179238724\\
25.662026882	1.10501131215929\\
25.682142019	1.145\\
25.701983929	1.18501054847626\\
25.722192764	1.23\\
25.742002487	1.27000984248155\\
25.762026787	1.31000954194998\\
25.782164574	1.35500922506085\\
25.802037954	1.40000892854296\\
25.822148085	1.44500865049314\\
25.842014313	1.49000838923813\\
25.862020493	1.53500814330088\\
25.882134438	1.58500788641571\\
25.902004719	1.63000766869362\\
25.92215848	1.68002976164114\\
25.942056656	1.73002890149269\\
25.962001324	1.7850280109847\\
25.982146978	1.8350272477541\\
26.001991034	1.88505968075284\\
26.022203445	1.93505813866147\\
26.042020321	1.99005653186034\\
26.062003136	2.04505501148502\\
26.082131386	2.10009523593574\\
26.102059603	2.15509280542625\\
26.122186184	2.21509029161341\\
26.14200139	2.2750879103894\\
26.162011147	2.33508565153401\\
26.182165861	2.3950835058511\\
26.202027321	2.45508146504347\\
26.222157955	2.51507952160563\\
26.241994619	2.57507766873157\\
26.26205945	2.64004261329244\\
26.282147408	2.69504174364703\\
26.302007675	2.75504083454311\\
26.32218647	2.81501776193331\\
26.342027903	2.87501739125175\\
26.362017393	2.93501703572569\\
26.382159233	2.99001672236126\\
26.40203166	3.05001639339857\\
26.422172785	3.10501610301782\\
26.442005873	3.1600158227452\\
26.462018013	3.2100155762862\\
26.482143641	3.26501531389977\\
26.502004623	3.31501508292195\\
26.52218914	3.36003348197604\\
26.542043686	3.40503303948728\\
26.562024593	3.45005797052745\\
26.58214736	3.49505722413811\\
26.602050304	3.53508840059199\\
26.622169733	3.57512587190997\\
26.642041922	3.61516942894797\\
26.662023783	3.65027738672008\\
26.682135582	3.68533919741453\\
26.702027559	3.71540711093683\\
26.722219467	3.74548061001522\\
26.742015123	3.77555956117765\\
26.762050629	3.80055588039434\\
26.78219223	3.83055152686921\\
26.802005053	3.85546689779591\\
26.822182894	3.88046388979462\\
26.842005968	3.90038780123208\\
26.862059593	3.92038582284958\\
26.882169008	3.9453833780762\\
26.902005434	3.96538144445147\\
26.922175646	3.99037905467639\\
26.941989899	4.01544829377742\\
26.962054014	4.03544607199749\\
26.982164621	4.0605202868598\\
27.001997948	4.08551710313395\\
27.022144079	4.11059606383308\\
27.042027235	4.13559246057926\\
27.062027931	4.16058890062453\\
27.082180977	4.18558538319313\\
27.101997614	4.21066799926092\\
27.122191906	4.2306648413695\\
27.14197278	4.25566093574194\\
27.162033796	4.28065707573031\\
27.182127476	4.30074412166081\\
27.202015162	4.32074067724505\\
27.222187519	4.34073726456693\\
27.242005348	4.36073388318985\\
27.262019634	4.37573136744019\\
27.282144785	4.39064061385124\\
27.302032948	4.40563843273594\\
27.322174788	4.41555489151703\\
27.341990471	4.42040722106007\\
27.362036943	4.42534179470919\\
27.382180452	4.4302821580572\\
27.402003288	4.43518037964636\\
27.422185183	4.44013794830746\\
27.441998482	4.44010135019461\\
27.462059498	4.44007038232504\\
27.482139826	4.44004504481655\\
27.502019405	4.44002533776554\\
27.52219367	4.44004504481655\\
27.542020559	4.43504509559937\\
27.562032223	4.43007054119909\\
27.582131863	4.42513841591424\\
27.60202384	4.42022906646251\\
27.622142792	4.41040814437848\\
27.641989231	4.40563843273594\\
27.662005186	4.40102544868807\\
27.682174444	4.39637634876724\\
27.702018023	4.39163978486396\\
27.722179651	4.38692660070806\\
27.742043257	4.38707761499612\\
27.762024403	4.38207998557762\\
27.782186508	4.38207998557762\\
27.802052259	4.37693100242624\\
27.822200298	4.37178739190277\\
27.842039585	4.37164728677875\\
27.862027645	4.36638580521694\\
27.88215971	4.36626270396091\\
27.902037382	4.3661453251123\\
27.922200918	4.37092667520287\\
27.942023039	4.37082658086545\\
27.961997271	4.3706435452917\\
27.982176065	4.37555996416459\\
28.002029896	4.38555868732822\\
28.022160769	4.39048118091856\\
28.042056322	4.40048008744501\\
28.062054634	4.40547954256968\\
28.082167864	4.41547845652088\\
28.101998329	4.42547737537997\\
28.122165918	4.43555238950009\\
28.142002344	4.45055052774373\\
28.162032843	4.46054929352877\\
28.182125807	4.47062915035457\\
28.20200634	4.48062774619807\\
28.222186565	4.49062634829486\\
28.24201417	4.50562426307387\\
28.26203227	4.51570869299604\\
28.282155037	4.52570712706865\\
28.302013397	4.53579651219055\\
28.322194576	4.54589100177292\\
28.342026711	4.56088807141767\\
28.362207413	4.57088612853132\\
28.382018805	4.58088419412672\\
28.401983261	4.59088226814846\\
28.422148466	4.60088035054162\\
28.442022324	4.61578270719062\\
28.462060928	4.62569184014673\\
28.482171297	4.635690347726\\
28.502037048	4.65068812112788\\
28.522174835	4.66060350169375\\
28.542027235	4.67560156557421\\
28.562031269	4.68560028171418\\
28.58214879	4.69568152667959\\
28.602038383	4.71076692269953\\
28.622171879	4.72585706512586\\
28.642048836	4.73595291361728\\
28.662041903	4.75105251496971\\
28.68214035	4.76604920243172\\
28.702005386	4.78094394445281\\
28.722162008	4.79594099213074\\
28.741998196	4.81084192215874\\
28.762242794	4.82574864658324\\
28.782024622	4.84066111187305\\
28.802007675	4.85557926513408\\
28.822199821	4.87057748116176\\
28.842037678	4.88557570814331\\
28.862053871	4.90065301771101\\
28.882201672	4.92065036351903\\
28.902004957	4.93073270417288\\
28.922200203	4.94573048194096\\
28.942040682	4.95081811421102\\
28.962020874	4.95581728880313\\
28.982197523	4.96581564297347\\
29.002066135	4.97581400375858\\
29.022214174	4.99572317087326\\
29.042006969	5.00572172618495\\
29.062025785	5.02063740973196\\
29.082149029	5.03063614267619\\
29.102013111	5.04563425150892\\
29.122197628	5.060632371552\\
29.142024517	5.07563050270604\\
29.162008762	5.09062864487285\\
29.182209492	5.10562679795537\\
29.202017546	5.12070551779733\\
29.222159147	5.13570345717118\\
29.242037773	5.15570072832006\\
29.262034893	5.17561832054876\\
29.282139063	5.19061653370773\\
29.302011013	5.21053979929143\\
29.322177172	5.23053773526203\\
29.341993809	5.24546709073653\\
29.362015963	5.26540121928045\\
29.382160664	5.29039932330254\\
29.402008772	5.31033897223143\\
29.42217803	5.33533738389617\\
29.442031622	5.35528243886352\\
29.462039232	5.38028112648401\\
29.482152462	5.40523126239757\\
29.502002716	5.44022977455916\\
29.522171021	5.47022851442241\\
29.542028189	5.50522706162062\\
29.561997652	5.53527325432087\\
29.582170963	5.57032315041058\\
29.602061987	5.60037721943799\\
29.622183561	5.64043438043561\\
29.642023087	5.67543170164173\\
29.662038803	5.71049253567501\\
29.682175398	5.74055746421896\\
29.702017307	5.77555408597305\\
29.7221632	5.81055074842308\\
29.742003441	5.84548116069156\\
29.762031317	5.88041665190486\\
29.782172918	5.91541418668211\\
29.80201149	5.9503550314246\\
29.822170258	5.9903526607371\\
29.841985941	6.02529874778006\\
29.862051249	6.06524937657142\\
29.882136822	6.09524814917326\\
29.902020931	6.13524653131396\\
29.922157049	6.17020258986688\\
29.942032099	6.21016304133796\\
29.962019682	6.25016199790053\\
29.982144356	6.28512728590281\\
30.001995087	6.32509683720336\\
30.022142649	6.36509622865201\\
30.042032719	6.40009570240946\\
30.062025785	6.44012422240441\\
30.082151651	6.4801234556141\\
30.102025986	6.51515540873739\\
30.122210503	6.55515446042273\\
30.142000198	6.59015364009065\\
30.162026405	6.63012066255208\\
30.18218565	6.66512002892671\\
30.202034473	6.70509134911673\\
30.222190619	6.74009087475829\\
30.242018938	6.77506642033862\\
30.262004852	6.8150660304945\\
30.282184839	6.85004562028604\\
30.302031279	6.89004535543853\\
30.322157621	6.92504512620676\\
30.341989279	6.9650448670486\\
30.362034082	7.00004464271479\\
30.382180929	7.04002840903359\\
30.40200758	7.07502826849476\\
30.422162294	7.11502810957202\\
30.441993237	7.15001573424842\\
30.462149858	7.19000695409956\\
30.481984854	7.22500692041191\\
30.502017498	7.2650068823092\\
30.522214174	7.30000684931186\\
30.54199481	7.34000681198594\\
30.562006474	7.37501525422151\\
30.58219862	7.41001518217068\\
30.602005482	7.44501511079729\\
30.622174025	7.48501503004503\\
30.641998529	7.5200149600915\\
30.662027597	7.55501489078612\\
30.682175398	7.59000658761242\\
30.702009439	7.62500655737423\\
30.722192287	7.66000163185361\\
30.742059469	7.69500162443128\\
30.76218152	7.73000161707616\\
30.782032251	7.76500160978734\\
30.801999331	7.80000160256394\\
30.822172165	7.83500159540507\\
30.842003107	7.87000635323759\\
30.862011671	7.90500632510816\\
30.882138491	7.94000629722672\\
30.902043819	7.97000627352325\\
30.922194719	8.00500624609375\\
30.941994905	8.04000621890307\\
30.962159157	8.07000619578449\\
30.982009649	8.10500154225772\\
31.001997709	8.13500153657023\\
31.02216053	8.17000152998762\\
31.041988611	8.2000015243901\\
31.062040567	8.2300060753319\\
31.082134008	8.26000605326655\\
31.10200572	8.29001357055584\\
31.122170448	8.32001352162363\\
31.141984224	8.35001347304302\\
31.162046432	8.38001342481025\\
31.182151079	8.40000595237884\\
31.202049017	8.4250014836794\\
31.222155809	8.44500148016565\\
31.242011547	8.46\\
31.262024164	8.47500147492612\\
31.282152891	8.48500147318785\\
31.302026033	8.49500147145367\\
31.322161198	8.495\\
31.341996908	8.50000147058811\\
31.362005234	8.50000147058811\\
31.382138968	8.50001323528381\\
31.402026415	8.49501324307385\\
31.422175169	8.49502354322812\\
31.442034721	8.49502354322812\\
31.462235212	8.49001325087305\\
31.48198843	8.48500589274987\\
31.501996994	8.48000589622437\\
31.522185087	8.47000590318567\\
31.541993618	8.46000591016342\\
};
\addplot [color=black, forget plot]
  table[row sep=crcr]{%
9.21000003814697	0\\
9.21000003814697	30\\
};
\node[right, align=left, circle,fill=white,draw=black,thick,line width=0.3mm,inner sep=1pt]
at (axis cs:6.21,22) {1};
\addplot [color=black, forget plot]
  table[row sep=crcr]{%
24.3599998950958	0\\
24.3599998950958	30\\
};
\node[right, align=left, circle,fill=white,draw=black,thick,line width=0.3mm,inner sep=1pt]
at (axis cs:21.36,22) {2};
\addplot [color=black, forget plot]
  table[row sep=crcr]{%
28.8900001049042	0\\
28.8900001049042	30\\
};
\node[right, align=left, circle,fill=white,draw=black,thick,line width=0.3mm,inner sep=1pt]
at (axis cs:25.89,22) {3};
\end{axis}
\end{tikzpicture}%

%% file: Images/Tikz/fig11-3-3.tikz
\begin{tikzpicture}

\begin{axis}[%
width=4.25cm,
height=1.683cm,
at={(0cm,0cm)},
scale only axis,
xmin=0.042082548,
xmax=31.541993618,
xlabel style={font=\color{white!15!black}},
xlabel={$t\;[\SI[per-mode=repeated-symbol]{}{\second}]$},
ymin=-250,
ymax=100,
ylabel style={font=\color{white!15!black}},
ylabel={$\delta\;[\text{Deg}]$},
axis background/.style={fill=white},
ylabel style={yshift=-0.1cm}
]
\addplot [color=blue, line width=1.0pt, forget plot]
  table[row sep=crcr]{%
0.042082548	10.3\\
0.062074423	10.3\\
0.08214736	10.3\\
0.102083445	10.3\\
0.122158766	10.3\\
0.142064571	10.3\\
0.162027836	10.3\\
0.182141781	10.1\\
0.202076912	10\\
0.222186804	9.89999999999998\\
0.24206996	9.89999999999998\\
0.262013435	9.79999999999995\\
0.282138109	9.70000000000005\\
0.302066565	9.60000000000002\\
0.322176218	9.5\\
0.342066765	9.5\\
0.36198473	9.39999999999998\\
0.38214159	9\\
0.402073145	8.5\\
0.422165871	8\\
0.44208169	7.70000000000005\\
0.462033033	7.70000000000005\\
0.482142448	7.89999999999998\\
0.502058506	8.10000000000002\\
0.522171021	8\\
0.542063236	7.70000000000005\\
0.562014103	7.39999999999998\\
0.582153797	6.89999999999998\\
0.602102041	6.39999999999998\\
0.622177362	5.89999999999998\\
0.642128944	5.79999999999995\\
0.662020683	5.70000000000005\\
0.682162285	5.70000000000005\\
0.702095509	5.79999999999995\\
0.722220182	5.79999999999995\\
0.742111921	5.79999999999995\\
0.762013197	5.60000000000002\\
0.782218933	5.39999999999998\\
0.802009821	5.20000000000005\\
0.822173595	5.10000000000002\\
0.842021942	4.79999999999995\\
0.862068653	4.60000000000002\\
0.882125616	4.29999999999995\\
0.902014017	4.10000000000002\\
0.922170401	4\\
0.94209528	3.89999999999998\\
0.96203804	3.79999999999995\\
0.982150078	3.70000000000005\\
1.001995802	6.29999999999995\\
1.022184372	6.29999999999995\\
1.041991234	6.29999999999995\\
1.062036276	6.29999999999995\\
1.082123518	6.29999999999995\\
1.102082253	6.29999999999995\\
1.122177362	6.20000000000005\\
1.141994715	6.20000000000005\\
1.162002563	6.20000000000005\\
1.182144642	6.10000000000002\\
1.202020168	6\\
1.222158432	5.89999999999998\\
1.241981506	5.89999999999998\\
1.262023687	5.89999999999998\\
1.282156467	5.79999999999995\\
1.302074909	5.79999999999995\\
1.322268009	5.60000000000002\\
1.342069626	5.60000000000002\\
1.362036943	5.5\\
1.382166862	5.39999999999998\\
1.401998997	5.29999999999995\\
1.422176838	5.29999999999995\\
1.442102194	5.29999999999995\\
1.462082386	5.39999999999998\\
1.482149124	5.39999999999998\\
1.50210166	5.39999999999998\\
1.522209644	5.29999999999995\\
1.542084455	5.29999999999995\\
1.56201601	5.20000000000005\\
1.582152605	5.10000000000002\\
1.602059126	5\\
1.622139215	4.70000000000005\\
1.6420784	4.60000000000002\\
1.662032366	4.5\\
1.682159185	4.5\\
1.702077627	4.5\\
1.722183228	4.5\\
1.742052317	4.5\\
1.761993408	4.5\\
1.782148838	4.5\\
1.802083254	4.5\\
1.822393179	4.5\\
1.842071772	4.5\\
1.862071514	4.5\\
1.882152319	4.60000000000002\\
1.902098656	4.70000000000005\\
1.922165871	4.79999999999995\\
1.942107439	4.79999999999995\\
1.96204114	4.79999999999995\\
1.982191563	4.79999999999995\\
2.002092361	4.79999999999995\\
2.022161007	4.79999999999995\\
2.042059898	4.79999999999995\\
2.062010765	4.79999999999995\\
2.082136154	4.79999999999995\\
2.102054596	4.79999999999995\\
2.122177124	4.79999999999995\\
2.142044306	4.79999999999995\\
2.162009954	4.79999999999995\\
2.182146549	4.79999999999995\\
2.202073097	4.79999999999995\\
2.222182274	4.79999999999995\\
2.242100477	4.79999999999995\\
2.262007952	4.79999999999995\\
2.282197952	4.79999999999995\\
2.302063227	4.79999999999995\\
2.322173834	4.79999999999995\\
2.342066526	4.79999999999995\\
2.362056971	4.89999999999998\\
2.38217783	4.89999999999998\\
2.402078152	4.89999999999998\\
2.422175407	4.89999999999998\\
2.442069769	4.89999999999998\\
2.462165833	4.89999999999998\\
2.482060909	5\\
2.502035141	5\\
2.522202015	5\\
2.542052746	5\\
2.561997175	5\\
2.582178116	5\\
2.60208106	5.10000000000002\\
2.622206688	5.10000000000002\\
2.642068863	5.20000000000005\\
2.662001371	5.20000000000005\\
2.682148933	5.20000000000005\\
2.702095985	5.29999999999995\\
2.722168922	5.29999999999995\\
2.742069483	5.29999999999995\\
2.762212515	5.29999999999995\\
2.782103539	5.29999999999995\\
2.801998615	5.29999999999995\\
2.822190285	5.39999999999998\\
2.842061758	5.5\\
2.86200285	5.60000000000002\\
2.882155657	5.79999999999995\\
2.902054071	5.79999999999995\\
2.922168732	5.79999999999995\\
2.942016125	5.70000000000005\\
2.962005377	5.60000000000002\\
2.982170582	5.70000000000005\\
3.002002239	5.70000000000005\\
3.022175312	5.70000000000005\\
3.042074442	5.70000000000005\\
3.062019825	5.70000000000005\\
3.082145929	5.70000000000005\\
3.102072716	5.70000000000005\\
3.122160673	5.60000000000002\\
3.142097235	5.5\\
3.162013769	5.39999999999998\\
3.182151079	5.29999999999995\\
3.202117682	5.29999999999995\\
3.222162008	5.5\\
3.24206996	5.60000000000002\\
3.262001038	5.5\\
3.282155991	5.39999999999998\\
3.302068949	5.20000000000005\\
3.322217226	5.20000000000005\\
3.342102528	5.29999999999995\\
3.361999273	5.29999999999995\\
3.382189989	5.20000000000005\\
3.402045965	5.10000000000002\\
3.422164679	5.10000000000002\\
3.442015648	5.20000000000005\\
3.462039471	5.29999999999995\\
3.482130527	5.39999999999998\\
3.502067804	5.60000000000002\\
3.522150517	5.79999999999995\\
3.542077541	5.60000000000002\\
3.56203866	5.70000000000005\\
3.582189322	5.79999999999995\\
3.602053165	5.79999999999995\\
3.622231483	5.60000000000002\\
3.64200592	5.60000000000002\\
3.662053108	5.5\\
3.682157516	5.39999999999998\\
3.702059746	5.20000000000005\\
3.722196341	5\\
3.742063999	4.89999999999998\\
3.762039423	4.79999999999995\\
3.782186747	4.70000000000005\\
3.802117586	4.39999999999998\\
3.822149277	4.20000000000005\\
3.842074156	4.10000000000002\\
3.86200738	4.29999999999995\\
3.882135153	4.5\\
3.902033567	4.60000000000002\\
3.922189236	4.70000000000005\\
3.942101002	4.70000000000005\\
3.962026596	4.70000000000005\\
3.982176304	4.79999999999995\\
4.002102137	4.89999999999998\\
4.022188663	5\\
4.042072058	4.89999999999998\\
4.062014341	4.79999999999995\\
4.082127094	4.70000000000005\\
4.10209775	4.70000000000005\\
4.12220645	4.70000000000005\\
4.14205718	4.70000000000005\\
4.161990643	4.60000000000002\\
4.182168007	4.5\\
4.202083349	4.39999999999998\\
4.222192049	4.39999999999998\\
4.242104053	4.5\\
4.262029648	4.5\\
4.282184124	4.5\\
4.302097559	4.39999999999998\\
4.322163105	4.39999999999998\\
4.342095852	4.29999999999995\\
4.36203599	4.39999999999998\\
4.382214546	4.39999999999998\\
4.402095795	4.39999999999998\\
4.422175646	4.39999999999998\\
4.442076445	4.39999999999998\\
4.461998224	4.39999999999998\\
4.48216033	4.5\\
4.502059221	4.60000000000002\\
4.522181511	4.70000000000005\\
4.542098999	4.70000000000005\\
4.562008858	4.70000000000005\\
4.582146883	4.70000000000005\\
4.602077723	4.70000000000005\\
4.622160435	4.79999999999995\\
4.642110348	4.79999999999995\\
4.662068844	4.79999999999995\\
4.682150602	4.89999999999998\\
4.702089787	5\\
4.722190619	5.10000000000002\\
4.742033005	5.20000000000005\\
4.76201272	5.29999999999995\\
4.78217864	5.39999999999998\\
4.802061319	5.5\\
4.822193384	5.5\\
4.842078686	5.5\\
4.862027645	5.60000000000002\\
4.882078886	5.60000000000002\\
4.902020931	5.60000000000002\\
4.922174454	5.60000000000002\\
4.94198966	5.70000000000005\\
4.962014675	5.70000000000005\\
4.982167959	5.70000000000005\\
5.002066851	5.70000000000005\\
5.022088051	5.70000000000005\\
5.042010069	5.70000000000005\\
5.062077045	5.70000000000005\\
5.082146645	5.70000000000005\\
5.101998568	5.79999999999995\\
5.122180223	6\\
5.142085791	6.10000000000002\\
5.162003517	6.20000000000005\\
5.182151556	6.29999999999995\\
5.202069044	6.39999999999998\\
5.222151041	6.60000000000002\\
5.24200058	6.70000000000005\\
5.262019396	6.79999999999995\\
5.28213954	6.79999999999995\\
5.301988125	6.79999999999995\\
5.32215786	6.79999999999995\\
5.341994524	6.89999999999998\\
5.36200428	7.10000000000002\\
5.382158279	7.20000000000005\\
5.401992321	7.29999999999995\\
5.422123909	7.60000000000002\\
5.442040682	7.79999999999995\\
5.462051153	8\\
5.48214221	8.10000000000002\\
5.502030611	8.20000000000005\\
5.522213697	8.29999999999995\\
5.542008877	8.39999999999998\\
5.562020063	8.60000000000002\\
5.58216095	8.70000000000005\\
5.602070093	8.79999999999995\\
5.62218833	9\\
5.642028093	9\\
5.662093639	9.10000000000002\\
5.682007074	9.20000000000005\\
5.702009678	9.20000000000005\\
5.722153664	9.20000000000005\\
5.741987944	9.20000000000005\\
5.761996508	9.29999999999995\\
5.782096148	9.60000000000002\\
5.80200243	9.70000000000005\\
5.822184324	9.89999999999998\\
5.842010736	10\\
5.862154245	10.1\\
5.882057905	10.2\\
5.902000189	10.3\\
5.922207832	10.4\\
5.942038298	10.5\\
5.962017298	10.6\\
5.982093573	10.6\\
6.002003193	10.7\\
6.02215147	10.9\\
6.042025089	11.1\\
6.062041759	11.2\\
6.082153082	11.4\\
6.102032185	11.5\\
6.122202635	11.7\\
6.142030954	11.8\\
6.162008762	11.8\\
6.182173014	11.9\\
6.202021122	11.9\\
6.222186804	12\\
6.241998672	12.1\\
6.262037039	12.2\\
6.282183409	12.3\\
6.302028894	12.4\\
6.322185278	12.4\\
6.342019558	12.5\\
6.362130165	12.5\\
6.382010698	12.6\\
6.402008057	12.6\\
6.422150373	12.6\\
6.441997766	12.6\\
6.462204456	12.7\\
6.482050896	12.8\\
6.501999855	12.9\\
6.52216506	13.1\\
6.541989326	13.2\\
6.562157154	13.3\\
6.582008123	13.5\\
6.602034807	13.6\\
6.6221416	13.8\\
6.642022371	13.8\\
6.662213326	13.9\\
6.681981802	14\\
6.702016592	13.9\\
6.722084761	14\\
6.742046356	14.1\\
6.762132645	14.2\\
6.782027721	14.2\\
6.802029848	14.3\\
6.822382689	14.4\\
6.842010736	14.4\\
6.862016201	14.4\\
6.882183313	14.5\\
6.901999235	14.5\\
6.922183514	14.6\\
6.942009211	14.6\\
6.962037086	14.7\\
6.982182264	14.6\\
7.001996994	14.6\\
7.022175312	14.5\\
7.042007923	14.6\\
7.062049866	14.6\\
7.082154036	14.6\\
7.102008104	14.6\\
7.122209072	14.6\\
7.142029524	14.6\\
7.162026167	14.7\\
7.182145119	14.8\\
7.201986551	14.9\\
7.22215271	15.1\\
7.242026567	15.2\\
7.262038469	15.5\\
7.282135963	15.8\\
7.30200696	15.9\\
7.322199821	15.8\\
7.341990471	15.7\\
7.362025499	15.6\\
7.382147789	15.5\\
7.402028084	15.5\\
7.422196388	15.5\\
7.442009211	15.4\\
7.462033749	15.4\\
7.482156515	15.4\\
7.502078533	15.4\\
7.522161245	15.5\\
7.54201889	15.5\\
7.562041998	15.5\\
7.582206249	15.4\\
7.602006674	15.4\\
7.622170448	15.2\\
7.641988993	15.2\\
7.662045956	15.2\\
7.682159185	15.2\\
7.702026844	15.3\\
7.722176552	15.5\\
7.742031097	15.8\\
7.762196302	16.2\\
7.782044888	16.5\\
7.80202055	16.6\\
7.822185278	16.5\\
7.841990948	16.3\\
7.8620193	16\\
7.882154465	15.8\\
7.902038813	15.4\\
7.922205448	15.2\\
7.941994429	15\\
7.962000608	14.9\\
7.982166052	14.9\\
8.002037287	14.9\\
8.02215147	15\\
8.042023897	15\\
8.062022448	15\\
8.082145929	15\\
8.102045536	15\\
8.122267246	15\\
8.142007351	14.9\\
8.162024975	14.7\\
8.182183027	14.6\\
8.202043295	14.6\\
8.222168446	14.6\\
8.242039204	14.6\\
8.262290478	14.6\\
8.282201052	14.5\\
8.302013636	14.5\\
8.322190762	14.4\\
8.342035055	14.2\\
8.361993313	14\\
8.382153273	13.7\\
8.402017832	13.5\\
8.422385931	13.3\\
8.442017555	13.3\\
8.462018967	13.3\\
8.482140779	13.3\\
8.502005577	13.3\\
8.522188663	13.3\\
8.542012691	13.3\\
8.562009573	13.3\\
8.582140446	13.3\\
8.60201478	13.1\\
8.622196436	12.8\\
8.641987324	12.3\\
8.662022591	11.7\\
8.682162523	11.3\\
8.70200491	11.1\\
8.722198009	11.2\\
8.741976738	11.4\\
8.761998653	11.6\\
8.782134056	11.5\\
8.802022219	11.5\\
8.822170258	11.4\\
8.841993093	11.4\\
8.86208415	11.3\\
8.882139921	11\\
8.902025938	10.6\\
8.922168255	10.1\\
8.941991568	9.60000000000002\\
8.96199131	9.10000000000002\\
8.982138872	8.70000000000005\\
9.002015829	8.29999999999995\\
9.022140265	8\\
9.04202199	7.89999999999998\\
9.062007427	7.89999999999998\\
9.082135201	8\\
9.102025509	7.89999999999998\\
9.122186422	7.79999999999995\\
9.142015934	7.60000000000002\\
9.161997557	7.10000000000002\\
9.182144642	6.60000000000002\\
9.202023745	6\\
9.222172022	5.5\\
9.242002964	5.10000000000002\\
9.2621696	4.70000000000005\\
9.2820158	4.20000000000005\\
9.302052498	3.79999999999995\\
9.322240829	3.5\\
9.3420403	3.20000000000005\\
9.362011194	2.89999999999998\\
9.382217407	2.60000000000002\\
9.402034044	2.29999999999995\\
9.42215085	2\\
9.442016363	1.60000000000002\\
9.462054968	1.20000000000005\\
9.482158661	0.799999999999955\\
9.502022028	0.600000000000023\\
9.522222281	0.100000000000023\\
9.542011261	-0.299999999999955\\
9.562206268	-0.799999999999955\\
9.582033873	-1.39999999999998\\
9.601994753	-2\\
9.622182846	-2.60000000000002\\
9.642109394	-3.29999999999995\\
9.66202116	-3.89999999999998\\
9.682155132	-4.39999999999998\\
9.702076912	-4.79999999999995\\
9.72217083	-5\\
9.7421	-5.10000000000002\\
9.762033224	-5.10000000000002\\
9.782195091	-5.29999999999995\\
9.80210948	-5.60000000000002\\
9.822205305	-6.10000000000002\\
9.842089415	-6.89999999999998\\
9.862030268	-7.79999999999995\\
9.882162809	-8.79999999999995\\
9.902081966	-9.5\\
9.922156811	-10.1\\
9.942095995	-10.6\\
9.962002039	-10.9\\
9.982184649	-11.4\\
10.002010584	-11.9\\
10.023465395	-12.5\\
10.042046547	-13.2\\
10.062082767	-13.8\\
10.082168818	-14.3\\
10.102005005	-14.8\\
10.122171879	-15.4\\
10.141993761	-16\\
10.162005424	-16.6\\
10.182159185	-17.2\\
10.20200634	-17.9\\
10.222197533	-18.7\\
10.242011786	-19.6\\
10.262054443	-20.4\\
10.282153845	-21\\
10.302008867	-21.5\\
10.322174549	-22.1\\
10.342010975	-22.8\\
10.362035513	-23.6\\
10.382130623	-24.3\\
10.402011156	-25.1\\
10.422178268	-25.7\\
10.442026138	-26.3\\
10.462051392	-26.9\\
10.482138157	-27.5\\
10.502046824	-28.1\\
10.52218771	-28.8\\
10.542002678	-29.7\\
10.562021255	-30.7\\
10.582208157	-31.6\\
10.602012396	-32.4\\
10.622630835	-33.1\\
10.642049551	-33.7\\
10.662068844	-34.4\\
10.682162046	-35.2\\
10.70199728	-36\\
10.722176313	-36.8\\
10.74203825	-37.5\\
10.762045622	-38\\
10.782156229	-38.4\\
10.802023888	-38.9\\
10.822166443	-39.5\\
10.842008352	-40.2\\
10.86204505	-40.9\\
10.882124424	-41.6\\
10.902003527	-42.1\\
10.922164202	-42.7\\
10.942017317	-43.3\\
10.962011099	-43.8\\
10.982141495	-44.4\\
11.001995325	-45\\
11.022160769	-45.6\\
11.041996956	-46.1\\
11.062013865	-46.6\\
11.082167149	-46.8\\
11.102060795	-47\\
11.122785091	-47.2\\
11.141993284	-47.5\\
11.162014008	-48\\
11.182183027	-48.5\\
11.20200181	-49\\
11.222160101	-49.2\\
11.242029905	-49.3\\
11.262011051	-49.4\\
11.282140493	-49.5\\
11.302013159	-49.8\\
11.322203159	-50.1\\
11.341996431	-50.1\\
11.361999273	-50.1\\
11.382134199	-50.1\\
11.402023792	-50\\
11.422180176	-50\\
11.442027569	-50.1\\
11.462018728	-50.1\\
11.482165337	-50.1\\
11.502015114	-50.1\\
11.522163391	-50.1\\
11.541991234	-50.1\\
11.56204319	-50.1\\
11.582142353	-50.1\\
11.601999283	-50.1\\
11.622165918	-50.1\\
11.642030716	-50.1\\
11.662001848	-50.1\\
11.682154179	-49.9\\
11.702039242	-49.8\\
11.722190619	-49.5\\
11.742001534	-49.3\\
11.76201725	-49\\
11.782306194	-48.7\\
11.80202961	-48.4\\
11.822197199	-48.1\\
11.842004776	-47.7\\
11.862013817	-47.3\\
11.882138729	-46.8\\
11.902037144	-46.1\\
11.922207832	-45.5\\
11.942010164	-45\\
11.962032795	-44.5\\
11.982157469	-44.3\\
12.002034426	-44\\
12.022168159	-43.6\\
12.041992426	-43.2\\
12.062012911	-42.6\\
12.082139969	-42\\
12.102028608	-41.4\\
12.122175932	-40.8\\
12.142016888	-40.3\\
12.162007093	-39.8\\
12.182150602	-39.4\\
12.201984882	-39\\
12.222195625	-38.5\\
12.242027998	-38.1\\
12.262045145	-37.6\\
12.282132149	-37.2\\
12.301986933	-36.8\\
12.322165728	-36.4\\
12.342005491	-35.8\\
12.362001896	-35.1\\
12.382138014	-34.3\\
12.402010679	-33.7\\
12.422173023	-33.2\\
12.441994429	-32.9\\
12.46200633	-32.8\\
12.482149124	-32.8\\
12.502034187	-32.5\\
12.522147179	-32.2\\
12.542021513	-31.7\\
12.562004566	-31.2\\
12.58214283	-30.9\\
12.602001905	-30.6\\
12.622149944	-30.4\\
12.641993761	-30.1\\
12.662008047	-29.9\\
12.682161331	-29.6\\
12.701981068	-29.4\\
12.722173214	-29.2\\
12.742001534	-29.1\\
12.762007236	-29\\
12.782156467	-29\\
12.801985979	-29\\
12.822182894	-29\\
12.842017889	-29\\
12.862043858	-29\\
12.88212204	-29.1\\
12.902052879	-29.1\\
12.922176838	-29.2\\
12.941979647	-29.3\\
12.962032318	-29.5\\
12.982141495	-29.7\\
13.002001524	-29.8\\
13.022152662	-30\\
13.042025328	-30.1\\
13.062027693	-30.2\\
13.082137346	-30.3\\
13.102001429	-30.4\\
13.122209072	-30.4\\
13.141997099	-30.3\\
13.162031651	-30.3\\
13.182130337	-30.4\\
13.202013254	-30.4\\
13.222183228	-30.4\\
13.242014885	-30.4\\
13.262023926	-30.6\\
13.282182693	-30.9\\
13.302033186	-31.2\\
13.322187185	-31.5\\
13.342011929	-31.8\\
13.362032652	-32.1\\
13.382173061	-32.5\\
13.402025938	-33\\
13.422155619	-33.4\\
13.442006588	-34\\
13.462030411	-34.4\\
13.482156277	-35\\
13.502064466	-35.5\\
13.522164583	-36\\
13.542022943	-36.5\\
13.561992407	-37\\
13.582232714	-37.4\\
13.60199976	-38\\
13.622158289	-38.4\\
13.642034531	-38.9\\
13.6620543	-39.4\\
13.682138205	-39.8\\
13.702019453	-40\\
13.722175598	-40.2\\
13.742042065	-40.5\\
13.762019157	-40.8\\
13.782135963	-41.1\\
13.801988363	-41.4\\
13.822141886	-41.5\\
13.841995239	-41.5\\
13.862030745	-41.4\\
13.882129192	-41.4\\
13.902008533	-41.4\\
13.922157764	-41.5\\
13.942016602	-41.5\\
13.96199894	-41.4\\
13.982160807	-41.4\\
14.002003908	-41.4\\
14.022171021	-41.4\\
14.042001486	-41.4\\
14.062019587	-41.4\\
14.082140923	-41.3\\
14.102076769	-41.3\\
14.122169018	-41.3\\
14.141991854	-41.3\\
14.162217855	-41.3\\
14.182018042	-41.3\\
14.202015162	-41.3\\
14.222146988	-41.3\\
14.242033482	-41.3\\
14.262063742	-41.3\\
14.282141447	-41.3\\
14.301989079	-41.3\\
14.322176218	-41.3\\
14.341989994	-41.3\\
14.36200428	-41.3\\
14.3821702	-41.3\\
14.402010918	-41.3\\
14.422150373	-41.3\\
14.442054033	-41.3\\
14.462033749	-41.3\\
14.482139349	-41.3\\
14.502010345	-41.4\\
14.522170782	-41.4\\
14.542027712	-41.6\\
14.562043428	-41.7\\
14.58217597	-41.9\\
14.602017879	-42.1\\
14.622169018	-42.3\\
14.642028809	-42.6\\
14.662008762	-42.9\\
14.682173729	-43.2\\
14.702037334	-43.5\\
14.722192049	-43.7\\
14.741998196	-44\\
14.76201582	-44.4\\
14.782176256	-44.8\\
14.802025318	-45.2\\
14.822154284	-45.6\\
14.8419981	-45.9\\
14.862028122	-46.1\\
14.882154465	-46.5\\
14.902001619	-46.9\\
14.922166586	-47.4\\
14.942019224	-47.9\\
14.962034941	-48.4\\
14.982142687	-48.8\\
15.002000093	-49.1\\
15.022146463	-49.5\\
15.042012453	-49.9\\
15.06201601	-50.2\\
15.082141876	-50.7\\
15.10201335	-51.3\\
15.122178078	-51.9\\
15.141991377	-52.6\\
15.162016153	-53.3\\
15.182161808	-53.9\\
15.201999187	-54.2\\
15.222200155	-54.4\\
15.242055893	-54.7\\
15.262008667	-55.2\\
15.282166243	-55.7\\
15.302016258	-56.2\\
15.322191	-56.7\\
15.341993332	-56.9\\
15.362011433	-57.1\\
15.382182598	-57.2\\
15.402018785	-57.5\\
15.422198772	-57.8\\
15.442034245	-58.1\\
15.462046862	-58.4\\
15.482173443	-58.7\\
15.502015114	-58.8\\
15.522414684	-58.7\\
15.542039871	-58.7\\
15.562013388	-58.9\\
15.58214283	-59.9\\
15.602000952	-62.5\\
15.622148275	-67.1\\
15.642017603	-72.8\\
15.662034512	-77.8\\
15.682162046	-80.7\\
15.702001095	-81\\
15.722177267	-79.6\\
15.741961718	-78.3\\
15.762009382	-78.5\\
15.782168627	-80.4\\
15.801988125	-83.3\\
15.822143555	-85.8\\
15.84201169	-86.9\\
15.862009764	-87\\
15.882174969	-86.7\\
15.901987791	-87\\
15.92216754	-88\\
15.941969395	-89.6\\
15.961994886	-91.1\\
15.982156992	-92\\
16.00205946	-91.9\\
16.022174835	-91.3\\
16.041994572	-90.9\\
16.062048912	-91\\
16.082163095	-91.5\\
16.102010965	-92.4\\
16.122182608	-93.3\\
16.141976595	-94.1\\
16.162011385	-94.6\\
16.182132483	-95\\
16.201993942	-95.2\\
16.222128391	-95.4\\
16.241983175	-95.6\\
16.261989355	-95.8\\
16.282131672	-95.9\\
16.301984072	-96\\
16.322142363	-96.1\\
16.341975689	-96.4\\
16.362000465	-96.8\\
16.38217926	-97.3\\
16.401986361	-97.9\\
16.422143936	-98.3\\
16.442050934	-98.4\\
16.462029934	-98.4\\
16.482128382	-98.4\\
16.501989365	-98.4\\
16.522183657	-98.5\\
16.541991472	-98.7\\
16.561990738	-98.9\\
16.582096577	-99\\
16.602063894	-99.1\\
16.622097969	-99.2\\
16.642044306	-99.3\\
16.662005901	-99.5\\
16.682119608	-99.7\\
16.70204854	-99.9\\
16.72213316	-100\\
16.742088079	-100.1\\
16.761964321	-100.1\\
16.782146931	-100.1\\
16.802073479	-100.1\\
16.822122812	-100.1\\
16.842083931	-100.2\\
16.861982107	-100.4\\
16.882124901	-100.6\\
16.902066469	-100.9\\
16.92209816	-101.1\\
16.942107677	-101.1\\
16.962004423	-101.1\\
16.982093573	-101\\
17.002074718	-101\\
17.022132635	-101\\
17.042076826	-101.1\\
17.061970949	-101.1\\
17.082075357	-101.2\\
17.102052212	-101.2\\
17.122137308	-101.2\\
17.142031193	-101.3\\
17.16200757	-101.3\\
17.182082891	-101.4\\
17.20207262	-101.4\\
17.222100496	-101.4\\
17.242043972	-101.4\\
17.261957645	-101.4\\
17.282083273	-101.4\\
17.302093744	-101.5\\
17.322118044	-101.5\\
17.34204793	-101.5\\
17.361960173	-101.6\\
17.382108212	-101.7\\
17.402024508	-101.8\\
17.422100306	-102\\
17.442066431	-102\\
17.462002516	-102.1\\
17.48214221	-102.1\\
17.501986027	-102.1\\
17.522157192	-102.2\\
17.54199338	-102.2\\
17.561994553	-102.3\\
17.582138062	-102.4\\
17.602009773	-102.6\\
17.622126341	-102.8\\
17.642013788	-103.1\\
17.662002325	-103.3\\
17.682146549	-103.4\\
17.701998472	-103.4\\
17.722178459	-103.4\\
17.742015362	-103.3\\
17.762011766	-103.4\\
17.782153606	-103.4\\
17.801993847	-103.5\\
17.822180986	-103.6\\
17.841971636	-103.7\\
17.861990213	-103.9\\
17.882103443	-104.1\\
17.901988268	-104.2\\
17.922172308	-104.3\\
17.941988468	-104.4\\
17.962006807	-104.4\\
17.982125759	-104.4\\
18.00197053	-104.4\\
18.022149801	-104.4\\
18.042065859	-104.5\\
18.062117338	-104.6\\
18.082151651	-104.6\\
18.10199523	-104.7\\
18.122134924	-104.8\\
18.142056704	-105\\
18.162050724	-105.3\\
18.18215251	-105.5\\
18.201961994	-105.7\\
18.222989321	-105.7\\
18.242043018	-105.6\\
18.262099743	-105.6\\
18.282104254	-105.6\\
18.301981449	-105.7\\
18.322139263	-105.7\\
18.34203124	-105.8\\
18.362072468	-105.9\\
18.382111073	-106\\
18.401993513	-106.2\\
18.422118425	-106.3\\
18.442046881	-106.4\\
18.462067366	-106.5\\
18.482110023	-106.5\\
18.501984835	-106.5\\
18.522121668	-106.6\\
18.542054415	-106.6\\
18.562091351	-106.7\\
18.582144976	-106.7\\
18.602002382	-106.8\\
18.622121572	-106.8\\
18.642043829	-107\\
18.662081718	-107.1\\
18.68214798	-107.2\\
18.701971769	-107.2\\
18.722103596	-107.2\\
18.742054224	-107.2\\
18.762120247	-107.3\\
18.782105923	-107.3\\
18.801943541	-107.4\\
18.822140217	-107.4\\
18.842042446	-107.4\\
18.862076283	-107.5\\
18.882124662	-107.5\\
18.901986361	-107.6\\
18.922109604	-107.7\\
18.942097187	-107.7\\
18.962079048	-107.7\\
18.98213625	-107.8\\
19.001996517	-107.8\\
19.022141933	-107.8\\
19.042103291	-107.8\\
19.062106609	-107.8\\
19.082149506	-107.9\\
19.10197401	-107.9\\
19.122107983	-108\\
19.142064571	-108\\
19.16207552	-108\\
19.182133913	-108\\
19.20197773	-108.1\\
19.22210598	-108.1\\
19.242041588	-108.1\\
19.262112141	-108.1\\
19.282094955	-108.2\\
19.301951408	-108.2\\
19.322146893	-108.2\\
19.342025757	-108.2\\
19.36207962	-108.2\\
19.382131338	-108.2\\
19.401996851	-108.3\\
19.42209363	-108.3\\
19.442054033	-108.3\\
19.462082148	-108.3\\
19.482102394	-108.3\\
19.501953125	-108.3\\
19.522129059	-108.4\\
19.54203248	-108.4\\
19.562086105	-108.4\\
19.582078218	-108.4\\
19.601973534	-108.4\\
19.62209034	-108.4\\
19.642031431	-108.4\\
19.662068129	-108.4\\
19.682100773	-108.5\\
19.701996803	-108.5\\
19.722111464	-108.5\\
19.742014885	-108.5\\
19.76205492	-108.5\\
19.78210783	-108.5\\
19.801952839	-108.5\\
19.822114706	-108.5\\
19.842041016	-108.6\\
19.862072468	-108.6\\
19.882095337	-108.7\\
19.901971579	-108.8\\
19.922175407	-108.8\\
19.942037582	-108.9\\
19.962098598	-108.9\\
19.982095718	-108.9\\
20.001986742	-108.9\\
20.022093058	-108.9\\
20.042049885	-108.9\\
20.062073946	-108.9\\
20.082108974	-109\\
20.102021933	-109\\
20.12212491	-109\\
20.142061234	-109.1\\
20.162065506	-109.1\\
20.182135582	-109.2\\
20.201997042	-109.3\\
20.222113132	-109.3\\
20.242067337	-109.3\\
20.262081623	-109.3\\
20.282110453	-109.4\\
20.301987171	-109.4\\
20.32218504	-109.4\\
20.342083931	-109.5\\
20.362050056	-109.5\\
20.382139683	-109.6\\
20.401967287	-109.7\\
20.422115803	-109.8\\
20.442018032	-109.8\\
20.462073326	-109.9\\
20.482110739	-109.9\\
20.501979351	-109.9\\
20.522124767	-109.9\\
20.542053938	-110\\
20.562147379	-110.1\\
20.582041979	-110.1\\
20.601985216	-110.1\\
20.622135162	-110.2\\
20.642071724	-110.2\\
20.662116528	-110.3\\
20.682121754	-110.3\\
20.702011585	-110.3\\
20.72214818	-110.3\\
20.742044687	-110.3\\
20.762064934	-110.4\\
20.782114506	-110.4\\
20.801952362	-110.4\\
20.822112322	-110.5\\
20.842067003	-110.5\\
20.862078905	-110.5\\
20.882071257	-110.6\\
20.901990891	-110.6\\
20.922109604	-110.7\\
20.942028046	-110.7\\
20.962088346	-110.7\\
20.982127428	-110.7\\
21.001998425	-110.7\\
21.022111893	-110.8\\
21.042070627	-110.8\\
21.0620718	-110.8\\
21.082098961	-110.8\\
21.101972342	-110.8\\
21.122126341	-110.8\\
21.142024517	-110.8\\
21.162059784	-110.8\\
21.182094336	-110.8\\
21.201954126	-110.9\\
21.222126961	-110.9\\
21.242029667	-110.9\\
21.262105465	-110.9\\
21.282117128	-110.9\\
21.301984787	-110.9\\
21.32212925	-110.9\\
21.34205842	-110.9\\
21.362068653	-110.9\\
21.382144451	-110.9\\
21.402006865	-110.9\\
21.422141314	-110.9\\
21.442042828	-110.9\\
21.462052345	-110.9\\
21.48212266	-110.9\\
21.501994848	-110.9\\
21.522117138	-110.9\\
21.542080402	-110.9\\
21.562087774	-110.9\\
21.582148314	-110.9\\
21.601935863	-110.9\\
21.622135878	-110.9\\
21.642077923	-110.9\\
21.662090302	-110.9\\
21.682119846	-110.9\\
21.702045679	-110.9\\
21.722146749	-110.9\\
21.742062569	-110.9\\
21.762056112	-110.9\\
21.782142401	-110.9\\
21.801938295	-110.9\\
21.822087765	-110.9\\
21.842089176	-110.9\\
21.862075567	-110.9\\
21.882115126	-110.9\\
21.901961088	-110.9\\
21.922130823	-110.9\\
21.942079544	-110.9\\
21.962067366	-110.9\\
21.982094288	-110.9\\
22.001972437	-110.9\\
22.022099018	-110.9\\
22.042035818	-110.9\\
22.062061071	-110.9\\
22.082083464	-110.9\\
22.101983786	-110.9\\
22.122139454	-110.9\\
22.142041922	-110.9\\
22.162076473	-110.9\\
22.182096958	-110.9\\
22.201950312	-110.9\\
22.222105026	-110.9\\
22.242027521	-110.9\\
22.262073755	-110.9\\
22.282116652	-110.9\\
22.301965952	-110.9\\
22.322112083	-110.9\\
22.342087507	-110.9\\
22.362069368	-110.9\\
22.382096529	-110.9\\
22.40201211	-110.9\\
22.422106504	-110.9\\
22.442033052	-110.9\\
22.462079525	-110.9\\
22.482102156	-110.9\\
22.502007484	-110.9\\
22.522216558	-110.9\\
22.542050838	-110.9\\
22.56208086	-110.9\\
22.582110882	-110.9\\
22.601999998	-110.9\\
22.62213397	-110.9\\
22.642079115	-110.9\\
22.662142754	-111\\
22.682119846	-111\\
22.701997519	-111\\
22.722173452	-111\\
22.742031813	-111\\
22.762136936	-111\\
22.782159567	-111.1\\
22.80195117	-111.1\\
22.822102785	-111.1\\
22.842041254	-111.1\\
22.862128019	-111.1\\
22.882138968	-111.1\\
22.901978731	-111.1\\
22.922153234	-111.1\\
22.942016125	-111.1\\
22.962067842	-111.1\\
22.982091665	-111.1\\
23.001998425	-111.1\\
23.022127628	-111.1\\
23.042078495	-111.1\\
23.062091589	-111.1\\
23.082093477	-111.1\\
23.1019876	-111.1\\
23.122159719	-111.2\\
23.142050505	-111.2\\
23.162089109	-111.2\\
23.182124615	-111.2\\
23.201952696	-111.2\\
23.222137928	-111.2\\
23.242037773	-111.2\\
23.262085438	-111.2\\
23.282083511	-111.2\\
23.301974058	-111.2\\
23.322146416	-111.2\\
23.34202981	-111.2\\
23.362091303	-111.2\\
23.382113695	-111.2\\
23.401981354	-111.3\\
23.42235136	-111.3\\
23.442028999	-111.3\\
23.4620924	-111.3\\
23.482110739	-111.3\\
23.501963854	-111.3\\
23.522117138	-111.3\\
23.542038918	-111.3\\
23.56214571	-111.3\\
23.582130194	-111.3\\
23.601999044	-111.3\\
23.622128725	-111.3\\
23.642071247	-111.3\\
23.662063122	-111.3\\
23.682090282	-111.3\\
23.701986551	-111.3\\
23.722147942	-111.3\\
23.742023468	-111.2\\
23.762069225	-111.2\\
23.782132864	-111.2\\
23.801974297	-111.2\\
23.822155952	-111.2\\
23.842038155	-111.2\\
23.86207509	-111.2\\
23.882091522	-111.3\\
23.901991844	-111.3\\
23.92219615	-111.3\\
23.942068815	-111.3\\
23.962062836	-111.3\\
23.982153893	-111.3\\
24.001996994	-111.3\\
24.02212429	-111.3\\
24.042058945	-111.3\\
24.062047243	-111.3\\
24.082148075	-111.3\\
24.101991177	-111.3\\
24.122166872	-111.3\\
24.142047405	-111.3\\
24.162062168	-111.3\\
24.182142496	-111.3\\
24.201944351	-111.3\\
24.222126722	-111.3\\
24.242063284	-111.3\\
24.262085915	-111.3\\
24.282108784	-111.3\\
24.302018881	-111.3\\
24.32211256	-111.3\\
24.342081308	-111.3\\
24.362098455	-111.3\\
24.382167339	-111.3\\
24.401962757	-111.3\\
24.422138929	-111.3\\
24.442104578	-111.3\\
24.462070942	-111.3\\
24.482081175	-111.3\\
24.501969337	-111.4\\
24.522120237	-111.4\\
24.542026281	-111.4\\
24.562080383	-111.4\\
24.582099199	-111.4\\
24.60195756	-111.4\\
24.622094393	-111.4\\
24.642066956	-111.4\\
24.662106752	-111.4\\
24.682098866	-111.4\\
24.701967239	-111.4\\
24.722149849	-111.4\\
24.742028475	-111.4\\
24.762083054	-111.4\\
24.782258987	-111.4\\
24.801982403	-111.4\\
24.822084904	-111.4\\
24.842042446	-111.4\\
24.862068176	-111.4\\
24.882104397	-111.5\\
24.901977301	-111.6\\
24.922142029	-111.7\\
24.941977024	-111.8\\
24.961987257	-111.9\\
24.982146025	-112\\
25.001965761	-112\\
25.022131443	-112\\
25.041999102	-112\\
25.06201458	-112.1\\
25.082130671	-112.2\\
25.102022648	-112.3\\
25.122161627	-112.7\\
25.141970634	-113.1\\
25.1620152	-113.5\\
25.182200909	-113.8\\
25.20199275	-114\\
25.222173214	-114\\
25.242016792	-114\\
25.262026548	-114.1\\
25.282164812	-114.3\\
25.302000761	-114.6\\
25.322451115	-114.8\\
25.341986418	-115.1\\
25.362017155	-115.4\\
25.382159948	-116\\
25.402005196	-116.8\\
25.422199011	-117.6\\
25.442036152	-118.1\\
25.462018251	-118.5\\
25.482153416	-118.6\\
25.501992464	-118.7\\
25.522176027	-118.9\\
25.541989326	-119.4\\
25.562006712	-120\\
25.582149982	-120.5\\
25.60199666	-121\\
25.622157097	-121.5\\
25.642018318	-122.1\\
25.662026882	-122.8\\
25.682142019	-123.5\\
25.701983929	-124.3\\
25.722192764	-124.9\\
25.742002487	-125.3\\
25.762026787	-125.7\\
25.782164574	-126\\
25.802037954	-126.4\\
25.822148085	-126.9\\
25.842014313	-127.6\\
25.862020493	-128.5\\
25.882134438	-129.7\\
25.902004719	-130.8\\
25.92215848	-132\\
25.942056656	-132.9\\
25.962001324	-133.5\\
25.982146978	-133.9\\
26.001991034	-134.1\\
26.022203445	-134.1\\
26.042020321	-134.1\\
26.062003136	-134\\
26.082131386	-133.9\\
26.102059603	-133.4\\
26.122186184	-131.8\\
26.14200139	-128.9\\
26.162011147	-125.2\\
26.182165861	-121.2\\
26.202027321	-117.9\\
26.222157955	-115.4\\
26.241994619	-113.7\\
26.26205945	-112.2\\
26.282147408	-110.1\\
26.302007675	-107.5\\
26.32218647	-104.2\\
26.342027903	-100.5\\
26.362017393	-96.6\\
26.382159233	-93\\
26.40203166	-90\\
26.422172785	-88.1\\
26.442005873	-87.6\\
26.462018013	-88\\
26.482143641	-89\\
26.502004623	-90\\
26.52218914	-91.1\\
26.542043686	-92.9\\
26.562024593	-95.7\\
26.58214736	-99.7\\
26.602050304	-104.6\\
26.622169733	-110\\
26.642041922	-115.3\\
26.662023783	-120.3\\
26.682135582	-125.5\\
26.702027559	-130.9\\
26.722219467	-137.3\\
26.742015123	-144.5\\
26.762050629	-152.3\\
26.78219223	-160.1\\
26.802005053	-167.1\\
26.822182894	-173.4\\
26.842005968	-178.6\\
26.862059593	-183.4\\
26.882169008	-188\\
26.902005434	-192.8\\
26.922175646	-198.1\\
26.941989899	-203.1\\
26.962054014	-207.6\\
26.982164621	-211\\
27.001997948	-213.5\\
27.022144079	-215.5\\
27.042027235	-217.5\\
27.062027931	-219.8\\
27.082180977	-222.3\\
27.101997614	-224.6\\
27.122191906	-226.4\\
27.14197278	-227.4\\
27.162033796	-227.8\\
27.182127476	-227.7\\
27.202015162	-227.4\\
27.222187519	-226.9\\
27.242005348	-226.2\\
27.262019634	-225\\
27.282144785	-222.8\\
27.302032948	-219.7\\
27.322174788	-216.3\\
27.341990471	-213\\
27.362036943	-210.1\\
27.382180452	-208\\
27.402003288	-206.5\\
27.422185183	-205.3\\
27.441998482	-203.9\\
27.462059498	-202.4\\
27.482139826	-200.9\\
27.502019405	-199.9\\
27.52219367	-199.4\\
27.542020559	-199.7\\
27.562032223	-200.2\\
27.582131863	-200.7\\
27.60202384	-200.9\\
27.622142792	-201.1\\
27.641989231	-201.4\\
27.662005186	-202.2\\
27.682174444	-204\\
27.702018023	-206.5\\
27.722179651	-209.4\\
27.742043257	-212.2\\
27.762024403	-214.8\\
27.782186508	-217.2\\
27.802052259	-219.5\\
27.822200298	-222.1\\
27.842039585	-224.7\\
27.862027645	-227\\
27.88215971	-228.8\\
27.902037382	-230\\
27.922200918	-230.7\\
27.942023039	-231.7\\
27.961997271	-233\\
27.982176065	-234.7\\
28.002029896	-236.5\\
28.022160769	-238\\
28.042056322	-238.9\\
28.062054634	-239.4\\
28.082167864	-239.4\\
28.101998329	-239.2\\
28.122165918	-239.1\\
28.142002344	-239.1\\
28.162032843	-239.2\\
28.182125807	-239.3\\
28.20200634	-239.1\\
28.222186565	-238.7\\
28.24201417	-237.9\\
28.26203227	-236.8\\
28.282155037	-235.3\\
28.302013397	-233.9\\
28.322194576	-232.3\\
28.342026711	-230.7\\
28.362207413	-228.9\\
28.382018805	-226.9\\
28.401983261	-224.9\\
28.422148466	-222.8\\
28.442022324	-220.9\\
28.462060928	-219\\
28.482171297	-217.2\\
28.502037048	-215.4\\
28.522174835	-213.5\\
28.542027235	-211.4\\
28.562031269	-209.3\\
28.58214879	-207.1\\
28.602038383	-204.8\\
28.622171879	-202.1\\
28.642048836	-199.4\\
28.662041903	-196.8\\
28.68214035	-194.2\\
28.702005386	-192.1\\
28.722162008	-190.1\\
28.741998196	-188.4\\
28.762242794	-186.4\\
28.782024622	-184.5\\
28.802007675	-182.2\\
28.822199821	-180\\
28.842037678	-177.6\\
28.862053871	-174.9\\
28.882201672	-171.9\\
28.902004957	-168.8\\
28.922200203	-165.9\\
28.942040682	-163.2\\
28.962020874	-161.2\\
28.982197523	-159.4\\
29.002066135	-157.7\\
29.022214174	-155.9\\
29.042006969	-154\\
29.062025785	-151.9\\
29.082149029	-149.8\\
29.102013111	-147.8\\
29.122197628	-145.7\\
29.142024517	-143.2\\
29.162008762	-140.5\\
29.182209492	-137.5\\
29.202017546	-134.6\\
29.222159147	-132\\
29.242037773	-129.9\\
29.262034893	-128\\
29.282139063	-126.2\\
29.302011013	-124.4\\
29.322177172	-122.3\\
29.341993809	-119.9\\
29.362015963	-117\\
29.382160664	-113.8\\
29.402008772	-110.5\\
29.42217803	-107.5\\
29.442031622	-104.9\\
29.462039232	-102.7\\
29.482152462	-101\\
29.502002716	-99.1\\
29.522171021	-97.1\\
29.542028189	-94.8\\
29.561997652	-92.4\\
29.582170963	-90\\
29.602061987	-87.7\\
29.622183561	-85.6\\
29.642023087	-83.4\\
29.662038803	-81.2\\
29.682175398	-79\\
29.702017307	-76.8\\
29.7221632	-74.7\\
29.742003441	-72.9\\
29.762031317	-71.2\\
29.782172918	-69.7\\
29.80201149	-67.9\\
29.822170258	-66.1\\
29.841985941	-63.9\\
29.862051249	-61.5\\
29.882136822	-58.9\\
29.902020931	-56.5\\
29.922157049	-54.2\\
29.942032099	-52.3\\
29.962019682	-50.5\\
29.982144356	-48.9\\
30.001995087	-47.1\\
30.022142649	-45.5\\
30.042032719	-43.6\\
30.062025785	-41.8\\
30.082151651	-40\\
30.102025986	-38.3\\
30.122210503	-36.6\\
30.142000198	-34.8\\
30.162026405	-32.9\\
30.18218565	-30.9\\
30.202034473	-28.9\\
30.222190619	-26.9\\
30.242018938	-25.3\\
30.262004852	-23.9\\
30.282184839	-22.4\\
30.302031279	-20.7\\
30.322157621	-18.6\\
30.341989279	-16.5\\
30.362034082	-14.5\\
30.382180929	-13\\
30.40200758	-11.7\\
30.422162294	-10.7\\
30.441993237	-9.89999999999998\\
30.462149858	-8.79999999999995\\
30.481984854	-7.39999999999998\\
30.502017498	-6\\
30.522214174	-4.5\\
30.54199481	-3\\
30.562006474	-1.89999999999998\\
30.58219862	-0.799999999999955\\
30.602005482	0.200000000000045\\
30.622174025	1.10000000000002\\
30.641998529	2\\
30.662027597	2.70000000000005\\
30.682175398	3.5\\
30.702009439	4.10000000000002\\
30.722192287	4.70000000000005\\
30.742059469	5.39999999999998\\
30.76218152	6\\
30.782032251	6.70000000000005\\
30.801999331	7.5\\
30.822172165	8.20000000000005\\
30.842003107	8.89999999999998\\
30.862011671	9.5\\
30.882138491	10\\
30.902043819	10.4\\
30.922194719	10.7\\
30.941994905	11\\
30.962159157	11.1\\
30.982009649	11.3\\
31.001997709	11.4\\
31.02216053	11.7\\
31.041988611	11.9\\
31.062040567	12.2\\
31.082134008	12.3\\
31.10200572	12.5\\
31.122170448	12.8\\
31.141984224	13\\
31.162046432	13.1\\
31.182151079	13.2\\
31.202049017	13.1\\
31.222155809	13.1\\
31.242011547	13.1\\
31.262024164	13.2\\
31.282152891	13.2\\
31.302026033	13.2\\
31.322161198	13.3\\
31.341996908	13.3\\
31.362005234	13.5\\
31.382138968	13.6\\
31.402026415	13.7\\
31.422175169	13.8\\
31.442034721	13.9\\
31.462235212	13.9\\
31.48198843	13.8\\
31.501996994	13.9\\
31.522185087	13.9\\
31.541993618	13.9\\
};
\addplot [color=black, forget plot]
  table[row sep=crcr]{%
9.21000003814697	-250\\
9.21000003814697	100\\
};
\node[right, align=left, circle,fill=white,draw=black,thick,line width=0.3mm,inner sep=1pt]
at (axis cs:6.21,-162.5) {1};
\addplot [color=black, forget plot]
  table[row sep=crcr]{%
24.3599998950958	-250\\
24.3599998950958	100\\
};
\node[right, align=left, circle,fill=white,draw=black,thick,line width=0.3mm,inner sep=1pt]
at (axis cs:21.36,12.5) {2};
\addplot [color=black, forget plot]
  table[row sep=crcr]{%
28.8900001049042	-250\\
28.8900001049042	100\\
};
\node[right, align=left, circle,fill=white,draw=black,thick,line width=0.3mm,inner sep=1pt]
at (axis cs:25.89,12.5) {3};
\end{axis}
\end{tikzpicture}%